\documentclass[twoside,11pt]{article}
\PassOptionsToPackage{pagebackref=true}{hyperref}
\usepackage{jmlr2e}
\usepackage{enumitem} 
\usepackage{amsmath} 
\usepackage{color}
\usepackage{verbatim}

\def\R{\mathbb{R}}
\def\norm#1{\|#1\|}
\def\half{\frac 1 2}

\def\argmin{\mathop{\rm arg\,min}}
\renewcommand*\backref[1]{\ifx#1\relax \else (cited on #1) \fi}

\definecolor{red}{rgb}{1,0,0}
\def\red#1{{\color{red}#1}}

\usepackage{lastpage}

\ShortHeadings{Why Line Search when you can Plane Search?}{Shea and Schmidt}
\firstpageno{1}

\begin{document}

\title{Why Line Search when you can Plane Search?\\
\large SO-Friendly Neural Networks allow Per-Iteration Optimization of\\ Learning and Momentum Rates for Every Layer}

\author{\name Betty Shea \email sheaws@cs.ubc.ca \\
       \addr        University of British Columbia
              \AND
       \name Mark Schmidt \email schmidtm@cs.ubc.ca \\
       \addr         University of British Columbia, Canada CIFAR AI Chair (Amii)
}

\editor{}

\maketitle

\begin{abstract}
  We introduce the class of SO-friendly neural networks, which include several models used in practice including networks with 2 layers of hidden weights where the number of inputs is larger than the number of outputs. SO-friendly networks have the property that performing a precise line search to set the step size on each iteration has the same asymptotic cost  during full-batch training as using a fixed learning. Further, for the same cost a plane-search can be used to set both the learning and momentum rate on each step. Even further, SO-friendly networks also allow us to use subspace optimization to set a learning rate and momentum rate for each layer on each iteration. We explore augmenting gradient descent as well as quasi-Newton methods and Adam with line optimization and subspace optimization, and our experiments indicate that this gives fast and reliable ways to train these networks that are insensitive to hyper-parameters.
  \end{abstract}


\section{Should we use Subspace Optimization in Machine Learning?}

To train machine learning (ML) models, we often try to minimize a function $f$ in terms of some parameters $w$. Starting from an initial guess $w_0$, a prototpyical algorithm for performing this minimization is \emph{gradient descent with momentum} (GD+M)~\citep{Polyak1964} which generates a sequence of guesses according to the formula\footnote{With the convention that $w_{-1}=w_0$.}
\begin{equation}
	w_{k+1} = w_k - \alpha_k\nabla f(w_k) + \beta_k(w_k - w_{k-1}).
	\label{eq:GDM}
\end{equation}
This update subtracts the gradient $\nabla f(w_k)$ multiplied by what we call the \emph{learning rate} $\alpha_k$, and then adds the momentum term $(w_k - w_{k-1})$ multiplied by what we call the \emph{momentum rate} $\beta^k$. Many variations exist including stochastic gradient descent (SGD) which uses a stochastic approximation of the gradient to reduce the iteration cost~\citep{Robbins1951} and quasi-Newton methods which pre-multiply the gradient by a matrix to improve the convergence rate~\citep[see][]{Nocedal2006}. The most popular variant in ML is the Adam optimizer which incorporates a variation on momentum and uses a scaling of a stochastic approximation of the gradient~\citep{Kingma2014}.

In the modern practice of ML, the learning rate $\alpha_k$ and momentum rate $\beta_k$ are often set to fixed constants for all $k$. This can result in poor performance of Adam and related methods~\citep{Vaswani2020}, so practitioners often run the algorithm with different choices of the constant learning and momentum rates. After running the algorithm with different hyper-parameter choices, we then choose the run giving the best performance~\citep{Zhang2022}. This ``hyper-parameter search'' approach to solving an optimization problem is expensive compared to traditional deterministic numerical optimization approaches like quasi-Newton methods. Popular quasi-Newton methods, such as those based on the limited-memory Broyden-Fletcher-Goldfarb-Shanno (L-BFGS) update~\citep{Nocedal1980}, typically give good performance ``out of the box''. Specifically, deterministic methods often give good performance without any hyper-parameter tuning using their default hyper-parameters. 

A key ingredient that allows L-BFGS to work well with default parameters is using a \emph{line search} (LS) to set the learning rate. In particular, they search for a value of $\alpha_k$ satisfying conditions guaranteeing sufficient progress in optimizing $f$ such as the strong Wolfe conditions~\citep{Wolfe1969,Nocedal2006}. This may require testing multiple learning rates on each iteration, but various heuristics exist that allow us to quickly find a suitable $\alpha_k$~\citep{More1994,Shanno1978}.
With these heuristics, the cost of using line search is typically  smaller than running the algorithm for multiple fixed learning rates. Further, since 
line searches allow us to find a good learning rate on each iteration, using line searches has the potential to drastically outperform any fixed learning rate. Several recent works highlight the advantages of using line search for modern ML problems within SGD~\citep{Truong2018,Vaswani2019} as well as Adam~\citep{Vaswani2020,galli2023don}. However, using line search increases the iteration cost of the method and these methods still use a fixed momentum rate.

In the context of many linear models like least squares and logistic regression, it is possible to use line searches ``for free''. In particular, for what we call linear-composition problems (LCPs) using a LS or even performing line optimization (LO) to numerically solve for the best step size does not change the asymptotic cost of the iterations compared to using a fixed step size. Further, for this same cost LCPs also allow us to use a \emph{plane search} to set the learning rate $\alpha_k$ and momentum rate $\beta_k$ on each iteration of GD+M~\eqref{eq:GDM} and related methods. In Section~\ref{sec:LCP} we review how we can optimize the learning and momentum rates for LCPs for the same cost as using a fixed $\alpha_k$ and $\beta_k$, and how this leads to significantly faster convergence in practice. Using a plane search is a special case of methods that use \emph{subspace optimization} (SO) to search along multiple directions on each iteration, and in Section~\ref{sec:related} we discuss the history of SO methods. 

LO and SO are not typically used for neural networks since performing LO or SO in general neural network model cannot be done ``for free'' as it can in linear models with suitable directions. In this work, we introduce the class of SO-friendly neural networks (Section~\ref{sec:SOfriendly}). The key property of these networks is that performing LO to set the learning rate, or SO to set the learning rate and momentum rate, does not increase the asymptotic cost of training compared to using a fixed learning and momentum rate. While SO-friendly networks are a restricted class, they include some networks used in practice such as networks with 2 layers of weights where the number of inputs is much larger than the number of outputs. A further appealing property of SO-friendly networks is that {\bf we can optimize over a separate learning rate and momentum rate for each layer}. For networks with 2 layers of weights this gives us 4 step sizes that are tuned ``for free'' on each iteration, and this flexibility can lead to even faster convergence in practice in some settings. In Section~\ref{sec:algorithms} we consider augmenting L-BFGS and Adam with LO and SO, and show that this leads to faster training for both LCPs and SO-friendly networks. 

\subsection{List of Contributions by Figure}

We have performed a wide variety of experiments that support our main claim that LO and SO can and should be used to improve the performance of various ML algorithms for suitable problem structures. Here we give of figures highlighting various observations from our experiments:
\begin{itemize}
\item Figure~\ref{fig:firstlogReg} illustrates how LO and SO can significantly improve the performance of GD+M on the LCP of logistic regression.
\item Figure~\ref{fig:NAGlogReg} shows that GD+M with LO and SO can even improve performance over accelerated gradient methods, but that performing SO over more than 2 directions only results in marginal performance gains.
\item Figure~\ref{fig:firstnn100} shows that LO and SO also lead to improved performance of GD+M for training 2-layer neural networks.
\item Figure~\ref{fig:firstSplitnn100} shows that using SO to set per-layer step sizes can sometimes substantially improve performance for training 2-layer networks, but sometimes makes performance much worse.
\item Figure~\ref{fig:firstSplitregnn100} shows that using per-layer step sizes consistently improves performance if regularization is added.
\item Figure~\ref{fig:QNlogReg} and~\ref{fig:QNnn100} show that LO can improve the performance of a quasi-Newton method, and that using SO to optimize the step sizes in a quasi-Newton method with momentum further improves performance.
\item Figures~\ref{fig:AdamlogReg} and~\ref{fig:Adamnn100} shows that LO improves the performance of Adam for LCPs though not always for neural networks, but that a multi-direction Adam method with step sizes set using SO typically improves performance over single-direction Adam methods.
\end{itemize}

\subsection{Comments on Limitations of the Applicability of Subspace Optimization}

In this work we restrict attention to deterministic optimization methods, to the restricted class of SO-friendly networks, and to differentiable objective functions. Before beginning, we briefly comment on these choices:
\begin{itemize}
\item {\bf Why focus on deterministic methods?} While stochastic methods do indeed become superior to deterministic as the number of examples increases, in many practical scenarios we have a finite dataset and we perform multiple passes through the data to train a model. For example, on the ImageNet dataset it is standard to perform 60 passes through the data (and this cost may be multiplied by the number of optimization hyper-parameter settings that are explored). It is not obvious that stochastic methods necessarily dominate deterministic methods in this common multi-pass setting, since it is possible that deterministic methods could exist that find accurate solutions in 10-20 passes through the data. Further, deterministic methods typically do not require a search over optimization hyper-parameters. The insensitivity of deterministic methods to hyper-parameter settings also means we can have more confidence that deterministic optimizers will work robustly in new scenarios. 
\item {\bf Why focus on SO-friendly networks?} Most deep neural networks are not SO-friendly and thus it is not efficient to use SO. Nevertheless, a variety of SO-friendly networks are used in practice (see Sections~\ref{sec:LCP} and~\ref{sec:SOfriendly})  including many networks with two hidden layers of weights or deep networks where we have sub-optimized individual layers. There has been recent interest in the ML literature on improved optimization methods for such networks~\citep{mishkin2022fast}, and we believe that practitioners in these settings would substantially reduce training times (or increase accuracy) by exploiting SO.  
\item {\bf Why focus on differentiable objectives?} Some objective functions arising in ML are non-differentiable, such as support vector machines which are LCPs and neural networks with rectified linear unit (ReLU) non-linear functions. Note that SO remains efficient for LCPs and SO-friendly networks even when the problem is non-differentiable. However, in non-differentiable cases it is less obvious which directions to use as search directions (since neither the gradient nor its negation necessarily point in a direction that we can search in order to improve the objective function).
\end{itemize}
Nevertheless, in Section~\ref{sec:discuss} we discuss how future work could use SO as a useful ingredient in stochastic methods and/or methods to train general deep neural networks.

\section{Line Search and Plane Search for Linear Composition Problems (LCPs)}
\label{sec:LCP}

Optimizing a function over a low-dimensional subspace is easier than optimizing the function over the full space. However, for a given problem it is possible that SO is still too expensive for it to be beneficial. Nevertheless, for certain problems we can perform LO or SO over a low-dimensional subspace without increasing the asymptotic iteration cost compared to using a fixed step size. We call these problems \emph{SO-friendly}. \citet{Narkiss2005} were the first to highlight an SO-friendly problem, for a class of LCPs. 

LCPs have the form $f(w)=g(Xw)$
for the $n \times d$ data matrix $X$ and a function $g$. For this problem setting, SO is efficient in the common case where the matrix multiplication with $X$ is much more expensive than evaluating $g$. For example, for a target vector $y$ the least squares objective is $f(w) = \half\norm{Xw-y}^2$, which can be written as an LCP with $g(m) = \half\norm{m - y}^2$. In this setting the matrix multiplication costs $O(nd)$ but evaluating $g$ only costs $O(n)$. Another canonical example is logistic regression, where $g(m) = \sum_{i=1}^n\log(1+\exp(-y_im_i))$ for binary labels $y_i \in \{-1,1\}$. Many other classic models in ML can be written as LCPs such as robust regression with the Huber loss and support vector machines. More advanced situations where LCPs arise include include neural tangent kernels (NTKs)~\citep{Jacot2018} and transfer learning by re-training the last layer of a pre-trained deep neural network~\citep{Donahue2014}.

\subsection{Efficient Line Search (LS) and Line Optimization (LO)}

The classic gradient descent (GD) iteration takes the form
\begin{equation}
w_{k+1} = w_k -\alpha_k \nabla f(w_k).
\label{eq:GD}
\end{equation}
To ensure that the step size $\alpha_k$ is sufficient small, standard deterministic optimizers impose the Armijo sufficient decrease condition
\begin{equation}
f(w_k - \alpha_k\nabla f(w_k)) \leq f(w_k) - \alpha_k\sigma \norm{\nabla f(w_k)}^2,
\label{eq:armijo}
\end{equation}
for a positive sufficient decrease factor $\sigma$.
LS codes may also ensure the step size is not too small by imposing an additional curvature condition as in the strong Wolfe conditions~\citep{Nocedal2006}. There exist approaches to finding an $\alpha_k$ satisfying these types of conditions that require only a small number of evaluations of $f$~\citep{More1994}. While these methods are effective for general smooth functions, in the special case of LCPs we can efficiently use LO to set $\alpha_k$ in order to decrease $f$ by a larger amount than LS methods.

For LCPs the gradient has the form  $\nabla f(w) = X^T\nabla g(Xw)$ and thus the gradient descent iteration can be written
\[
w_{k+1} = w_k - \alpha_kX^T\nabla g(Xw_k).
\]
If we use a fixed step size, then for large $n$ and $d$ the \emph{cost of this update is dominated by the 2 matrix multiplications with $X$ required for each iteration}. But instead of using a a fixed step size, consider the line optimization (LO) problem of finding the $\alpha_k$ that maximally decreases $f$,
\begin{align}
	& \argmin_{\alpha}f(w_k - \alpha X^T\nabla g(Xw_k)))\label{GD:LO}\\
	\equiv & \argmin_{\alpha}g(X(w_k - \alpha X^T\nabla g(Xw_k) ))\nonumber\\
	\equiv & \argmin_{\alpha}g(\underbrace{Xw_k}_{m_k} - \alpha \underbrace{X(X^T\nabla g(m_k))}_{d_k})\nonumber\\
	\equiv & \argmin_{\alpha}g(m_k - \alpha d_k)\nonumber,
\end{align}
where we have defined $m_k = Xw_k$ and $d_k = X(X^T\nabla g(m_k))$. The key to efficiently optimizing over $\alpha$ is tracking the ``memory'' $m_k \in \R^n$ across iterations. Notice that computing $d_k$ requires 2 matrix multiplications, that we have $m_{k+1} = m_k - \alpha_kd_k$, and that given $m_k$ and $d_k$ we can evaluate any value of $\alpha$ in $O(n)$ time plus the inexpensive cost of evaluating $g$. This allows us  to numerically \emph{solve the LO problem with a total of only 2 matrix multiplications per iteration}. Thus, in terms of the bottleneck operation the cost of gradient descent with LO for LCPs is the same as the cost of using a fixed step size. In the case of logistic regression, the iteration cost with a fixed step size is $O(nd)$ while the iteration cost if we use bisection to solve the LO to an accuracy of $\epsilon$ over a bounded domain is only $O(nd + n\log(1/\epsilon))$. Thus, for large $d$ we can perform LO to a high accuracy ``for free''.

\begin{figure}
\includegraphics[width=\textwidth]{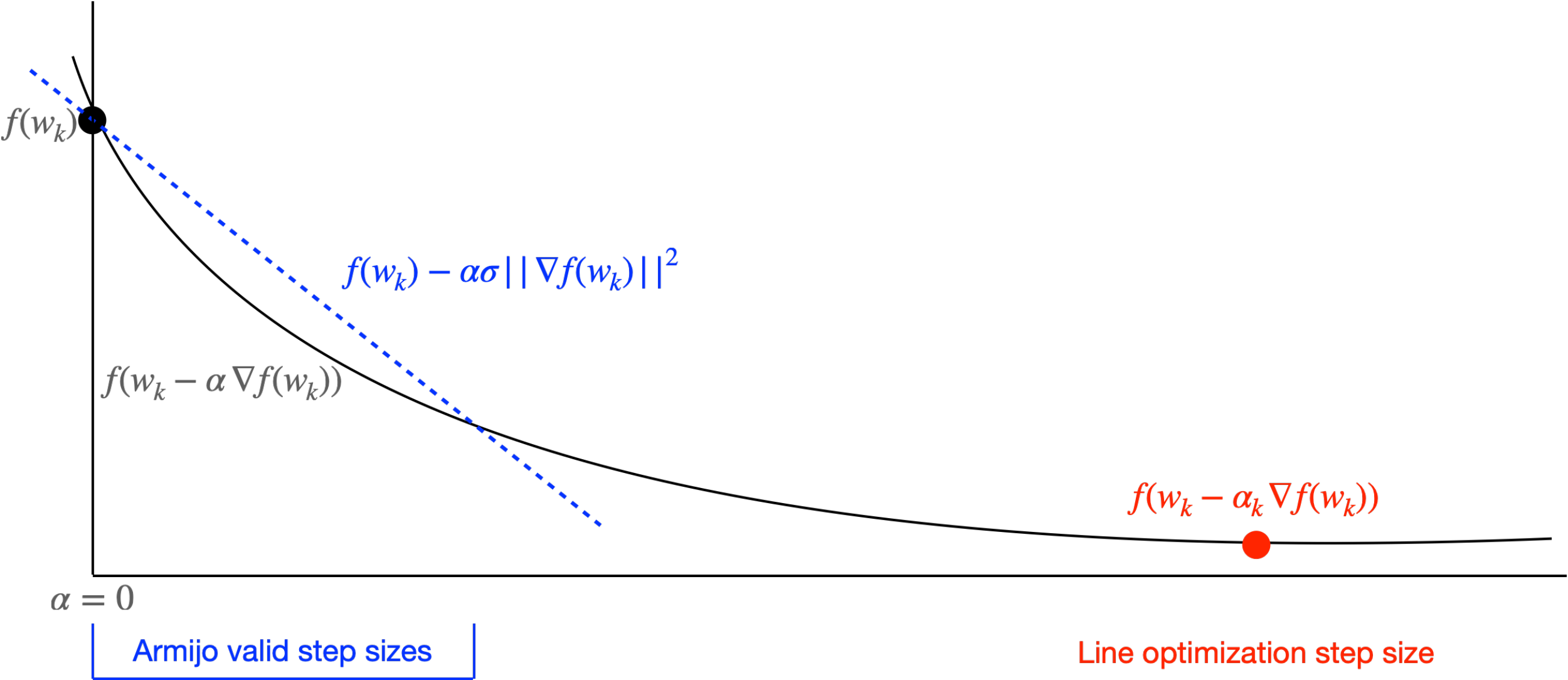}
\caption{Valid step sizes allowed by Armijo condition (in blue) and update minimizing function (in red) by line optimization. For a fixed $\sigma$, the improvement of line optimization over the step sizes allowed by the Armijo condition can be made arbitrarily large even for convex functions.}
\label{fig:LS_LO}
\end{figure}

Note that the the value of $f$ achieved by LO may be substantially smaller than the value achieved by using the strong Wolfe conditions. Indeed, for non-quadratic functions the Armijo condition~\eqref{eq:armijo} used as part of the strong Wolfe conditions may exclude the optimal step size (see Figure~\ref{fig:LS_LO}). Indeed, even for convex problems the decrease in $f$ achieved by LO can be arbitrarily larger than the largest decrease for any step size satisfying the Armijo condition. Unfortunately, if $g$ is non-convex then the LO may be difficult to solve. However, applying a local optimizer to the LO problem is still likely to achieve a greater decrease in the function than line searches.

\subsection{Efficient Plane Search (PS) for the Memory Gradient Method}

The GD+M update~\eqref{eq:GDM} for LCPs has the form
\[
w_{k+1} = w_k - \alpha_kX^T\nabla g(Xw_k) + \beta_k(w_k - w_{k-1}).
\]
With a fixed learning rate $\alpha_k$ and momentum rate $\beta_k$, notice that the dominant cost of this update is again the 2 matrix multiplications with $X$. Consider instead using a SO to set $\alpha_k$ and $\beta_k$ to maximally decreases the function $f$,
\begin{align}
	& \argmin_{\alpha,\beta}f(w_k - \alpha X^T\nabla g(Xw_k) + \beta(w_k - w_{k-1}))\label{GD:SO}\\
	\equiv & \argmin_{\alpha,\beta}g(X(w_k - \alpha X^T\nabla g(Xw_k) + \beta(w_k - w_{k-1})))\nonumber\\
	\equiv & \argmin_{\alpha,\beta}g(\underbrace{Xw_k}_{m_k} - \alpha \underbrace{X(X^T\nabla g(m_k))}_{d_k} + \beta (m_k - m_{k-1}))\nonumber\\
	\equiv & \argmin_{\alpha,\beta}g(\underbrace{(1+\beta_k)m_k - \alpha d_k + \beta m_{k-1}}_\text{potential $m_{k+1}$}).\nonumber
\end{align}
Using the GD+M update~\eqref{eq:GDM} while optimizing $\alpha_k$ and $\beta_k$ is called the \emph{memory gradient} method~\citep{Miele1969}. For LCPs adding the momentum term requires us to store $m_{k-1}$, but still only requires 2 matrix multiplications total per iteration (the same as using a fixed learning and momentum rate). If we solve the two-dimensional SO to accuracy $\epsilon$ using a cutting plane method the cost for logistic regressions iterations with optimized $\alpha_k$ and $\beta_k$ values is $O(nd + n\log(1/\epsilon))$. Thus, using PS to optimize the learning and momentum rate does not increase the asymptotic cost compared to using LO to optimize the learning rate. However, note that optimizing both rates ``for free'' is somewhat specific to the use of momentum as the second direction; if we use directions that are not in the span of the previous parameters and gradients this would require additional matrix multiplications.

 Note that the memory gradient method does not require $\alpha_k$ or $\beta_k$ to be positive. Thus, the method could set $\beta_k$ close to 0 to ``reset'' the momentum or could even use negative momentum. A common variation on LCPs is functions of the form $f(w) = g(Xw) + g_0(w)$ for a regularization function $g_0$. The memory gradient remains efficient in this setting in the typical case where evaluating $g_0$ is much less expensive than matrix multiplications with $X$. A typical choice is L2-regularization, $g_0(w) = (\lambda/2)\norm{w}^2$ for a positive regularization parameter, which only costs $O(d)$ to evaluate.
 
\subsection{LO and SO for LCPs in Practice}
\label{sec:LCPexp}


LCPs are widely used in ML, but is not common to fit these models with LO or SO. We believe that this is because it is not common knowledge that this leads to impressive performance gains compared to standard approaches to set the step size(s). To help give the reader context, in this section we present empirical results comparing a variety of ways to set the step size(s) for linear and logistic regression. In particular, our first set of experiments compares the following methods:
\begin{itemize}
\item \texttt{GD(1/L)}: gradient descent~\eqref{eq:GD} with a step size of $\alpha_k = 1/L_k$, where $L_k$ is an estimate of the maximum curvature of the function. We initialize to $L_0 = 1$ and each subsequent iteration is initialized with the previous estimate $L_k = L_{k-1}$. But on each iteration the value $L_k$ is doubled until we satisfy the inequality $f(w_k - (1/L_k)\nabla f(w_k)) \leq f(x_k) - (1/2L_k)\norm{\nabla f(w_k)}^2$. This backtracking approach corresponds to using the Armijo condition~\eqref{eq:armijo} with $\sigma = 1/2$. This approach guarantees a similar theoretical convergence rate to using a fixed step size of $\alpha_k = 1/L$, where $L$ is the Lipschitz constant of the gradient~\citep[see][]{beck2009fast}. In particular, this approach is within a constant factor of the worst-case optimal fixed step size for gradient descent for minimizing convex functions.
\item \texttt{GD(LS)}: gradient descent~\eqref{eq:GD} using a line search to find a learning rate satisfying the strong Wolfe conditions. We use a standard implementation of the line search~\citep[see][Algorithm~3.5]{Nocedal2006}. The Armijo sufficient decrease parameter $\sigma$ in~\eqref{eq:armijo} is set to 0.0001 which allows larger step sizes than the \texttt{GD(1/L)} method, while the curvature parameter is set to 0.9 requiring that the step sizes are large enough to slightly decrease the magnitude of the directional derivative along the line. We initialize the line search with the previous step size $\alpha_{k-1}$, double the step size during the initial ``bracketing phase'', and then take the midpoint of the current bracket during the ``zoom'' phase.
\item \texttt{GD+M(LS)}: a non-linear conjugate gradient (CG) variation on the gradient descent with momentum~\eqref{eq:GDM} update,
\[
w_{k+1} = w_k +\alpha_kd_k,
\]
where 
\[
d_k = -\nabla f(w_k) + \eta_k(w_k - w_{k-1}).
\]
We set $\eta_k$ using the non-negative variant of the formula of~\citet{polak1969note},
\[
\eta_k = \max\left\{0,\frac{\nabla f(w_k)^T(\nabla f(w_k) - \nabla f(w_{k-1})}{\norm{\nabla f(w_k)}^2}\right\}.
\]
We set $\alpha_k$ using a line search to satisfy the strong Wolfe conditions. We reset the momentum rate to $\eta_k=0$ on iterations where the directional derivative $\nabla f(w_k)^Td_k$ of the search direction $d_k$ is not negative.
\item \texttt{GD(LO)}: gradient descent~\eqref{eq:GD} using LO~\eqref{GD:LO} to set the learning rate on each iteration. We discuss how we numerically solve the LO in Appendix~\ref{app:SOsolve}.
\item \texttt{GD+M(LO)}: the non-linear CG method \texttt{GD+M(LS)} but using LO to set the step size~$\alpha_k$. Note that with exact computation this is equivalent to the linear conjugate gradient method for the special case of strongly-convex quadratic functions.
\item \texttt{GD+M(SO)}: gradient descent with momentum~\eqref{eq:GDM} where the learning rate and momentum rate are optimized on each iteration using SO~\eqref{GD:SO}. We again refer to Appendix~\ref{app:SOsolve} for how we numerically solve the SO problems. Note that with exact computation this method is also equivalent to the linear conjugate gradient method for strongly-convex quadratic functions.
\end{itemize}

\begin{figure}
\includegraphics[width=0.24\textwidth]{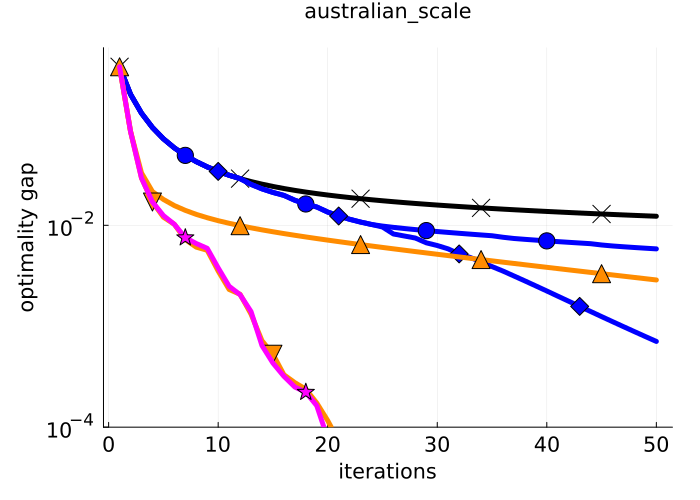}
\includegraphics[width=0.24\textwidth]{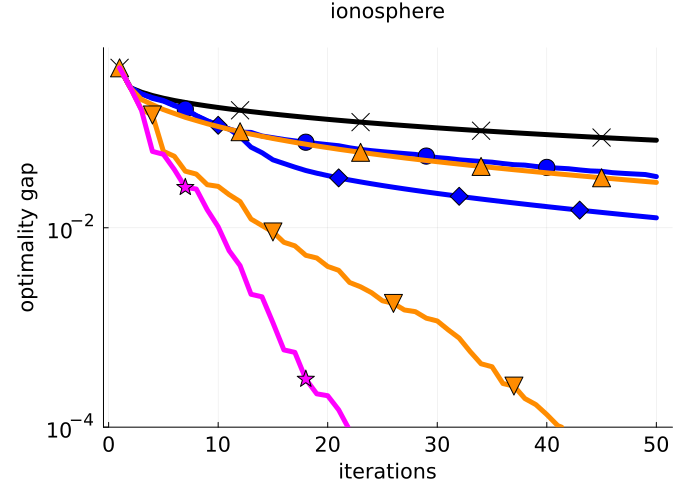}
\includegraphics[width=0.24\textwidth]{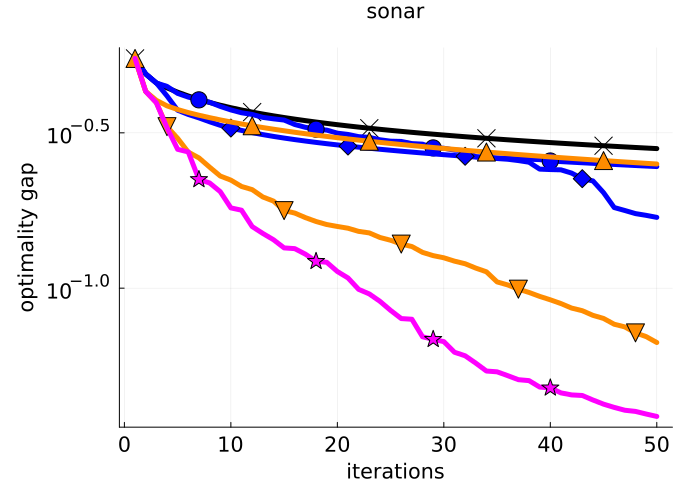}
\includegraphics[width=0.24\textwidth]{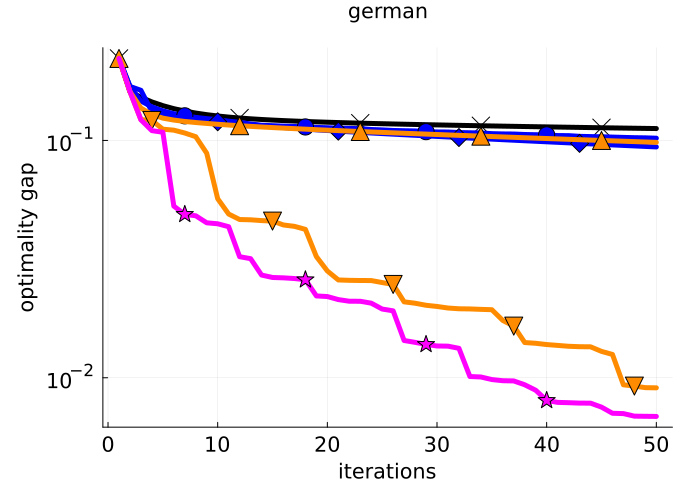}
\includegraphics[width=0.24\textwidth]{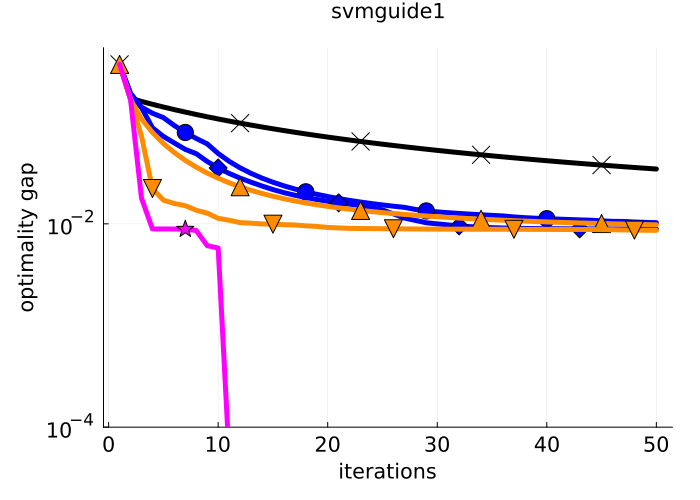}
\includegraphics[width=0.24\textwidth]{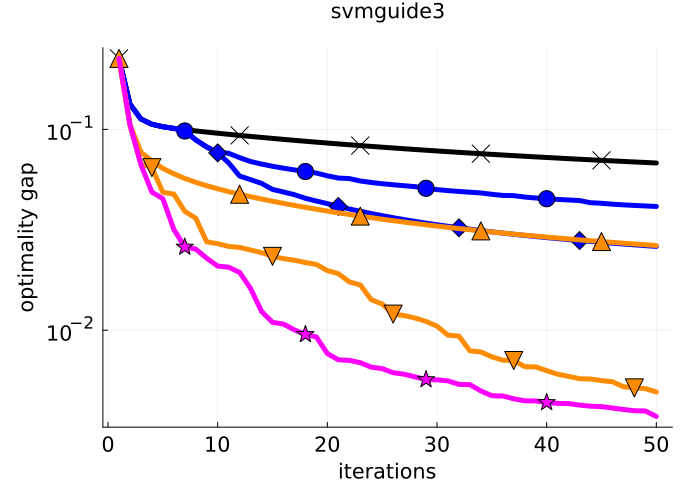}
\includegraphics[width=0.24\textwidth]{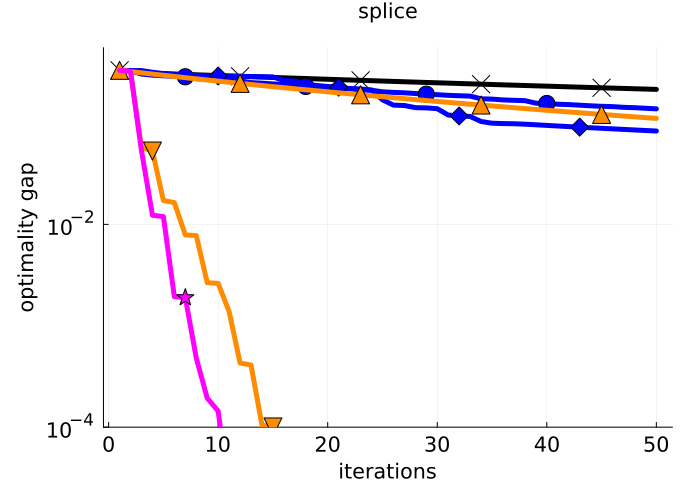}
\includegraphics[width=0.24\textwidth]{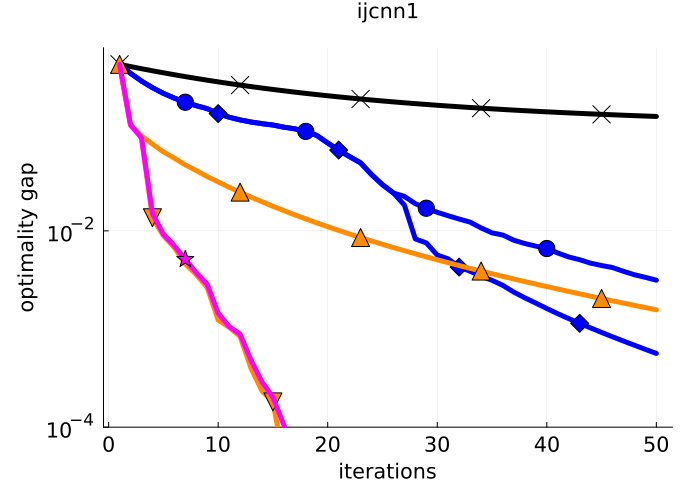}
\includegraphics[width=0.24\textwidth]{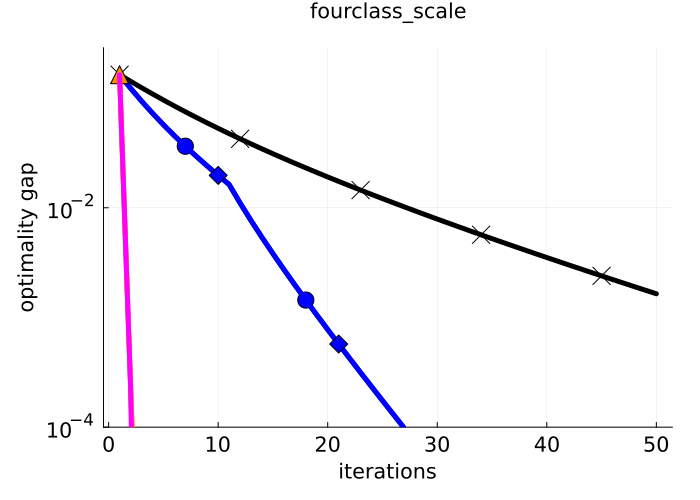}
\includegraphics[width=0.24\textwidth]{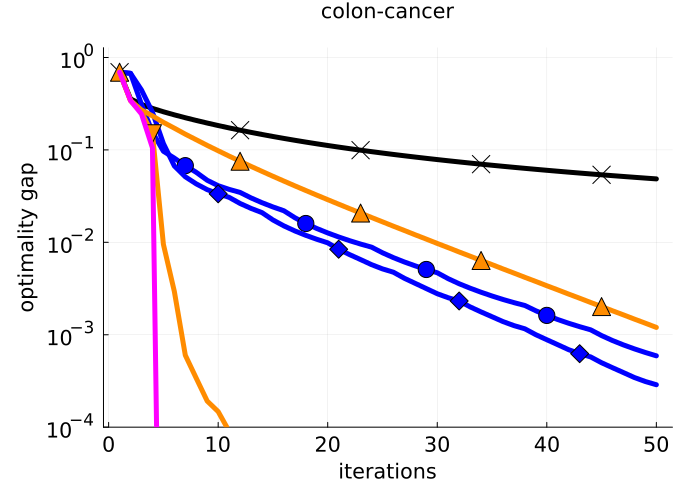}
\includegraphics[width=0.24\textwidth]{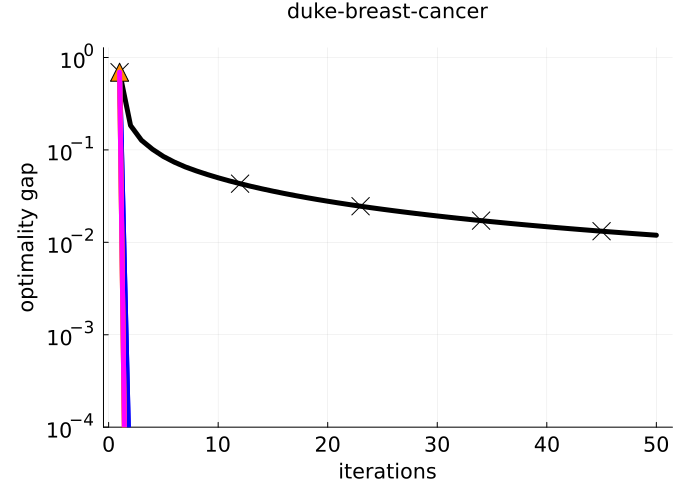}
\includegraphics[width=0.24\textwidth]{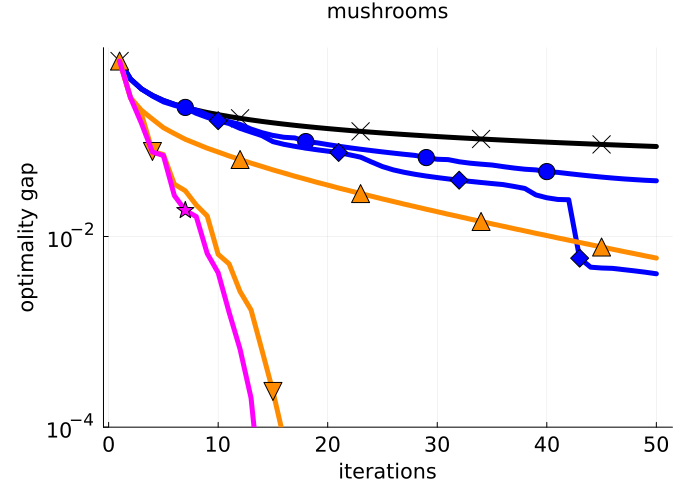}
\includegraphics[width=0.24\textwidth]{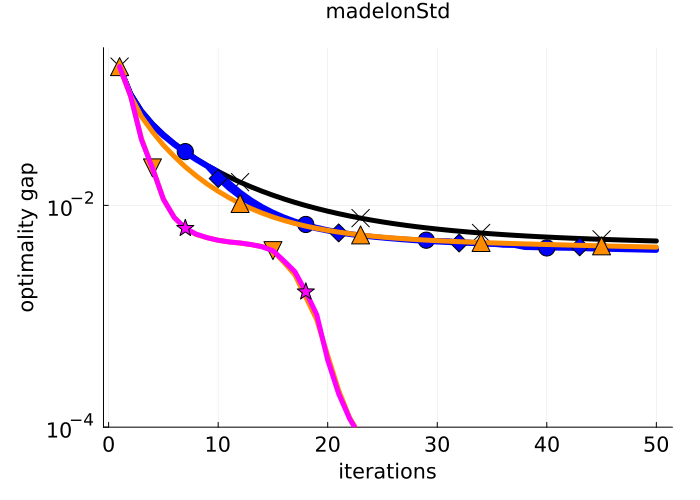}
\includegraphics[width=0.24\textwidth]{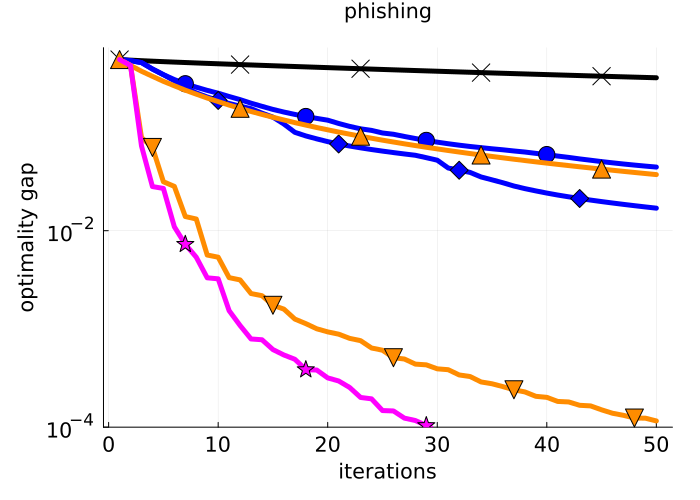}
\includegraphics[width=0.24\textwidth]{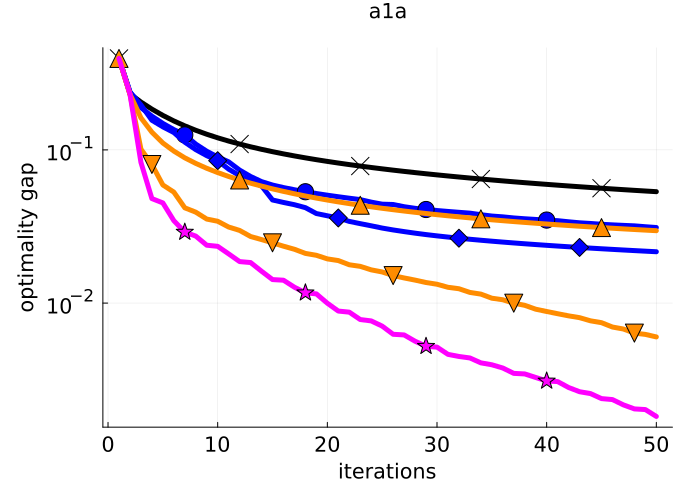}
\includegraphics[width=0.24\textwidth]{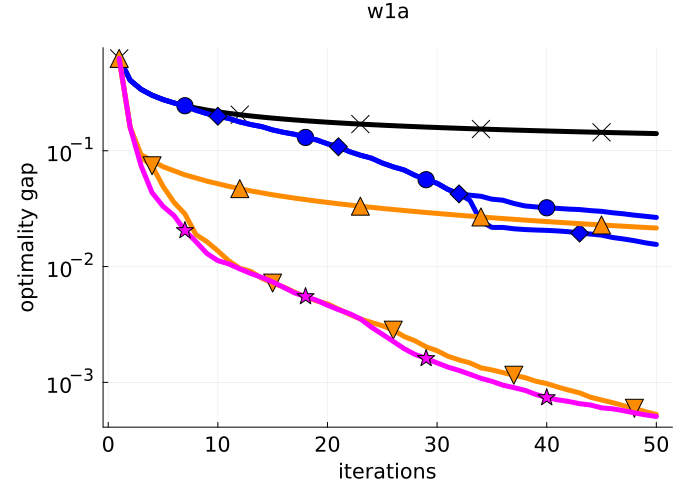}
\includegraphics[width=\textwidth]{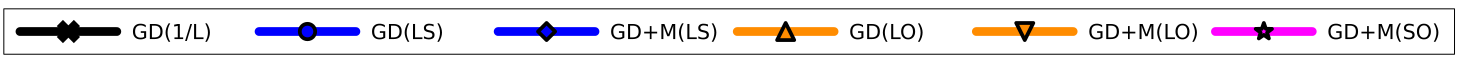}
\caption{Performance of different gradient-based methods for fitting logistic regression models. Each plot is a different dataset. The black line only backtracks, the blue lines use a line search that can decrease or increase the step size to satisfy the strong Wolfe conditions, the orange lines use LO, and the magenta line uses SO.  The \texttt{GD} methods use the gradient direction, the \texttt{GD+M(L*)} methods use the gradient direction and momentum with the non-linear conjugate gradient relationship between the parameters, and the \texttt{GD+M(SO)} method optimizes the learning rate and momentum rate. We see that LS methods tend to dominate the 1/L method, GD+M methods tend to dominate GD methods, LO methods tend to dominate LS methods, and the best performance on every dataset was achieved with SO.}
\label{fig:firstlogReg}
\end{figure}

We explored the performance of these methods for logistic regression on 16 datasets drawn from the  \texttt{libsvm} ~\citep{Chang2011} collection and the UCI ML Repository ~\citep{Dua2019}. These datasets were selected in~\citet{shea2023greedy} to represent a variety of optimization challenges. We did not add a bias term or standardize the datasets, which makes these datasets particularly-challenging for optimizers to fit (an exception is the ``madelon'' dataset, where we standardized the features as they were poorly scaled and the optimizers were not able to make any reasonable progress). Figure~\ref{fig:firstlogReg} shows the sub-optimality against the number of iterations of the different methods on these datasets.\footnote{We approximated the optimal solution by running an variation on the non-monotonic Barzilai-Borwein method of~\citet{Raydan1997} for 5,000 iterations.}
While in our experiments we plot the performance in terms of the number of iterations, we remind the reader that for LCPs the {\bf iteration costs of all of these methods are dominated by the cost of the 2 matrix multiplications required per iteration}. Thus, as the size of the data grows the runtimes of the iterations of all methods we compare will only differ by a small constant factor. This makes the comparison fair since we are most interested in performance on larger datasets.

We observed several trends across the datasets in this experiment:
\begin{enumerate}
\item \texttt{GD(1/L)} was always the worst method, even though theoretically it is close to optimal. The superiority of practical step sizes highlights the limitations of selecting algorithms purely based on theoretical convergence rates.
\item LO methods tended to outperform LS methods, indicating that a more precise line search improves performance. The \texttt{GD(LO)} method outperformed \texttt{GD(LS)} on most datasets while the \texttt{GD+M(LO)} method outperformed \texttt{GD+M(LS)} on all datasets.
\item Adding momentum tended to help, despite momentum not improving the theoretical rate for this class of problems, and momentum helped more when using LO. \texttt{GD+M(LS)} outerperformed \texttt{GD(LS)} on most problems, while after the first iteration \texttt{GD+M(LO)} dominated \texttt{GD(LO)} across all iterations across all problems.
\item The memory gradient method \texttt{GD+M(SO)}, that optimizes both the learning rate and momentum rate, performed as well or better than all other methods across every iteration on every dataset.
\end{enumerate}

\begin{figure}
\includegraphics[width=0.24\textwidth]{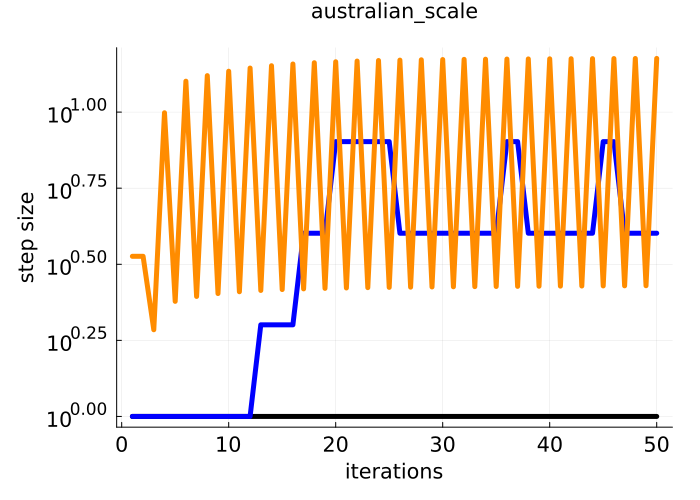}
\includegraphics[width=0.24\textwidth]{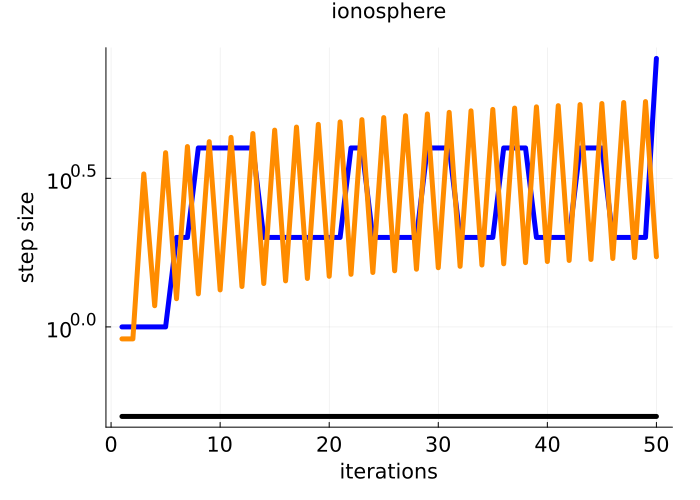}
\includegraphics[width=0.24\textwidth]{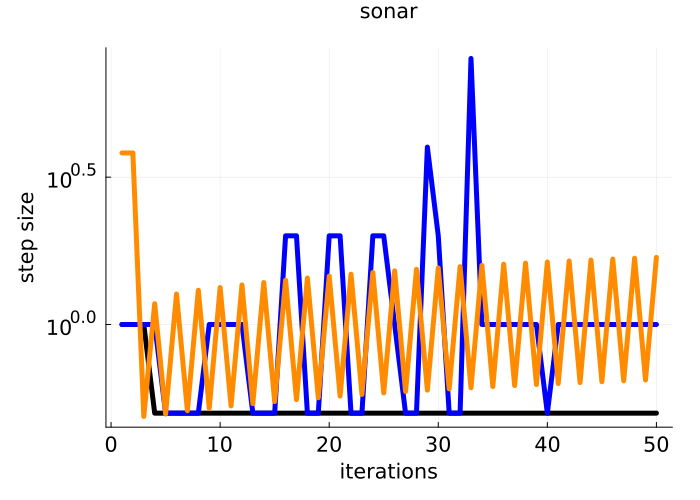}
\includegraphics[width=0.24\textwidth]{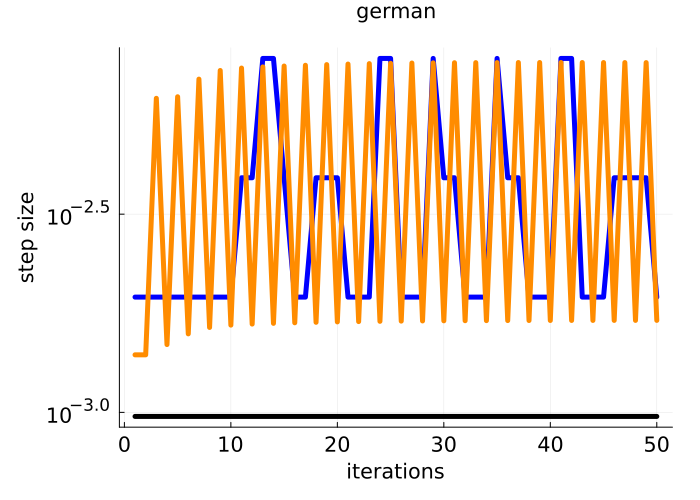}
\includegraphics[width=0.24\textwidth]{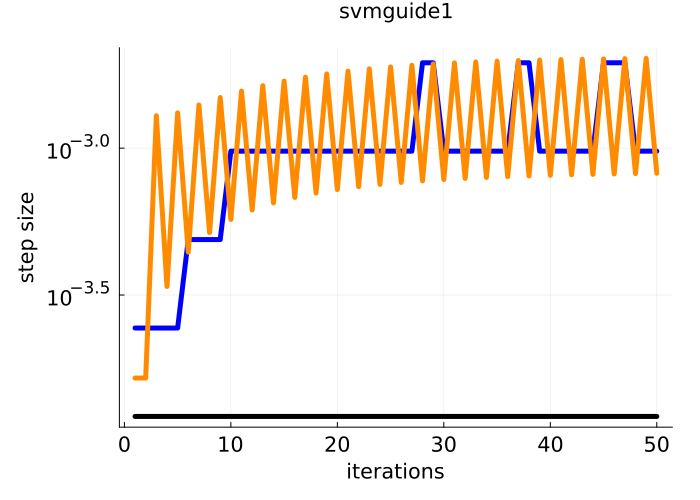}
\includegraphics[width=0.24\textwidth]{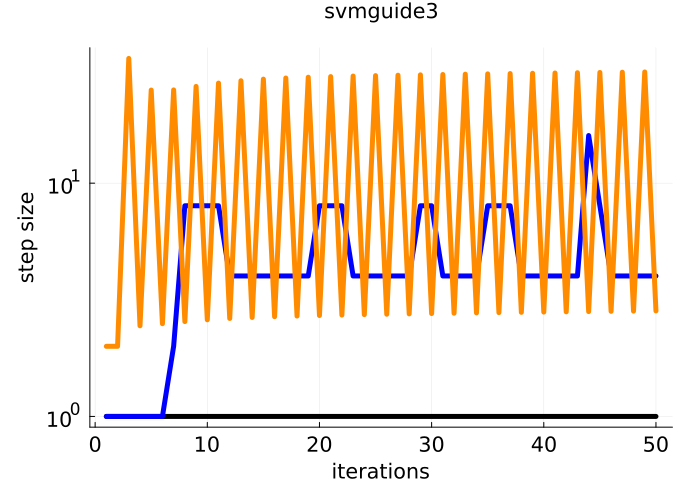}
\includegraphics[width=0.24\textwidth]{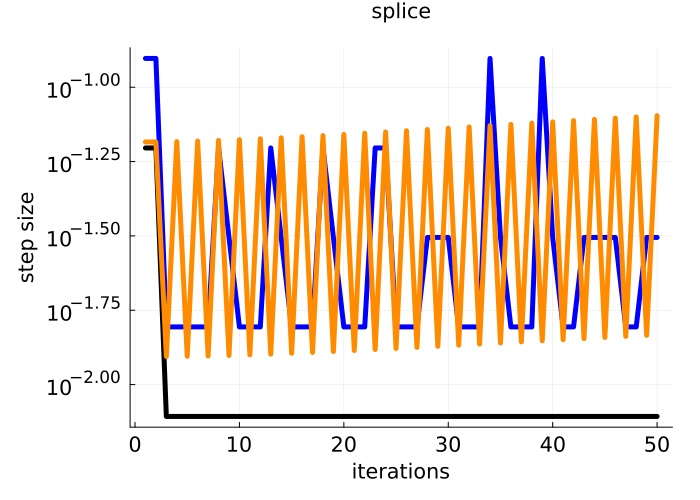}
\includegraphics[width=0.24\textwidth]{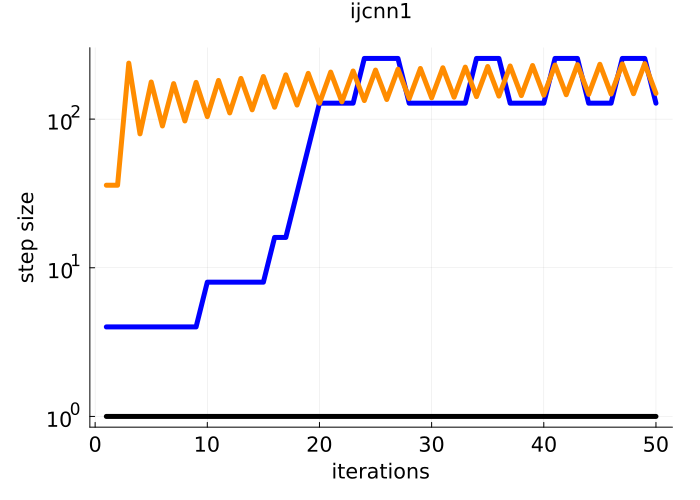}
\includegraphics[width=0.24\textwidth]{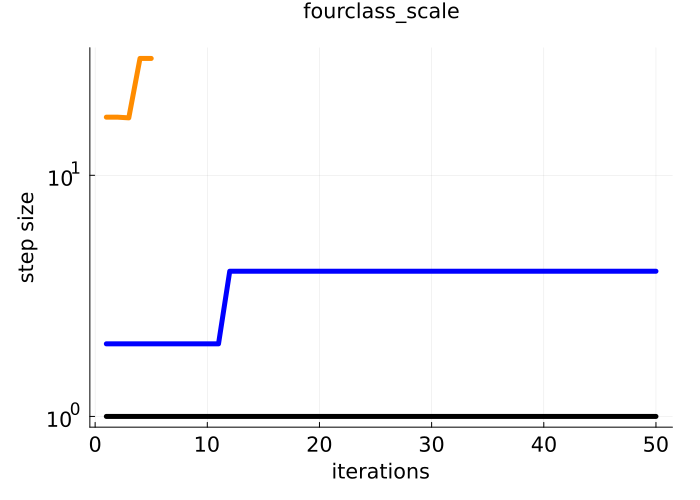}
\includegraphics[width=0.24\textwidth]{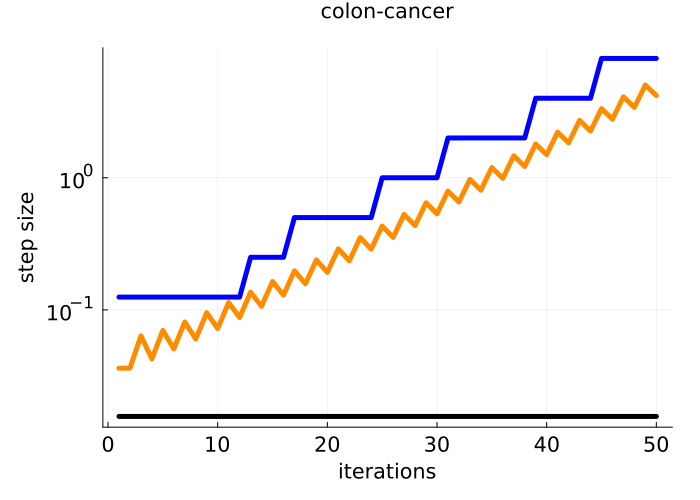}
\includegraphics[width=0.24\textwidth]{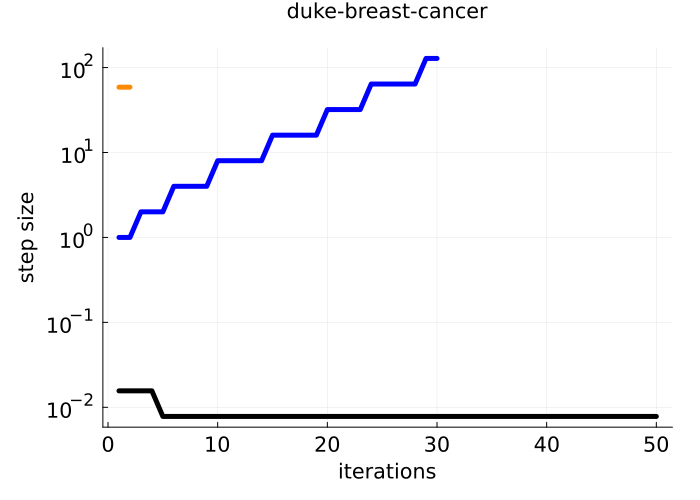}
\includegraphics[width=0.24\textwidth]{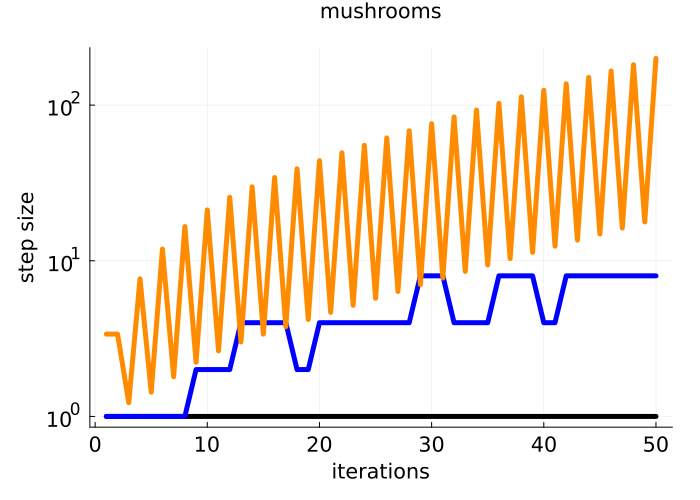}
\includegraphics[width=0.24\textwidth]{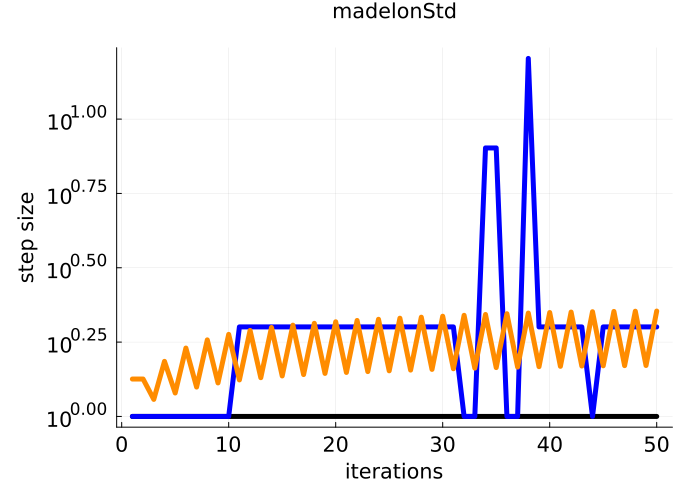}
\includegraphics[width=0.24\textwidth]{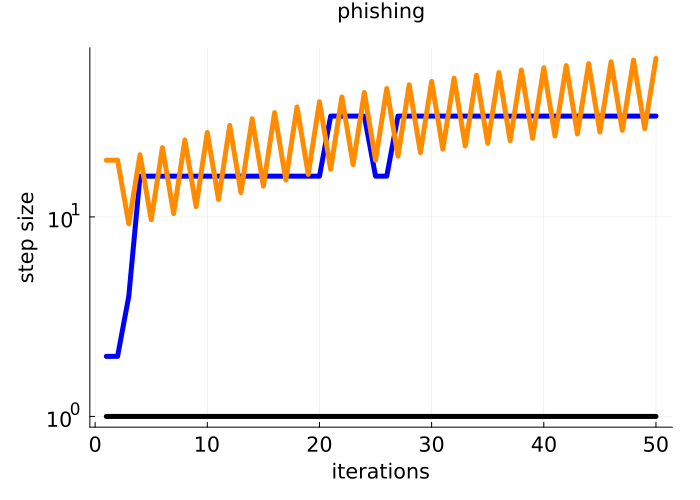}
\includegraphics[width=0.24\textwidth]{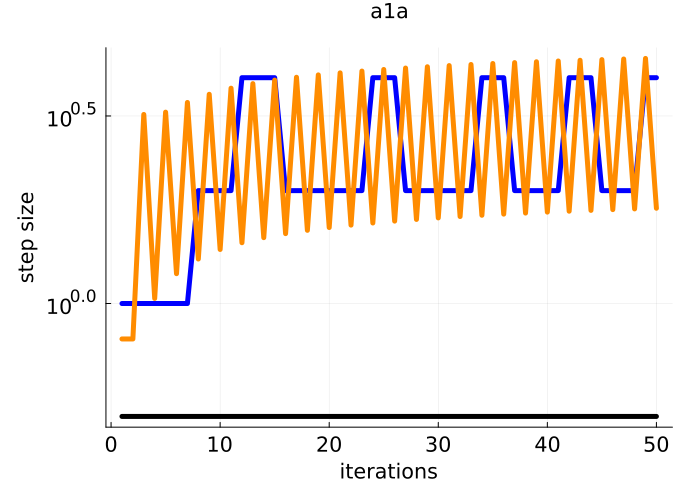}
\includegraphics[width=0.24\textwidth]{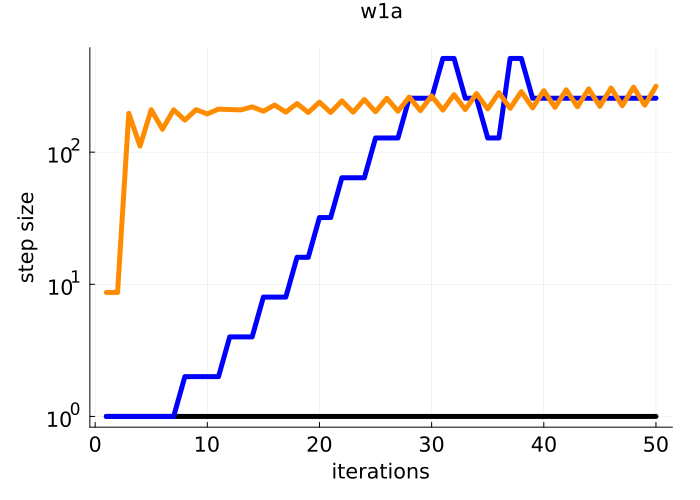}
\begin{center}
\includegraphics[width=.5\textwidth]{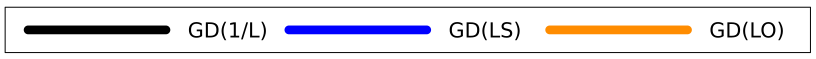}
\end{center}
\caption{Step sizes of different gradient descent methods for fitting logistic regression models. Note that the \texttt{GD(LO)} step sizes tended to lead to the fastest convergence while the \texttt{GD(1/L)} step sizes always converged slowest.}
\label{fig:LOsteplogReg}
\end{figure}

To investigate why LO and SO improve performance, we plotted the step sizes used by the different methods. In Figure~\ref{fig:LOsteplogReg} we show the step sizes used by the different methods that only use the gradient direction. In these plots we see that the theoretically-motivated 1/L method tended to use much smaller step sizes than the other methods, either always accepting the initial step size or decreasing the step size within the first few iterations and then using a small constant step size. In contrast, the LO step sizes tended to oscillate (with a period of 2) while often slowly increasing. Only two exceptions to this pattern were observed, and on these problems the LO method quickly solved the problem using large step sizes. The LS method tended to start with small step sizes before growing into a similar range to the LO method's oscillations, and tended to also have oscillations but with a longer period.

\begin{figure}
\includegraphics[width=0.24\textwidth]{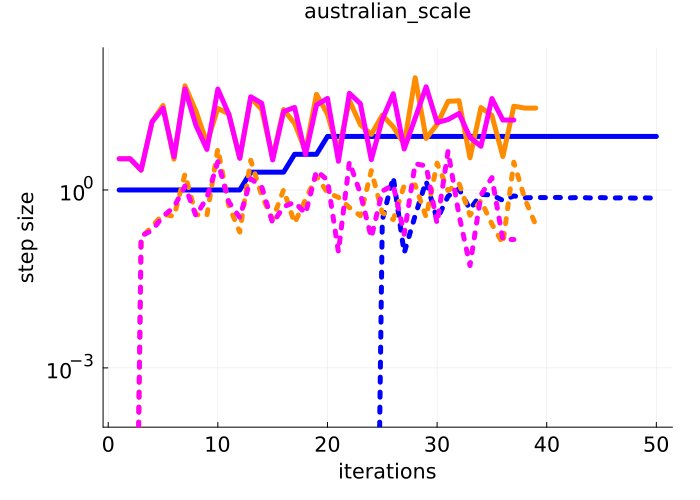}
\includegraphics[width=0.24\textwidth]{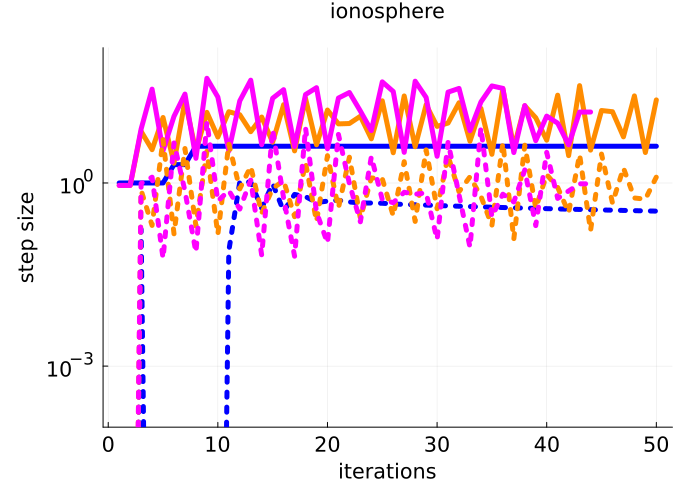}
\includegraphics[width=0.24\textwidth]{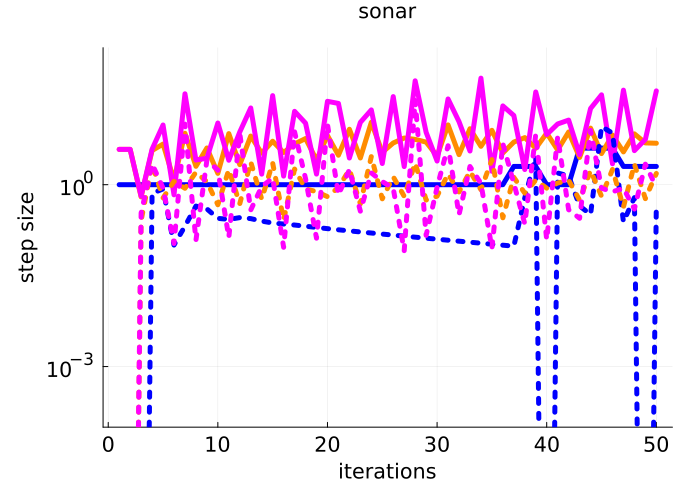}
\includegraphics[width=0.24\textwidth]{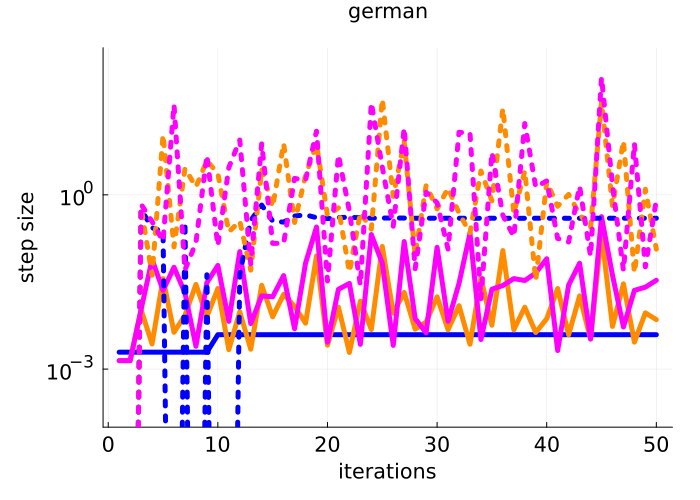}
\includegraphics[width=0.24\textwidth]{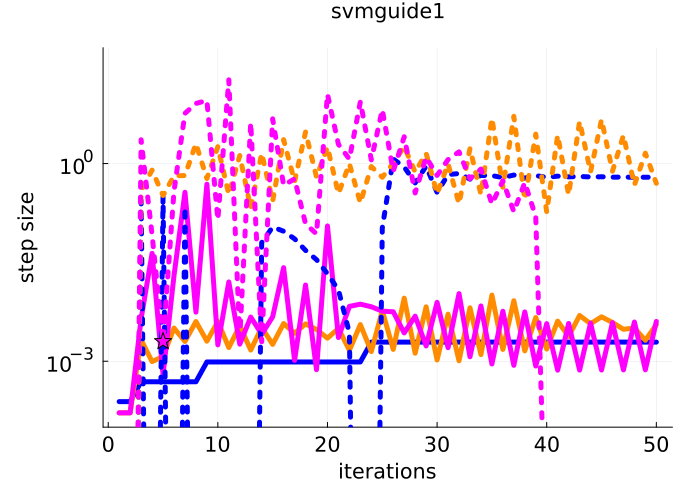}
\includegraphics[width=0.24\textwidth]{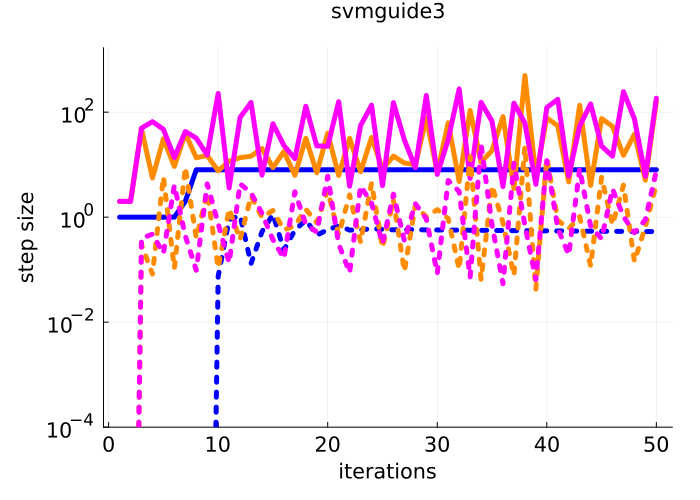}
\includegraphics[width=0.24\textwidth]{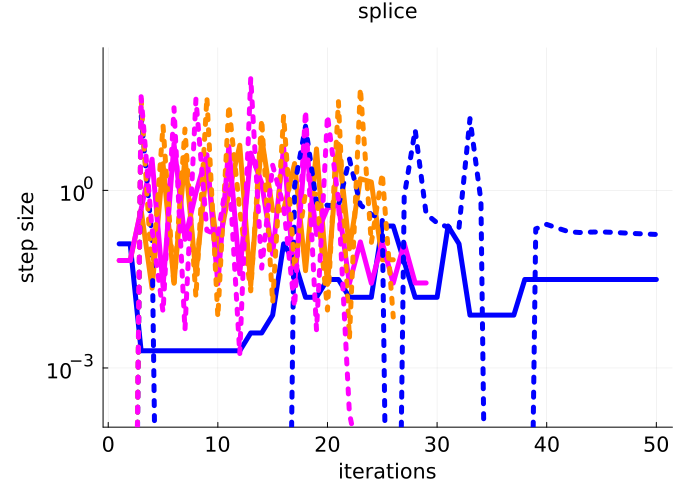}
\includegraphics[width=0.24\textwidth]{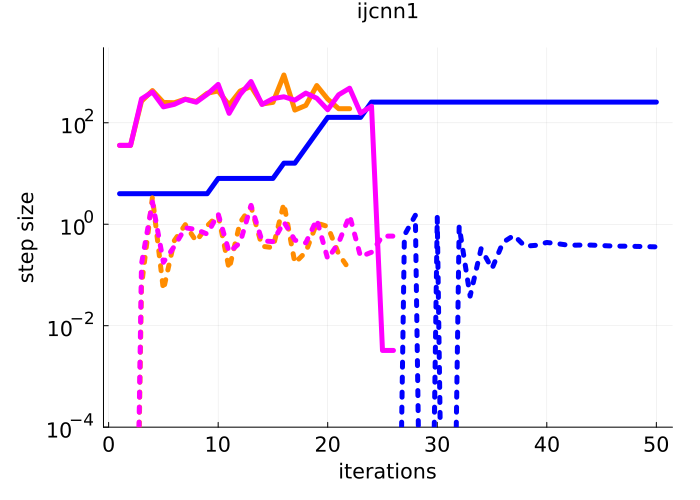}
\includegraphics[width=0.24\textwidth]{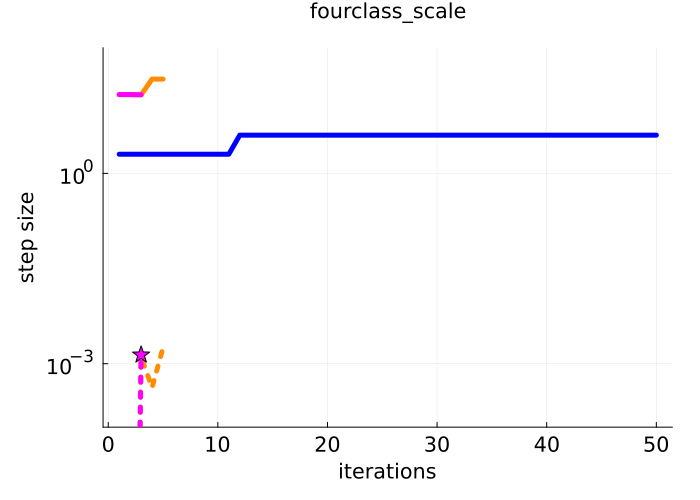}
\includegraphics[width=0.24\textwidth]{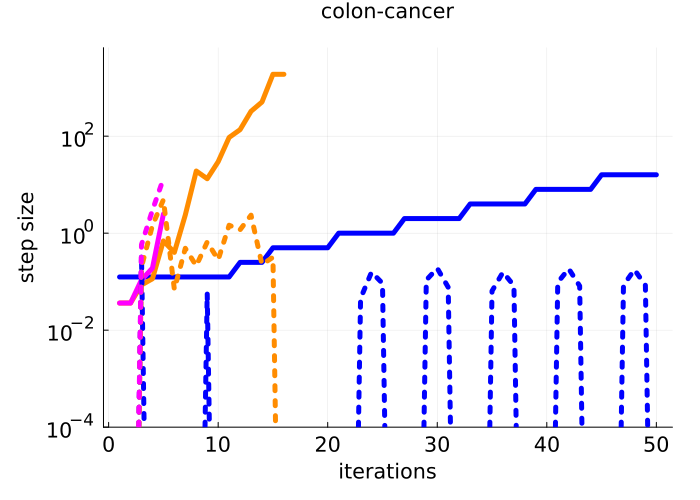}
\includegraphics[width=0.24\textwidth]{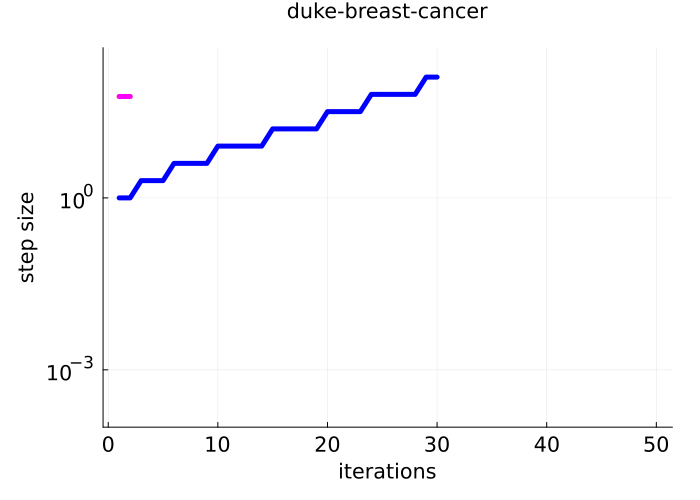}
\includegraphics[width=0.24\textwidth]{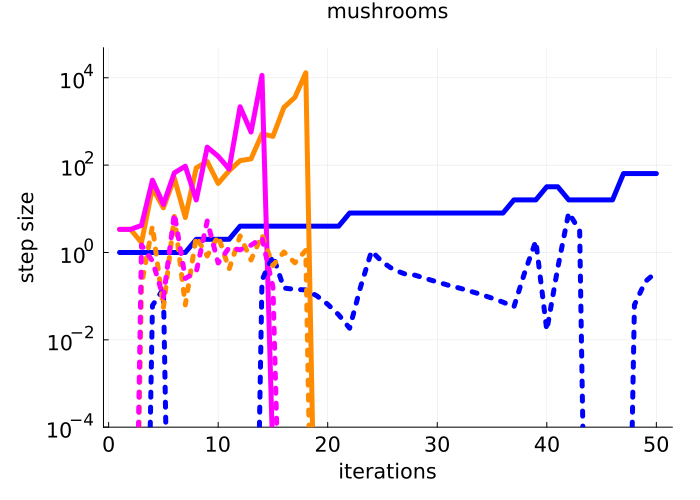}
\includegraphics[width=0.24\textwidth]{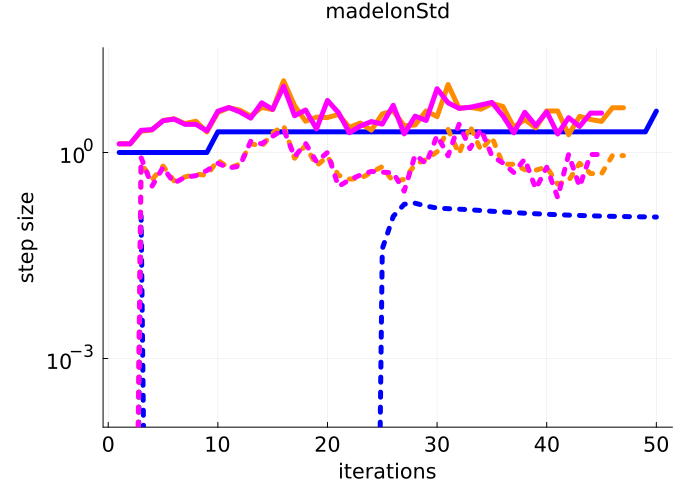}
\includegraphics[width=0.24\textwidth]{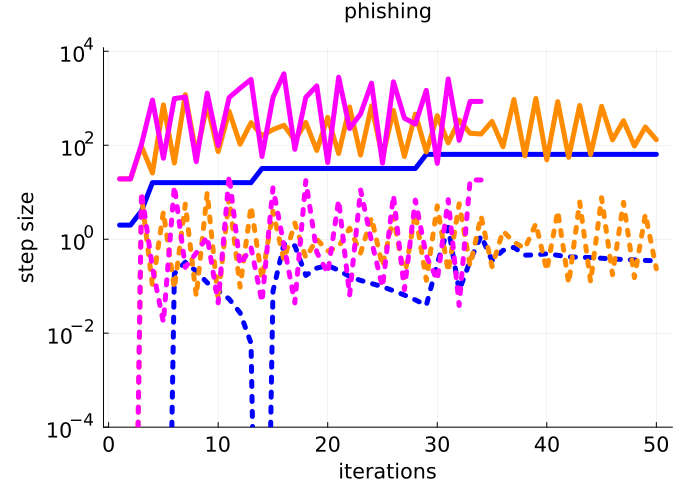}
\includegraphics[width=0.24\textwidth]{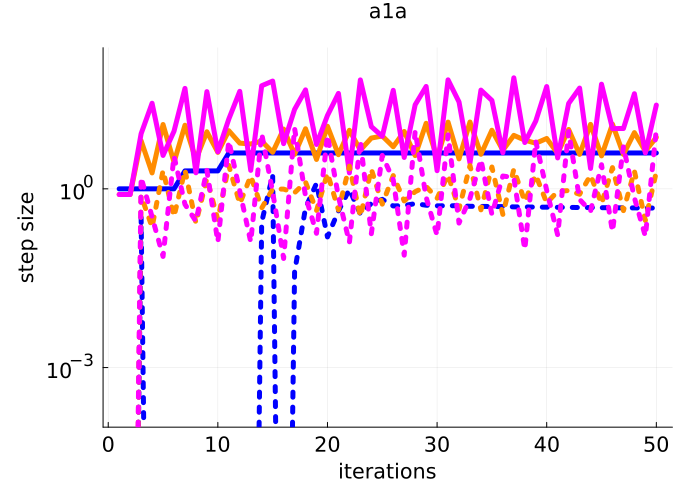}
\includegraphics[width=0.24\textwidth]{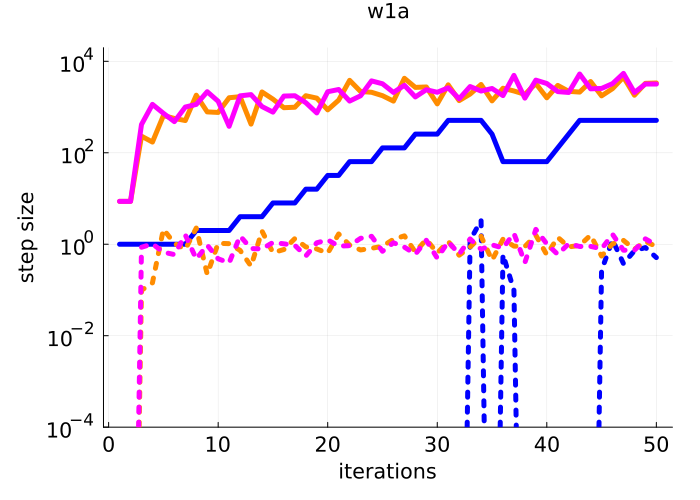}
\begin{center}
\includegraphics[width=.5\textwidth]{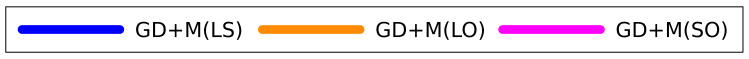}
\end{center}
\caption{Step sizes of different gradient plus momentum methods for fitting logistic regression models. The solid lines are the learning rates and the dashed lines are the momentum rates. Note that the \texttt{GD+M(SO)} step sizes tended to lead to the fastest convergence while the \texttt{GD+M(LS)} step sizes usually converged slowest. A star is used to indicate the iteration where the momentum rate was negative.}
\label{fig:SOsteplogReg}
\end{figure}

In Figure~\ref{fig:SOsteplogReg}, we plot the learning rate and momentum rate for the methods that incorporate momentum. Note that the optimal learning and momentum rates found by the \texttt{GD+M(SO)} method tended not be constant, often differing by orders of magnitude between iterations. The optimal learning rate and momentum rate do seem to show oscillations, but the oscillations seem less regular than the \texttt{GD(LO)} method where momentum is not used. We see that the non-linear conjugate gradient method with a precise line optimization \texttt{GD+M(LO)} often closely tracks the learning rate and momentum rates found by the \texttt{GD+M(SO)} method, while the imprecise line search \texttt{GD+M(LS)} finds very different values (and indeed often resets the momentum to zero on early iterations). We finally note that on one iteration, the optimal momentum rate used by the \texttt{GD+M(SO)} method was negative (absolute value of the rate is shown).

\section{The Scattered 50+ Year History of Subspace Optimization Methods}
\label{sec:related}

The use of SO in both theoretical and practical works has a long history. Indeed, the idea of SO arises in a variety of contexts in the numerical optimization literature. In this section we review a variety of the most closely related works. 

\textbf{Memory Gradient Method}: the memory gradient method \texttt{GD+M(SO)} where SO is used to set the learning rate and momentum rate in the GD+M update~\eqref{eq:GDM} was first explored by~\citet{Miele1969}. They proposed a particular method to approximately solve the SO problem and presented numerical results showing that SO could improve performance on a simple test function. However, Miele and Cantrell's work had relatively little impact. This is likely because numerical optimization work in the 1960s did not have the modern focus on exploiting problem structure (like LCPs) to quickly solve the SO.

\textbf{Conjugate Gradient Methods}: the conjugate gradient (CG) method was proposed in 1952 for solving positive-definite linear systems~\citep{Hestenes1952}, but it can equivalently be written as using the memory-gradient method to minimize strongly-convex quadratic functions~\citep{Cantrell1969}. Due to its outstanding empirical and theoretical properties, various non-linear CG methods have been proposed for general smooth optimization. The first non-linear CG method is due to~\citet{Fletcher1964}, but numerous variants have appeared over the years~\citep[see][]{Hager2006}. Non-linear CG methods typically use a LS along a direction that is a weighted combination of the gradient and momentum directions, and have been explored in some neural network settings~\citep{lecun2002efficient}. However, by fixing the weighting between the learning rate and momentum rate, CG methods may make less progress per iteration than the memory gradient method. Current non-asymptotic analyses of non-linear CG methods show that their worst case performance is not better than gradient descent with a line search~\citep{gupta2023nonlinear}. Further, it was already identified by Fletcher and Reeves that on some problems we need to ``reset'' the momentum rate to 0 for non-linear CG to obtain good practical performance. It is known that using LO and resetting non-linear CG every $n$ iterations leads to asymptotic quadratic convergence (as in Newton's method) in terms of $n$-iteration cycles~\citep{cohen1972rate}. The fact that non-linear CG's performance is improved by this resetting mechanism suggests that we might improve performance by using the memory gradient method's approach of optimizing the momentum rate on each iteration (where some iterations could use small or zero or even negative momentum rates).

\textbf{Accelerated Gradient Methods}: SO was important in the development of first-order methods achieving the accelerated $O(1/k^2)$ error rate for minimizing convex functions~\citep{Nemirovski1982,Nemirovski1983}. Nemirovski showed the $O(1/k^2)$ rate could be achieved using SO over a subspace containing the current gradient and specific weighted combinations of previous gradients. Nemirovski later showed that we could achieve this rate with a LS instead of SO, while~\citet{Nesterov1983} eventually showed the rate could be achieved with a fixed learning rate and a modified momentum. Nesterov's accelerated gradient (NAG) algorithm can also be used to achieve accelerated convergence rates for strongly-convex~\citep{Nesterov2013} and non-convex~\citep{Carmon2017} objectives. 

\begin{figure}
\includegraphics[width=0.24\textwidth]{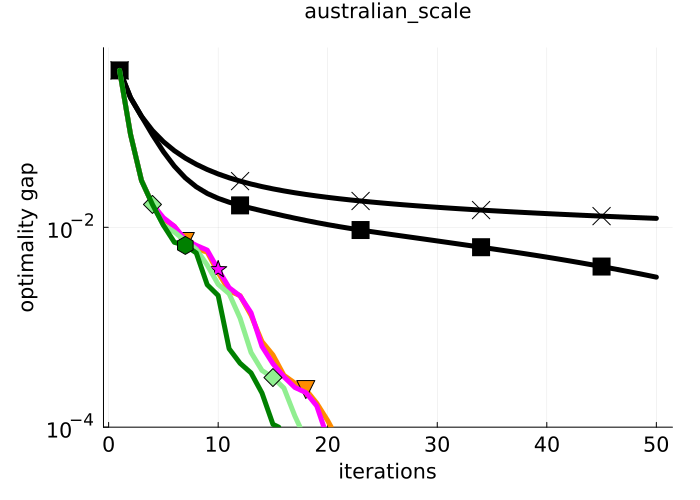}
\includegraphics[width=0.24\textwidth]{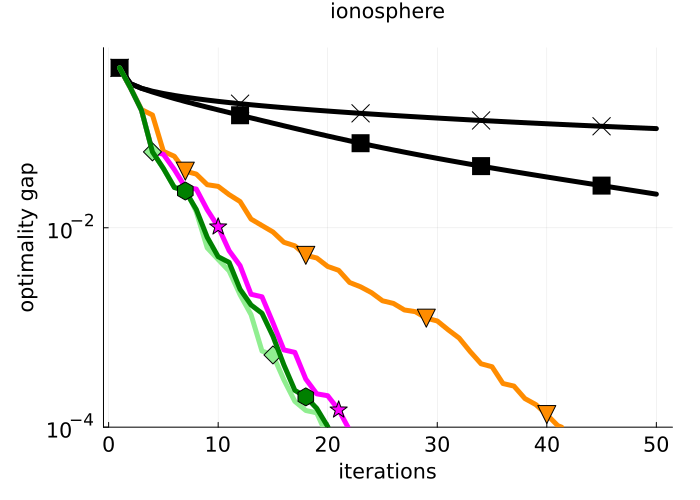}
\includegraphics[width=0.24\textwidth]{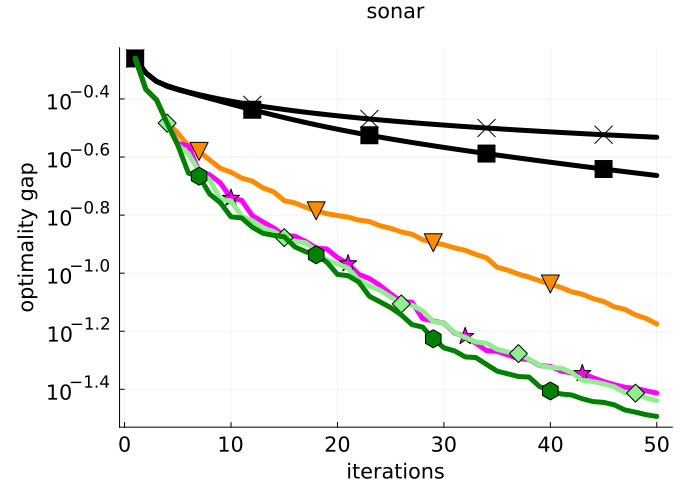}
\includegraphics[width=0.24\textwidth]{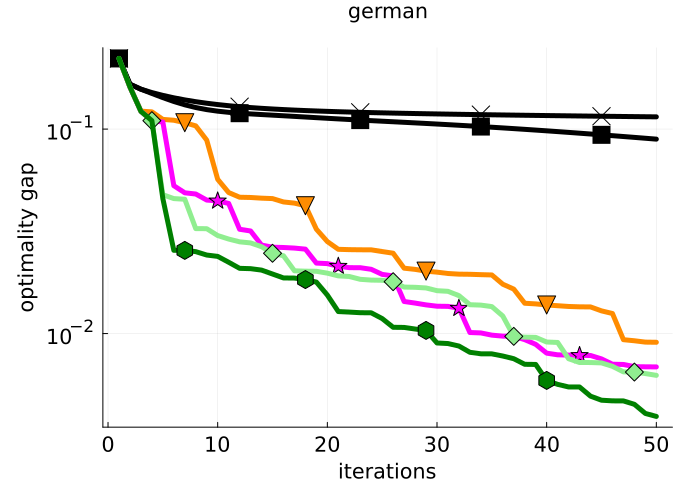}
\includegraphics[width=0.24\textwidth]{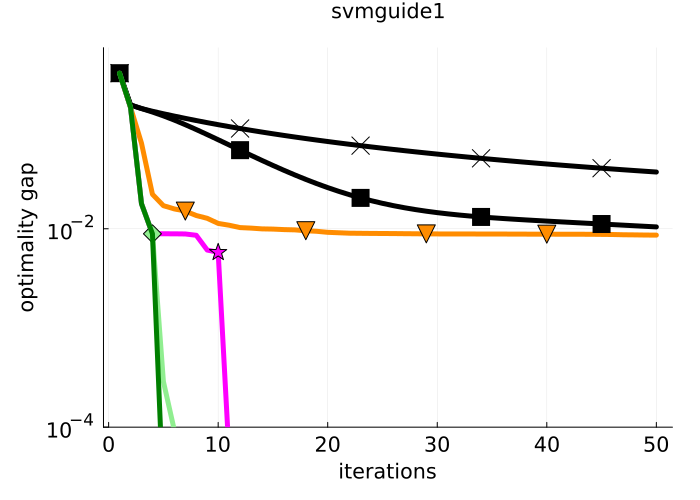}
\includegraphics[width=0.24\textwidth]{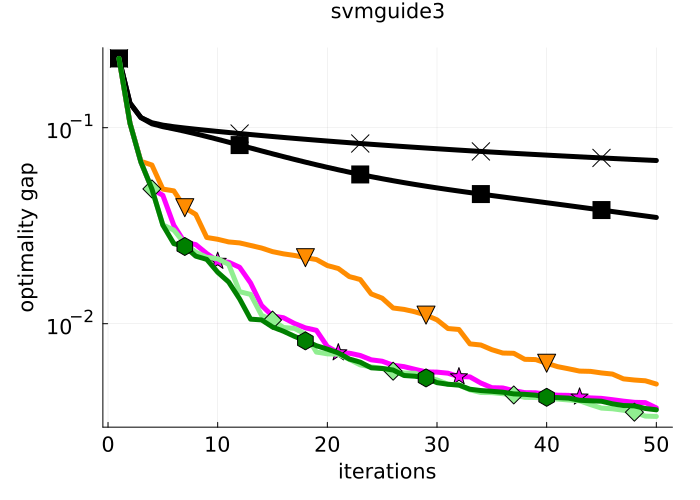}
\includegraphics[width=0.24\textwidth]{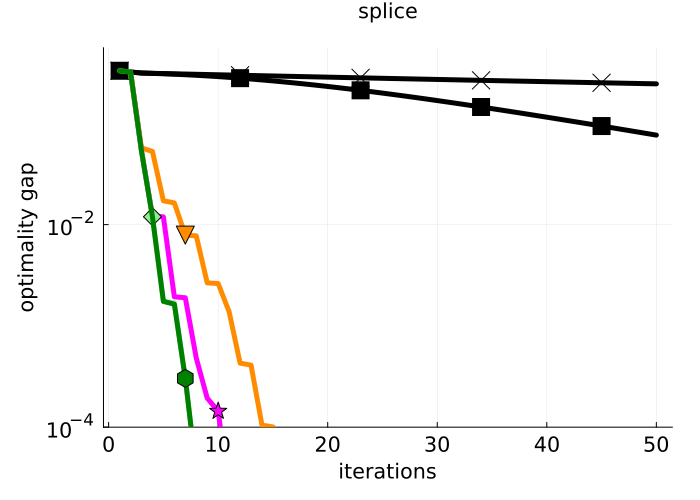}
\includegraphics[width=0.24\textwidth]{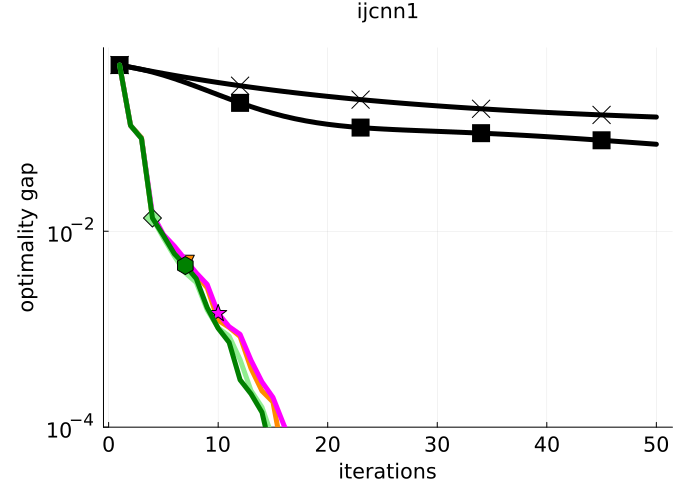}
\includegraphics[width=0.24\textwidth]{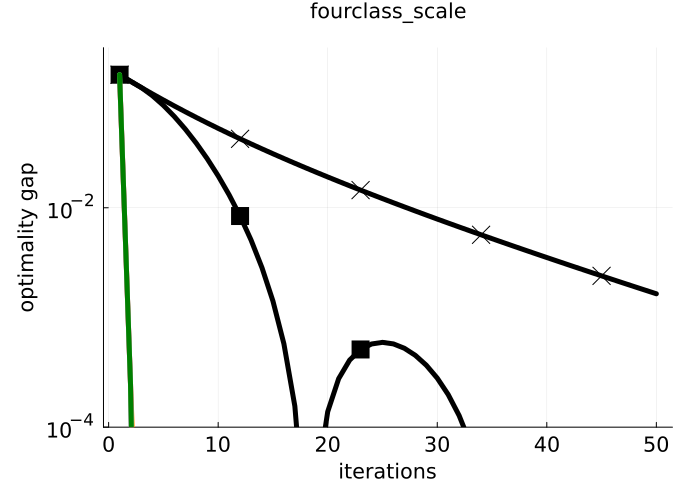}
\includegraphics[width=0.24\textwidth]{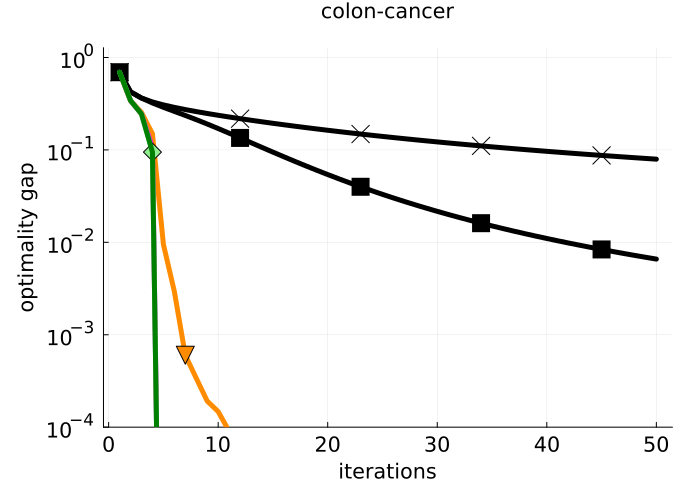}
\includegraphics[width=0.24\textwidth]{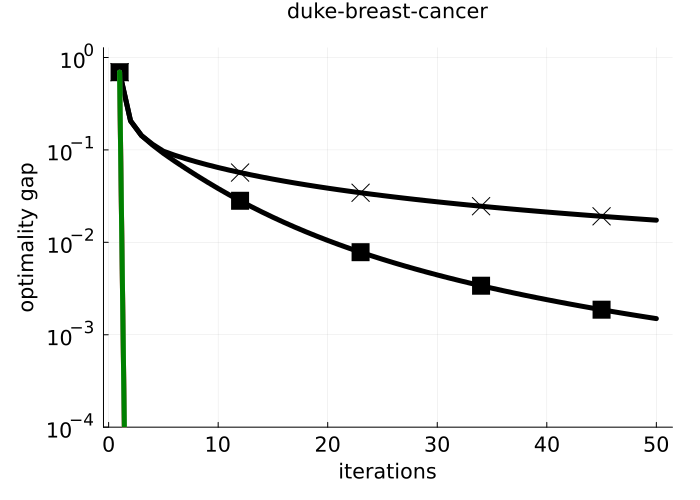}
\includegraphics[width=0.24\textwidth]{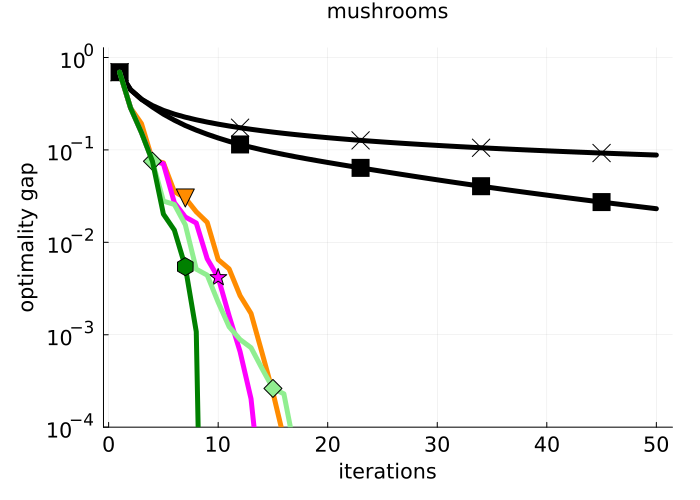}
\includegraphics[width=0.24\textwidth]{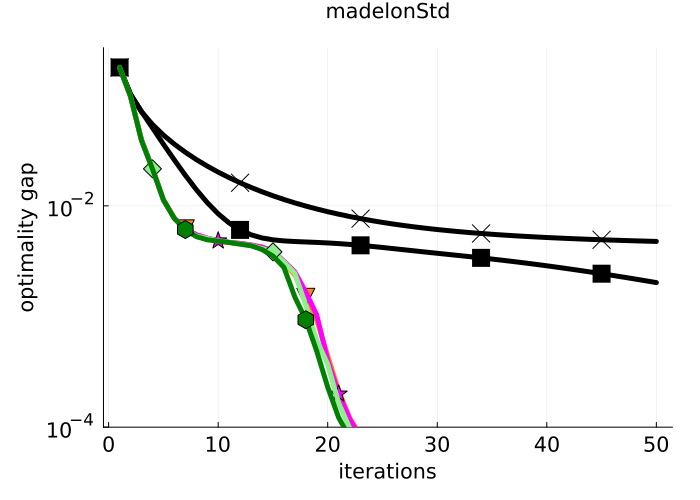}
\includegraphics[width=0.24\textwidth]{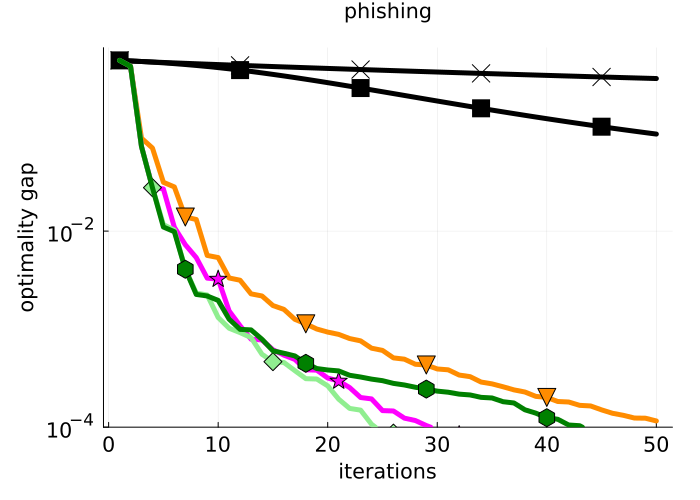}
\includegraphics[width=0.24\textwidth]{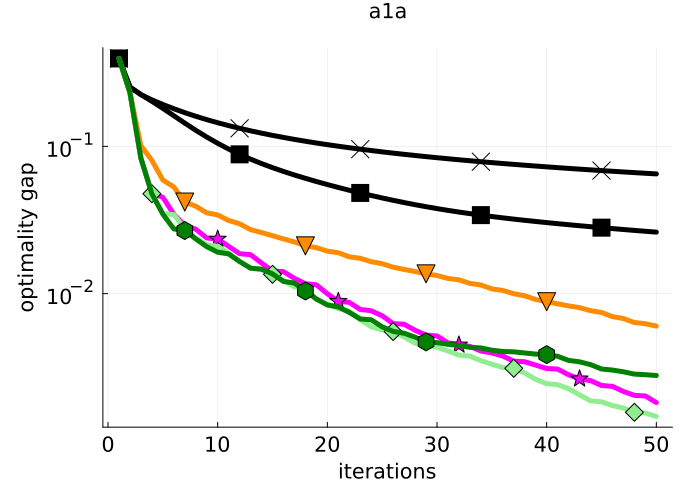}
\includegraphics[width=0.24\textwidth]{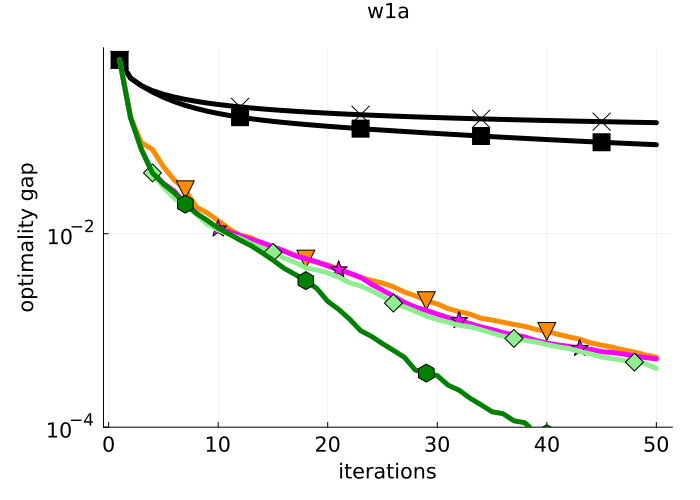}
\includegraphics[width=\textwidth]{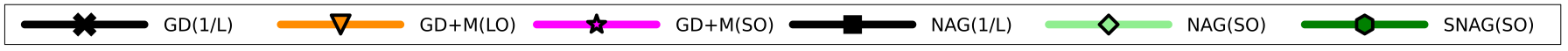}
\caption{Performance of different gradient-based methods for fitting logistic regression models. The black lines only backtrack, the orange line uses LO, the magenta line uses a two-dimensional SO, the light green line uses a three-dimensional SO, and the dark green line uses a four-dimensional SO.  The \texttt{NAG(1/L)} is an implementation of Nesterov's accelerated gradient method. The \texttt{NAG(SO)} adds ``gradient momentum'' to the memory gradient method \texttt{GD+M(SO)}, while the \texttt{SNAG(SO)} method further adds scaling of the parameter vector. We see that acceleration on its own tends to be less effective than using LO with an appropriate direction. We also see that adding additional directions improves performance but that only small gains tend to be observed by optimizing over more than 2 directions.}
\label{fig:NAGlogReg}
\end{figure}

However, convergence rates only describe the worst case performance and it has been highlighted that NAG often underperforms non-linear CG methods in practice~\citep{Narkiss2005}. In Figure~\ref{fig:NAGlogReg}, we add a NAG variant to the previous experiment. In particular, the \texttt{NAG(1/L)} method implements the backtracking version of NAG described by~\citet{beck2009fast}. In this experiment we see that the accelerated \texttt{NAG(1/L)} dominates the non-accelerated \texttt{GD(1/L)}, but that \texttt{NAG(1/L)} is consistently outperformed by the non-accelerated non-linear CG method with LO \texttt{GD+M(LO)} and the non-accelerated memory gradient method~\texttt{GD+M(SO)}.

\textbf{Sequential Subspace Optimization}: \citet{Narkiss2005} proposed sequential subspace optimization (SESOP), which uses SO over multiple directions in order to obtain the $O(1/k^2)$ rate for convex optimization but also obtain CG-like practical performance. They highlight LCPs as a problem class where SO over the gradient and momentum directions can be performed without significantly increasing the cost of the method. SESOP has been applied to various LCP problems~\citep{Narkiss2005, Narkiss2005_SVM,Elad2007}, and has been combined with diagonal scaling and a Hessian-free Newton approach~\citep{Zibulevsky2008}. 

\textbf{Supermemory Method}: the supermemory gradient method~\citep{Cragg1969} uses the most-recent momentum direction $(w_k - w_{k-1})$ and momentum directions $(w_t - w_{t-1})$ from previous iterations $t$. A variation on this idea is to use several previous gradient directions, and examples of this type of method include the methods of~\citet{Elad2007} and~\citet{Wang2006}. It is known that using SO over all previous gradient directions has appealing convergence properties~\citep{Drori2020}, but this can impose significant additional computational costs. We note that for LCPs these supermemory methods also only requires 2 matrix multiplications per iteration. However,
in our experiments we did not find that  additional momentum terms improved performance enough to justify the increased memory requirements and increased SO cost.

\textbf{Multi-Linear Problems}: The linear structure in LCPs is what allows us to use LO and SO within the gradient and memory gradient methods without changing their iteration cost. A related problem is multi-linear problems where the function is linear in certain subsets of the variables. \citet{sorber2016exact} discuss performing LO and SO for a supermemory method in the special case of tensor factorization models. In Appendix~\ref{app:MF} we consider variations on the memory gradient method for matrix factorization, while Appendix~\ref{app:logdet} discusses using LO or SO for log-determinant problems if we use low-rank directions. Unfortunately, it does not seem possible to implement LO and SO for generic multi-linear models without changing the iteration cost. We also note that typical neural networks are not multi-linear. In the case of \citet{sorber2016exact}, they only perform an SO every 4th iteration to keep costs low in their application.

\textbf{Acceleration and SO}: NAG can be written as augmenting the GD+M method with a third direction that is a form of ``gradient momentum'', $(\nabla f(w_k) - \nabla f(w_{k-1}))$~\citep{Shi2021}. 
A CG method based on these three directions was proposed by~\citet{Andrei2014} that gave excellent numerical performance. But we could also consider a version of NAG that uses these 3 directions,
\begin{equation}
x_{k+1} = x_k - \alpha_k \nabla f(w_k) + \beta_k(w_k - w_{k-1}) + \gamma_k(\nabla f(w_k) - \nabla f(w_{k-1})),
\label{eq:NAGSO}
\end{equation}
and performs 3-dimensional SO (this still only requires 2 matrix multiplications per iteration for LCPs). This is appealing since different choices of $\{\alpha_k, \beta_k, \gamma_k\}$ allow NAG to achieve accelerated convergence rates in convex, strongly-convex, and non-convex settings~\citep{Nesterov2013,Li2022}. We thus might expect in some cases that SO would adapt to the problem difficulty. Indeed, several authors have shown that SO allows a single algorithm to achieve optimal rates in several settings~\citep{Drori2020,Nesterov2021,Guminov2023}.

In Figure~\ref{fig:NAGlogReg} we consider using~\eqref{eq:NAGSO} with a three-dimensional subspace optimization, under the name \texttt{NAG(SO)}. In this experiment we found that \texttt{NAG(SO)} typically only gave a small improvement over the memory gradient method. This lack of improvement is consistent with the experiments of~\citet{Karimi2017}. They propose a restarted variant of the  memory gradient method called CGSO that can add a correction step to achieve the accelerated rate in the strongly-convex setting. This correction step adds directions to the memory gradient method, but they found that in practice the method applied the memory gradient step in almost every case.

\textbf{Scaled Gradient and Memory Methods}: \citet{sorber2016exact} consider a scaled gradient method,
\[
x_{k+1} = \delta_k x_k - \alpha_k\nabla f(w_k),
\]
where on each iteration we estimate a scaling factor $\delta_k$ for the original variable. For LCPs we can search for an optimal $\alpha_k$ and $\delta_k$ while still only requiring 2 matrix multiplications per iteration. However,~\citet{sorber2016exact} found that adding the scaling factor $\delta_k$ did not significantly improve performance in their tensor factorization experiments. In preliminary experiments we found that this scaled gradient method improved performance over the gradient method with LO, but did not tend to outperform the memory gradient method which also uses 2 directions.

We could also use a scaled variant of the memory gradient method
\begin{equation}
x_{k+1} = \delta_k x_k - \alpha_k\nabla f(w_k) + \beta_k(w_k - w_{k-1}),
\label{eq:scaledMG}
\end{equation}
and solving for the three scaling factors still only requires 2 matrix multiplications. In preliminary experiments we found that this 3-direction approach can improve the performance of the memory gradient method, but tended to be outperformed by the 3-direction \texttt{NAG(SO)} method that has momentum and gradient momentum~\eqref{eq:NAGSO}. On the other hand, a further small performance gain could be obtained by combining all 4 directions into the update
\[
x_{k+1} = \delta_kx_k - \alpha_k \nabla f(w_k) + \beta_k(w_k - w_{k-1}) + \gamma_k(\nabla f(w_k) - \nabla f(w_{k-1})).
\]
We refer to this method as \texttt{SNAG(SO)} in Figure~\ref{fig:NAGlogReg}.

\textbf{Second-Order SO}: Using LO to set the step size in Newton's method does not increase the iteration cost for many problem structures and preserves  the superlinear convergence rate of Newton's method~\citep{shea2023greedy}. Recent work gives explicit non-asymptotic convergence rates of BFGS with LO~\citep{jin2024non}.~\citet{Conn1994} discuss using SO with the gradient direction and the Newton direction.  This method improves on the global convergence properties of Newton's method while maintaining its superlinear convergence rate~\citep{zibulevsky2013speeding,shea2023greedy}. We note that it is common to use SO to minimize quadratic approximations of the function, as in trust region methods~\citep[see][]{Nocedal2006}, but in this work we focus on methods that use SO to directly minimize the function.
An approximate second-order method has also been proposed in a majorization-minorization framework for certain problem structures~\citep{Chouzenoux2010}. However, second-order information typically significantly increases the iteration cost. Thus, in Section~\ref{sec:QN} we consider using SO within quasi-Newton methods that cheaply approximate second-order information.

\textbf{Global Optimization with SO}: 
\citet{wang2016bayesian} consider global optimization, and search for an approximate solution by optimizing over a random subspace. Recent works have explored iterative variations where a different subspace is used on each iteration~\citep{cartis2022dimensionality,cartis2023bound} or growing subspaces are used~\citep{cartis2023global}. But in this work we restrict attention to the typical ML setting where the dimension is high and only a local minimum may be found. 

\textbf{SGD with SO}:
\citet{richardson2016seboost} consider augmenting an SGD method with SO. In particular, they alternate between taking a large number of SGD steps with a small mini-batch and then taking an SO step with a large mini-batch. The SO step uses the current gradient and differences between previous iterates (as in the supermemory method). 
This method  improved the performance of three different SGD methods (SGD with momentum, NAG, and AdaGrad) in terms of runtime on several deep learning tasks. In this work we restrict focus to deterministic methods since these are the key ingredient behind the most effective SGD methods. In particular, SGD with momentum is a stochastic version of Polyak's deterministic GD+M method while it has recently been shown that the performance advantage of Adam on language models is due to its deterministic properties~\citep{kunstner2022noise}. We expect that by improving the deterministic ingredients of stochastic methods that this work will lead to faster stochastic methods in the future. 

\section{Line Search and [Hyper-]Plane Search for Neural Networks}
\label{sec:SOfriendly}

While SO has a long history and numerous applications, the work of~\citet{richardson2016seboost} is the only previous work we are aware of that considers using SO to train neural networks. The lack of works using SO to train neural networks is sensible because for general neural networks we cannot evaluate the objective at many points in a subspace for a lower cost than evaluating the objective at arbitrary points. Nevertheless, there exist neural networks that are used in practice where a linear operator with the parameters is the bottleneck operation. These networks allow us to efficiently implement SO to set the learning rate and momentum rate, and indeed allow us to optimize over separate rates for each layer. In these settings we may expect to see the same performance gains that SO offers for LCPs.

\subsection{Definition: SO-Friendly Neural Networks}

We can write typical neural network objective functions in the form $f(W,v) = g(h(XW) v)$ for non-linear functions $g$ and $h$. Here, $X$ is the data matrix (in $\R^{n \times d}$), $W$ is the set of weights in the first hidden layer (in $\R^{d \times r}$ with $r$ hidden units), and $v$ is the concatenation of the parameters in the second and higher layers of the network. We say that {\bf a neural network is SO-friendly if the cost of evaluating $f$ is dominated by the cost of matrix multiplications with $X$.} The dominant computational cost in SO-friendly networks is similar to LCPs, and this allows us to use LO and SO efficiently even though the parameters $v$ apply to a non-linear function of the input. 

There are a variety of possible SO-friendly neural network structures, but we will first discuss the simple case of a network with two layers of hidden weights. This type of network has been popularized in practice under the name extreme learning machine (ELM)~\citep{Huang2006}, and ELMs are among the best performing out-of-the-box classifiers~\citep{Fernandez2014}. However, in ELMs the first layer of weights is set randomly while we consider using SO to fit both layers efficiently. Two-layer networks are also popular in reinforcement learning within the trust-region policy optimization and proximal policy optimization frameworks~\citep{schulman2015trust,schulman2017proximal}, where simple networks have been sufficient to solve a number of continuous control tasks~\citep{rajeswaran2017towards}. Finally, we note that having an efficient method for fitting 2-layer networks would allow us to replace 1-layer networks in settings where these degenerate linear models are used. For example, for transfer learning we could train the last two layers of a pre-trained deep network instead of only training the last layer.

\subsection{Example: 2-Layer Networks with a Single Output - Tied Step Size(s)}
\label{sec:SOfriendly_2layerNN_tiedStep}

A classic example of an SO-friendly neural network is a fully-connected neural network with two layers of weights and a single output. 
In this setting $v$ has $r$ elements and $f$ has the simplified form $f(W,v) = g(h(XW)v)$, where the activation function $h$ applies a non-linear operation such as a sigmoid function element-wise and the loss function $g$ maps the $n$ predictions to a scalar loss. The gradient descent update for a 2-layer neural network with a single output takes the form
\begin{align*}
	W_{k+1} & = W_k - \alpha_kX^T\underbrace{\text{diag}(\nabla g(h(XW_k)v_k)) h'(XW_k)\text{diag}(v_k)}_{R_k},\\
	v_{k+1} & = v_k - \alpha_k\underbrace{h(XW_k)^T\nabla g(h(XW_k)v_k)}_{\nabla_v f(W_k,v_k)},
\end{align*}
where $h'$ is the matrix containing the derivative with respect to each input and ``diag'' makes a diagonal matrix from a vector. With a fixed step size the computational cost of this update is dominated by the two multiplications with $X$ per iteration: one to compute $XW_k$ and another to compute $X^TR_k$. Consider now performing LO to set the step size,
\begin{align*}
& \argmin_\alpha f(W_k - \alpha X^TR_k,v_k - \alpha\nabla_v f(W_k,v_k))\\
\equiv & \argmin_\alpha g(h(X(W_k - \alpha X^TR_k)(v_k - \alpha\nabla_v f(W_k,v_k))))\\
\equiv & \argmin_\alpha g(h(\underbrace{XW_k}_{M_k} - \alpha\underbrace{X(X^TR_k)}_{D_k})(v_k - \alpha_k\nabla_v f(W_k,v_k)))\\
\equiv & \argmin_\alpha g(h(\underbrace{M_k - \alpha D_k}_\text{potential $M_{k+1}$})(v_k - \alpha\nabla_v f(W_k,v_k))).
\end{align*}
Analogous to LCPs, if we track the $n \times r$ product $M_k = XW_k$ then we can perform LO using only two of the bottleneck matrix multiplications per iteration: one to compute the $d$ by $r$ product $X^TR_k$, and another to pre-mutiply this matrix by $X$.
These two matrix multiplications cost $O(ndr)$, and given $M_k$ and $D_k$ evaluating the LO objective only costs $O(nr)$. Similar to the LCP case, we can also optimize the momentum rate without additional matrix multiplications with $X$. 

For neural networks the LO and SO problems are non-convex, and we did find that it was possible to find poor local optima of the SO problem in particular. Thus, care should be taken in how we try to solve the SO problem and we should at least use a method that guarantees we decrease $f$ compared to using a step size (or step sizes) of zero (otherwise LO/SO can result in worse performance). Nevertheless,  we found that the simple generic method used in our LCP experiments (see Appendix~\ref{app:SOsolve}) typically performed well and we  used this method in all our neural network experiments.

\subsection{LO and SO for 2-Layer Networks in Practice (Tied Step Sizes)}

\begin{figure}
\includegraphics[width=0.24\textwidth]{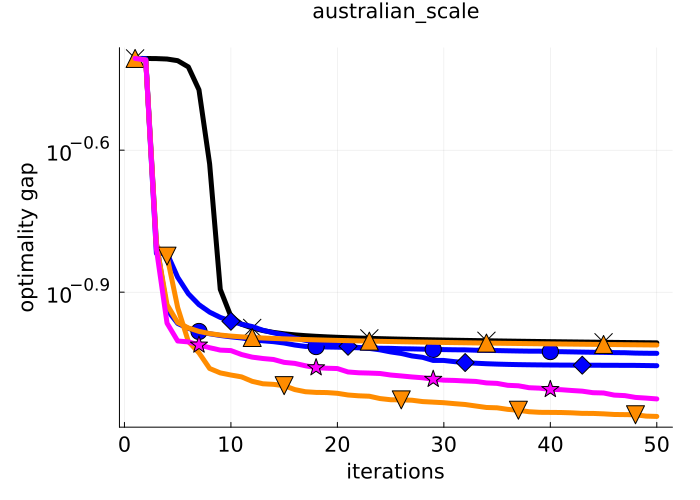}
\includegraphics[width=0.24\textwidth]{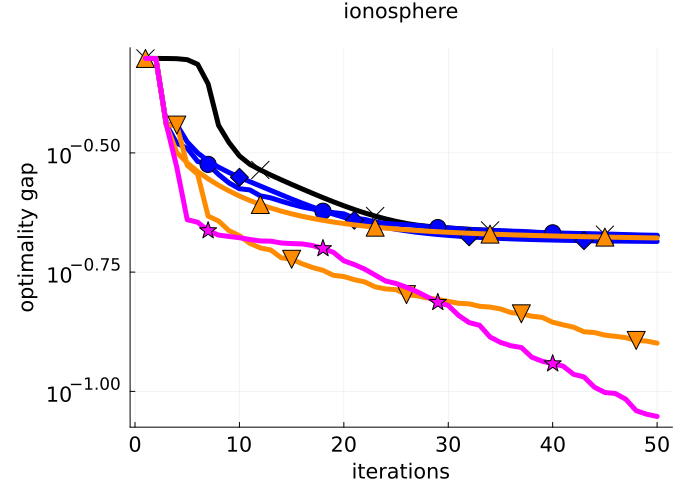}
\includegraphics[width=0.24\textwidth]{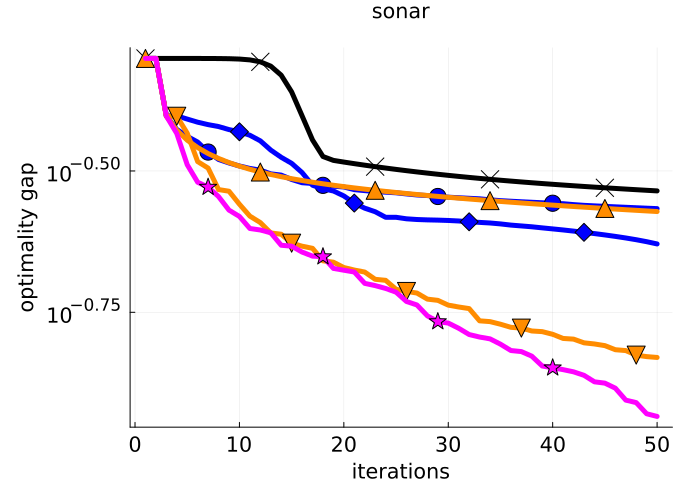}
\includegraphics[width=0.24\textwidth]{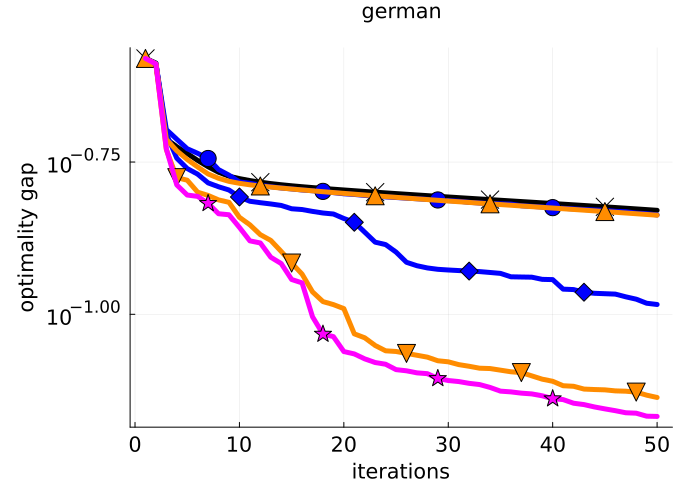}
\includegraphics[width=0.24\textwidth]{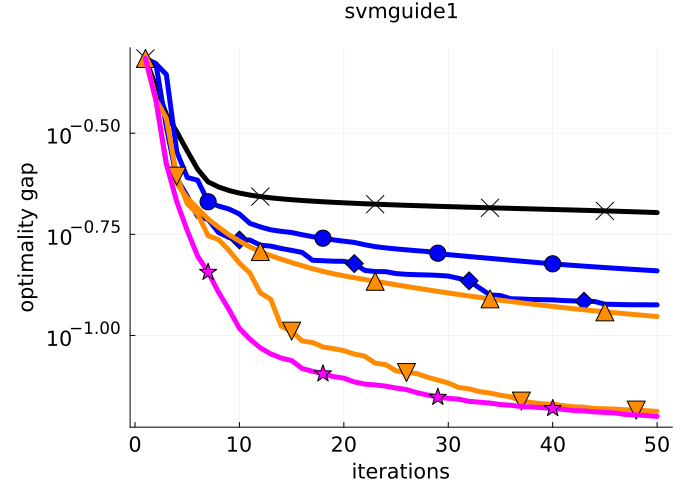}
\includegraphics[width=0.24\textwidth]{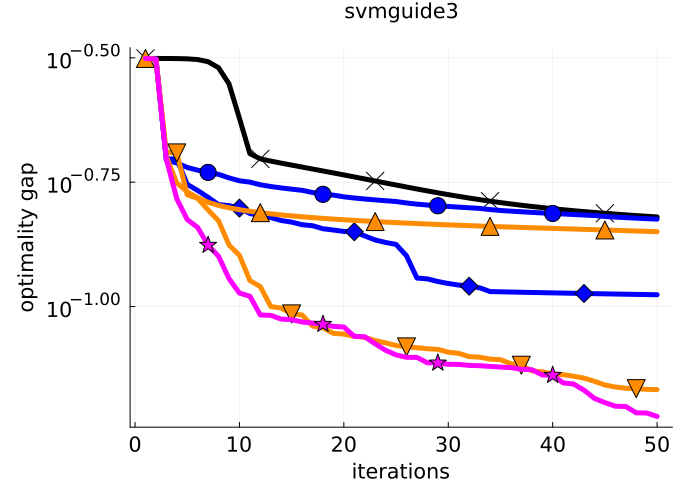}
\includegraphics[width=0.24\textwidth]{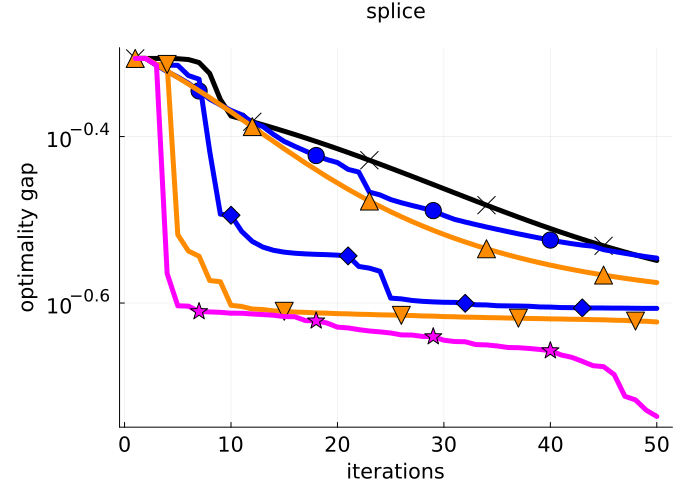}
\includegraphics[width=0.24\textwidth]{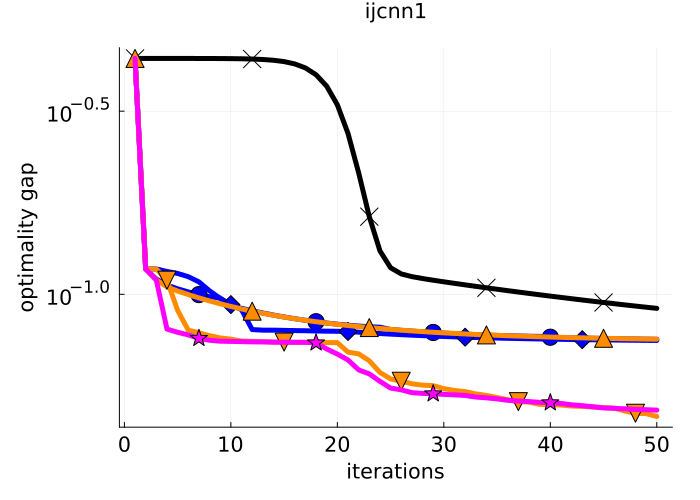}
\includegraphics[width=0.24\textwidth]{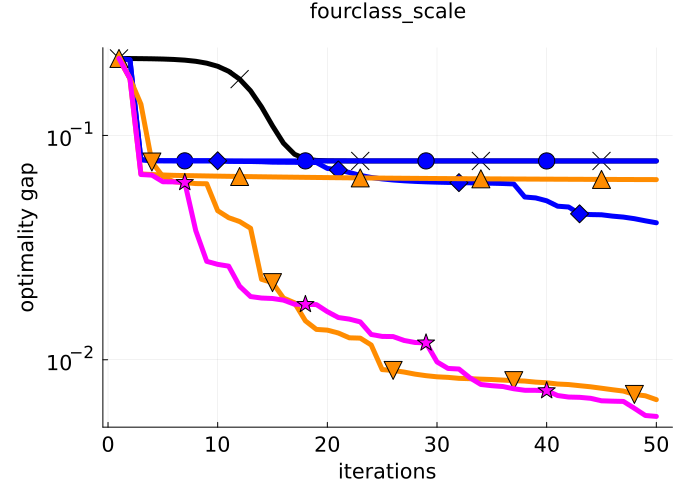}
\includegraphics[width=0.24\textwidth]{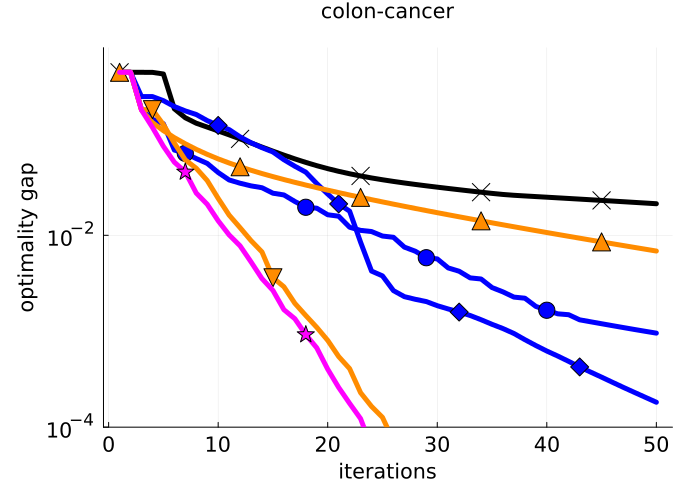}
\includegraphics[width=0.24\textwidth]{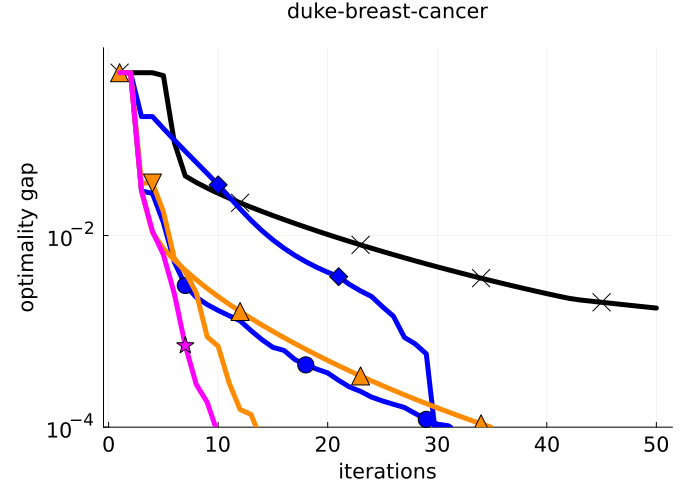}
\includegraphics[width=0.24\textwidth]{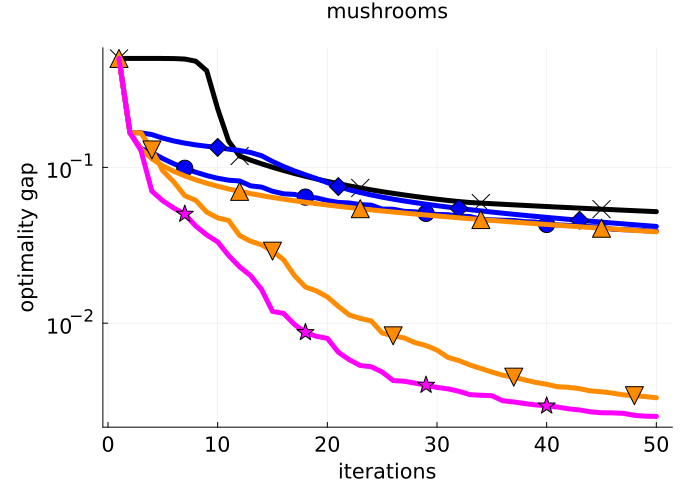}
\includegraphics[width=0.24\textwidth]{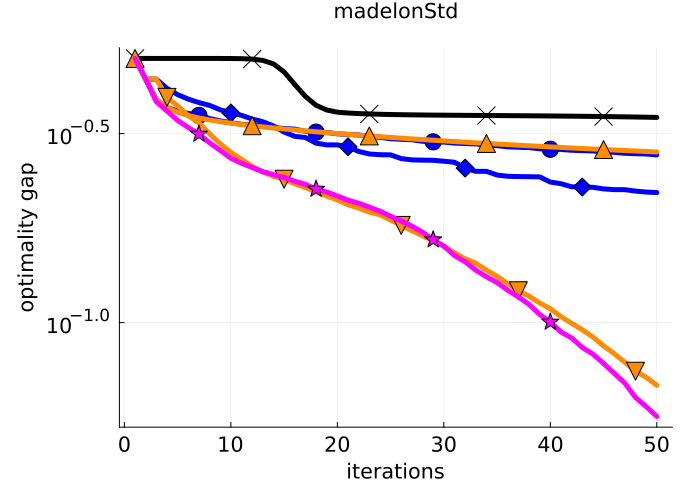}
\includegraphics[width=0.24\textwidth]{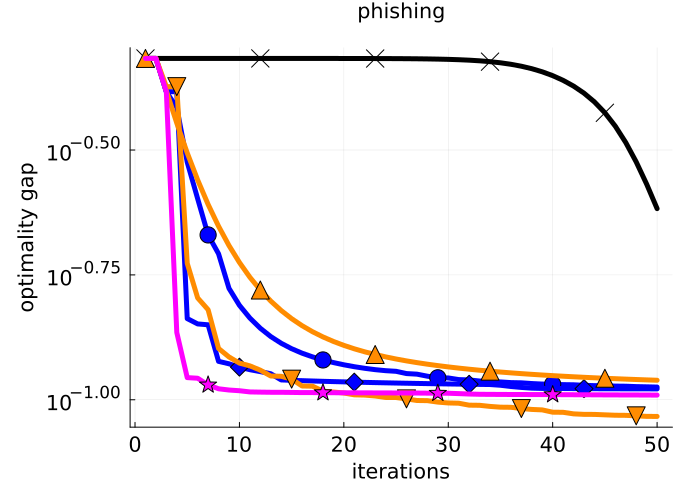}
\includegraphics[width=0.24\textwidth]{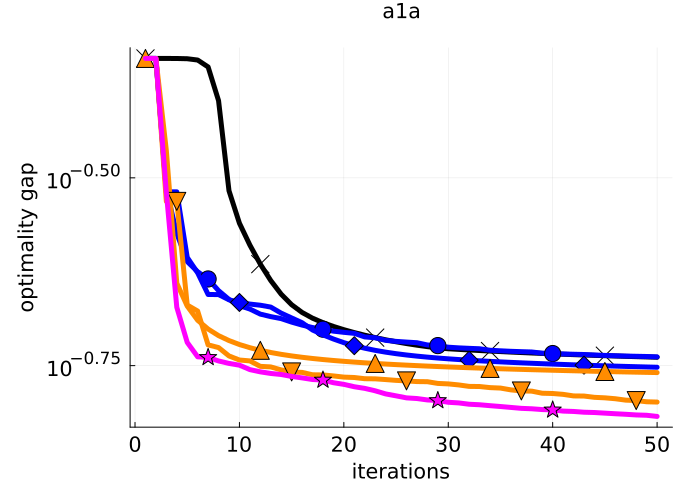}
\includegraphics[width=0.24\textwidth]{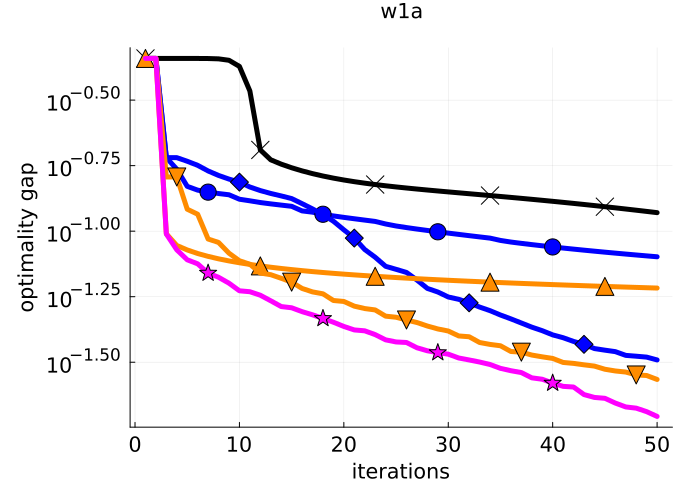}
\includegraphics[width=\textwidth]{firstLegend.png}
\caption{Performance of different gradient-based methods for 2-layer neural networks.  The black line only backtracks, the blue lines use a line search that can decrease or increase the step size to satisfy the strong Wolfe conditions, the orange lines use LO, and the magenta line uses SO.  The \texttt{GD} methods use the gradient direction, the \texttt{GD+M(L*)} methods use the gradient direction and momentum with the non-linear conjugate gradient relationship between the parameters, and the \texttt{GD+M(SO)} method optimizes the learning rate and momentum rate. We see that the 1/L method has better performance for training neural networks than for training linear models, but that the \texttt{GD+M(LO)} and \texttt{GD+M(SO)} methods that exploit momentum and optimized step sizes tend to outperform the other methods..}
\label{fig:firstnn100}
\end{figure}

In this section we consider training 2-layer networks using the methods and datasets from Section~\ref{sec:LCPexp}. We emphasize that the iteration costs of all methods in this setting are dominated by the cost of the two products with $X$ on each iteration, and thus on large datasets all methods will have similar runtimes. For these networks we use $r=100$ hidden units, use $h=\tanh$ as the activation function, use the squared loss objective $g(h(XW)v)=\norm{h(XW)v-y}_2^2$, and initialize all elements of $W$ and $v$ with a sample from a standard normal distribution divided by the total number of weights in the network  $r(d + 1)$.

\begin{figure}
\includegraphics[width=0.24\textwidth]{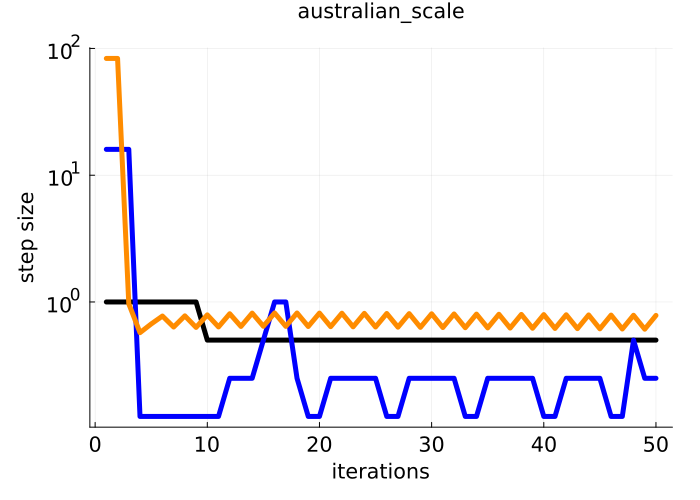}
\includegraphics[width=0.24\textwidth]{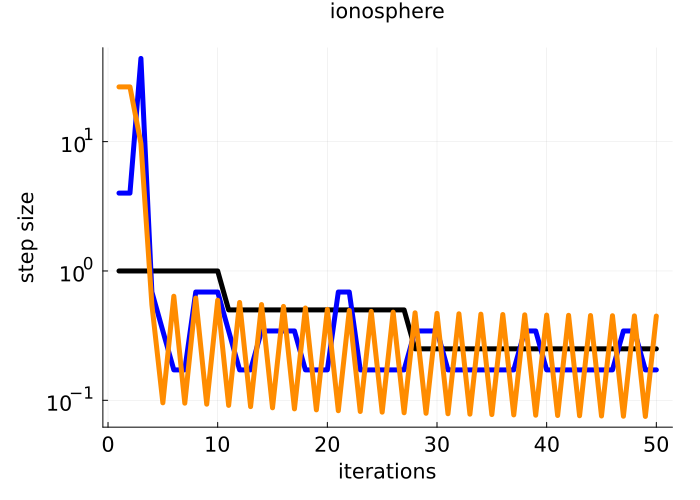}
\includegraphics[width=0.24\textwidth]{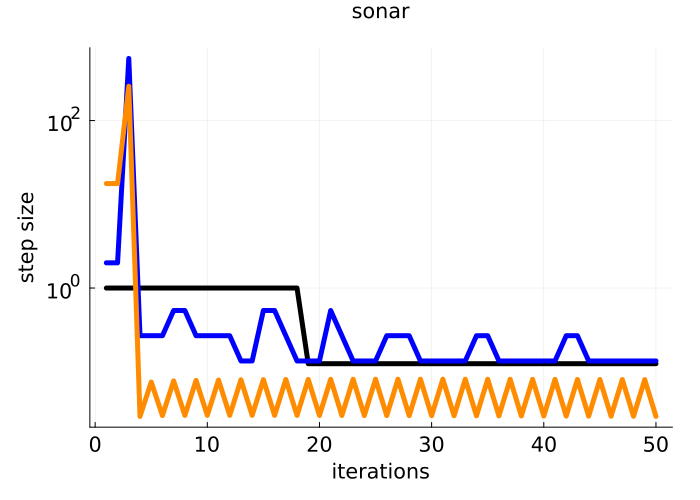}
\includegraphics[width=0.24\textwidth]{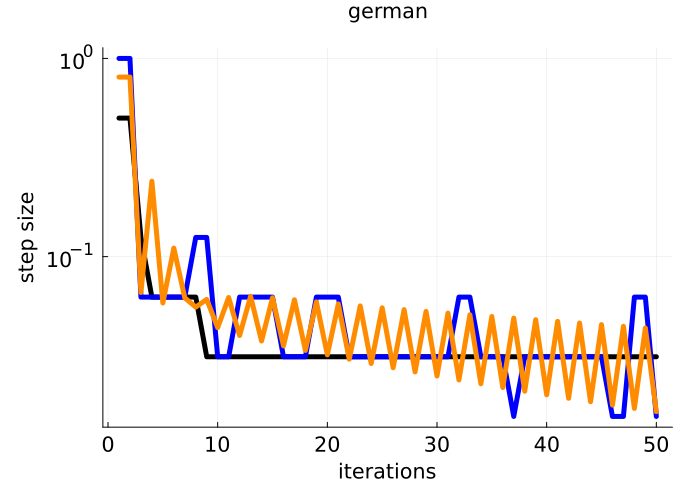}
\includegraphics[width=0.24\textwidth]{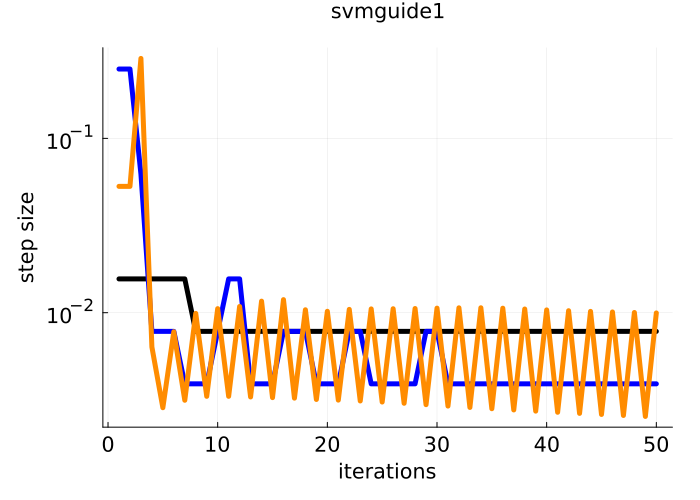}
\includegraphics[width=0.24\textwidth]{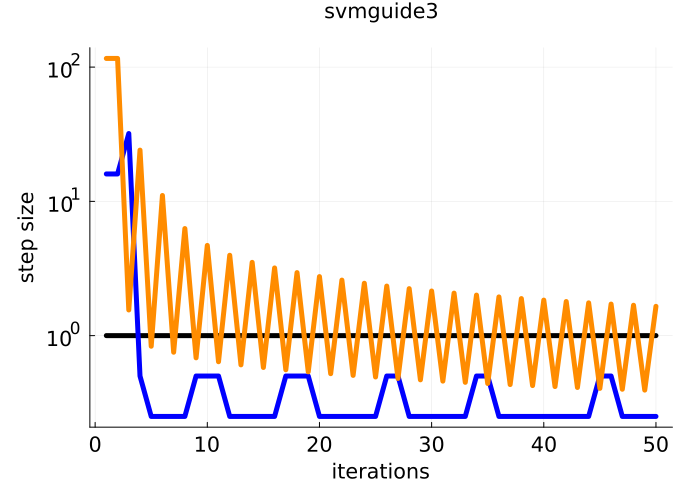}
\includegraphics[width=0.24\textwidth]{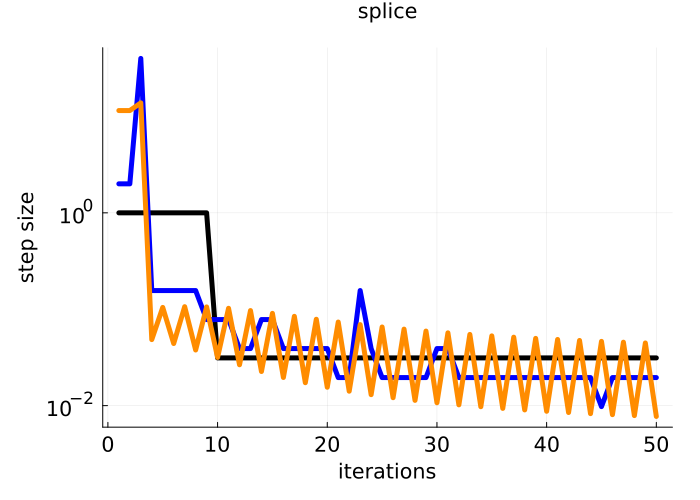}
\includegraphics[width=0.24\textwidth]{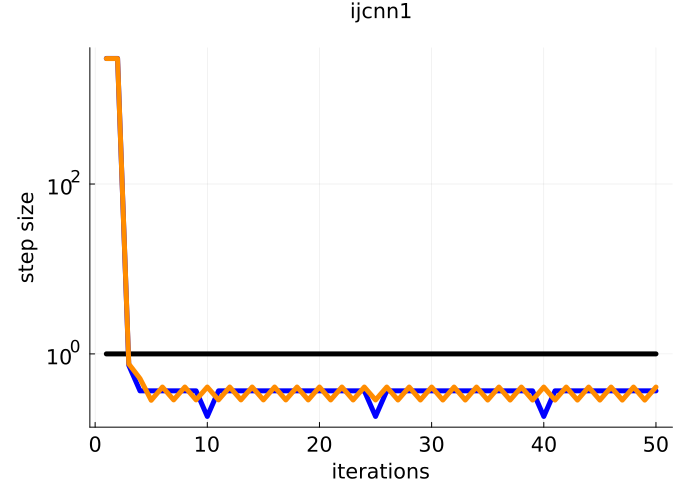}
\includegraphics[width=0.24\textwidth]{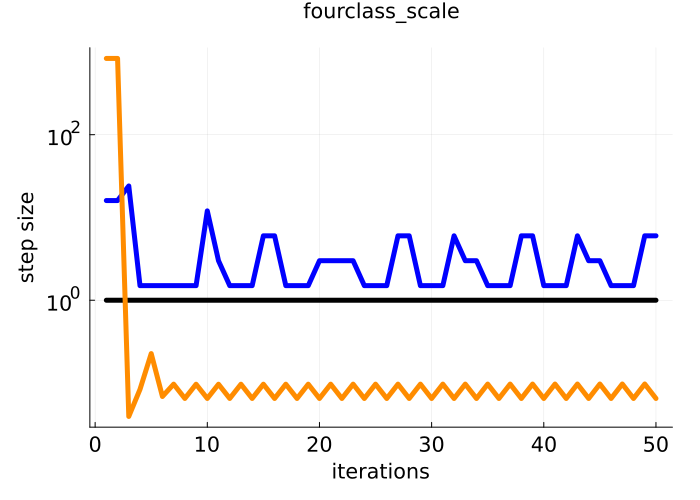}
\includegraphics[width=0.24\textwidth]{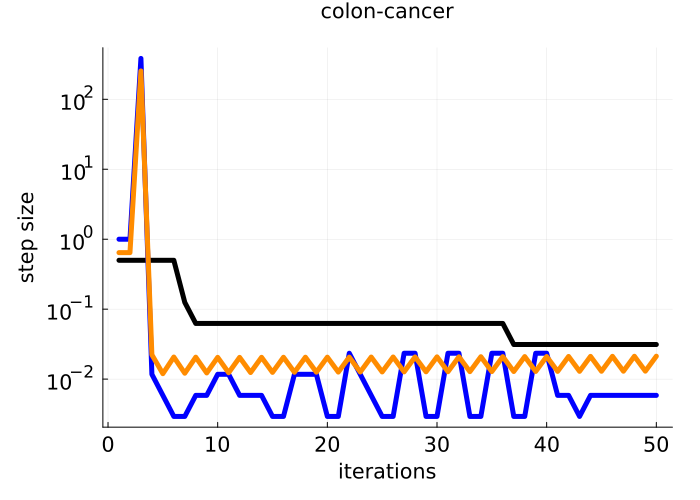}
\includegraphics[width=0.24\textwidth]{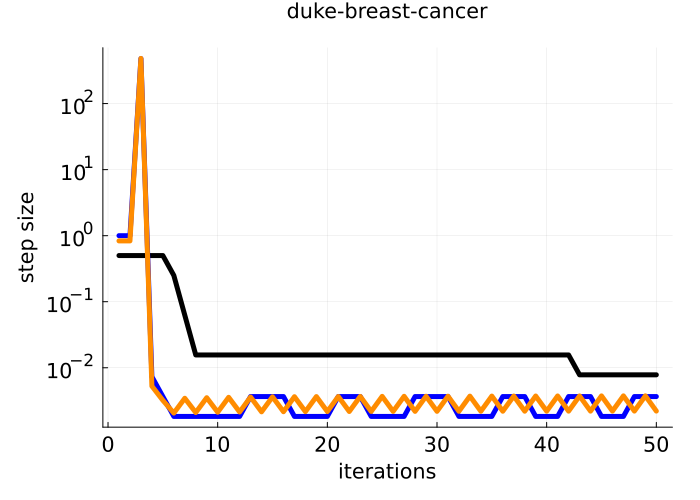}
\includegraphics[width=0.24\textwidth]{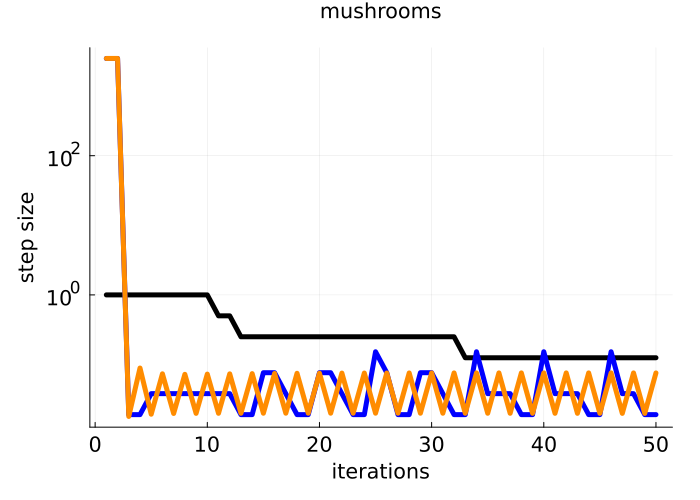}
\includegraphics[width=0.24\textwidth]{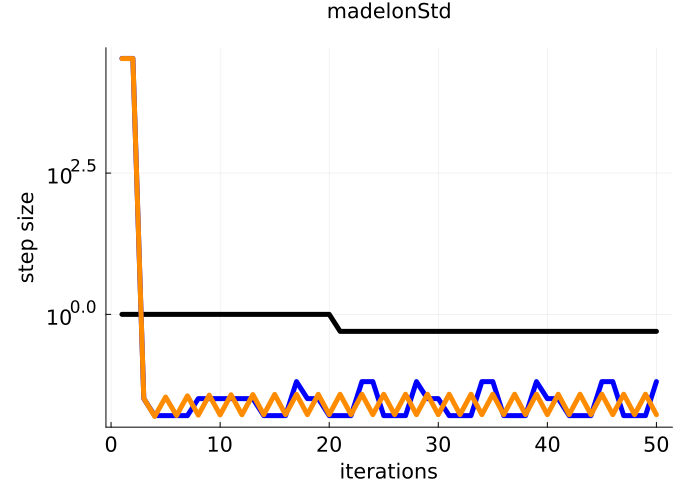}
\includegraphics[width=0.24\textwidth]{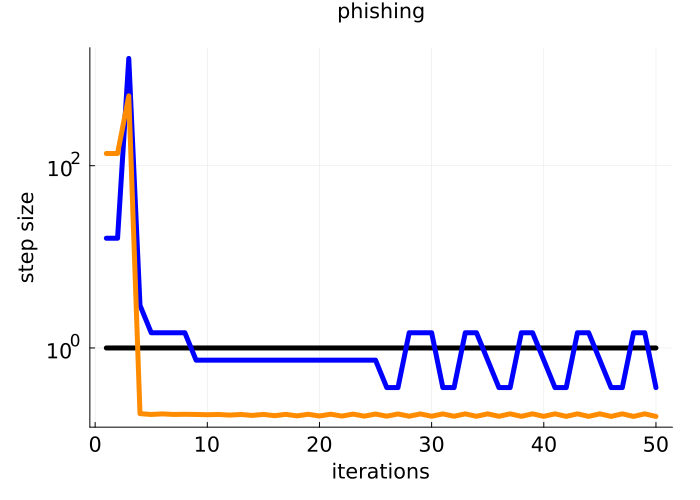}
\includegraphics[width=0.24\textwidth]{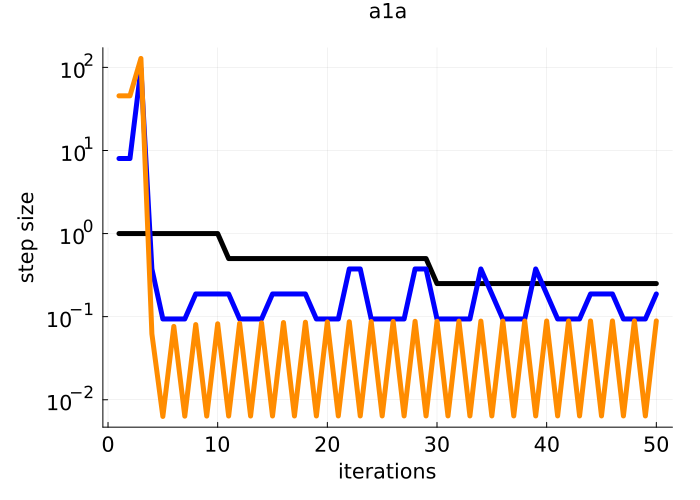}
\includegraphics[width=0.24\textwidth]{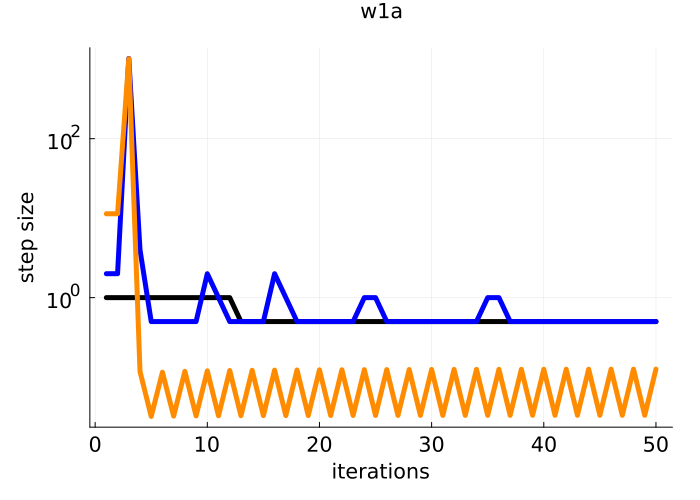}
\begin{center}
\includegraphics[width=.5\textwidth]{LOstepLegend.png}
\end{center}
\caption{Step sizes of different gradient descent methods for fitting 2-layer neural networks. The basic \texttt{GD(1/L)} tended to find step sizes closer to the line search and LO methods on these problems.}
\label{fig:LOstepnn100}
\end{figure}

We plot the results of the 2-layer training experiment in Figure~\ref{fig:firstnn100}. We highlight two observations from this experiment:
\begin{enumerate}
\item The essentially-fixed step size method \texttt{GD(1/L)} performed much better for neural networks than for it did for the linear model. In particular, \texttt{GD(1/L)} was often not substantially worse than the line search \texttt{GD(LS)} method and in some cases performed better than the line search method. This counter-intuitive behaviour is perhaps explained by the edge of stability phenomenon~\citep{cohen2020gradient}; unlike linear models where a constant step size must adapt to the maximum curvature of the function, in neural networks it appears empirically that the curvature of the function can adapt to the step size. In Figure~\ref{fig:LOstepnn100} we plot the step sizes of the different gradient descent methods, and we see that the \texttt{GD(1/L)} method often found steps on a similar scale to the \texttt{GD(LS)} method (unlike for linear models where the \texttt{GD(1/L)} step sizes were typically too small). That the curvature of the network may be adapting to the step size may explain why classic line searches have had little impact on the training of neural networks.
\item The methods using LO and SO with momentum, \texttt{GD+M(LO)} and \texttt{GD+M(SO)}, consistently outperformed the other methods. Thus, it appears that \textbf{neural network training can be sped up by per-iteration step size tuning but that it must be done precisely and must use an appropriate per-iteration momentum}. We did not find obvious patterns in the step sizes used by  \texttt{GD+M(LO)} and \texttt{GD+M(SO)} that were not seen in the LCP experiments, although we note that the \texttt{GD+M(SO)} method used negative momentum rates slightly more frequently than for linear models.
\end{enumerate}

\subsection{Example: 2-Layer Networks with a Single Output - Per-Layer Step Sizes}
\label{sec:SOfriendly_2layerNN_varyStep}

While LO and SO appear to be helpful for training neural networks, we can further take advantage of the layer structure to efficiently use different step sizes in different layers. First consider the case of a 2-layer network and a variant of gradient descent where we have a step size for each layer:
\begin{align*}
	W_{k+1} & = W_k - \alpha_k^1X^TR^k,\\
	v_{k+1} & = v_k - \alpha_k^2\nabla_v f(W_k,v_k),
\end{align*}
where $\alpha_k^1$ is the step size for the first layer and $\alpha_k^2$ is the step size for the second layer. We might expect better performance by updating the different layers at different rates. Observe that using SO to set both step sizes requires only the same two matrix multiplications with $X$ that we need to set a single step size,
\[
 \argmin_{\alpha^1, \alpha^2} g(h(\underbrace{M_k - \alpha^1D_k}_\text{potential $M_{k+1}$})(v_k - \alpha^2\nabla_v f(W_k,v_k))).
\]
The GD+M update~\eqref{eq:GDM} with a separate learning and momentum rate for each layer has the form
\begin{align*}
	W_{k+1} & = W_k - \alpha_k^1X^TR_k + \beta_k^1(W_k - W_{k-1}) \\
	v_{k+1} & = v_k - \alpha_k^2\nabla_vf(W_k, v_k) + \beta_k^2(v_k - v_{k-1}),
\end{align*}
and optimizing over all 4 step sizes again only requires the same two matrix multiplications,
\begin{align*}
& \argmin_{\alpha^1, \alpha^2, \beta^1, \beta^2} f(W_k - \alpha^1 X^TR_k + \beta^1(W_k - W_{k-1}),v_k - \alpha^2\nabla_v f(W_k,v_k) + \beta^2(v_k - v_{k-1}))\\
\equiv & \argmin_{\alpha^1, \alpha^2, \beta^1, \beta^2} g(h(X(W_k - \alpha^1 X^TR_k + \beta^1(W_k - W_{k-1}))(v_k - \alpha^2\nabla_v f(W_k,v_k) + \beta^2(v_k - v_{k-1}))))\\
\equiv & \argmin_{\alpha^1, \alpha^2, \beta^1, \beta^2} g(h(\underbrace{(1+\beta^1)M_k - \alpha^1 D_k  - \beta^1M_{k-1}}_\text{potential $M_{k+1}$})((1+\beta^2)v_k - \alpha^2\nabla_v f(W_k,v_k)- \beta^2 v_{k-1})).
\end{align*}
Thus, at each iteration we can numerically search for a learning rate for each layer and a momentum rate for each layer for the same asymptotic cost as using fixed step sizes. We could also consider adding previous iterations or gradients as additional directions as in supermemory methods, or use a scaling of each layer as in the scaled memory gradient method~\eqref{eq:scaledMG}.


\subsection{LO and SO for 2-Layer Networks in Practice (Per-Layer Step Sizes)}

In this section, we repeat our previous experiment but include methods with per-layer step sizes:
\begin{itemize}
\item \texttt{GD(SB)}: Gradient descent where SO is used to set a learning rate for each layer (2 step sizes for 2-layer networks).
\item \texttt{GD+M(SB)}: Gradient descent with momentum using the non-linear CG direction, but where SO is used to set a step size for each layer (2 step sizes for 2-layer networks). We use the PR+ formula separately for each layer, but only reset if the combined direction across layers is not a descent direction.
\item \texttt{GD+M(SO+SB)}: Gradient descent with momentum where SO is used to set a separate learning rate and momentum rate for each layer (4 step sizes for 2-layer networks).
\end{itemize}
We once again repeat that the iteration costs of all methods are dominated by the cost of the two products with $X$ on each iteration, although we note that a 4-dimensional SO problem may take more time to solve than a 2-dimensional or 1-dimensional problem.

\begin{figure}
\includegraphics[width=0.24\textwidth]{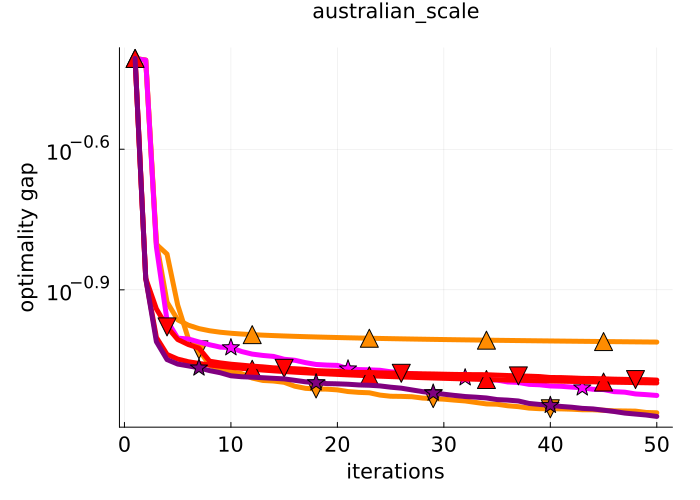}
\includegraphics[width=0.24\textwidth]{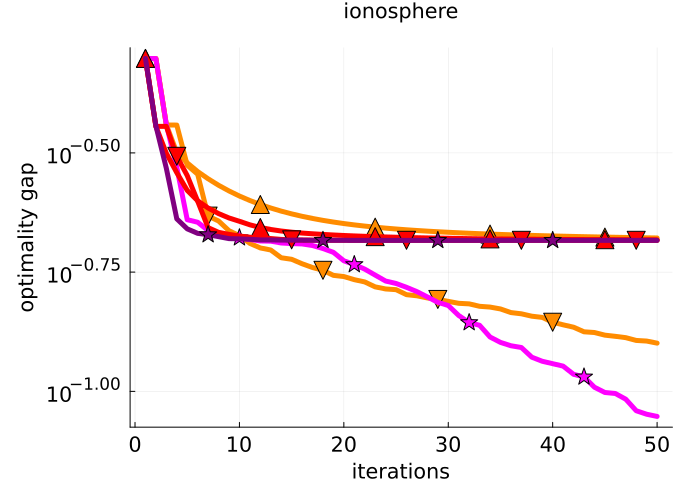}
\includegraphics[width=0.24\textwidth]{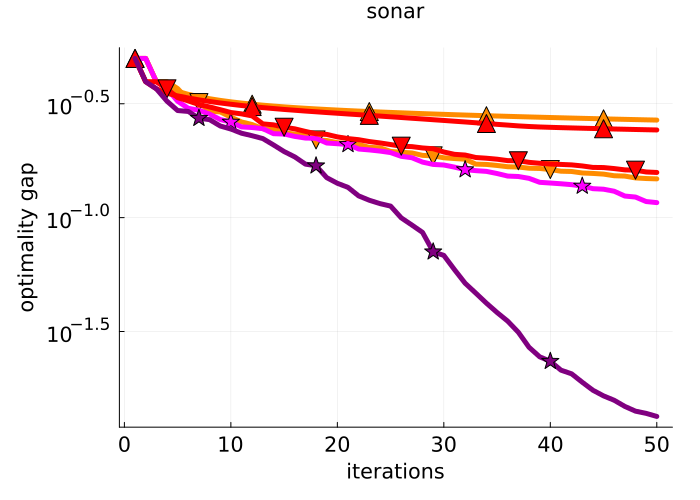}
\includegraphics[width=0.24\textwidth]{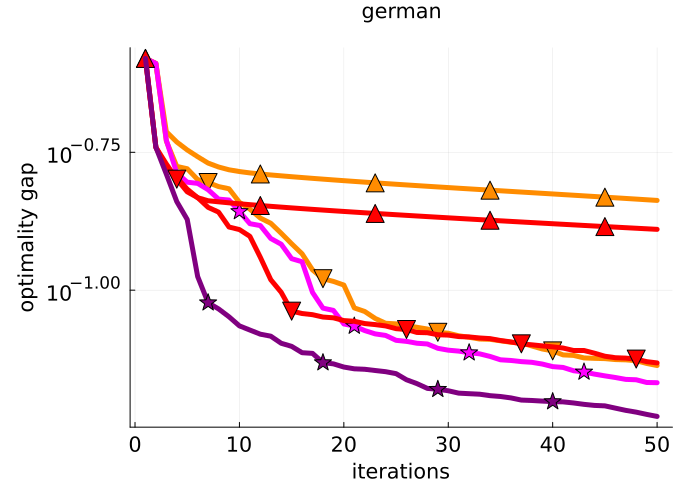}
\includegraphics[width=0.24\textwidth]{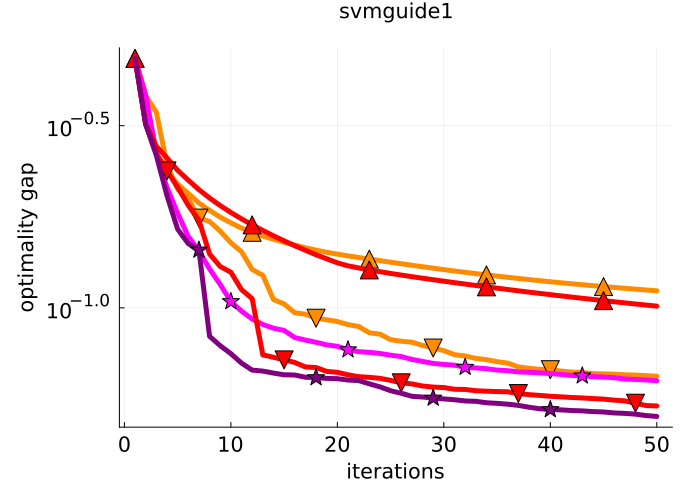}
\includegraphics[width=0.24\textwidth]{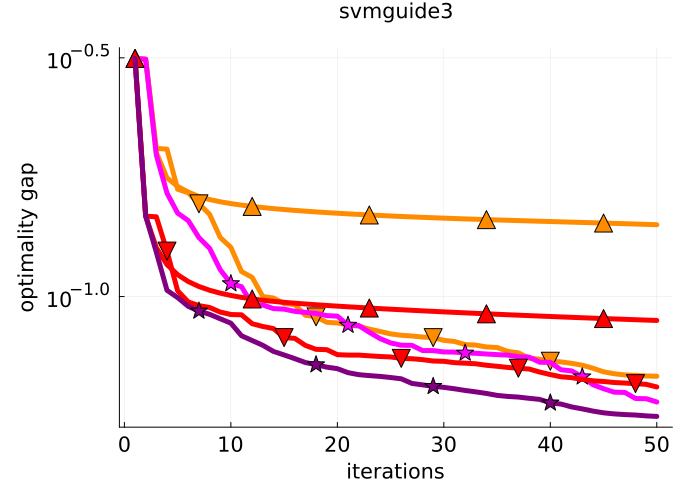}
\includegraphics[width=0.24\textwidth]{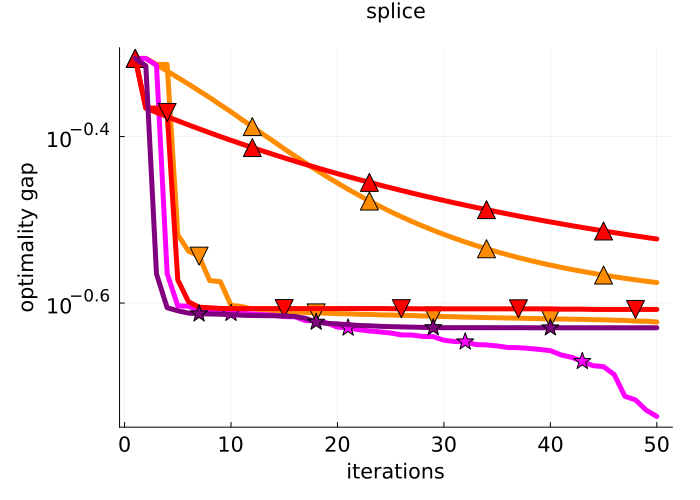}
\includegraphics[width=0.24\textwidth]{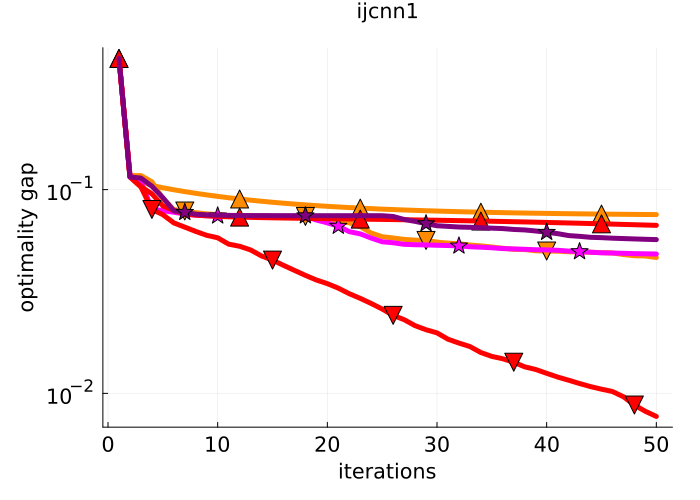}
\includegraphics[width=0.24\textwidth]{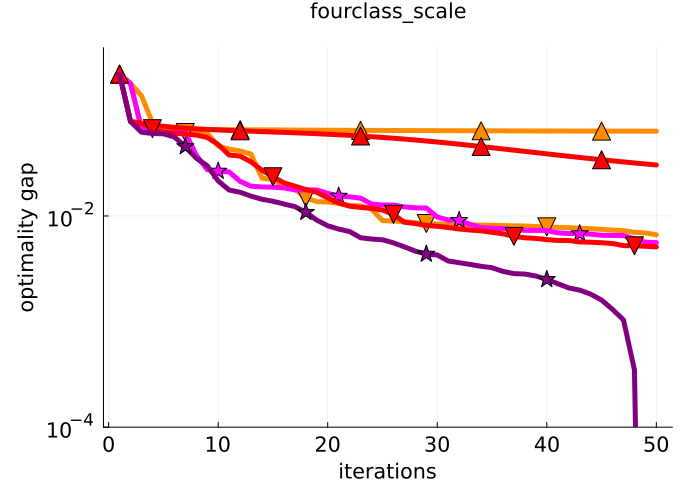}
\includegraphics[width=0.24\textwidth]{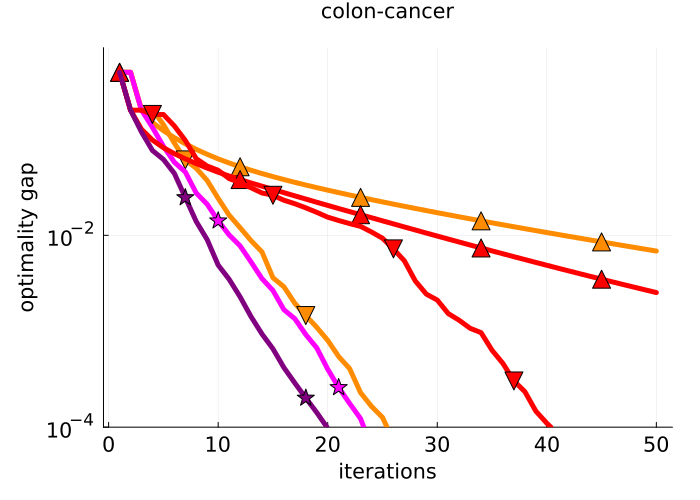}
\includegraphics[width=0.24\textwidth]{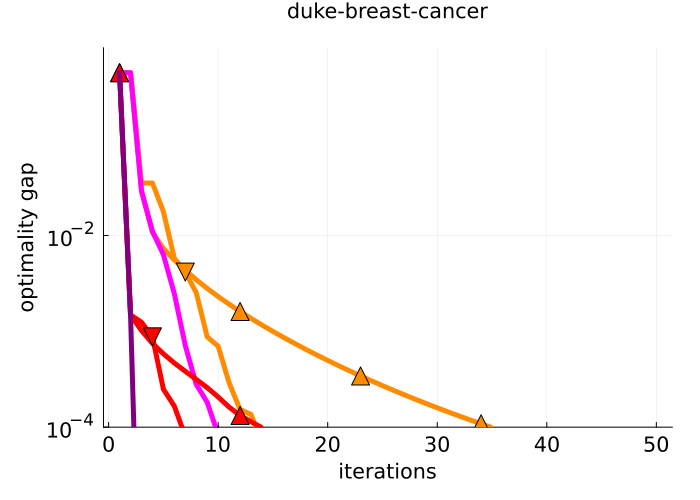}
\includegraphics[width=0.24\textwidth]{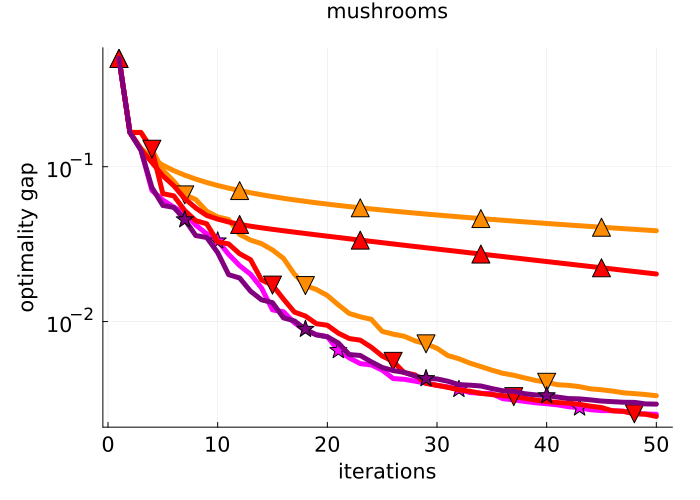}
\includegraphics[width=0.24\textwidth]{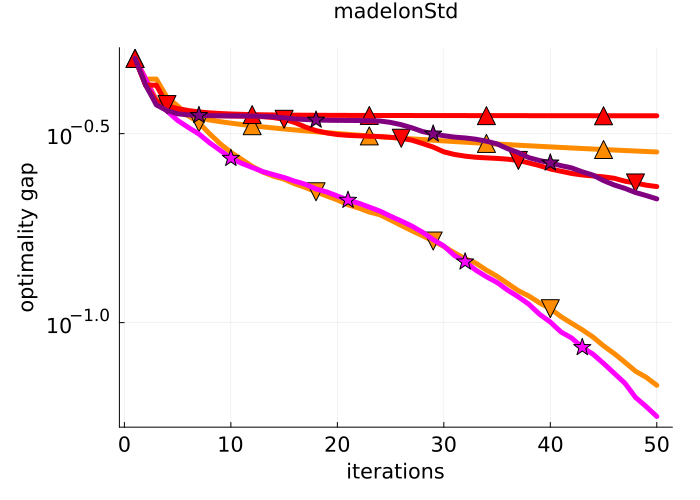}
\includegraphics[width=0.24\textwidth]{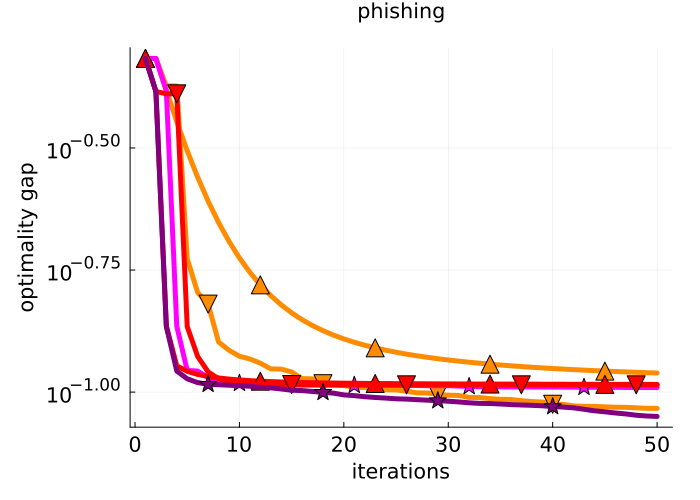}
\includegraphics[width=0.24\textwidth]{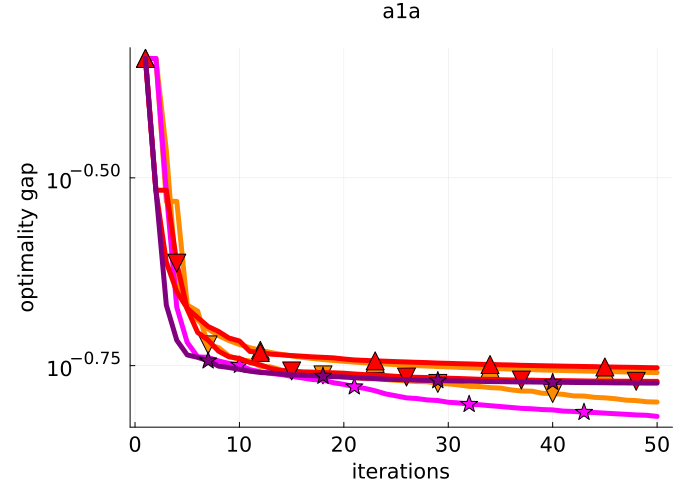}
\includegraphics[width=0.24\textwidth]{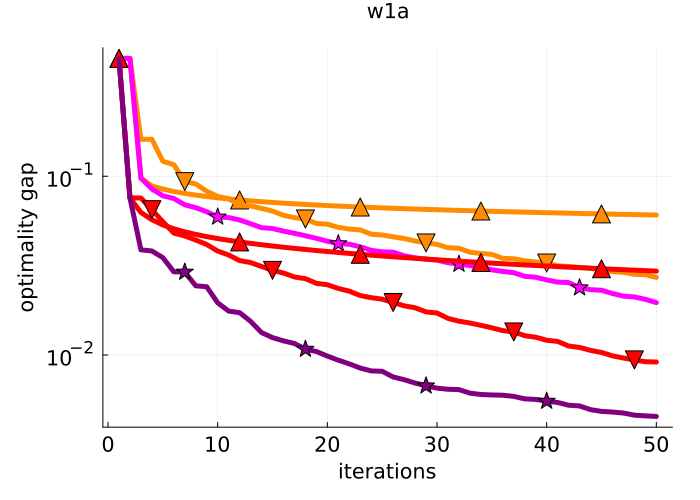}
\includegraphics[width=\textwidth]{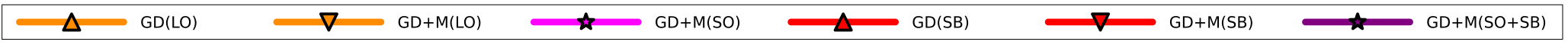}
\caption{Performance of different gradient-based methods for 2-layer neural networks. The red \texttt{GD(SB)} and \texttt{GD+M(SB)} use SO to a step size for each layer while the purple \texttt{GD+M(SO+SB)} method set a learning and momentum rate for each layer. We see that in some cases per-layer step sizes substantially improve performance while in some cases per-layer step sizes harm performance.
}
\label{fig:firstSplitnn100}
\end{figure}

\begin{figure}
\includegraphics[width=0.24\textwidth]{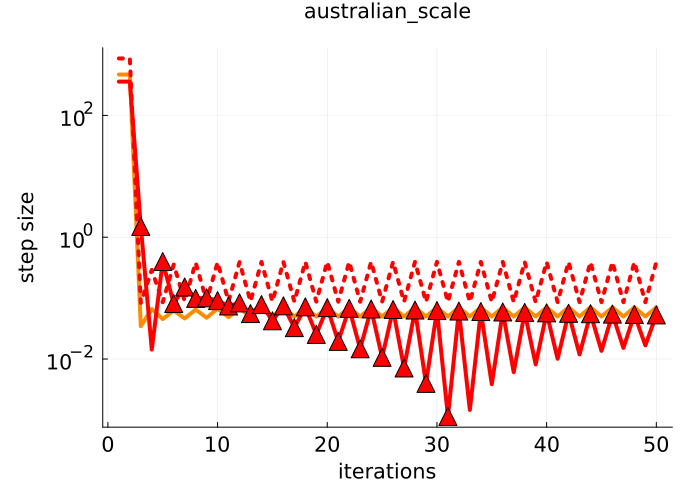}
\includegraphics[width=0.24\textwidth]{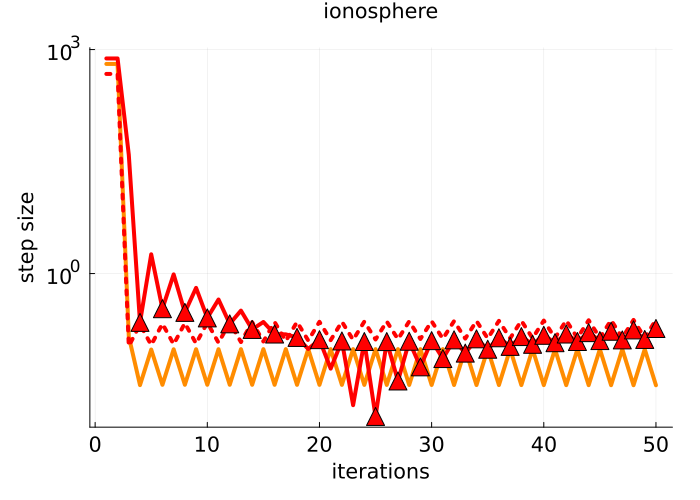}
\includegraphics[width=0.24\textwidth]{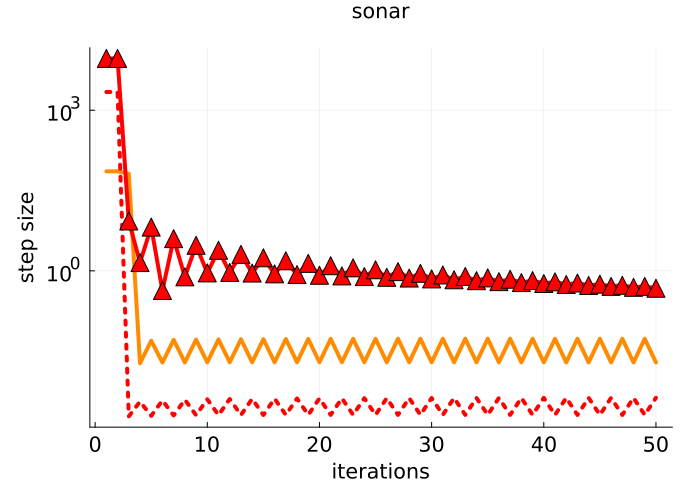}
\includegraphics[width=0.24\textwidth]{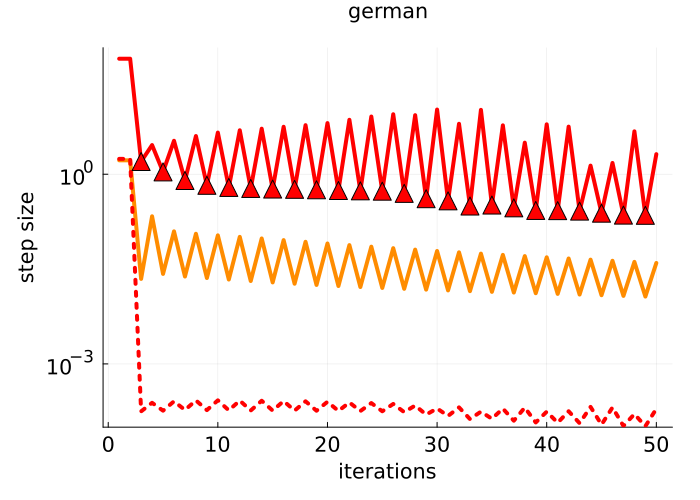}
\includegraphics[width=0.24\textwidth]{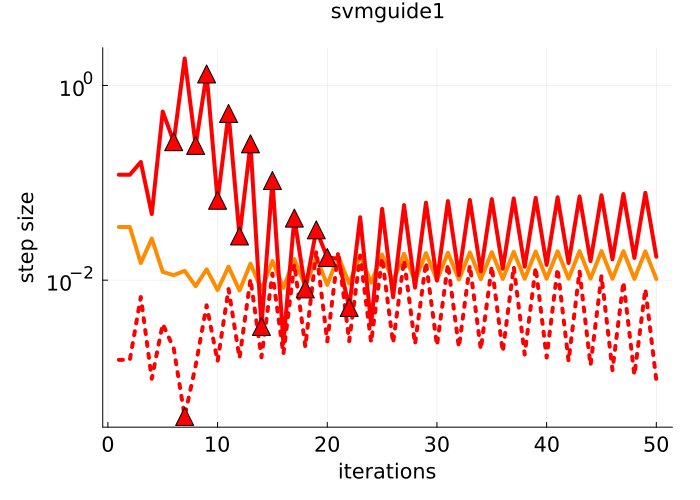}
\includegraphics[width=0.24\textwidth]{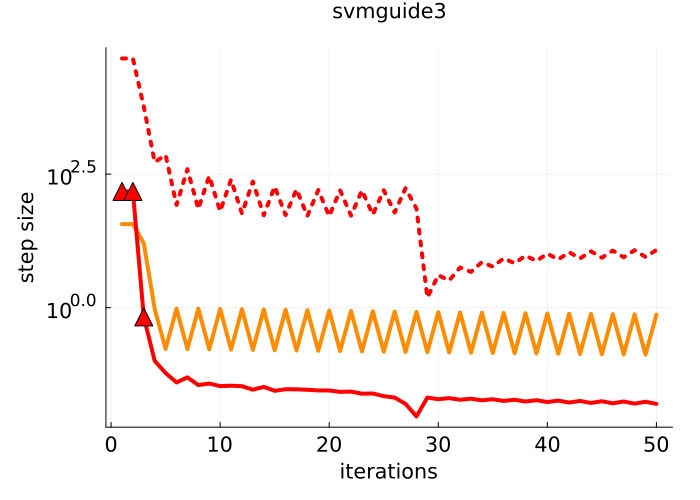}
\includegraphics[width=0.24\textwidth]{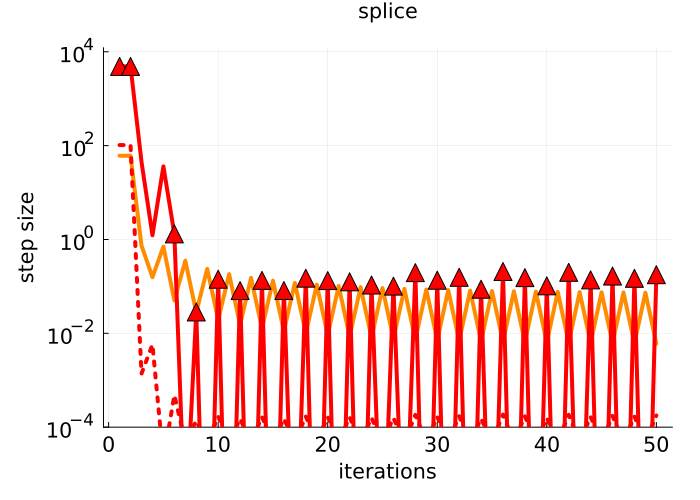}
\includegraphics[width=0.24\textwidth]{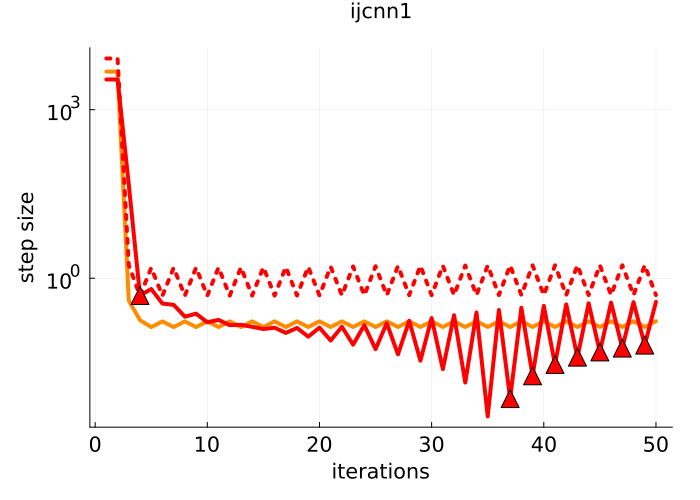}
\includegraphics[width=0.24\textwidth]{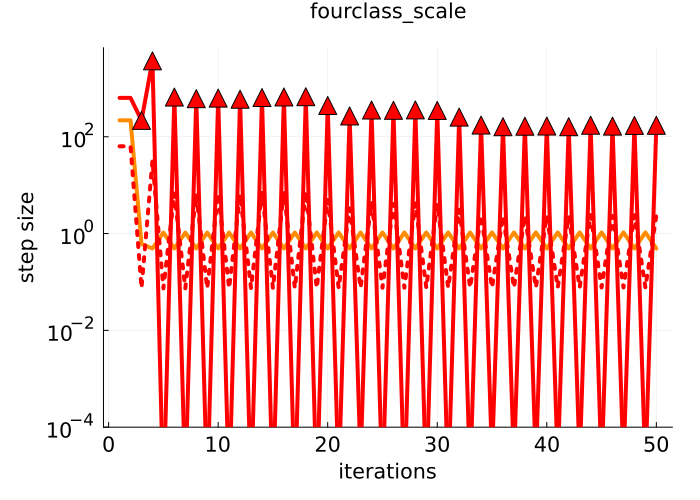}
\includegraphics[width=0.24\textwidth]{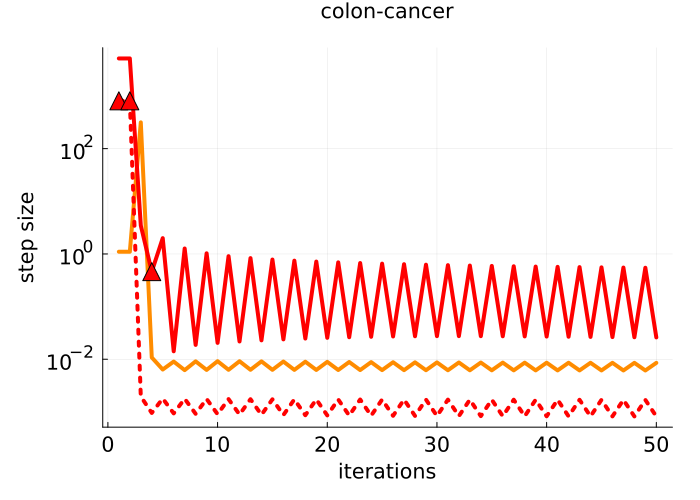}
\includegraphics[width=0.24\textwidth]{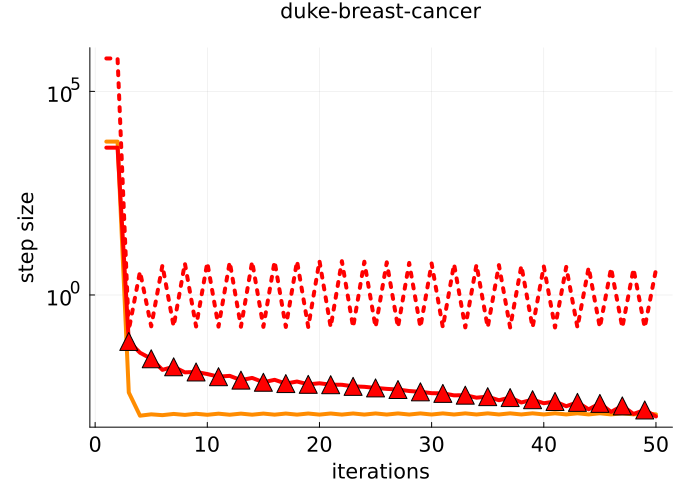}
\includegraphics[width=0.24\textwidth]{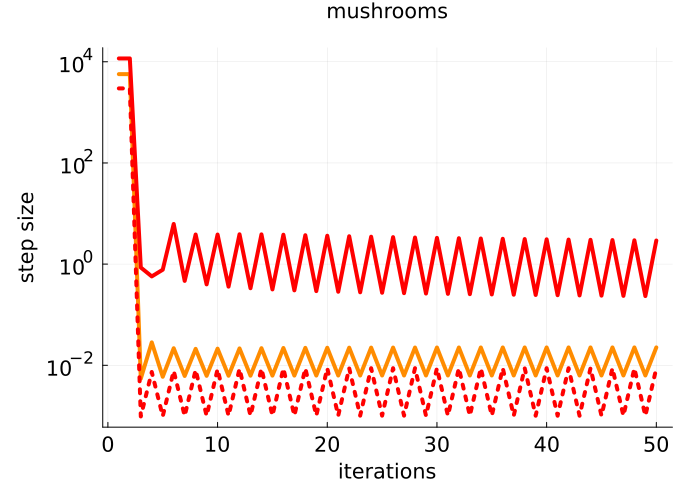}
\includegraphics[width=0.24\textwidth]{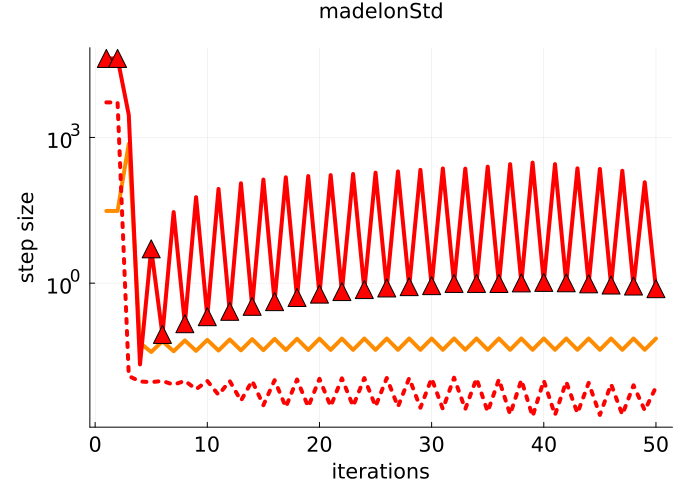}
\includegraphics[width=0.24\textwidth]{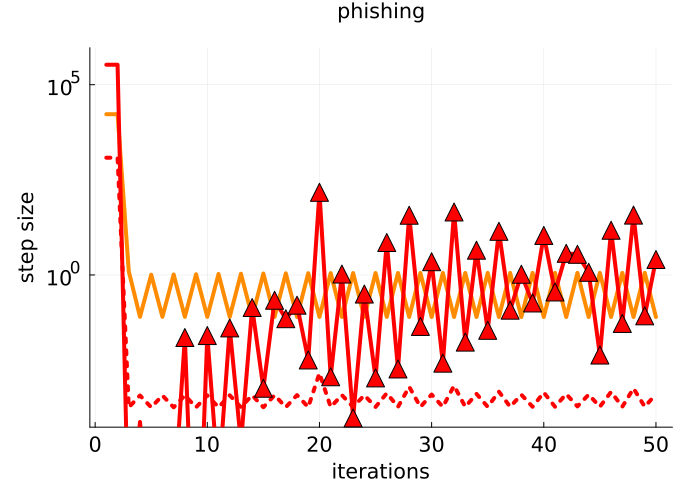}
\includegraphics[width=0.24\textwidth]{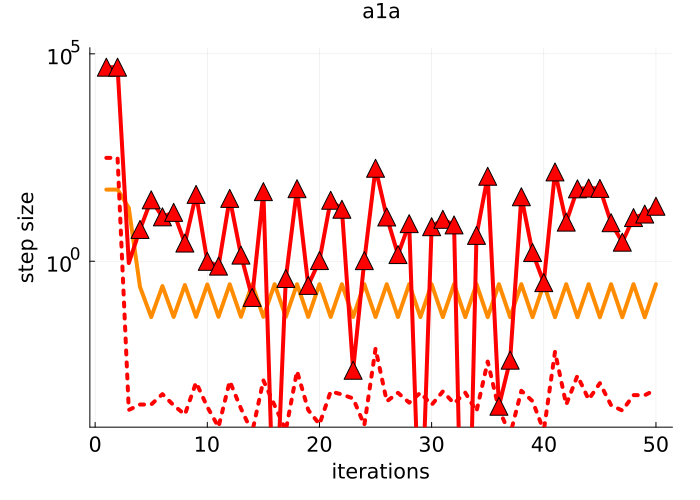}
\includegraphics[width=0.24\textwidth]{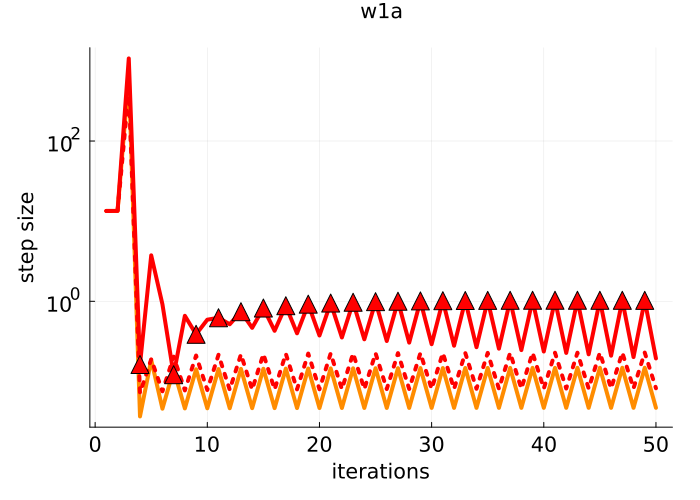}
\begin{center}
\includegraphics[width=.32\textwidth]{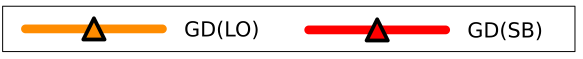}
\end{center}
\caption{Step sizes of gradient descent methods for fitting 2-layer neural networks, optimizing a single step size for both layers (\texttt{GD(LO)}) or optimizing a step size for each layer (\texttt{GD(SB)}). For the per-layer method, the solid line is for the output layer and the dashed line is for the input layer. We plot the absolute value, and include markers to indicate iterations where the step was negative. We see that both methods tend to have cyclic behaviour, we see that the per-layer method tends to use larger step sizes (for for at least one of the layers) and tends to have larger oscillations. We also see that the output layer often uses negative step sizes, in some cases using a negative step size on every iteration but in many cases using a negative step size on every second iteration.
}
\label{fig:GDLOSBstepSizePlotnn100}
\end{figure}

Figure~\ref{fig:firstSplitnn100} shows the performance of these methods with per-layer step sizes and the corresponding methods with step sizes tied across layers. In these plots we observed two conflicting behaviours:
\begin{itemize}
\item On some datasets using per-layer step sizes significantly improves performance.
\item On some datasets using per-layer step sizes significantly harms performance.
\end{itemize}
Unfortunately, based on this performance it does not appear that per-layer step sizes are a reliable option if we want a method that is free of hyper-parameters. A potential source of the poor performance of per-layer step sizes  is that the optimized per-layer step sizes are often very large. We compare the learning rates for gradient descent with optimized tied and per-layer learning rates in Figure~\ref{fig:GDLOSBstepSizePlotnn100}, where we see that the per-layer rates are often much larger for at least one of the layers and have larger oscaillations (sometimes oscillating between positive and negative step sizes). These large step sizes lead to a large initial reduction in the objective, but they make the norm of the parameters much larger than with tied step sizes. This large norm seems to be the cause of the slow progress on later iterations for some datasets.

\begin{figure}
\includegraphics[width=0.24\textwidth]{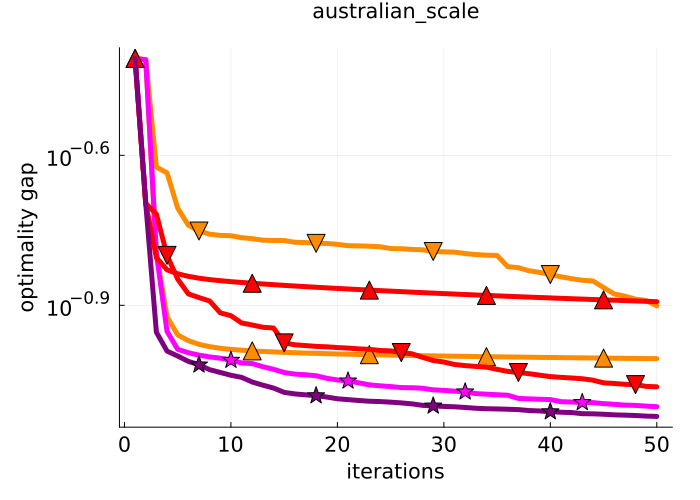}
\includegraphics[width=0.24\textwidth]{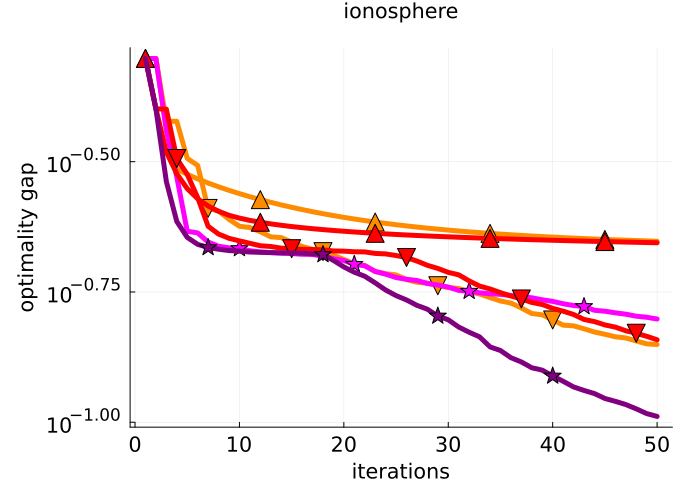}
\includegraphics[width=0.24\textwidth]{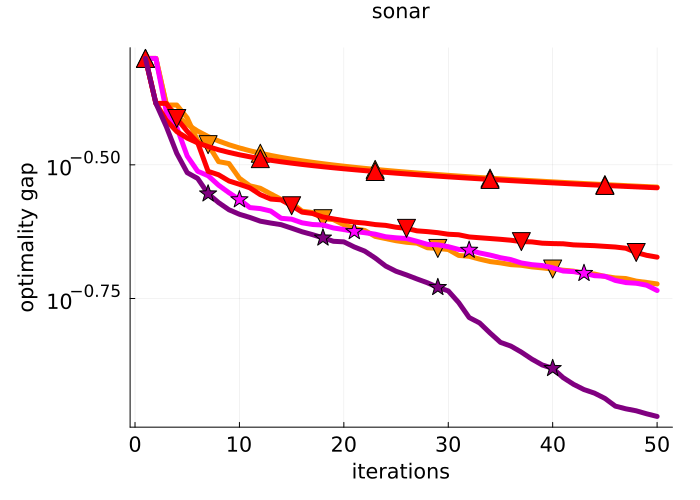}
\includegraphics[width=0.24\textwidth]{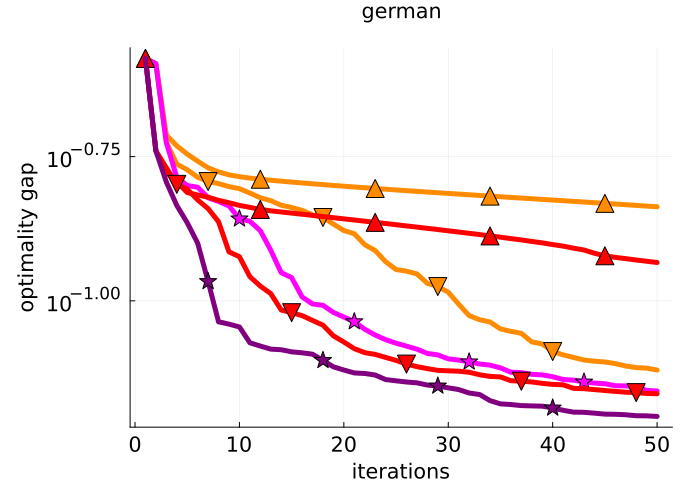}
\includegraphics[width=0.24\textwidth]{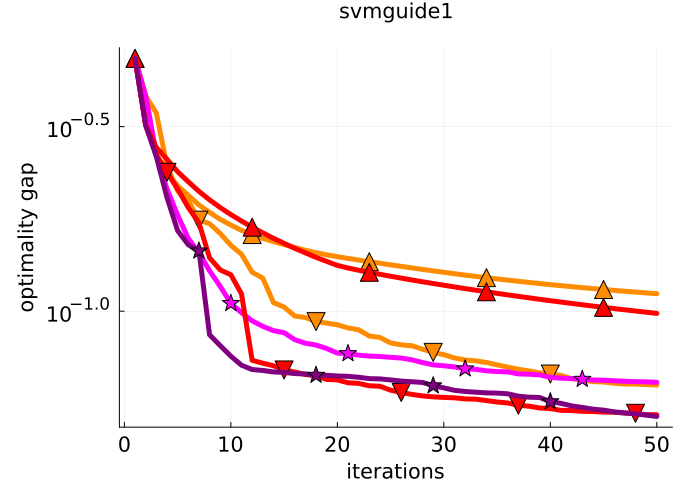}
\includegraphics[width=0.24\textwidth]{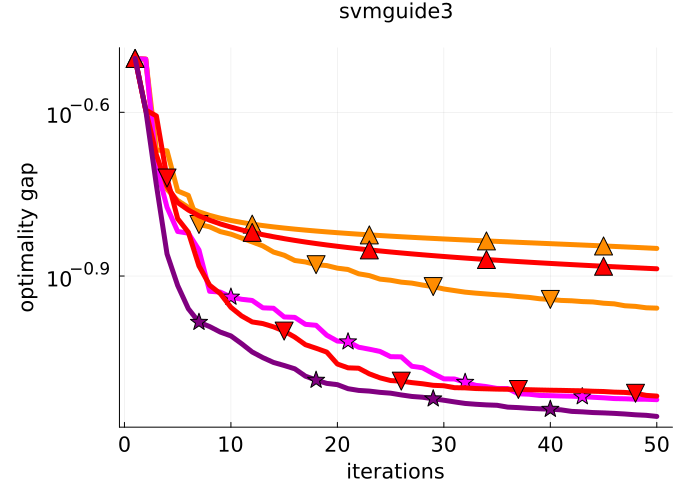}
\includegraphics[width=0.24\textwidth]{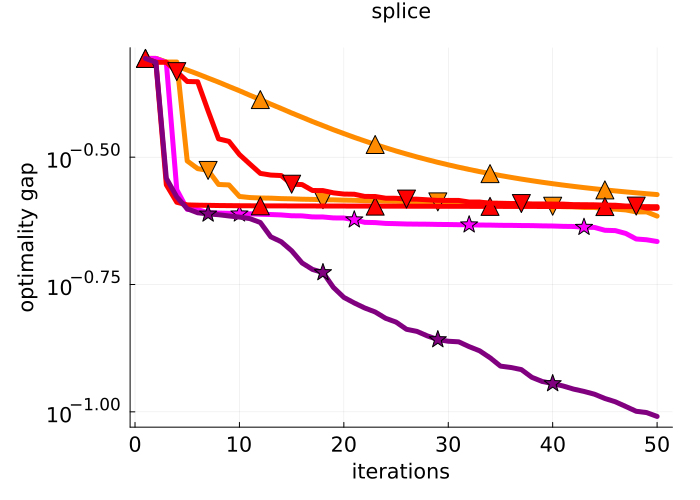}
\includegraphics[width=0.24\textwidth]{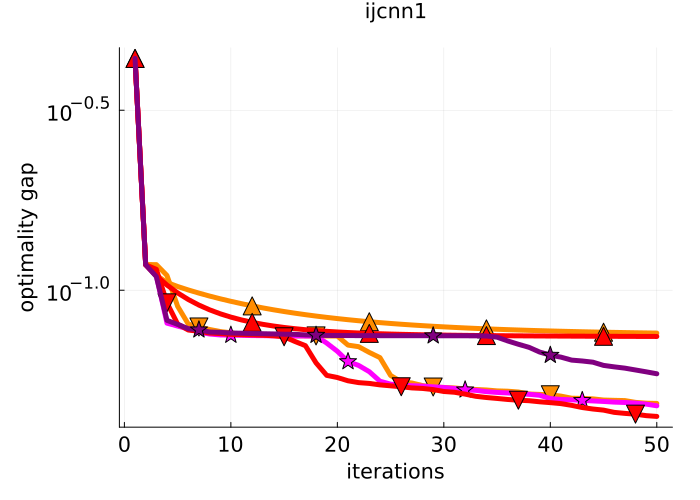}
\includegraphics[width=0.24\textwidth]{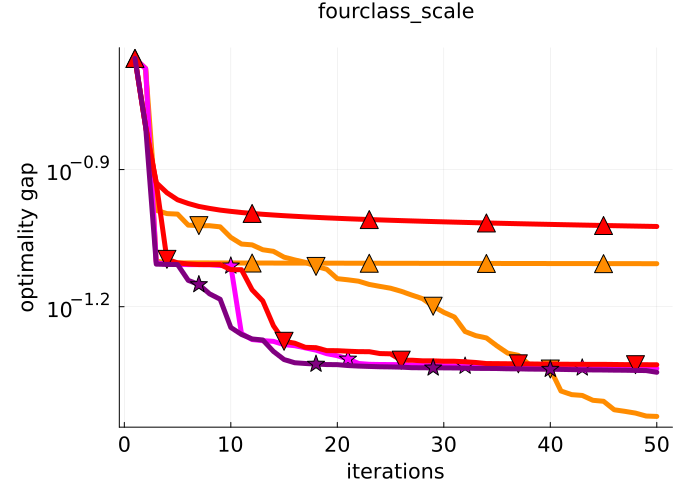}
\includegraphics[width=0.24\textwidth]{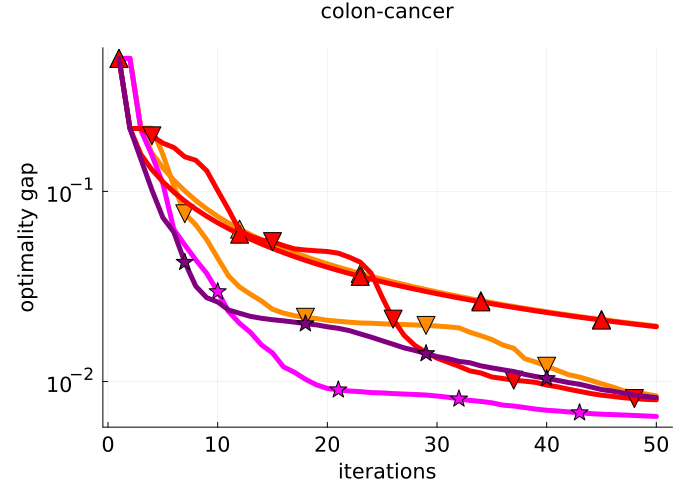}
\includegraphics[width=0.24\textwidth]{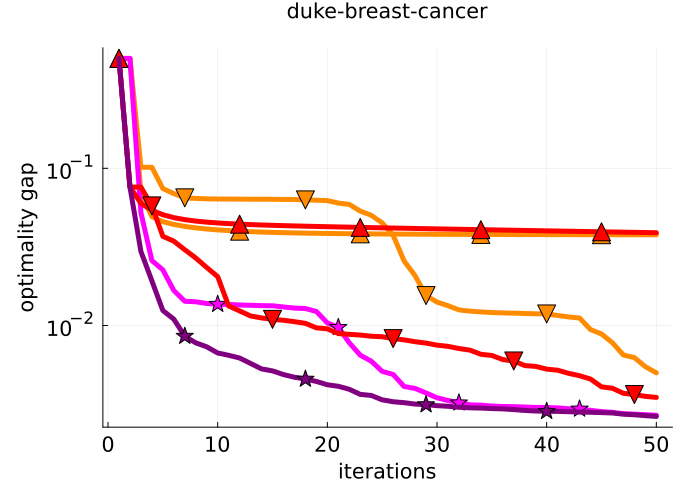}
\includegraphics[width=0.24\textwidth]{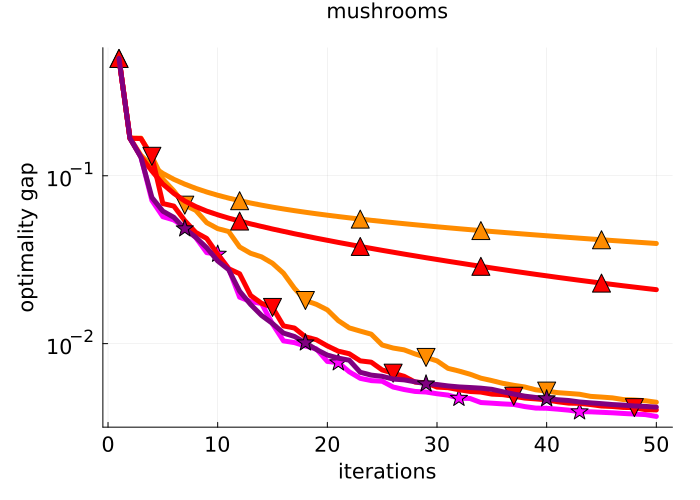}
\includegraphics[width=0.24\textwidth]{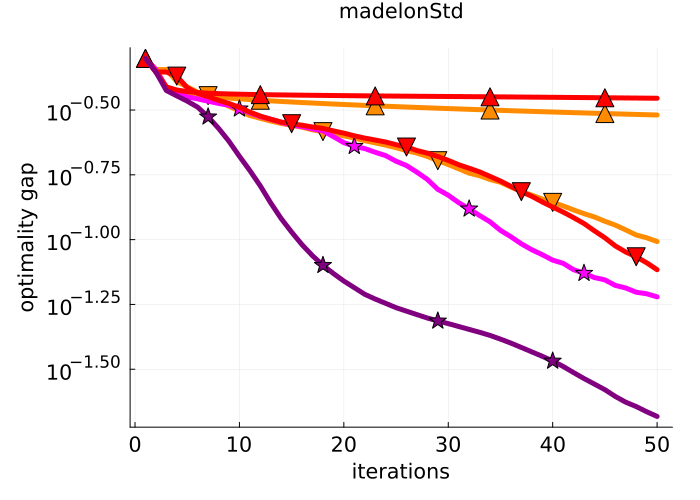}
\includegraphics[width=0.24\textwidth]{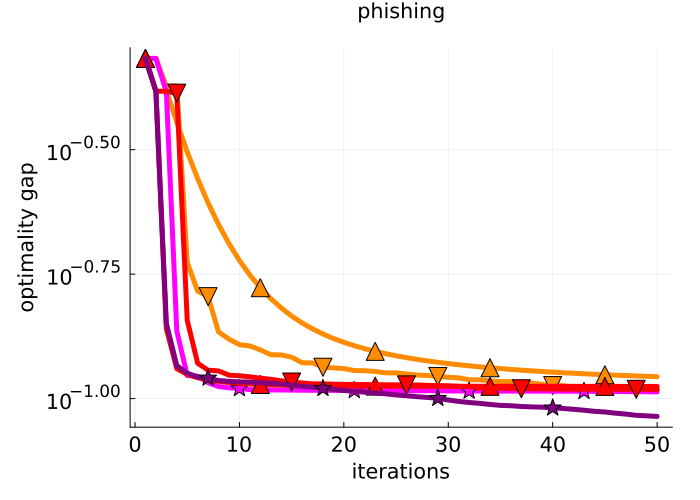}
\includegraphics[width=0.24\textwidth]{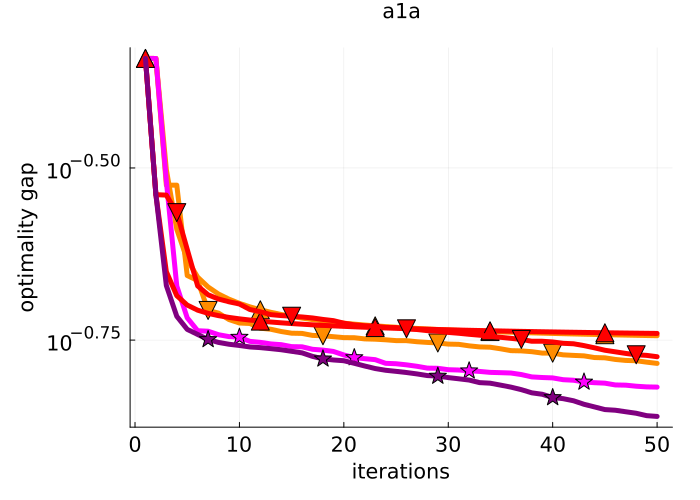}
\includegraphics[width=0.24\textwidth]{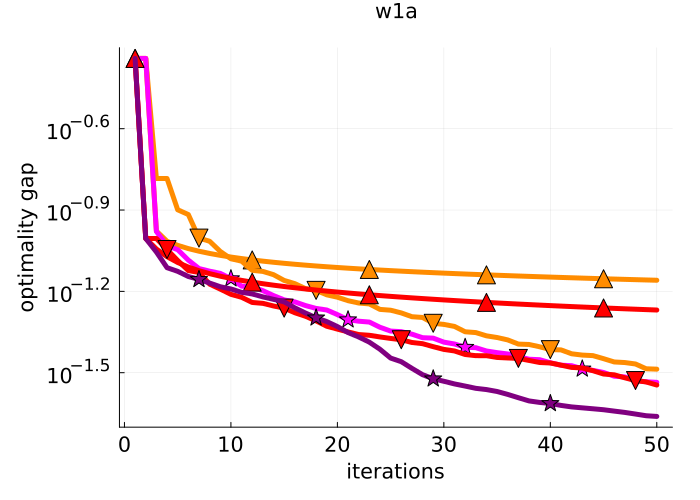}
\includegraphics[width=\textwidth]{firstSplitLegend.png}
\caption{Performance of different gradient-based methods for regularized 2-layer neural networks. The red \texttt{GD(SB)} and \texttt{GD+M(SB)} use SO to a step size for each layer while the purple \texttt{GD+M(SO+SB)} method set a learning and momentum rate for each layer. We see that per-layer step sizes consistently improve performance.
}
\label{fig:firstSplitregnn100}
\end{figure}

\begin{figure}
\includegraphics[width=0.24\textwidth]{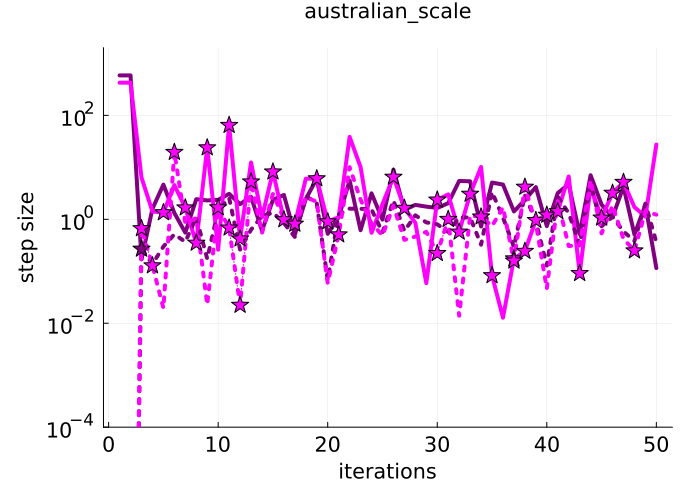}
\includegraphics[width=0.24\textwidth]{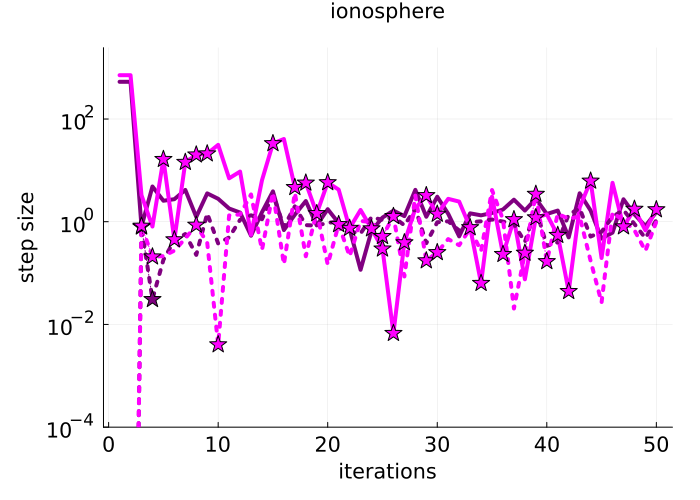}
\includegraphics[width=0.24\textwidth]{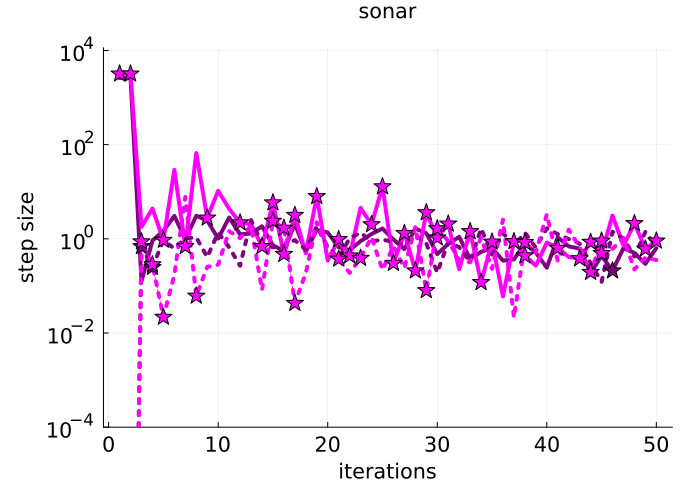}
\includegraphics[width=0.24\textwidth]{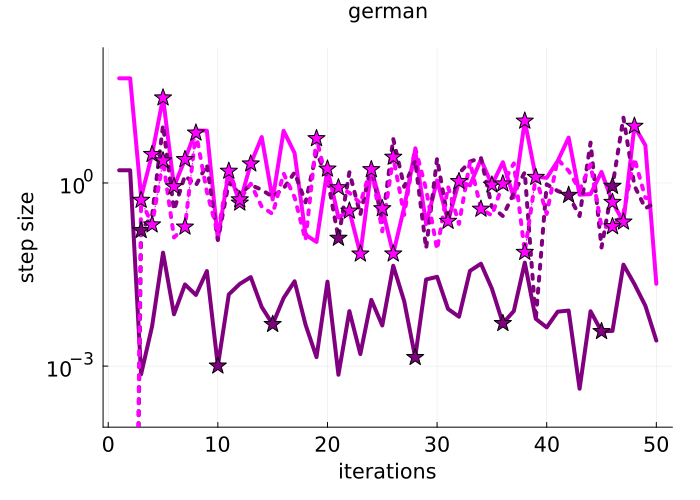}
\includegraphics[width=0.24\textwidth]{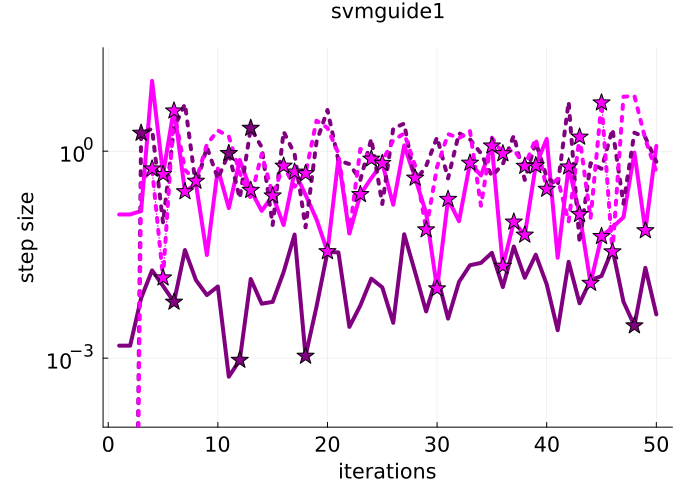}
\includegraphics[width=0.24\textwidth]{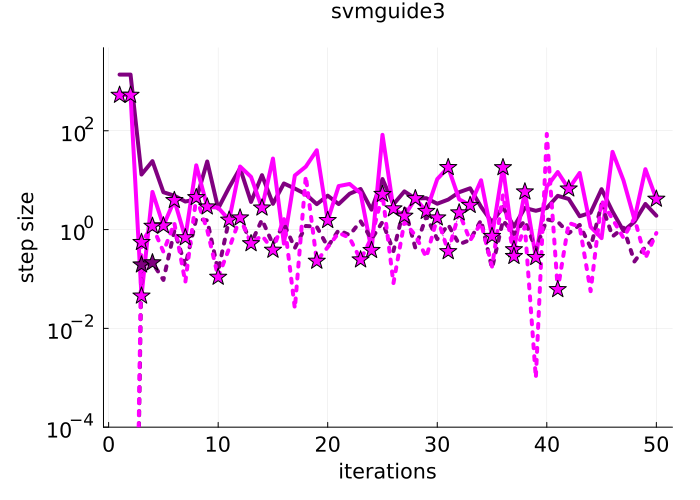}
\includegraphics[width=0.24\textwidth]{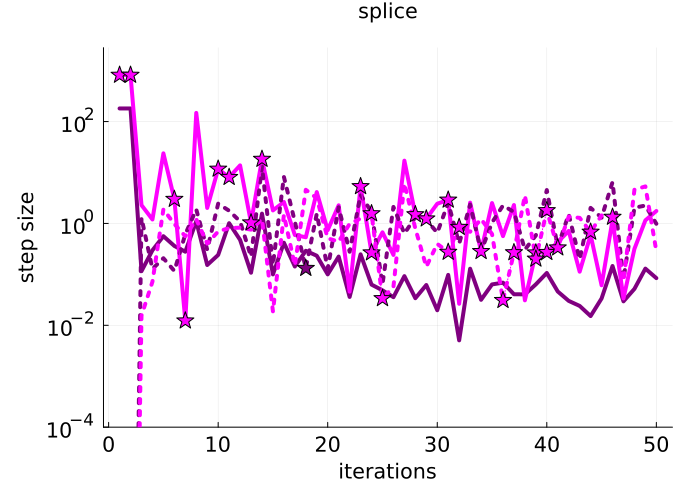}
\includegraphics[width=0.24\textwidth]{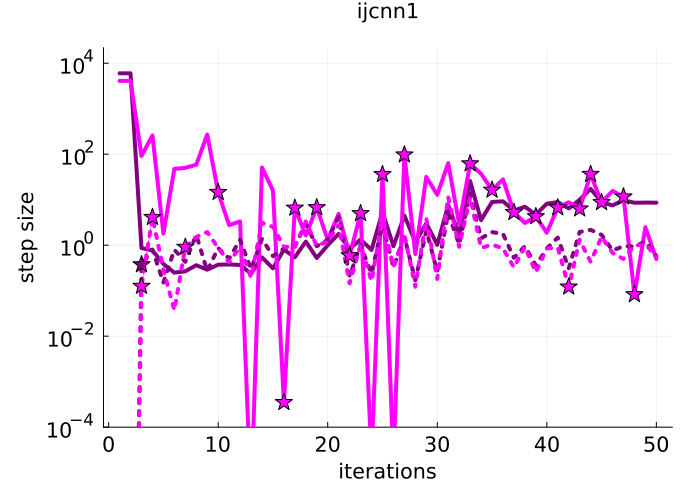}
\includegraphics[width=0.24\textwidth]{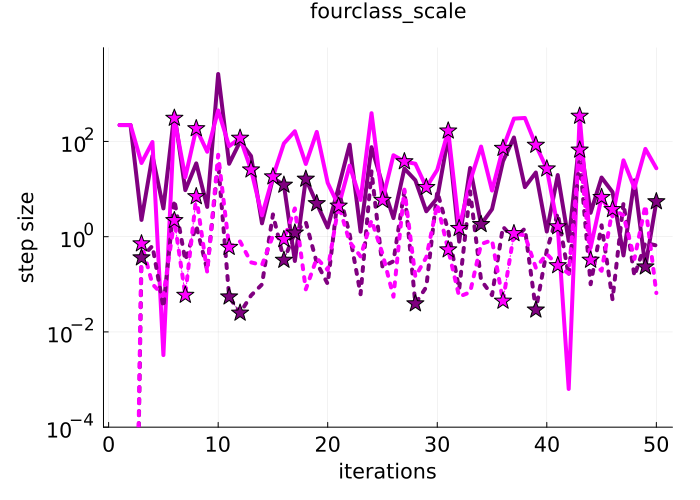}
\includegraphics[width=0.24\textwidth]{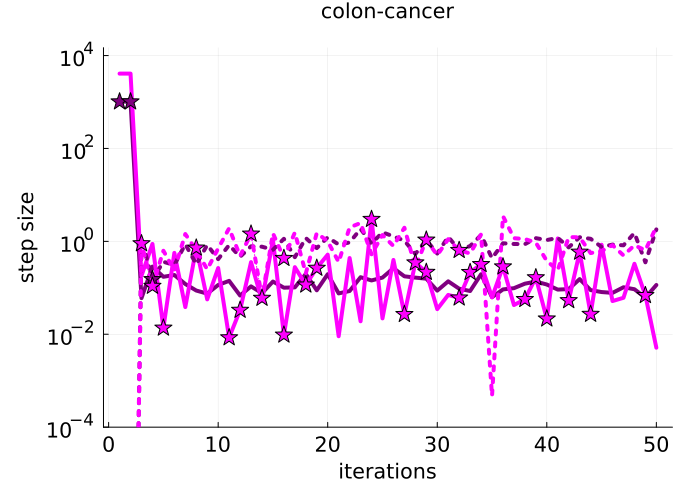}
\includegraphics[width=0.24\textwidth]{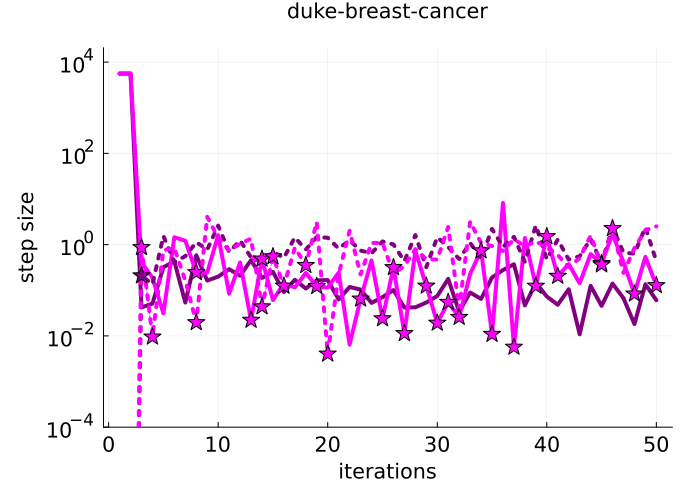}
\includegraphics[width=0.24\textwidth]{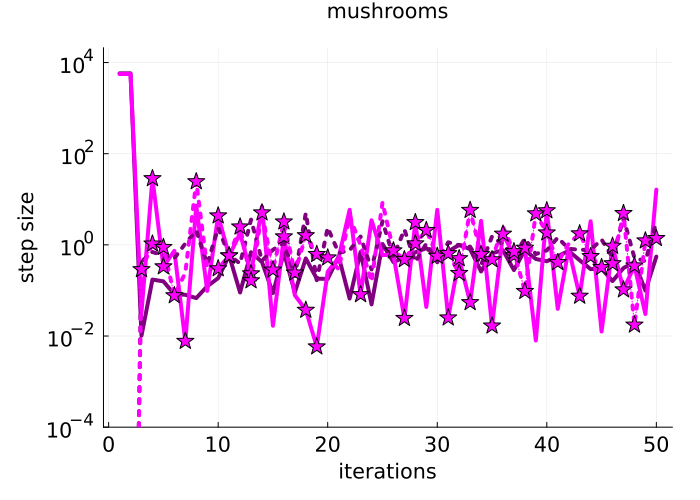}
\includegraphics[width=0.24\textwidth]{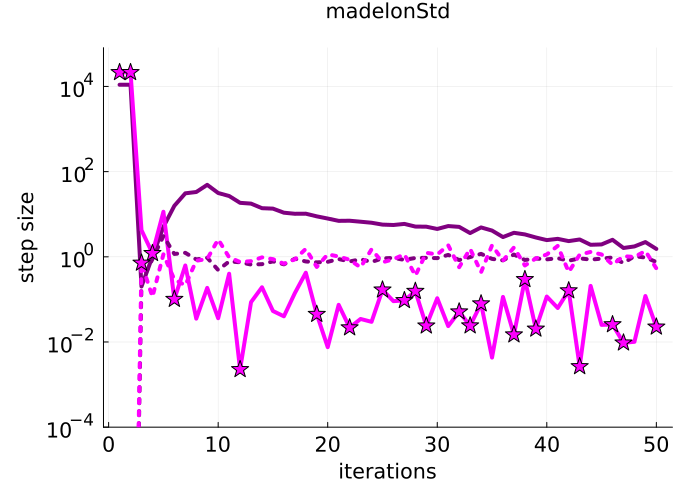}
\includegraphics[width=0.24\textwidth]{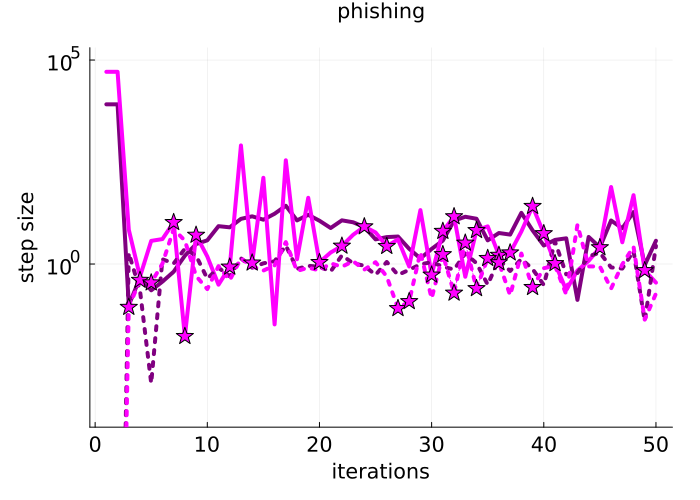}
\includegraphics[width=0.24\textwidth]{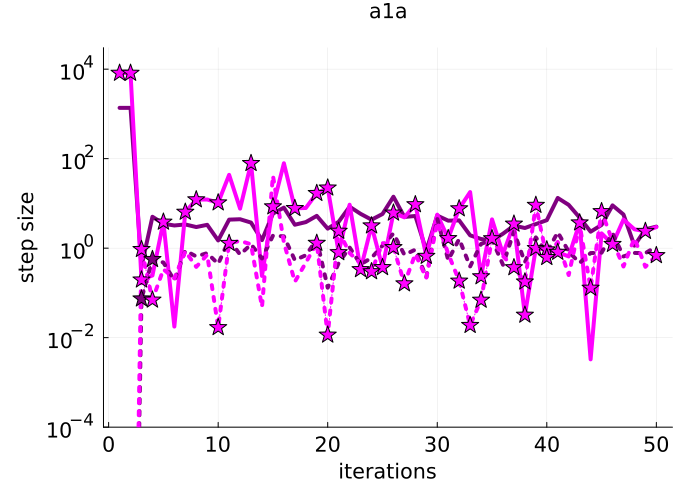}
\includegraphics[width=0.24\textwidth]{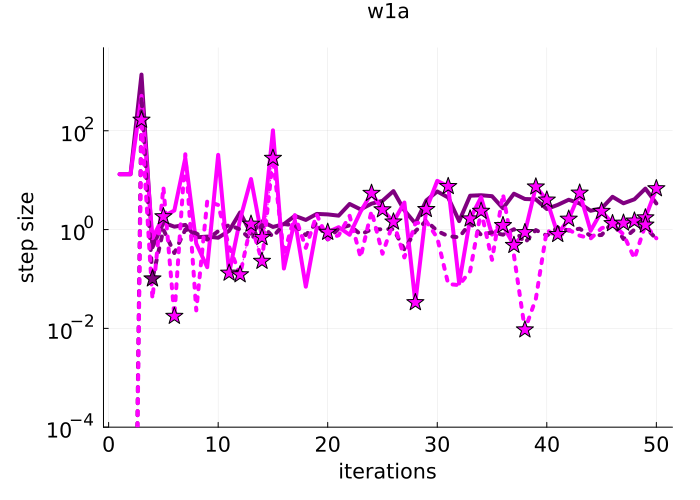}
\begin{center}
\includegraphics[width=.32\textwidth]{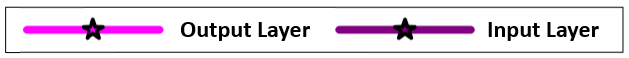}
\end{center}
\caption{Step sizes of gradient descent methods for fitting 2-layer neural networks, optimizing a learning rate and momentum rate for each layer (\texttt{GD+M(SO+SB)}). The solid lines are the learning rates and the dashed lines are the momentum rates. We plot the absolute value, and include markers to indicate iterations where the step was negative.
}
\label{fig:GDSOSBstepSizePlotregnn100}
\end{figure}

To explore whether controlling the norm of the parameters would make per-layer step sizes viable, we repeated this experiment using L2-regularization of the parameters. In particular, we add to the objective a term of the form $(\lambda/2)(\norm{W}_F^2 + \norm{v}^2)$ where we set $\lambda = 1/n$. Figure~\ref{fig:firstSplitregnn100} shows the performance of the same methods on the regularized objective. In this plot, we see that per-layer step sizes typically lead to a performance improvement (and in many cases a large improvement). We plot the four learning and momentum rates used by best-performing method (\texttt{GD+M(SO+SB)}) in Figure~\ref{fig:GDSOSBstepSizePlotregnn100}, showing the wide range of per-layer learning and momentum rates that lead to this strong performance.
It is possible that other strategies for controlling the scale of the parameters, such as trust region methods, would also allow per-layer step sizes to be effective if we do not want to use a regularized objective.

\subsection{Other Examples of SO-Friendly Networks}

Above we consider fully-connected 2-layer neural networks with a single output. But there exists a variety of other settings where neural networks are SO-friendly. Here we give three examples.

{\bf Fully-connected 2-layer networks with a small number of outputs}: if we have a 2-layer network with $c$ outputs, the  $r \times 1$ vector $v$ is replaced by a $r \times c$ matrix $V$ and the the cost of forward and backward propagation through the last layer of weights changes from $O(r)$ to $O(rc)$ for each of the $n$ examples. In order to remain SO efficient, the $O(nrc)$ cost associated with applying the last layer must be dominated by the $O(ndr)$ cost of applying the first layer. Thus, 2-layer networks with multiple outputs  are SO-friendly provided that $c\ll d$, and we can use SO in cases where the number of outputs is significantly smaller than the number of inputs. Note that such situations arise in common ML datasets such as MNIST~\citep{lecun1998gradient} which has 784 inputs and 10 outputs, CIFAR-10~\citep{krizhevsky2009learning} which has1024 inputs and 10 outputs, and ImageNet~\citep{deng2009imagenet} which has 181,503 inputs for an average image and 1,000 outputs. Note that if we fix the weights in the first layer of a 2-layer network, then the optimization problem becomes an LCP. Thus, even if the number of labels is large it could still make sense to do SO on only the second layer on a subset of the iterations.

\begin{figure}
\centering
\includegraphics[width=.7\textwidth]{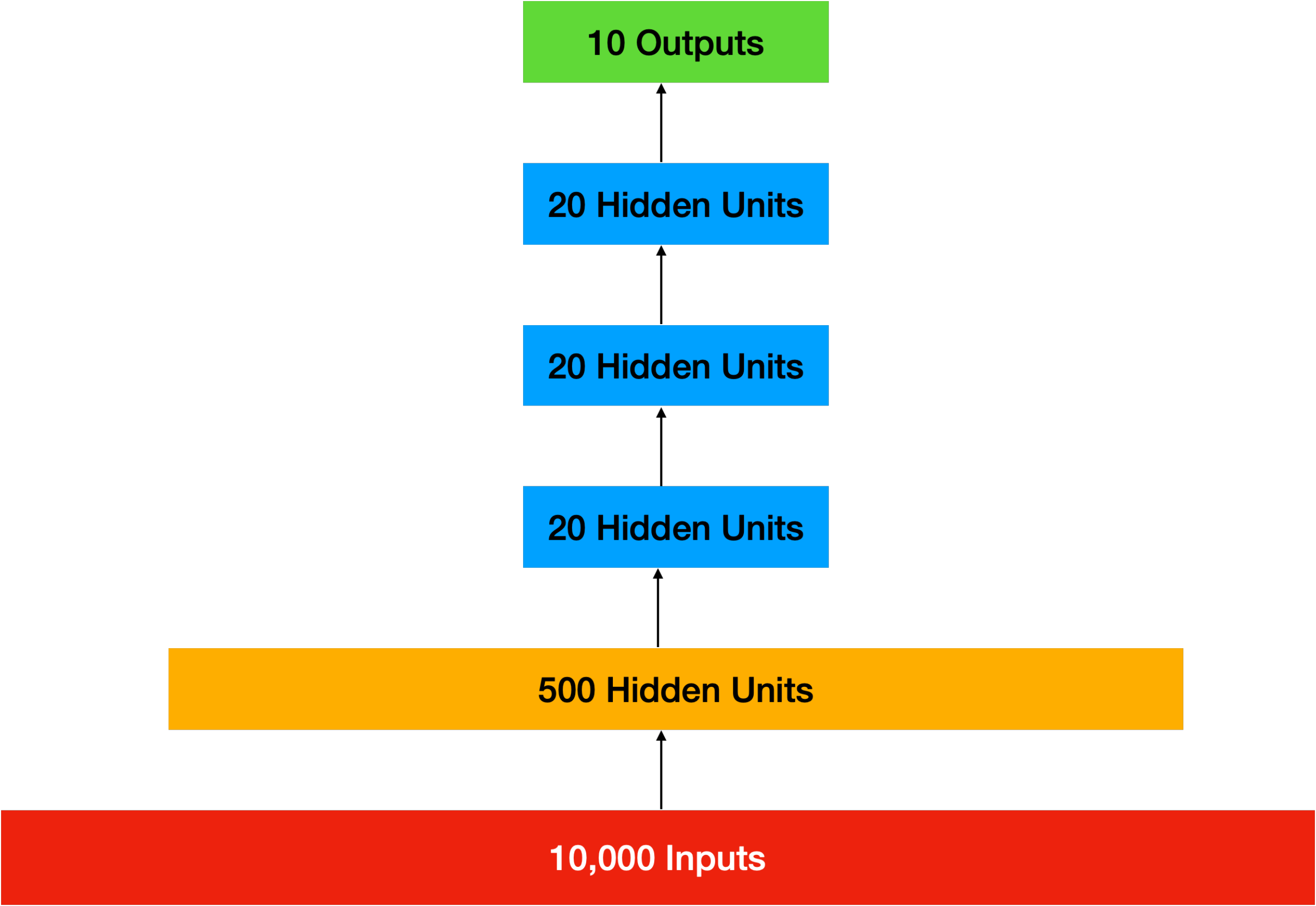}
\caption{Example structure of a 5-layer SO-friendly neural network ($d = 10000, r = 500, \rho = 20, \ell = 5$). For the same asymptotic cost as using a fixed learning rate and momentum rate, we could numerically optimize a learning rate and momentum rate for each of the 5 layers on each step (10 step sizes).}
\label{fig:SO-friendly}
\end{figure}

{\bf Wide-then-narrow deep networks}: consider a fully-connected network with $\ell$ layers where the first layer has $r$ outputs and the remaining layers have at most $\rho$ outputs. In this setting the cost of applying the network to all examples is $O(ndr + nr\rho + n\rho^2\ell)$. Such networks are SO-friendly if we have $\rho\ll d$ and $\rho\ell\ll r$. Such networks must have a large number of inputs and can have a large number of hidden units in the first layer, but must have a relatively smaller number of units in subsequent layers. Thus, they can perform a large-dimensional linear transformation followed by a number of lower-dimensional non-linear transforms. We give an example of a 5-layer network that is SO-friendly in Figure~\ref{fig:SO-friendly}.
While being SO-friendly limits the possible structures available, note that this structure allows efficient per-layer optimization of the learning and momentum rates for all the layers.

{\bf Convolutional neural networks (CNNs) with large strides and/or pooling operators}: convolutional neural networks are a popular variant of neural networks that include convolution and pooling layers~\citep[see][]{lecun1998gradient,krizhevsky2012imagenet}. Convolution layers may include a ``stride'' variable that controls the size of the output of a layer compared to the input, and pooling layers often also result in a dimensionality reduction. If a CNN is setup with large-enough strides or pooling areas, it is possible to make SO-friendly CNNs where applying the first layer is the dominant cost.

{\bf Non-linear neural tangent kernels}: the NTK approach to training deep neural network optimizes a linearized approximation to the network~\citep{Jacot2018}. This linearization allows us to efficiently use SO via the linear CG method if we use the squared loss function. However, the linearization may lose important information about the non-linearity present in the network. We could alternately consider training with a linearization of only the first $l$ layers, and maintaining the non-linearity in the deep layers. If $l$ is chosen large enough, the network will be SO-friendly (with the standard NTK corresponding to setting $l$ to the number of layers). This would give a better approximation than the NTK, and would allow optimizing shared step sizes for the first $l$ layers and then separate step sizes for the layers beyond layer $l$.

\section{Augmenting Quasi-Newton and Adam with Subspace Optimization}
\label{sec:algorithms}

The previous sections considered using SO to set step sizes within the GD+M update. However, SO is also suitable for many of the common variants on this update. In this section, we discuss using SO to improve on the performance of two of the most empirically effective methods.

\subsection{Augmenting Quasi-Newton with Subspace Optimization}
\label{sec:QN}

Quasi-Newton updates can be written in the form
\begin{equation}
	\label{eq:scaledGD}
	w_{k+1} = w_k - \alpha_kB_k\nabla f(w_k),
\end{equation}
where the learning rate is chosen using the strong Wolfe conditions, and $B_k$ is chosen to satisfy a variant of the secant equations, $B_k(\nabla f(w_k) - \nabla f(w_{k-1})) = (w_k - w_{k-1})$. One of the most successful quasi-Newton methods is the L-BFGS method, which does not explicitly store $B_k$ but implicitly forms the matrix based on the $p$ most-recent values of $(w_k-w_{k-1})$ and $(\nabla f(w_k) -\nabla f(w_{k-1}))$~\citep{Nocedal1980}. The L-BFGS method only requires $O(dp)$ memory and allows multiplications with $B_k$ to be performed in $O(dp)$,  so with small $p$ the L-BFGS update has a similar cost to the GD+M update. But  in practice the approximate second-order information in $B_k$ often allows the L-BFGS update to make substantially more progress per iteration. For precise details on the method, see~\citet[][Chapters~6-7]{Nocedal2006}. The computational savings for LO in LCPs and SO-friendly networks can also be used with updates of the form~\eqref{eq:scaledGD},
\begin{align*}
	& \argmin_{\alpha}f(w_k - \alpha B_kX^T\nabla g(Xw_k))\\
	\equiv & \argmin_{\alpha}g(X(w_k - \alpha B_kX^T\nabla g(Xw_k)))\\
	\equiv & \argmin_{\alpha}g(m_k - \alpha \underbrace{X(B_kX^T\nabla g(m_k))}_{d_k})\\
	\equiv & \argmin_{\alpha}g(\underbrace{(1+\beta)m_k - \alpha d_k)}_\text{potential $m_{k+1}$}).
\end{align*}

We can consider adding directions to quasi-Newton methods and using SO to set the corresponding step sizes. A natural update to consider is a hybrid with the gradient method and quasi-Newton method
\[
w_{k+1} = w_k - \alpha_k^1\nabla f(w_k) - \alpha_k^2B_k\nabla f(w_k).
\]
This method would guarantee at least as much progress as gradient descent on each iteration, which could lead to more progress than a quasi-Newton update if the matrix $B_k$ is badly scaled. However, this update would require 3 matrix multiplications with $X$ for LCPs and SO-friendly neural networks. Thus, on large problems adding the gradient direction would increase the iteration cost by around 50\% compared to the standard quasi-Newton method.

Alternately, we could consider adding a momentum term to the quasi-Newton update,
\begin{equation}
	\label{eq:scaledGDM}
	w_{k+1} = w_k - \alpha_kB_k\nabla f(w_k) + \beta_k(w_k - w_{k-1}).
\end{equation}
The computational savings for SO in LCPs and SO-frinedly networks are preserved for this ``quasi-Newton with momentum'' method. In the case of LCPs solving for $\alpha_k$ and $\beta_k$ corresponds to solving
\begin{align*}
	& \argmin_{\alpha,\beta}f(w_k - \alpha B_kX^T\nabla g(Xw_k) + \beta(w_k - w_{k-1}))\\
	\equiv & \argmin_{\alpha,\beta}g(X(w_k - \alpha B_kX^T\nabla g(Xw_k) + \beta(w_k - w_{k-1})))\\
	\equiv & \argmin_{\alpha,\beta}g(m_k - \alpha \underbrace{X(B_kX^T\nabla g(m_k))}_{d_k} + \beta (m_k - m_{k-1}))\\
	\equiv & \argmin_{\alpha,\beta}g(\underbrace{(1+\beta)m_k - \alpha d_k + \beta m_{k-1}}_\text{potential $m_{k+1}$}),
\end{align*}
where $m_k = Xw_k$ and $d_k=X(B_kX^T\nabla g(m_k))$. Similar to the GD+M update this only requires 2 matrix multiplications with $X$ per iteration.
Thus, for LCPs we can optimally set the learning rate of the L-BFGS method and use a non-zero momentum term with an optimally set momentum rate. Further, for SO-friendly networks we could additionally set these step sizes on a per-layer basis. 

\begin{figure}
\includegraphics[width=0.24\textwidth]{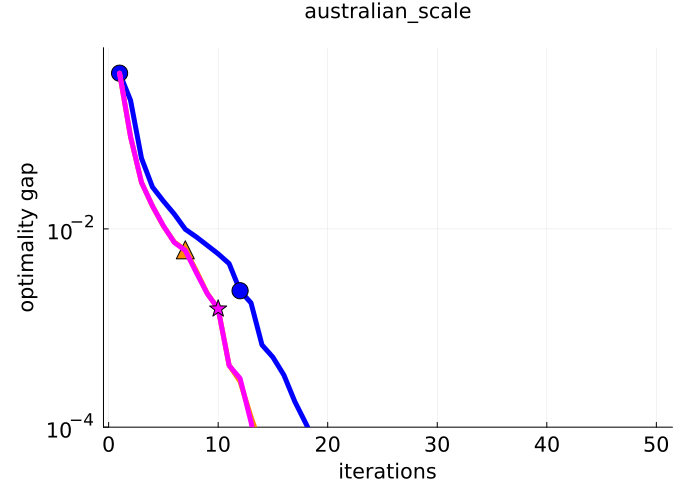}
\includegraphics[width=0.24\textwidth]{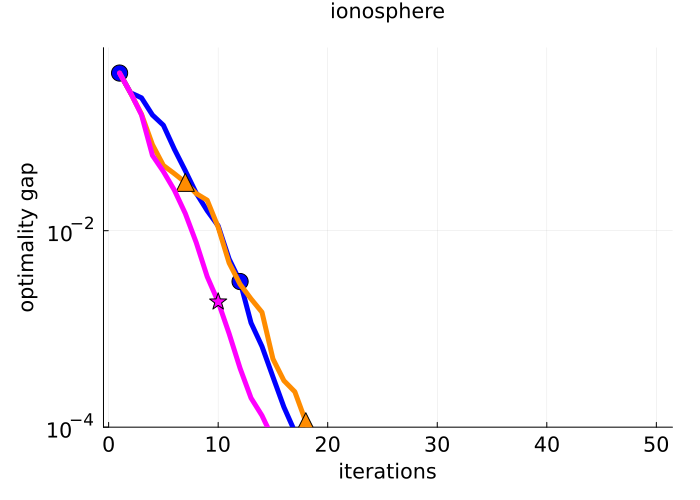}
\includegraphics[width=0.24\textwidth]{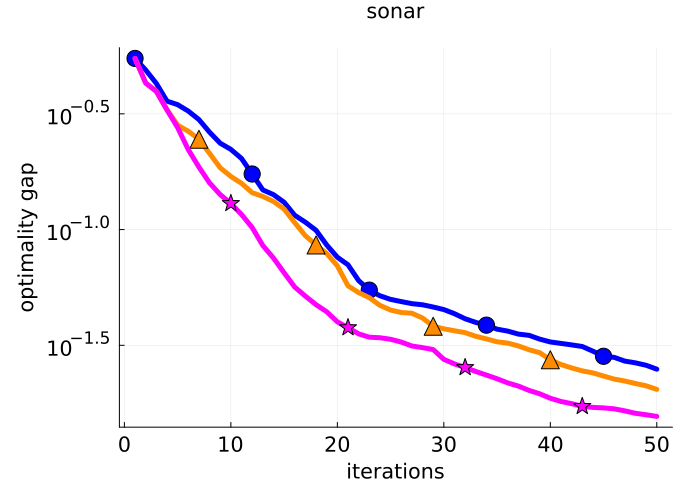}
\includegraphics[width=0.24\textwidth]{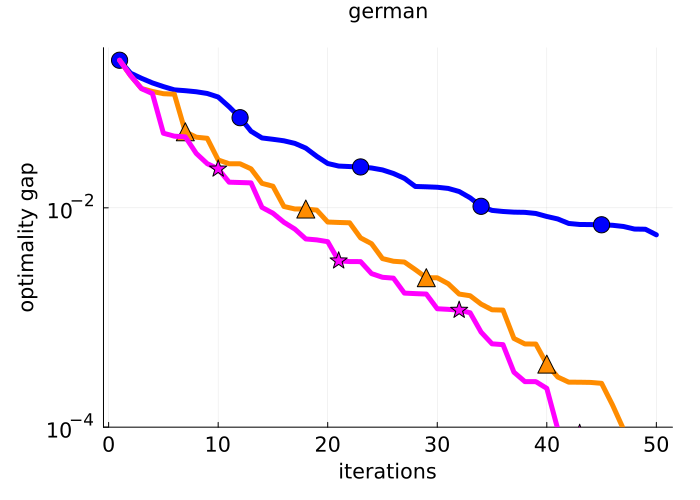}
\includegraphics[width=0.24\textwidth]{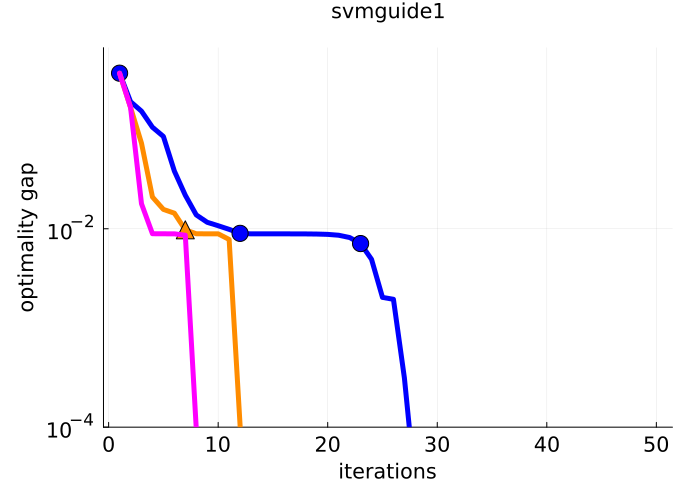}
\includegraphics[width=0.24\textwidth]{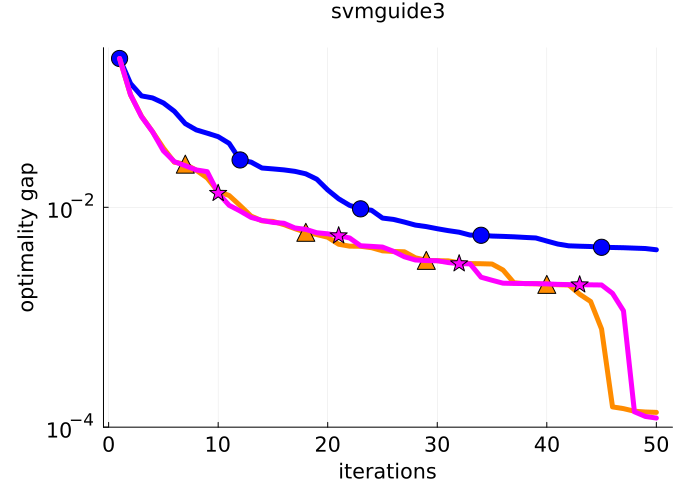}
\includegraphics[width=0.24\textwidth]{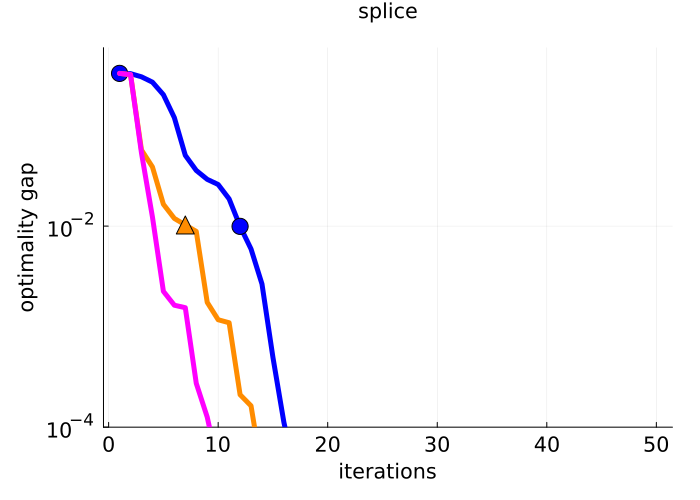}
\includegraphics[width=0.24\textwidth]{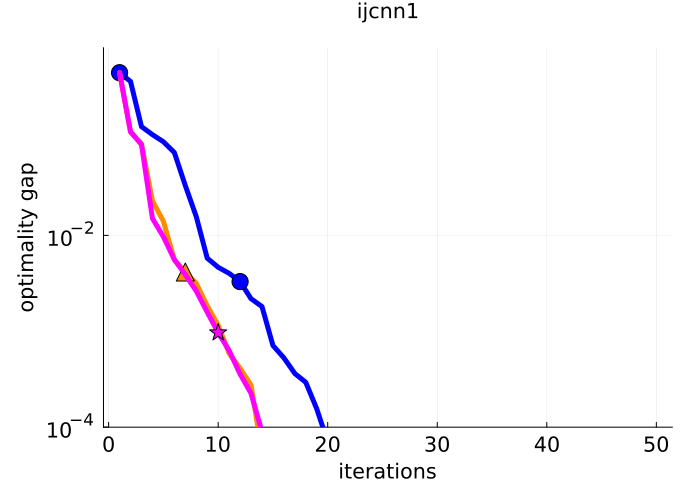}
\includegraphics[width=0.24\textwidth]{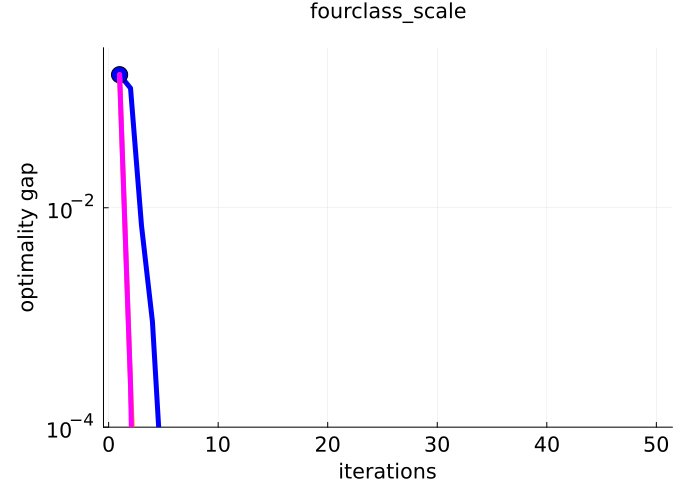}
\includegraphics[width=0.24\textwidth]{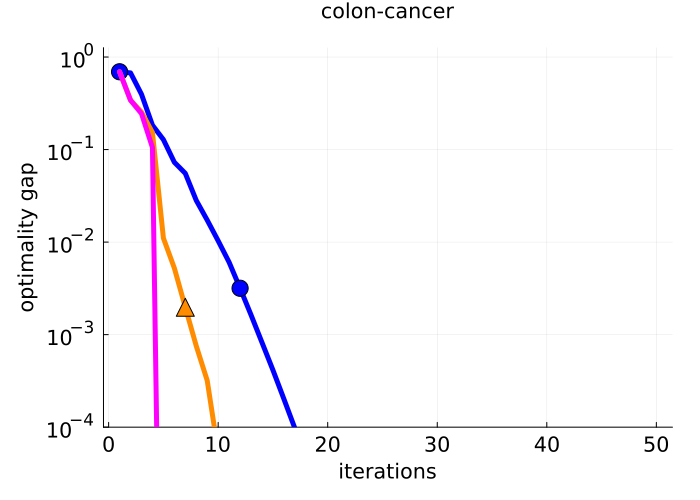}
\includegraphics[width=0.24\textwidth]{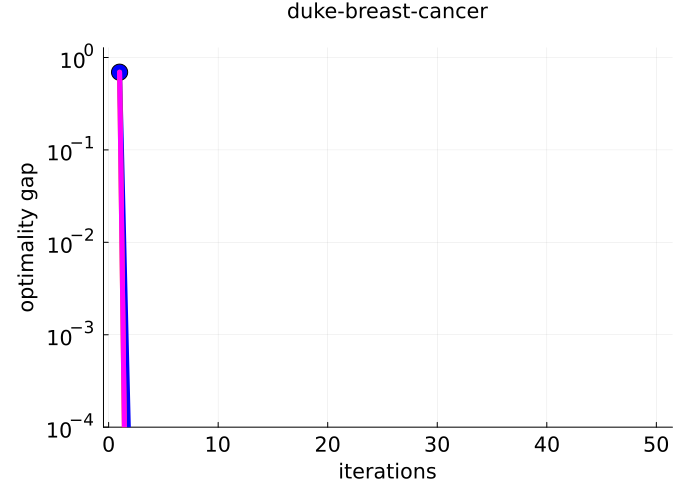}
\includegraphics[width=0.24\textwidth]{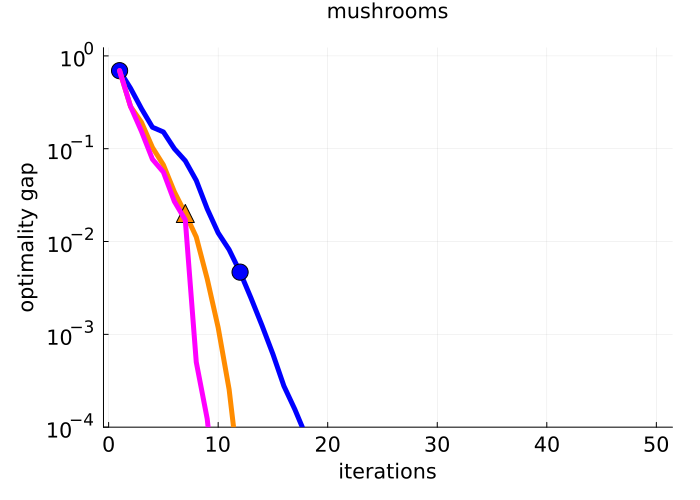}
\includegraphics[width=0.24\textwidth]{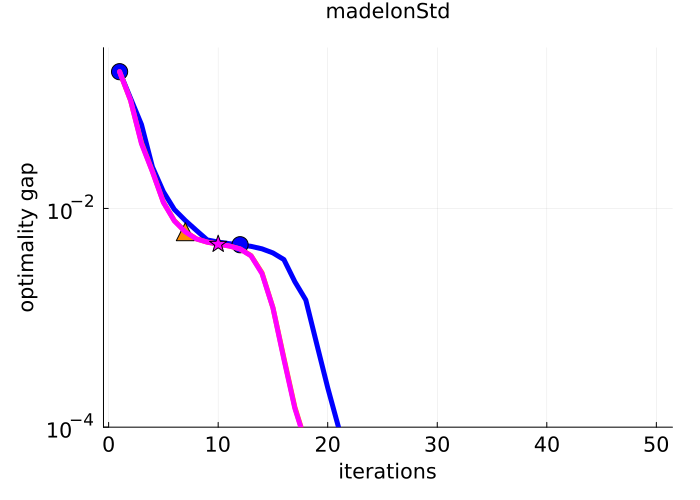}
\includegraphics[width=0.24\textwidth]{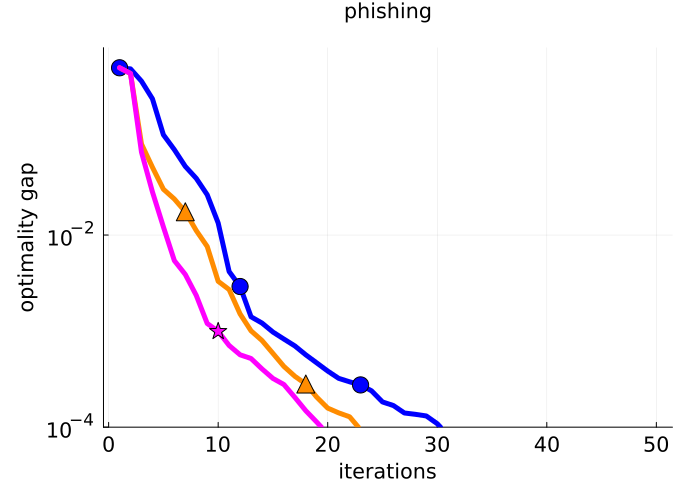}
\includegraphics[width=0.24\textwidth]{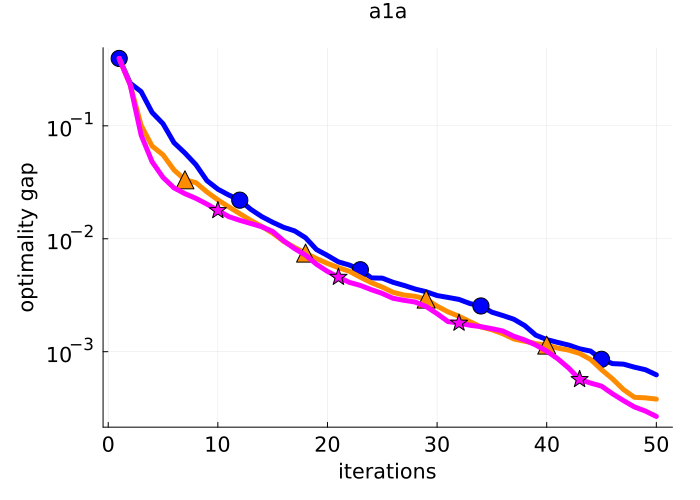}
\includegraphics[width=0.24\textwidth]{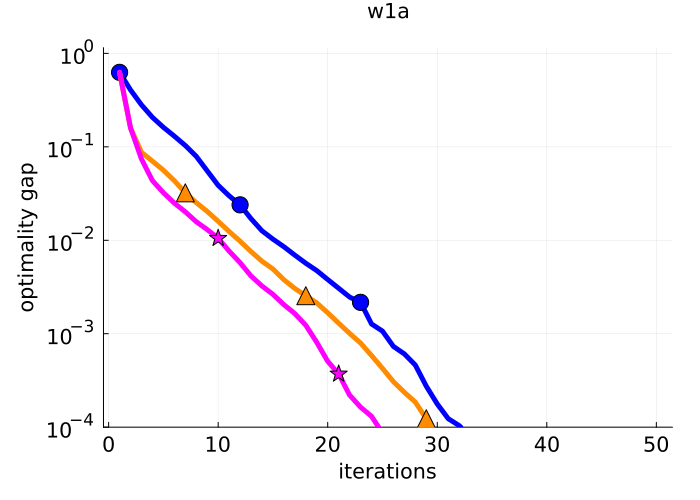}
\centering
\includegraphics[width=.5\textwidth]{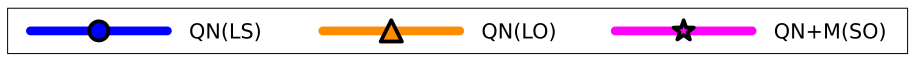}
\caption{Performance of different quasi-Newton methods for fitting logistic regression models. The blue line uses a line search initialized with 1, the orange line uses LO, and the magenta line uses SO to optimize the learning and momentum rates.}
\label{fig:QNlogReg}
\end{figure}

\begin{figure}
\includegraphics[width=0.24\textwidth]{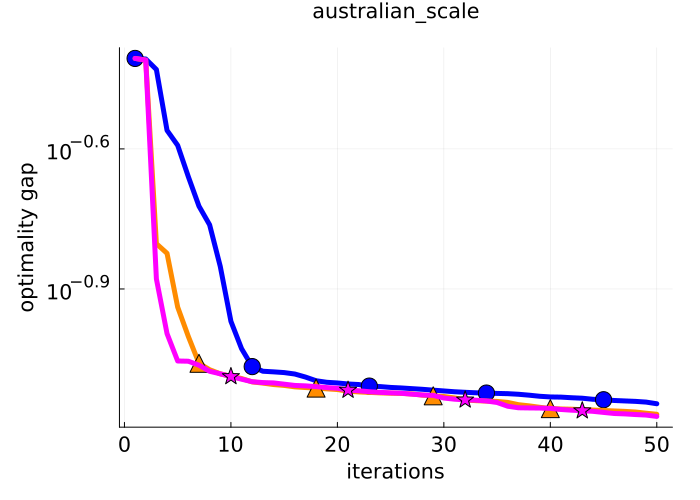}
\includegraphics[width=0.24\textwidth]{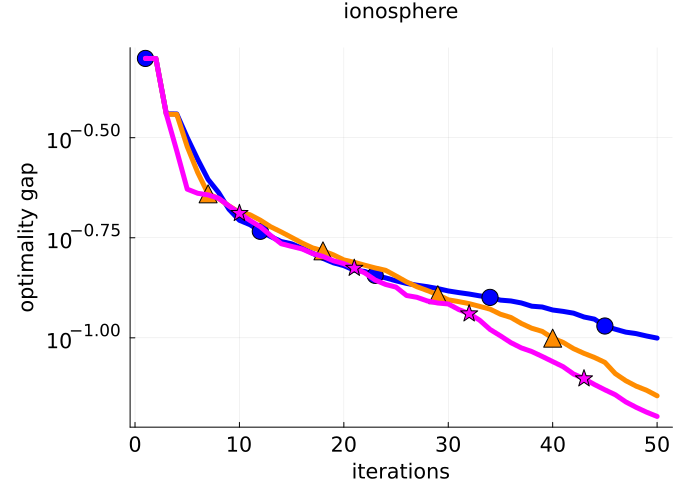}
\includegraphics[width=0.24\textwidth]{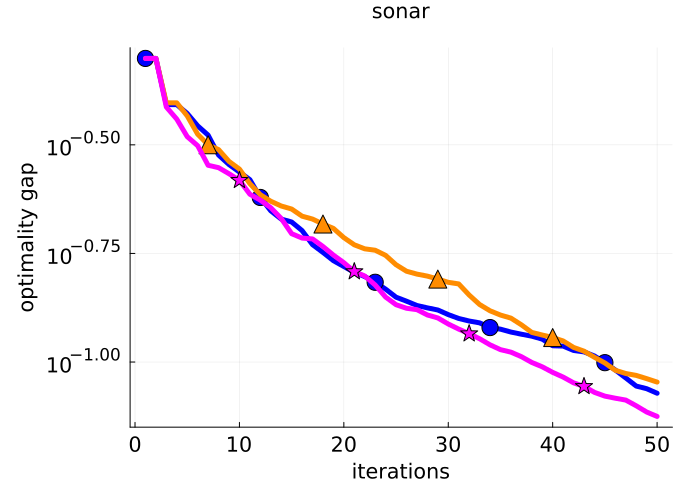}
\includegraphics[width=0.24\textwidth]{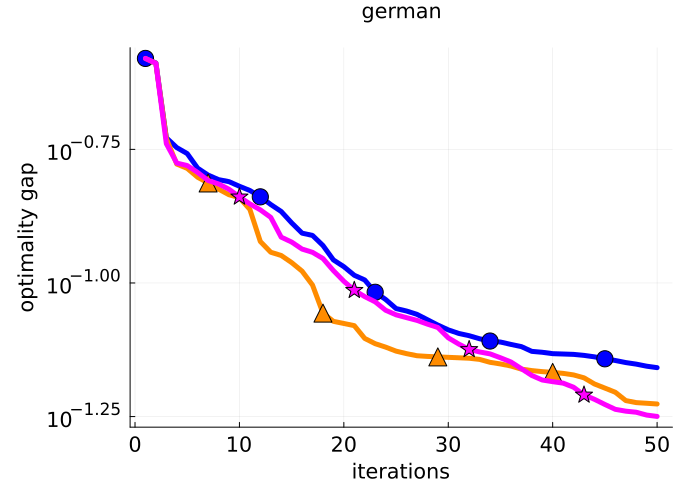}
\includegraphics[width=0.24\textwidth]{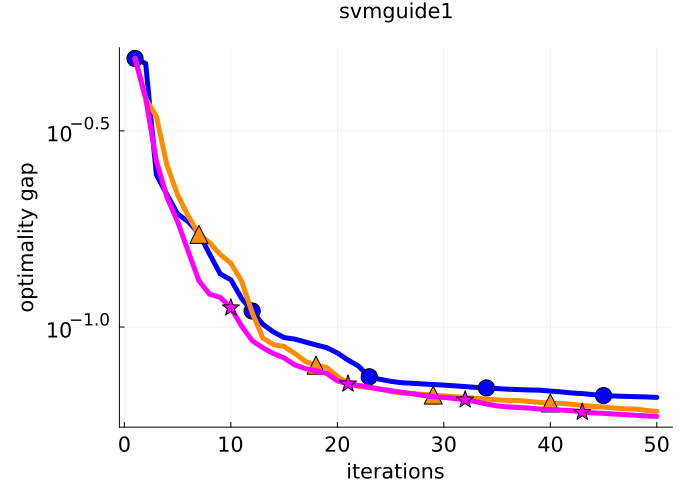}
\includegraphics[width=0.24\textwidth]{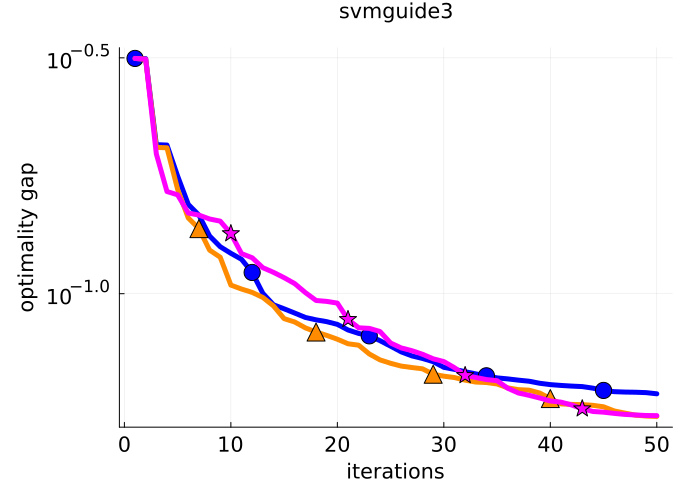}
\includegraphics[width=0.24\textwidth]{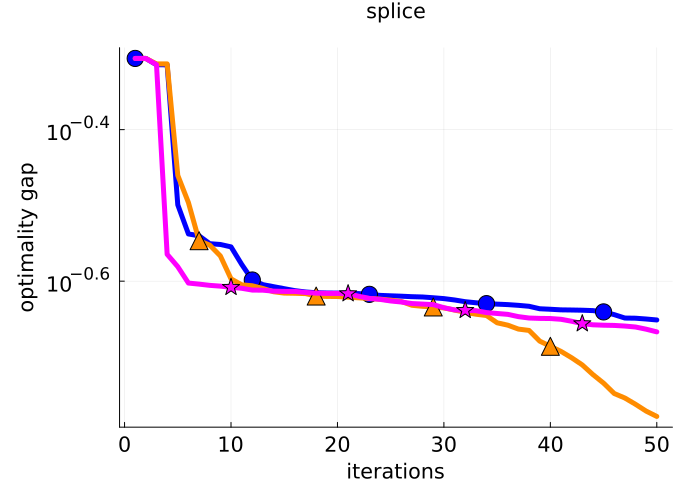}
\includegraphics[width=0.24\textwidth]{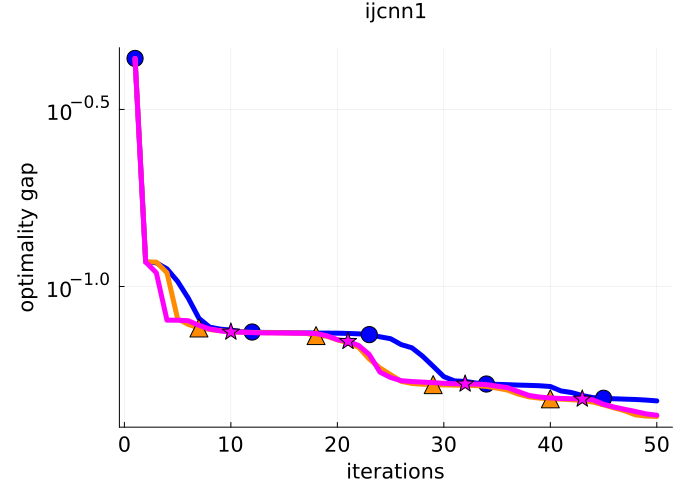}
\includegraphics[width=0.24\textwidth]{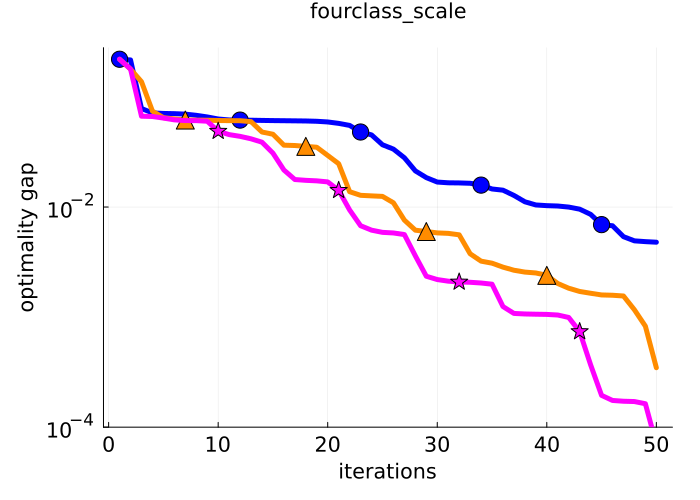}
\includegraphics[width=0.24\textwidth]{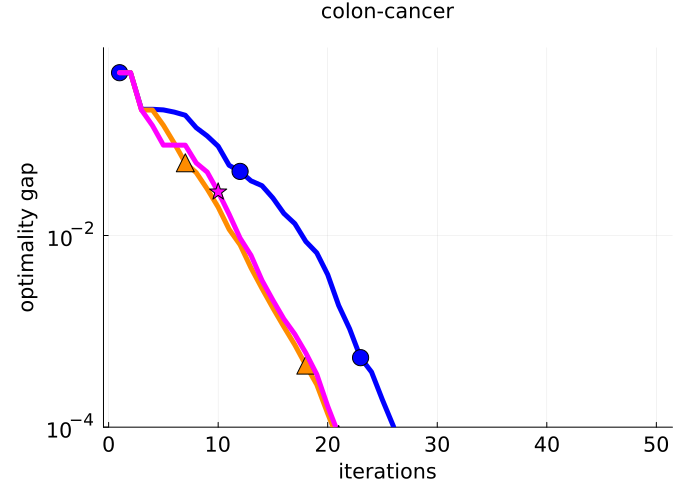}
\includegraphics[width=0.24\textwidth]{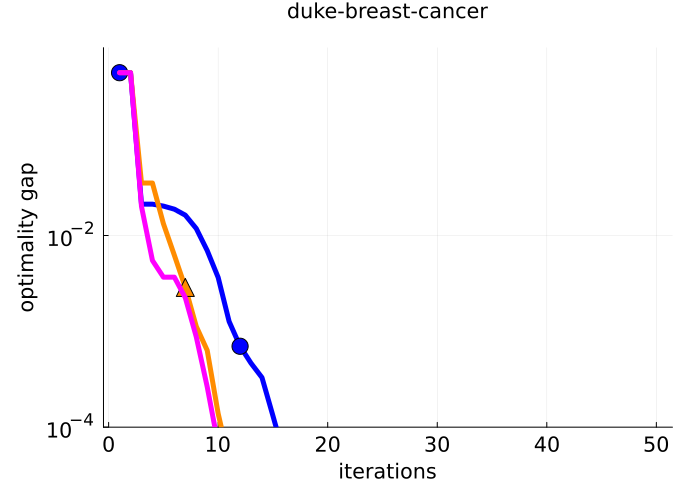}
\includegraphics[width=0.24\textwidth]{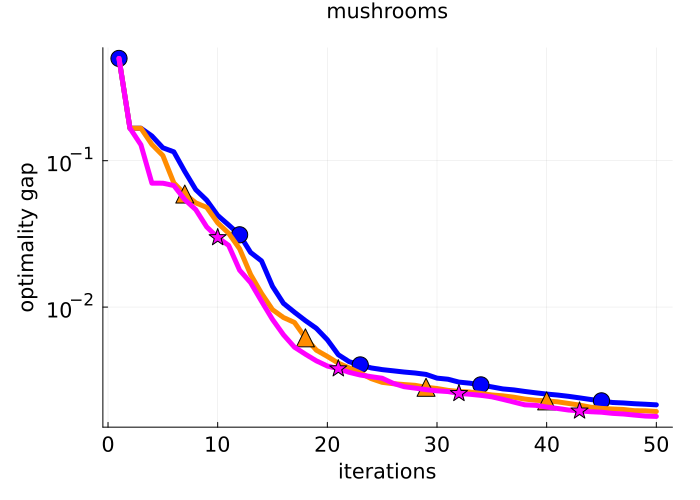}
\includegraphics[width=0.24\textwidth]{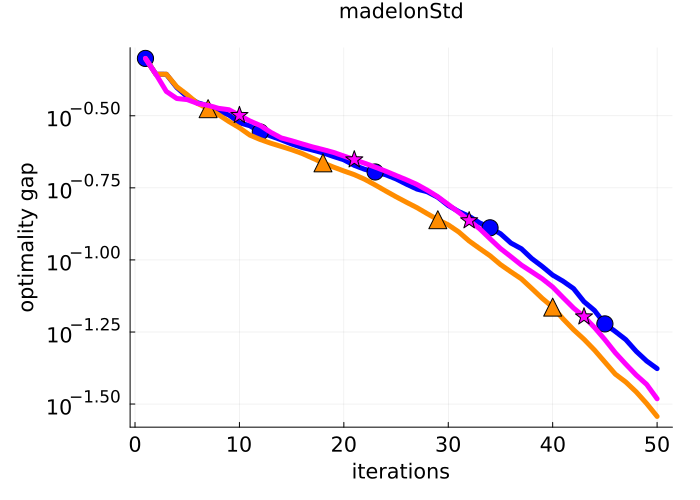}
\includegraphics[width=0.24\textwidth]{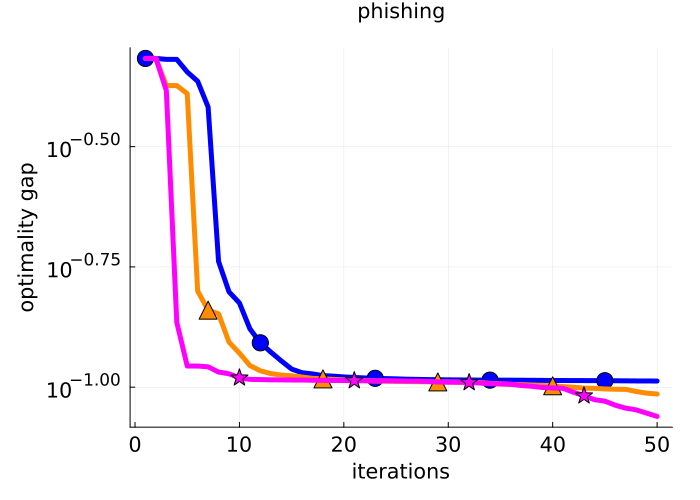}
\includegraphics[width=0.24\textwidth]{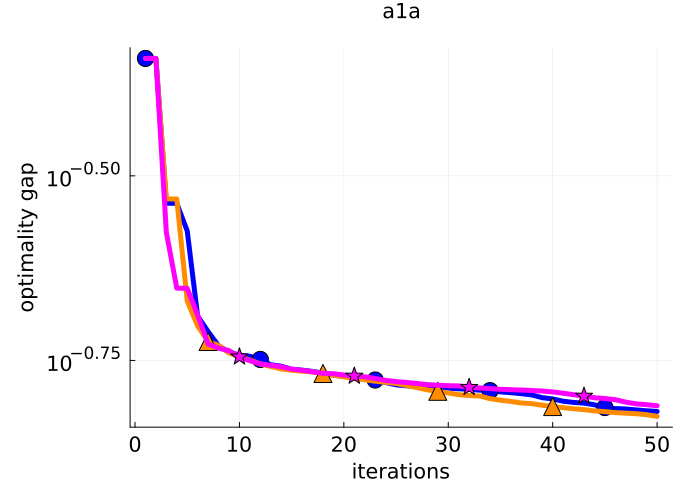}
\includegraphics[width=0.24\textwidth]{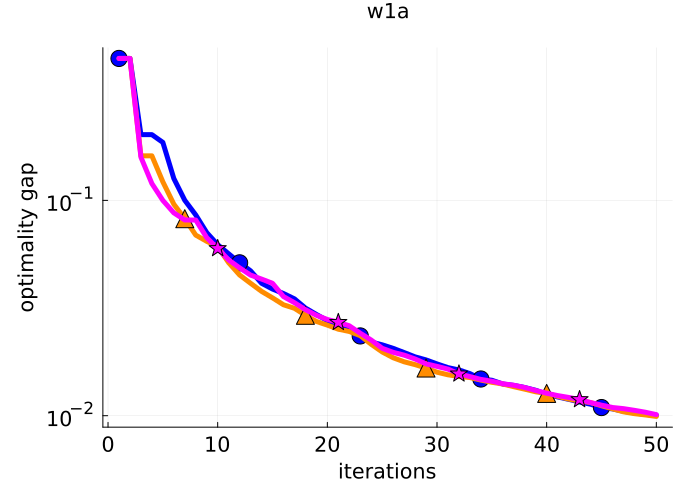}
\centering
\includegraphics[width=.5\textwidth]{QNLegend.png}
\caption{Performance of different quasi-Newton methods for 2-layer neural networks. The blue line uses a line search initialized with 1, the orange line uses LO, and the magenta line uses SO to optimize the learning and momentum rates}
\label{fig:QNnn100}
\end{figure}

In Figures~\ref{fig:QNlogReg} and~\ref{fig:QNnn100} we plot the performance of methods that use L-BFGS directions where the step size is set using the Wolfe conditions (\texttt{QN(LS)}), the step size is set using LO (\texttt{QN(LO)}), or we add a momentum term and set both step sizes using SO (\texttt{QN+M(SO)}). For all methods, we use the method of~\citet{Shanno1978} to improve the scaling of $B_k$, and for the \texttt{QN(LS)} method we initialize the line search with a step size of $\alpha_k=1$. While the differences are smaller than what we see for gradient methods, in these figures we see that using LO often improves performance and that using SO with the additional momentum term often further improves performance.

\begin{figure}
\includegraphics[width=0.24\textwidth]{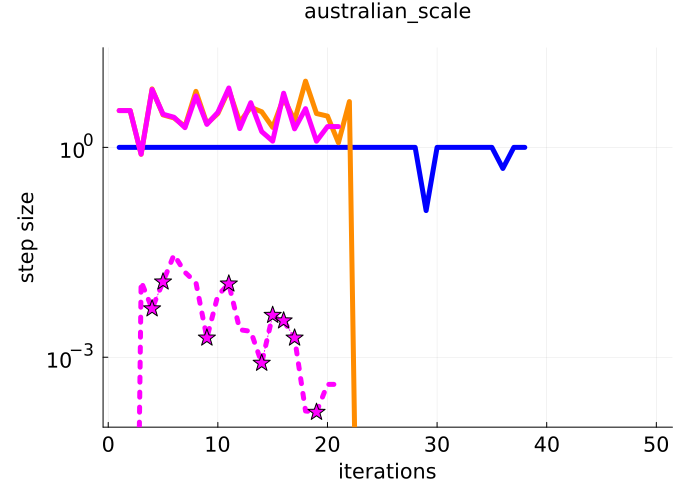}
\includegraphics[width=0.24\textwidth]{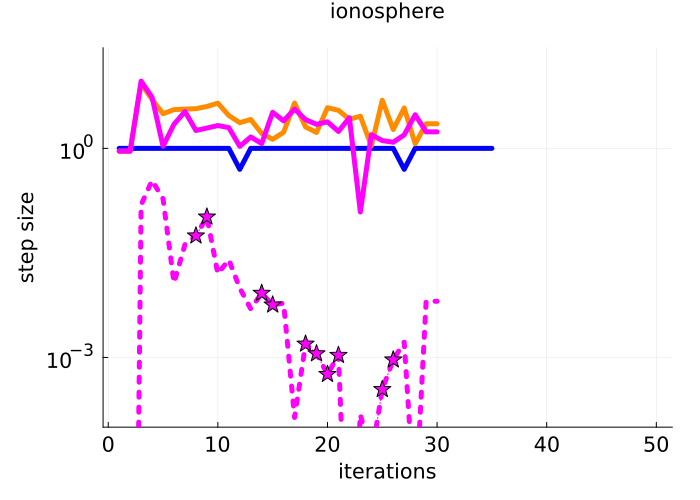}
\includegraphics[width=0.24\textwidth]{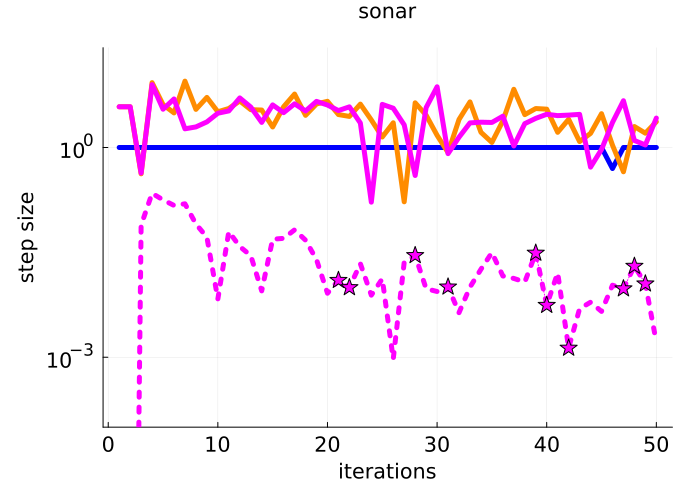}
\includegraphics[width=0.24\textwidth]{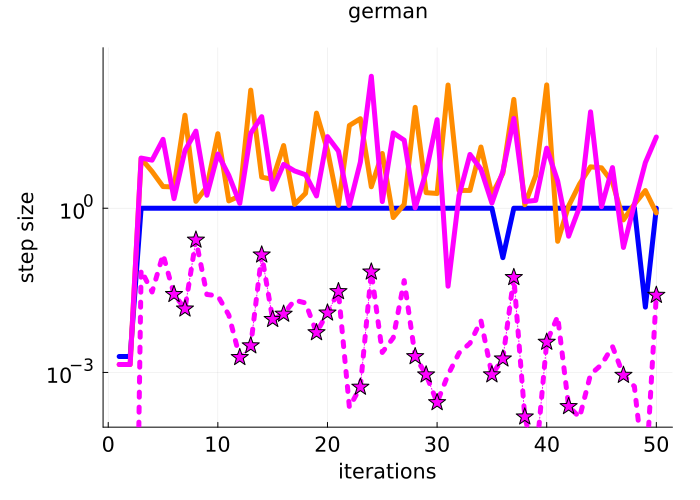}
\includegraphics[width=0.24\textwidth]{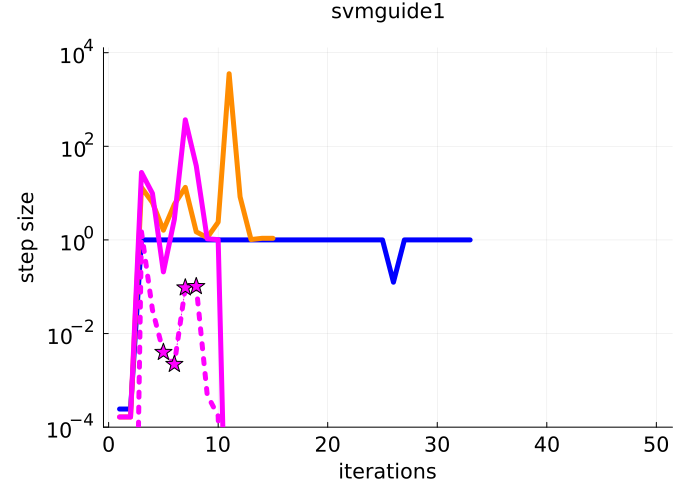}
\includegraphics[width=0.24\textwidth]{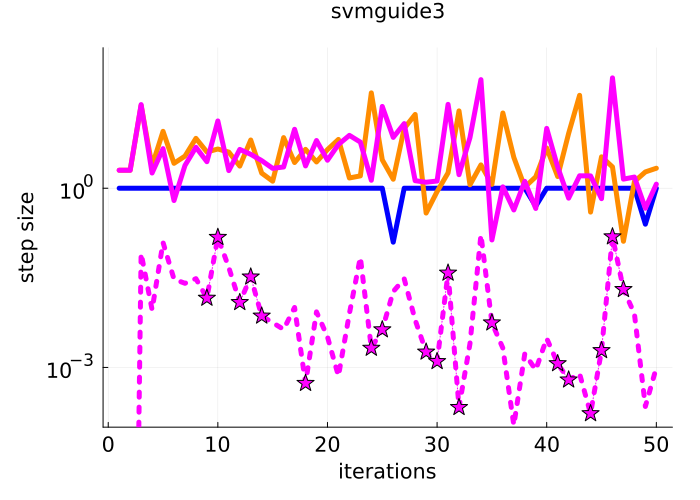}
\includegraphics[width=0.24\textwidth]{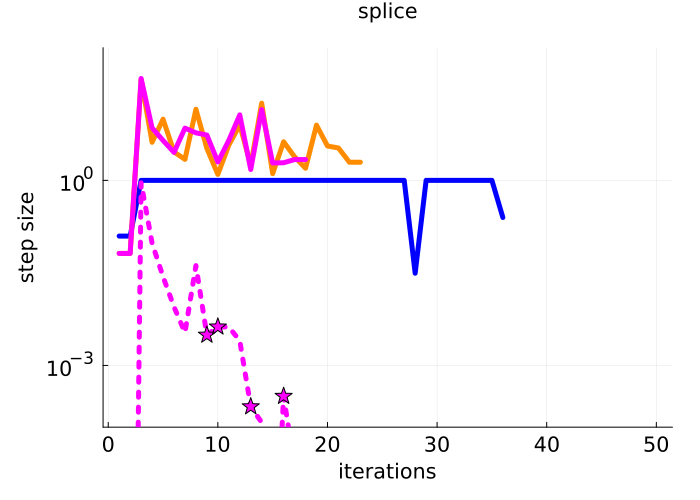}
\includegraphics[width=0.24\textwidth]{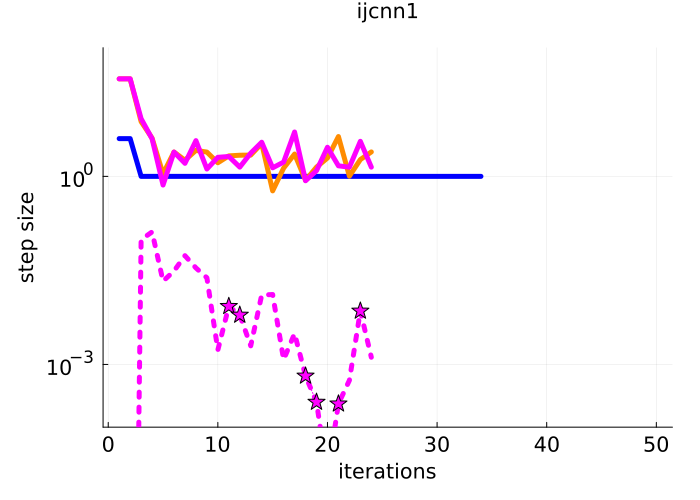}
\includegraphics[width=0.24\textwidth]{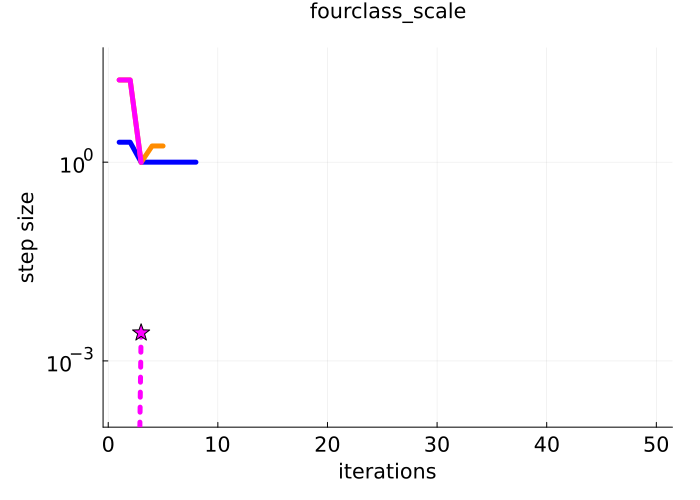}
\includegraphics[width=0.24\textwidth]{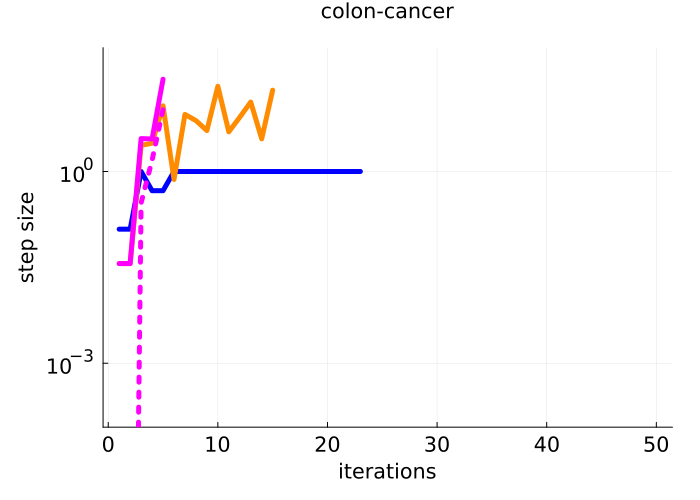}
\includegraphics[width=0.24\textwidth]{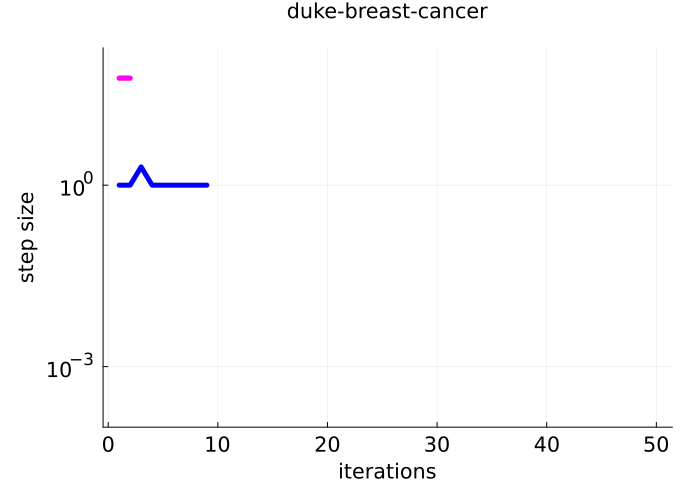}
\includegraphics[width=0.24\textwidth]{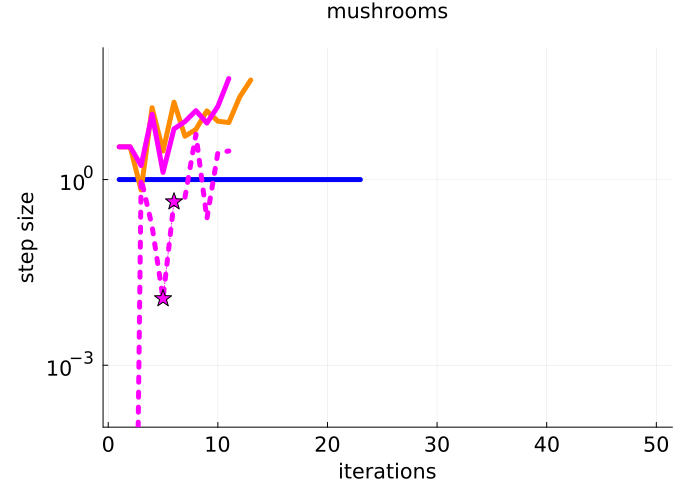}
\includegraphics[width=0.24\textwidth]{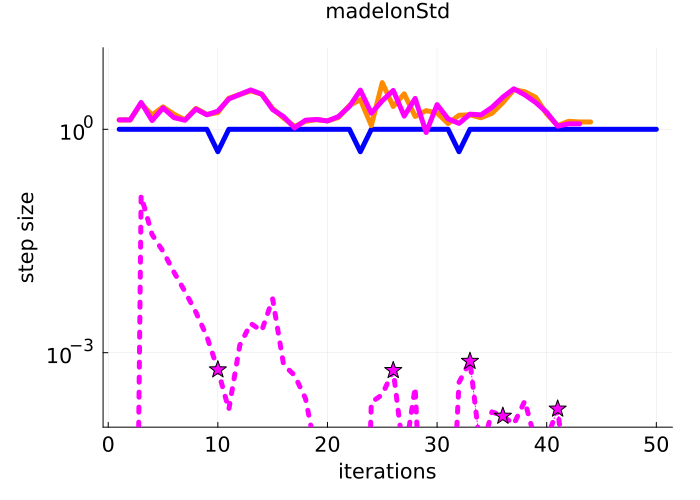}
\includegraphics[width=0.24\textwidth]{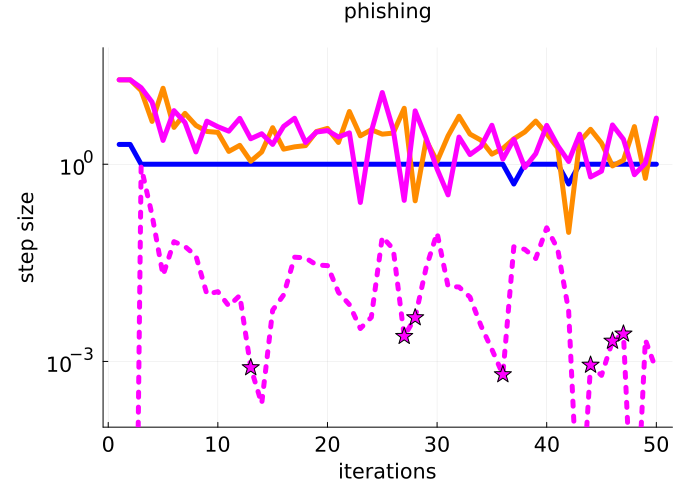}
\includegraphics[width=0.24\textwidth]{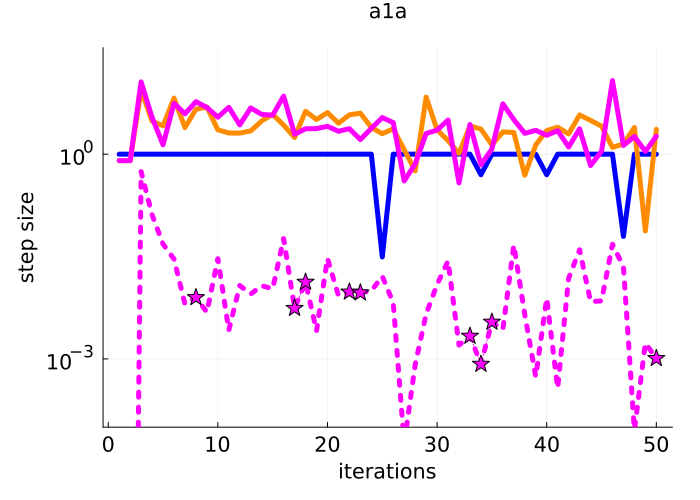}
\includegraphics[width=0.24\textwidth]{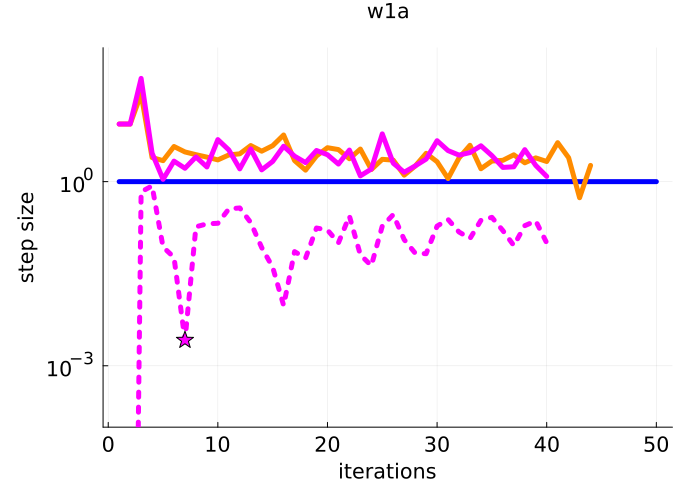}
\centering
\includegraphics[width=.67\textwidth]{QNLegend.png}
\caption{Step sizes of different quasi-Newton methods for fitting logistic regression models. We plot the absolute value, and include markers to indicate iterations where the step was negative. The dashed line is the momentum rate for the SO method. We see that the line search typically uses a learning rate of 1, while the LO and SO methods typically use slightly larger values. Note that the momentum rate was frequently negative in the SO method.}
\label{fig:QNsteplogReg}
\end{figure}

\begin{figure}
\includegraphics[width=0.24\textwidth]{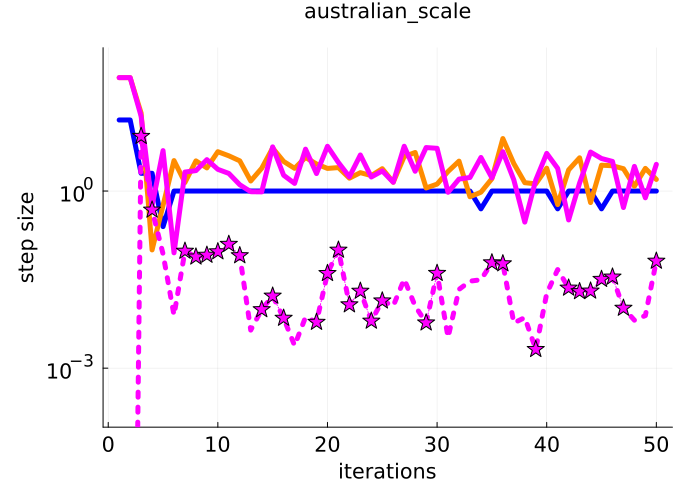}
\includegraphics[width=0.24\textwidth]{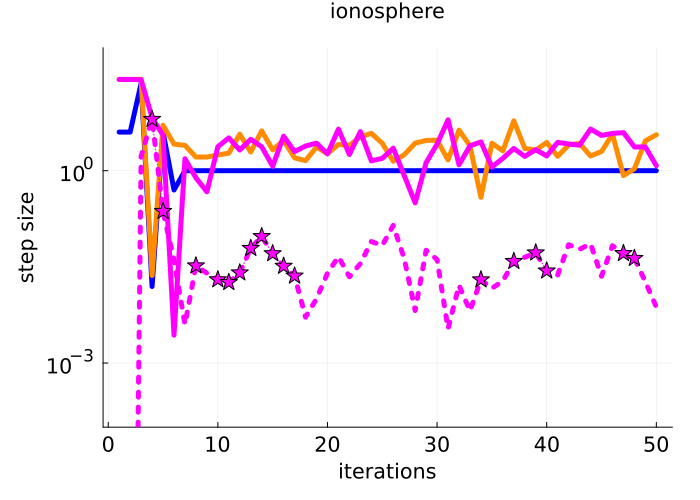}
\includegraphics[width=0.24\textwidth]{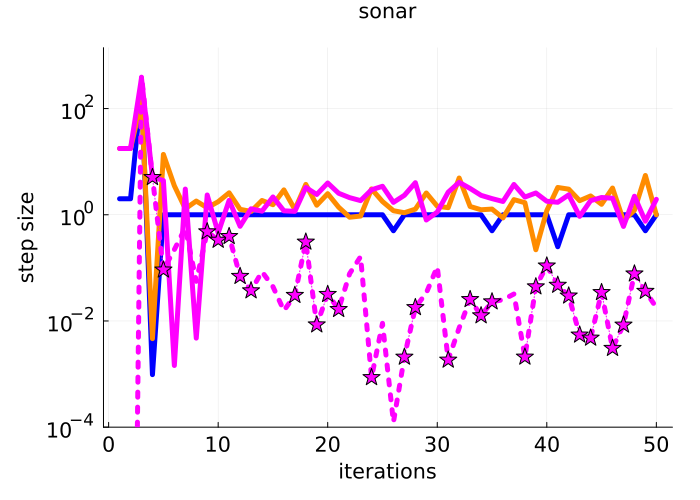}
\includegraphics[width=0.24\textwidth]{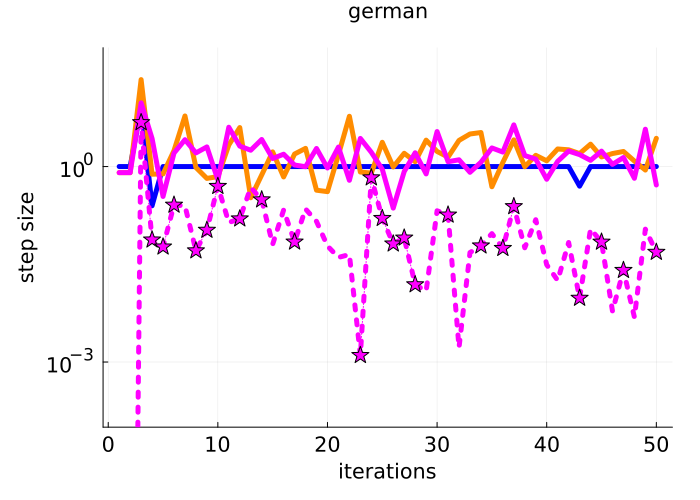}
\includegraphics[width=0.24\textwidth]{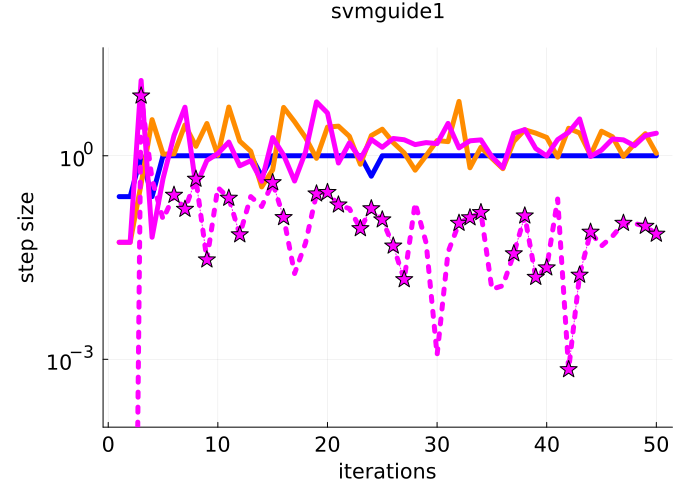}
\includegraphics[width=0.24\textwidth]{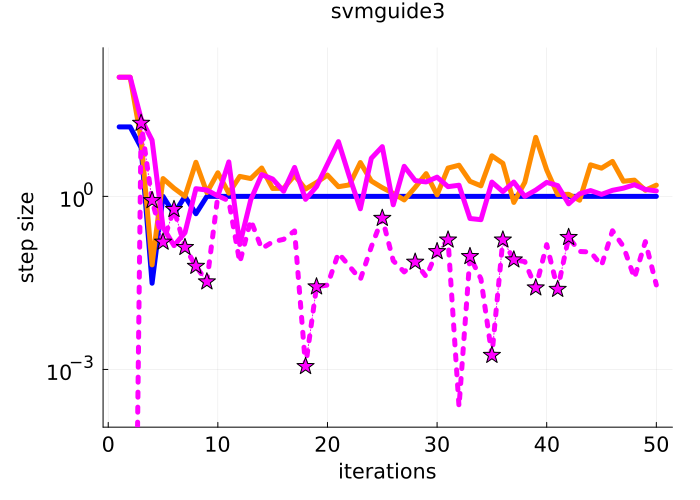}
\includegraphics[width=0.24\textwidth]{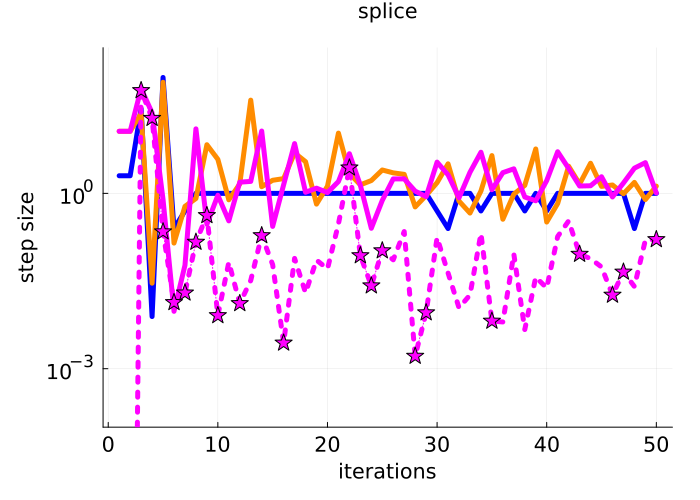}
\includegraphics[width=0.24\textwidth]{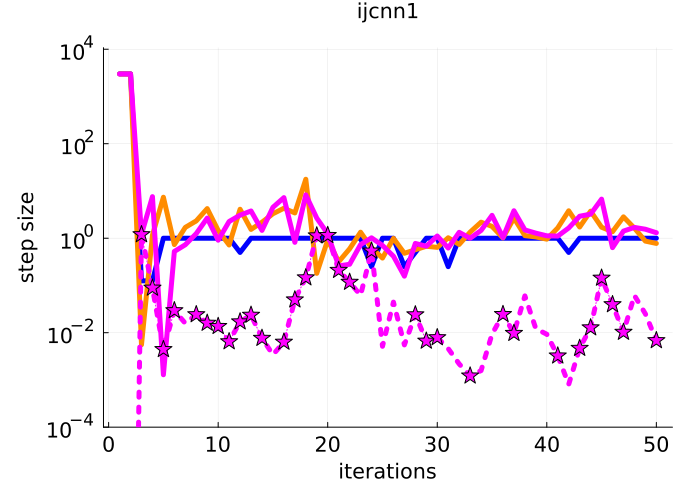}
\includegraphics[width=0.24\textwidth]{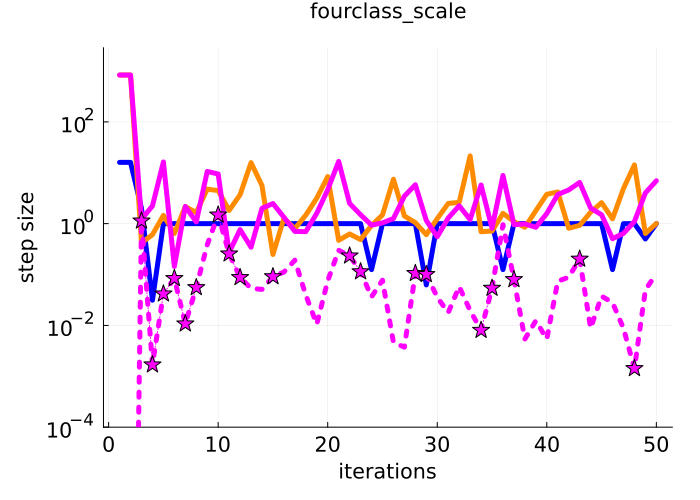}
\includegraphics[width=0.24\textwidth]{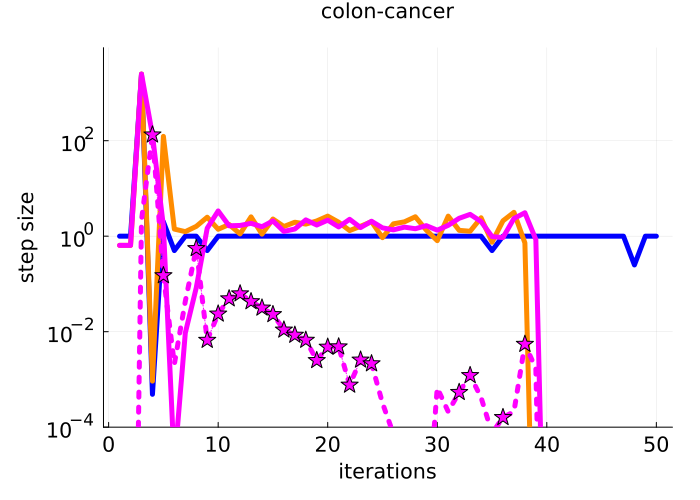}
\includegraphics[width=0.24\textwidth]{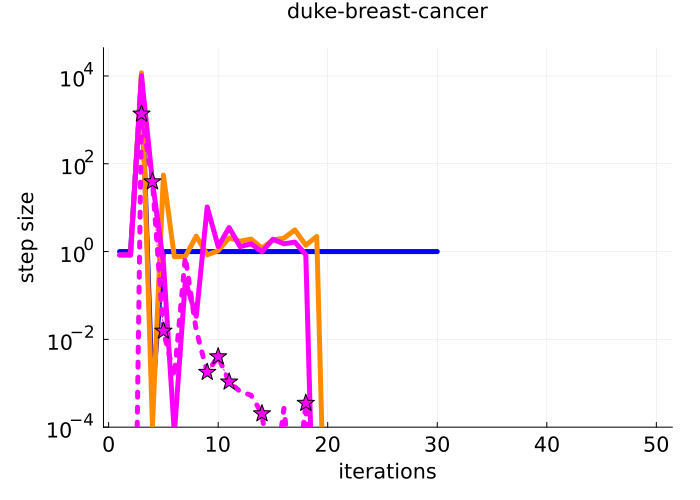}
\includegraphics[width=0.24\textwidth]{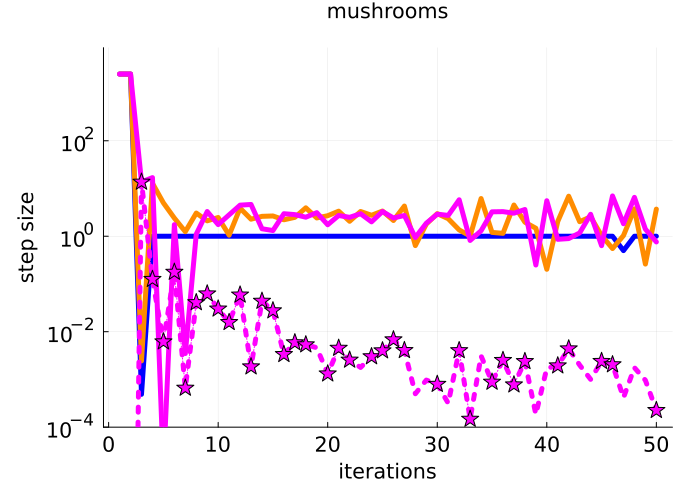}
\includegraphics[width=0.24\textwidth]{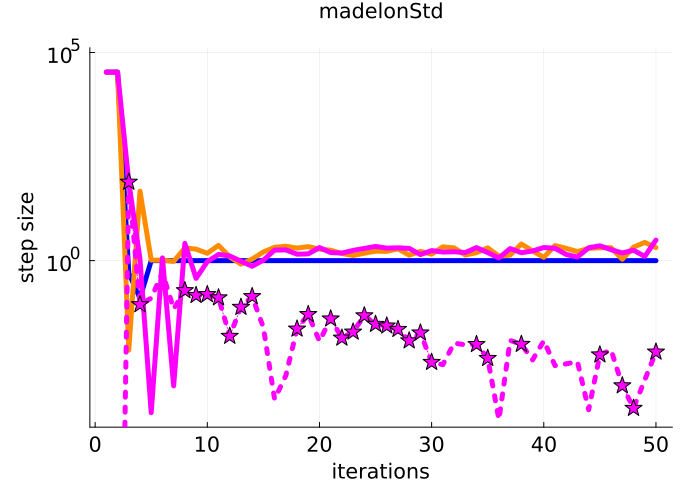}
\includegraphics[width=0.24\textwidth]{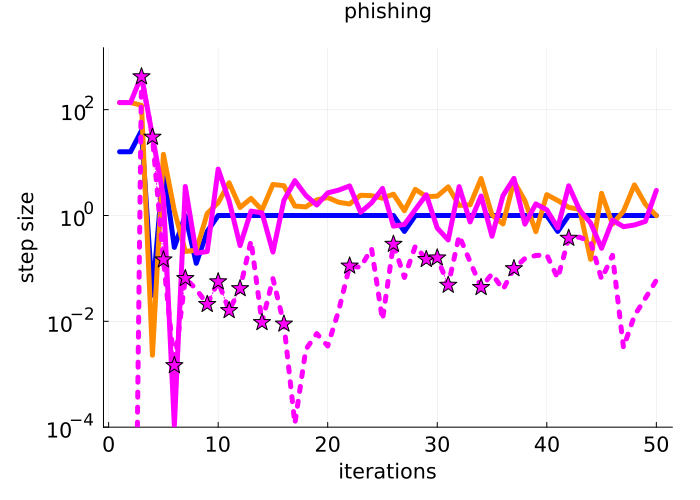}
\includegraphics[width=0.24\textwidth]{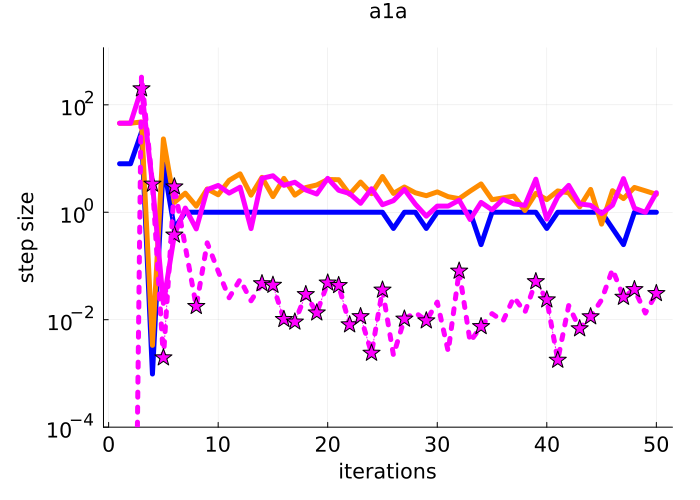}
\includegraphics[width=0.24\textwidth]{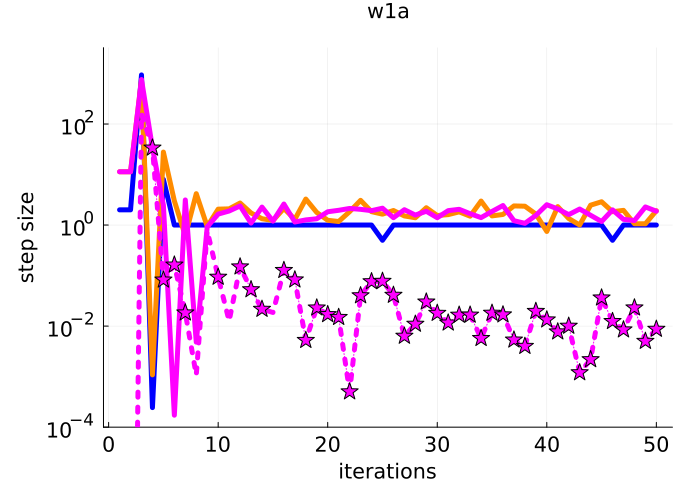}
\centering
\includegraphics[width=.67\textwidth]{QNLegend.png}
\caption{Step sizes of different quasi-Newton methods for fitting = 2-layer neural networks. We plot the absolute value, and include markers to indicate iterations where the step was negative. The dashed line is the momentum rate for the SO method. We see that the line search typically uses a learning rate of 1, while the LO and SO methods typically use slightly larger values. Note that the momentum rate was frequently negative in the SO method.}
\label{fig:QNstepnn100}
\end{figure}

It may appear surprising that we can improve the performance of quasi-Newton methods with LO and SO, since asymptotically if quasi-Newton methods converge to a strict minimizer then the choice of $\alpha_k=1$ is optimal and momentum is not needed. However, we note that for any finite iteration quasi-Newton can benefit from using step sizes other than 1 and can benefit from momentum. In Figures~\ref{fig:QNsteplogReg} and~\ref{fig:QNstepnn100} we plot the step sizes. In these plots we that the line search \texttt{QN(LS)} method typically accepts the initial step size of 1, while the LO and SO methods typically use step sizes that are slightly larger than 1. We also see that the momentum rate used by the \texttt{QN+M(SO)} method is typically much smaller than the step size, and that negative momentum steps are often used (we were surprised that $\alpha_k$ was also negative on one iteration on the phishing dataset).

\subsection{Augmenting Adam with Subspace Optimization}

The Adam optimizer uses the update
\begin{equation}
\begin{aligned}
	\mu_{k+1} & = \beta_k^1\mu_k + (1-\beta_k^1)\nabla f(w_k)\\
	v_{k+1} & = \beta_k^2 v_{k-1} + (1-\beta_k^2)\nabla f(w_k)\circ \nabla f(w^k)\\
	w_{k+1} & = w_k - \alpha_k \underbrace{V_{k+1}^{-1} \mu_{k+1}}_{d_k},
	\end{aligned}
	\label{eq:adam}
\end{equation}
for a learning rate $\alpha_k$, gradient momentum rate $\beta_k^1$,  squared-gradient momentum rate $\beta_k^2$, and $V_k$ being a diagonal matrix where diagonal $i$ given by $(\sqrt{(v_k)_i} + \epsilon)$ for a small $\epsilon$ (we use $\circ$ for element-wise multiplication).\footnote{We omit the bias correction factors in the update of $w_{k}$ for simplicity and because they made the performance of the method worse in our experiments. But these factors do not affect the efficiency of SO.} 
In many applications the three learning rates $\{\alpha_k,\beta_k^1,\beta_k^2\}$ are set to fixed values across iterations, with a default choice being $\alpha_k = 0.001$, $\beta_k^1 = 0.99$, $\beta_k^2 = 0.999$, and $\epsilon=10^{-8}$. However, for over-parameterized problems recent works show that the out-of-the-box performance of Adam is improved by using a line search to set $\alpha_k$ on each iteration~\citep{Vaswani2020,galli2023don}. For LCPs and SO-friendly networks we can further improve the performance of Adam using LO. For example, with 2 matrix multiplications per iteration we could use LO to set $\alpha_k$: one to compute $\nabla f(w^k)$ and one to compute $XV_{k+1}^{-1}\mu_k$. 

Unfortunately, even for LCPs it does not appear efficient to use SO to set the learning and momentum rates in Adam. It would require 3 matrix multiplications per iteration to optimize $\alpha_k$ and $\beta_k^1$: one to compute $\nabla f(w^k)$, one to compute $XV_{k+1}{-1}\mu_k$, and one to compute $XV_{k+1}^{-1}\nabla f(w_k)$. Unfortunately, this increase the iteration by 50\% in terms of the bottleneck matrix multiplication operations. Further, it does not appear that the structure of the algorithm allows to efficiently use SOs to optimize all three parameters even for LCPs. 

Nevertheless, for amenable problem structures it is possible to add additional directions to Adam that can be efficiently optimized with SO. For example, we explored adding an additional momentum term $\beta_k(w_k - w_{k-1})$ to Adam and found this improved performance. But we found that a larger performance increase could be obtained by using multiple Adam directions,
\begin{equation}
	w_{k+1}  = w_k - \alpha_k^1 d_k - \alpha_k^2 d_{k-1},
	\label{eq:adam2}
\end{equation}
where $d_k$ is defined by the Adam algorithm~\eqref{eq:adam}
Despite Adam usually being considered as a method that works with stochastic gradients, we found that this update was often an effective deterministic optimization method when using SO to set the step sizes.

\begin{figure}
\includegraphics[width=0.24\textwidth]{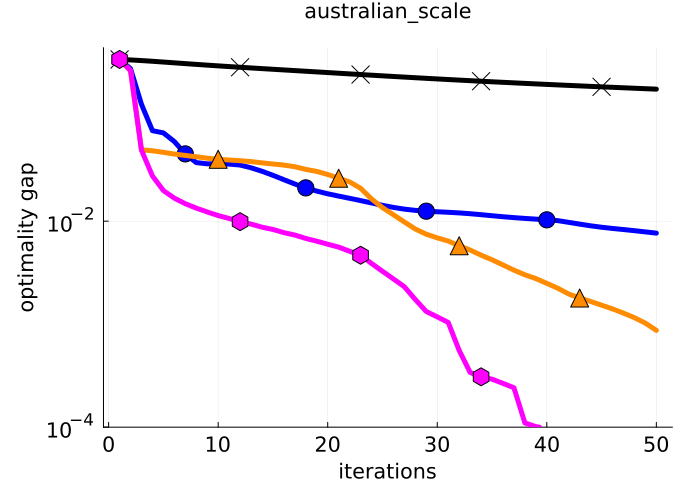}
\includegraphics[width=0.24\textwidth]{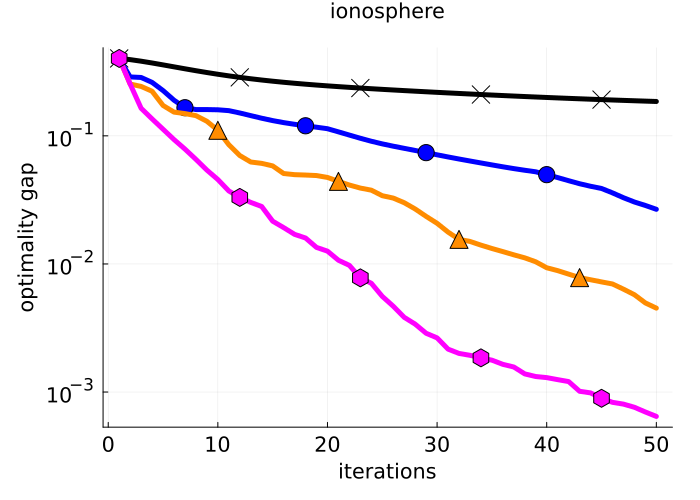}
\includegraphics[width=0.24\textwidth]{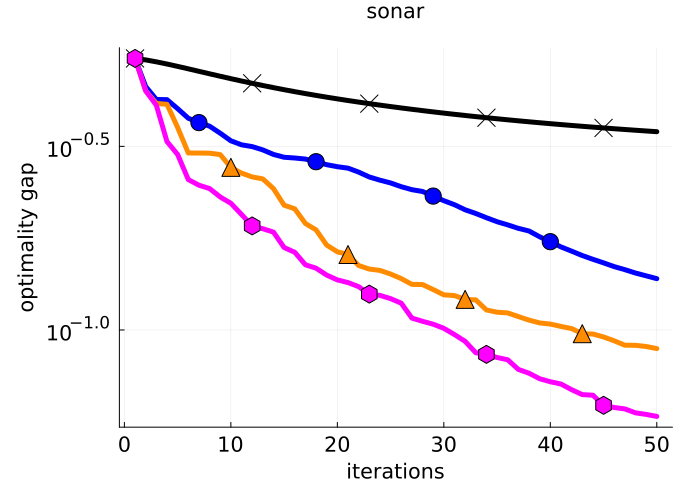}
\includegraphics[width=0.24\textwidth]{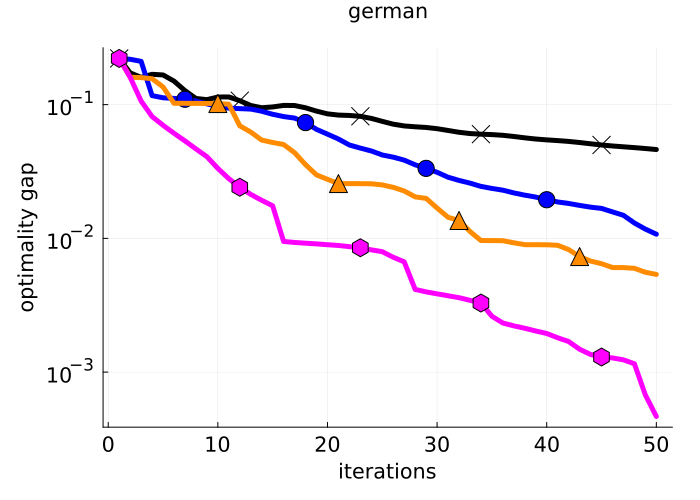}
\includegraphics[width=0.24\textwidth]{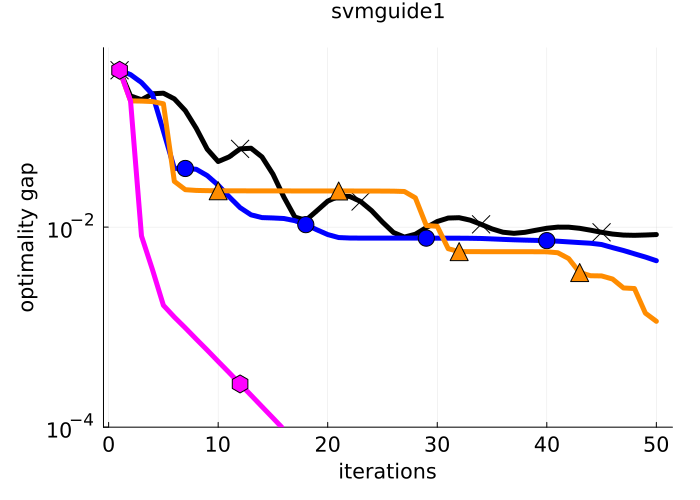}
\includegraphics[width=0.24\textwidth]{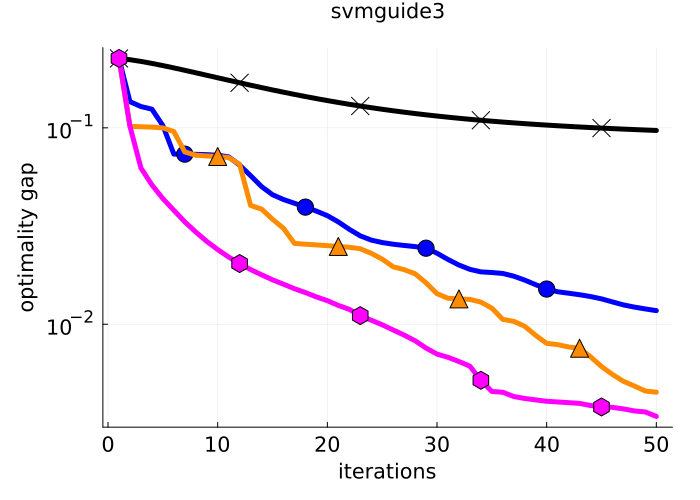}
\includegraphics[width=0.24\textwidth]{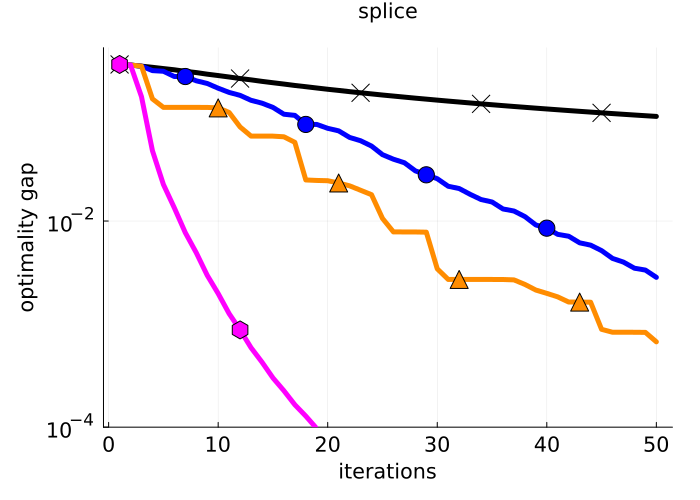}
\includegraphics[width=0.24\textwidth]{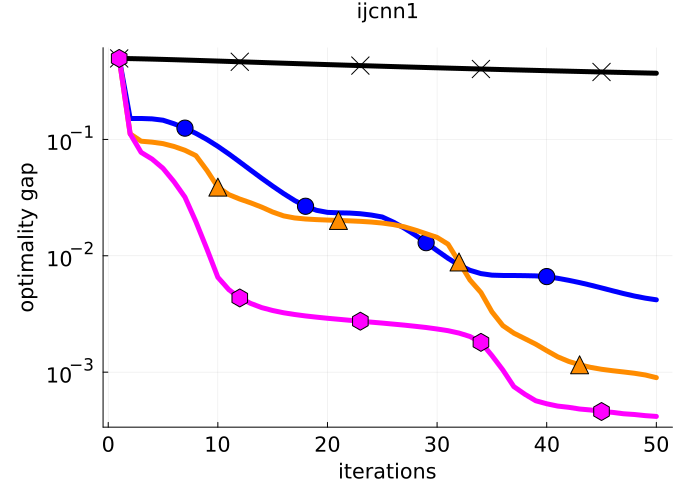}
\includegraphics[width=0.24\textwidth]{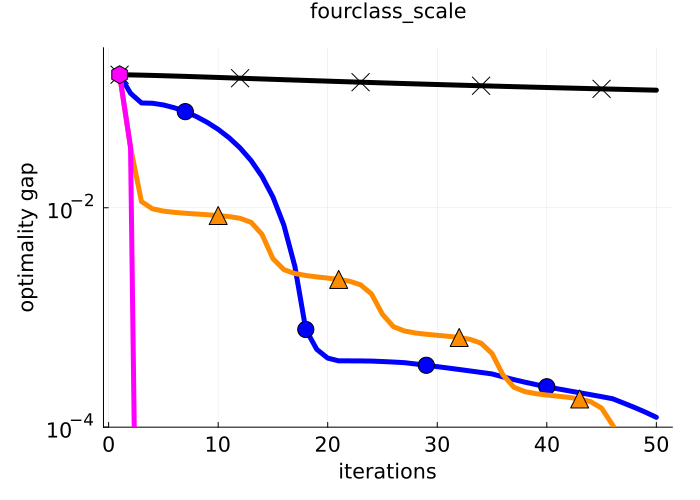}
\includegraphics[width=0.24\textwidth]{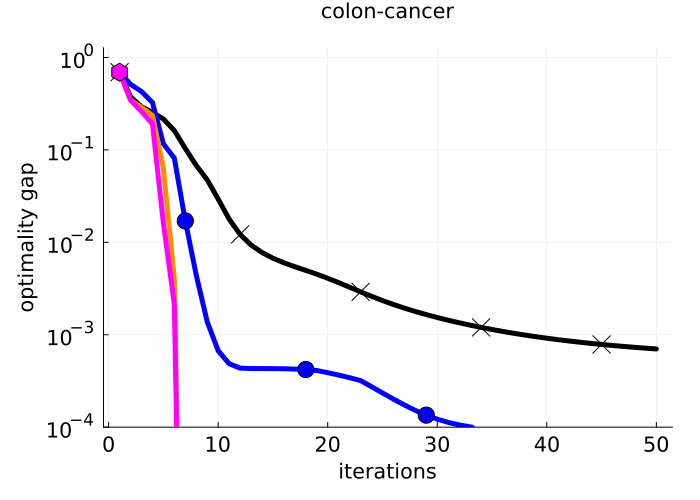}
\includegraphics[width=0.24\textwidth]{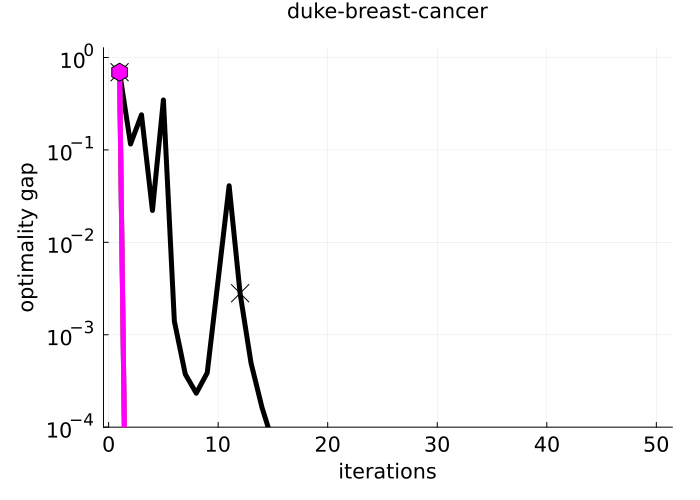}
\includegraphics[width=0.24\textwidth]{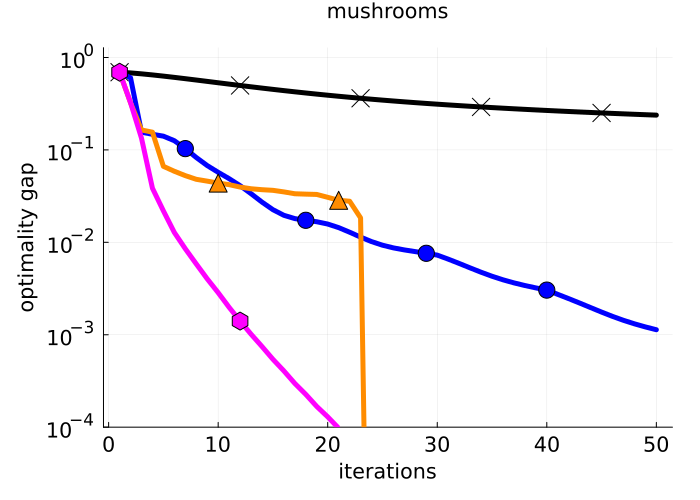}
\includegraphics[width=0.24\textwidth]{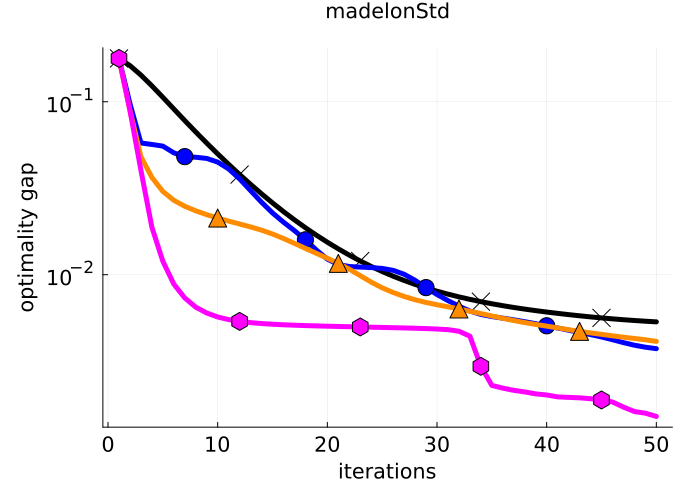}
\includegraphics[width=0.24\textwidth]{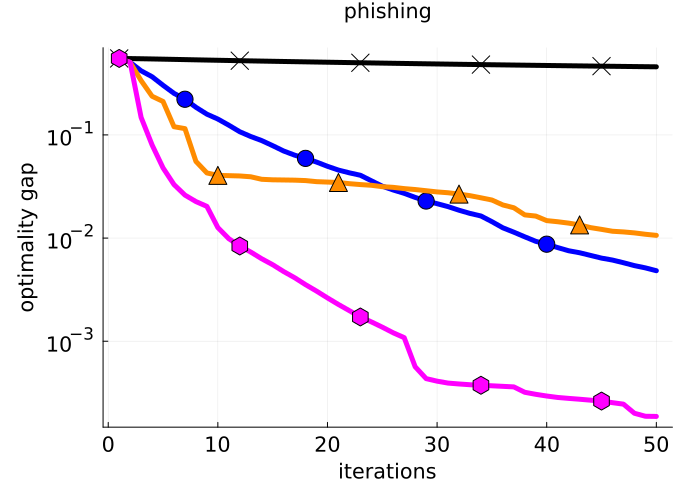}
\includegraphics[width=0.24\textwidth]{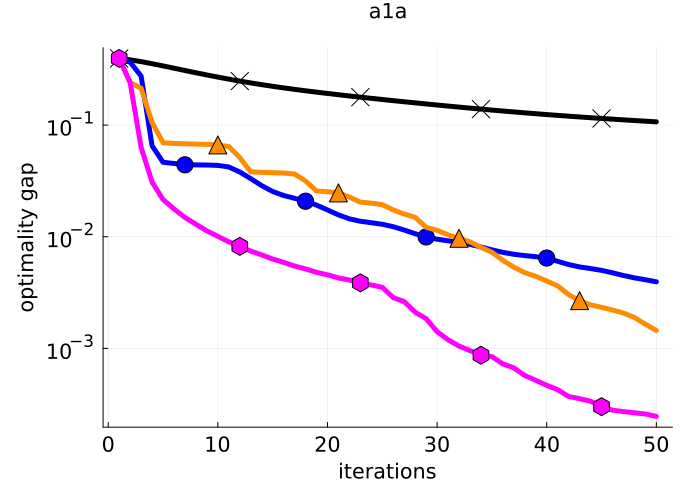}
\includegraphics[width=0.24\textwidth]{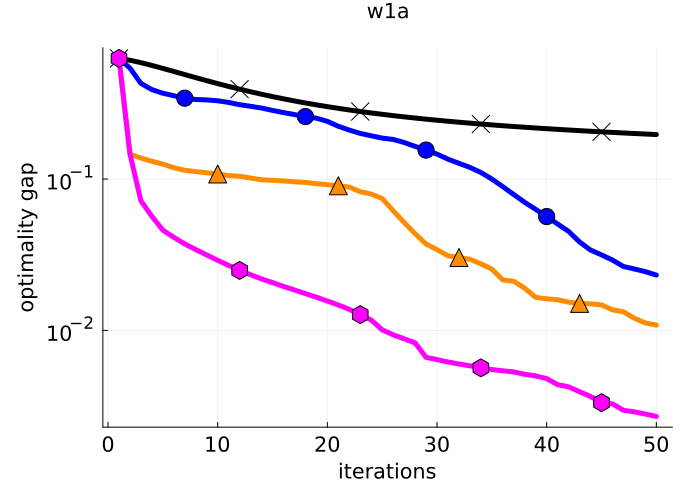}
\centering
\includegraphics[width=.67\textwidth]{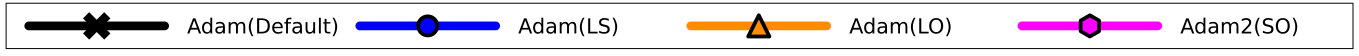}
\caption{Performance of different variations on Adam for fitting logistic regression models. We see that the default Adam settings (black line) tend to perform poorly, while using a line search (blue) or line optimization (orange) improves the performance. The multi-direction method using SO(magenta) dominated the single-direction methods.}
\label{fig:AdamlogReg}
\end{figure}

\begin{figure}
\includegraphics[width=0.24\textwidth]{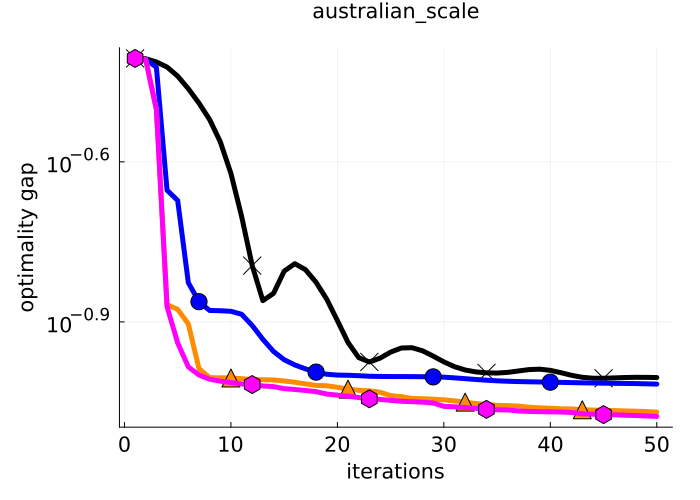}
\includegraphics[width=0.24\textwidth]{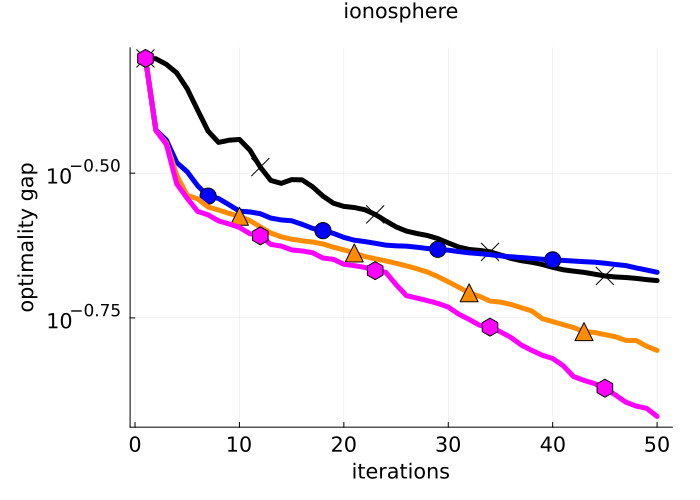}
\includegraphics[width=0.24\textwidth]{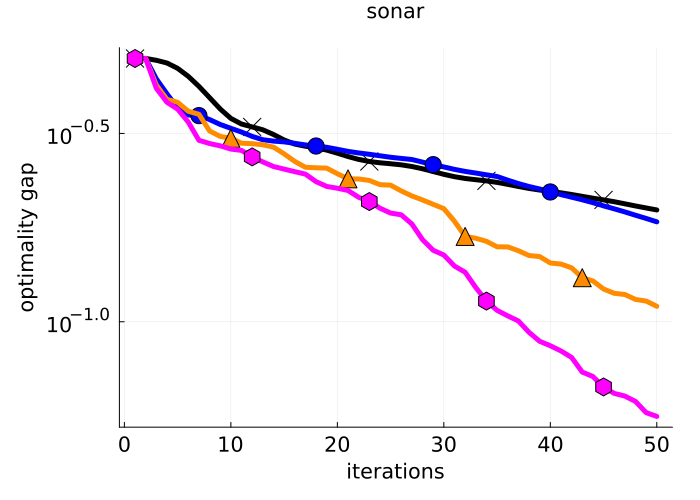}
\includegraphics[width=0.24\textwidth]{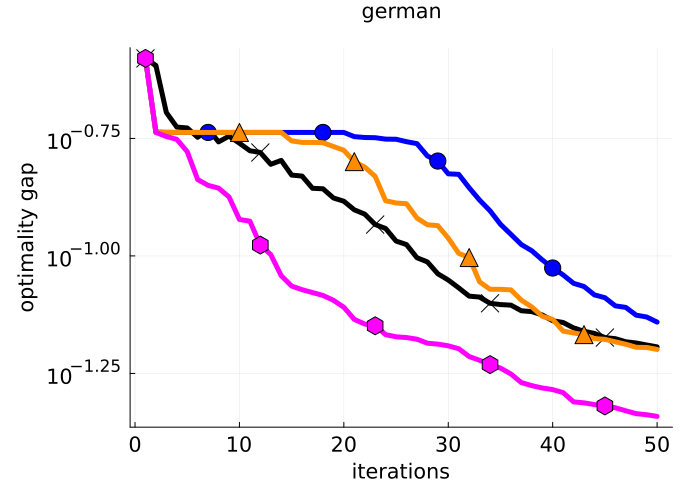}
\includegraphics[width=0.24\textwidth]{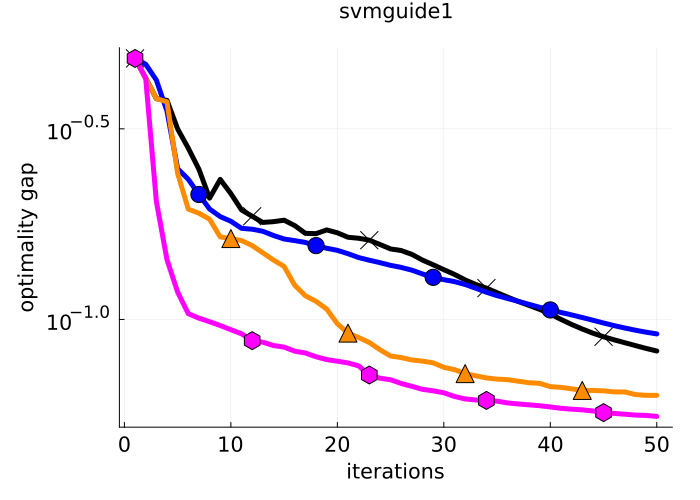}
\includegraphics[width=0.24\textwidth]{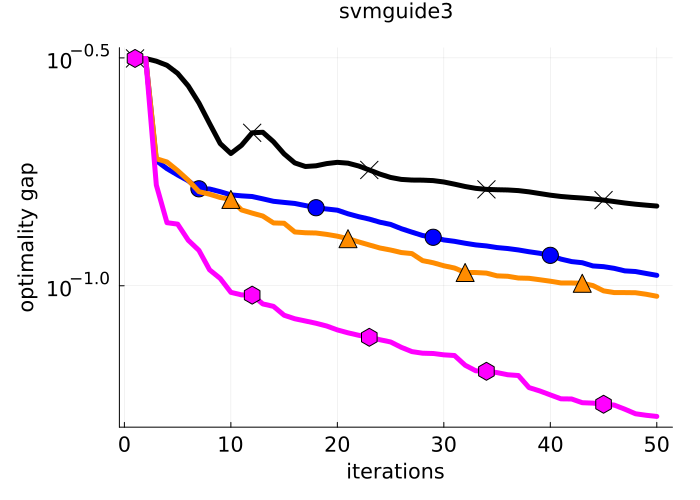}
\includegraphics[width=0.24\textwidth]{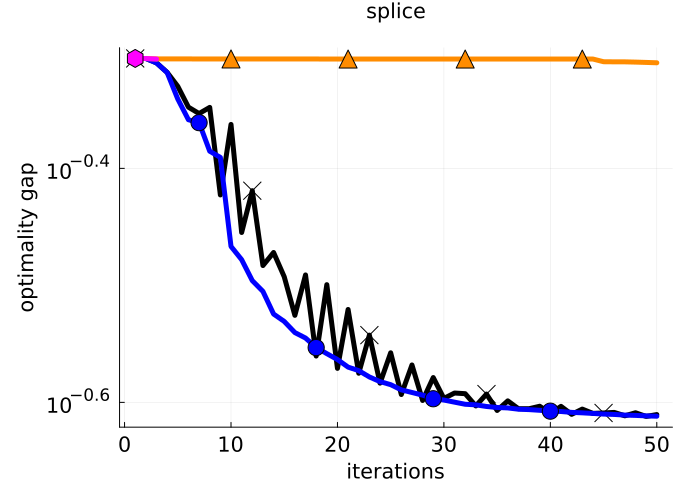}
\includegraphics[width=0.24\textwidth]{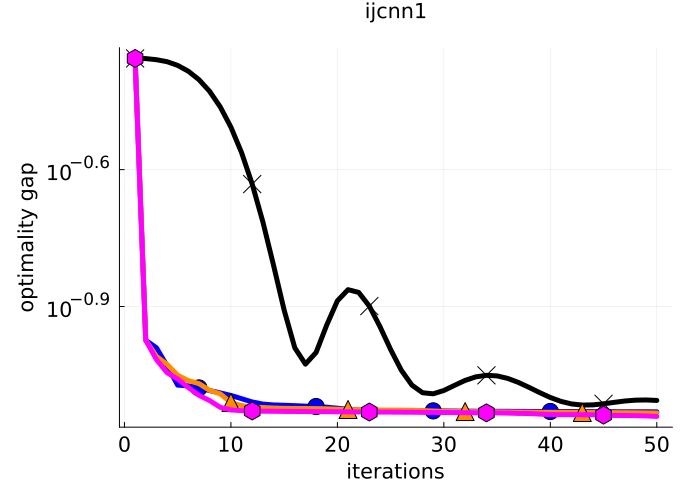}
\includegraphics[width=0.24\textwidth]{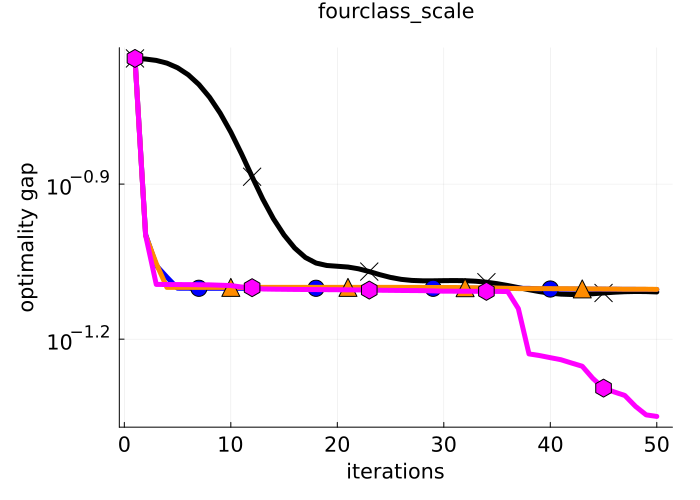}
\includegraphics[width=0.24\textwidth]{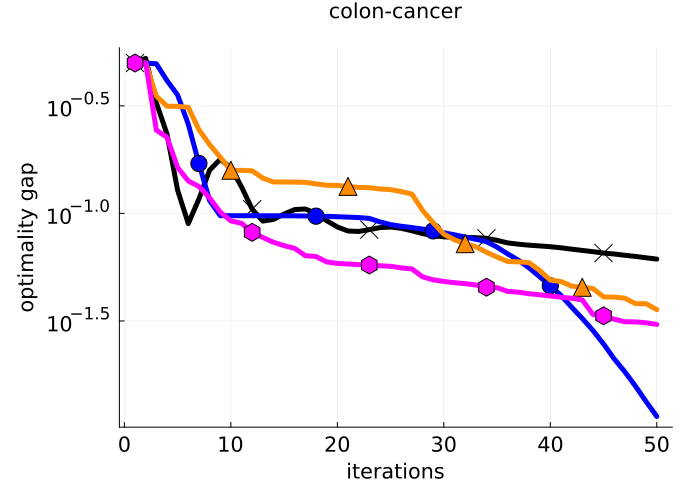}
\includegraphics[width=0.24\textwidth]{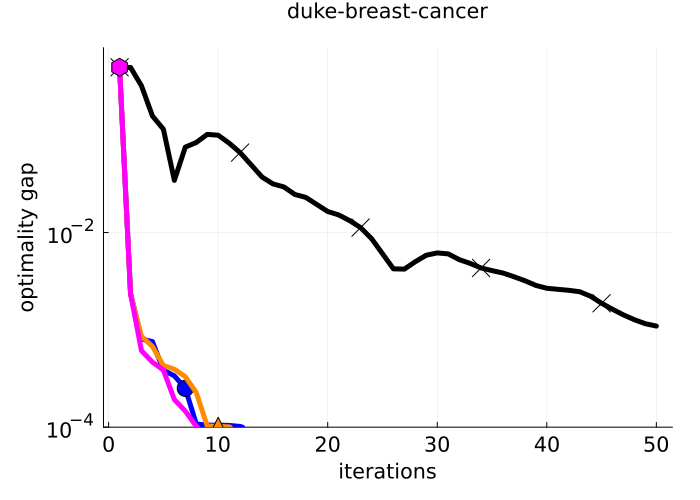}
\includegraphics[width=0.24\textwidth]{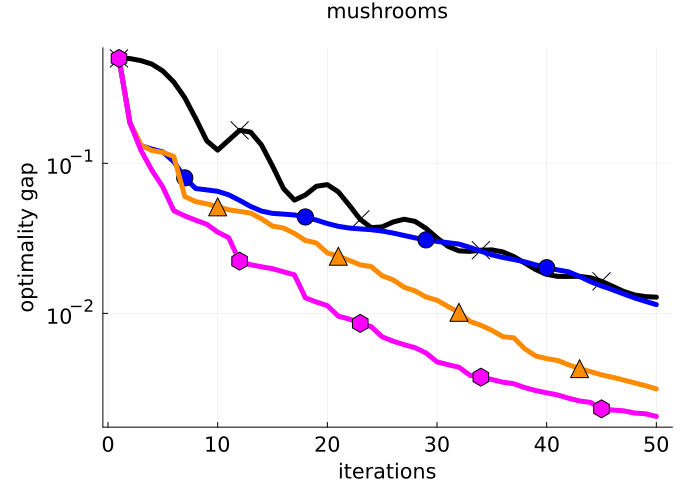}
\includegraphics[width=0.24\textwidth]{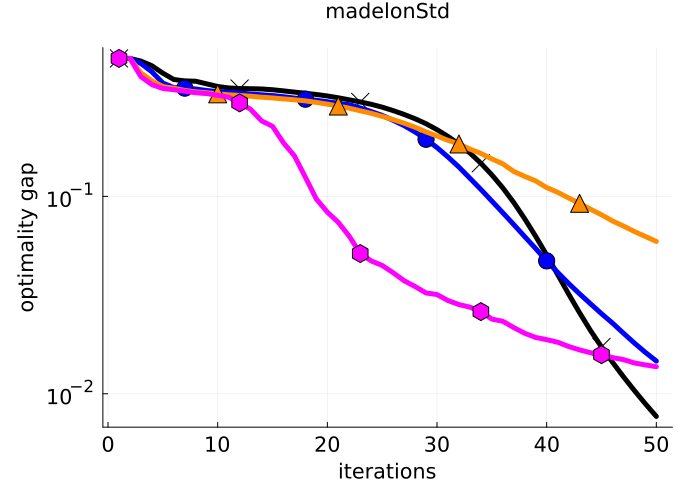}
\includegraphics[width=0.24\textwidth]{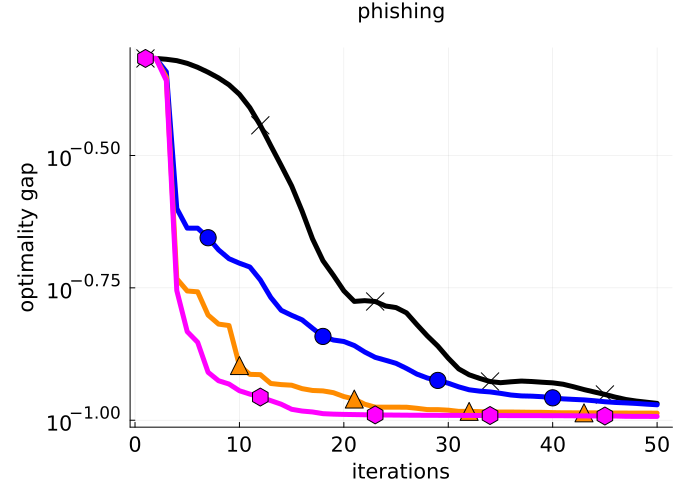}
\includegraphics[width=0.24\textwidth]{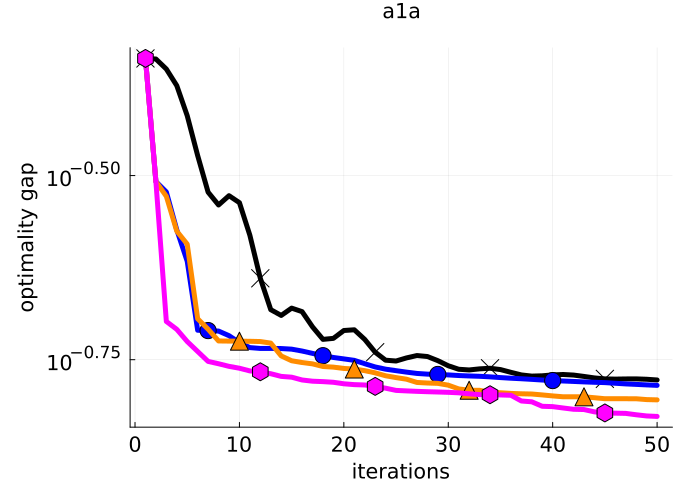}
\includegraphics[width=0.24\textwidth]{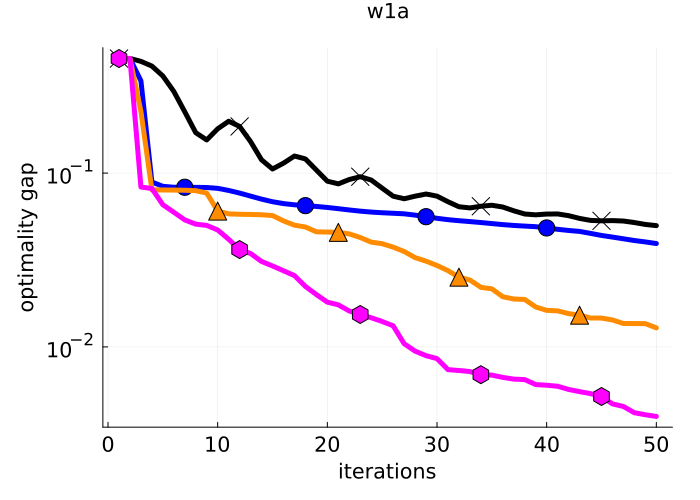}
\centering
\includegraphics[width=.67\textwidth]{AdamLegend.png}
\caption{Performance of different variations on Adam for fitting 2-layer neural networks. We see that the default Adam settings (black line) are more competitive for neural networks than for linear models. The multi-direction method using SO(magenta) often outperformed the single-direction methods.}
\label{fig:Adamnn100}
\end{figure}

We plot the performance of a variety of Adam variants for logistic regression and 2-layer networks in Figures~\ref{fig:AdamlogReg} and~\ref{fig:Adamnn100}. The methods we compare are using a step size of $\alpha=10^{-3}$ as in PyTorch (\texttt{Adam(Default)}), using the strong Wolfe line search to set $\alpha_k$ initialized with the previous step size (\texttt{Adam(LS)}), using LO to set $\alpha_k$ (\texttt{Adam(LO)}), or using SO for update~\eqref{eq:adam2} to optimize over two Adam step sizes (\texttt{Adam2(SO)}). The other Adam hyper-parameters are fixed at their default values.
The Adam direction $d_k$ may not be a descent direction, and on iterations where this happened we made the \texttt{Adam(LS)} search along the direction $-d_k$ instead of $d_k$.
For logistic regression we see that the default step size tends to perform poorly, while the more clever step sizes perform better and the multi-direction method with SO tends to perform the best. For neural networks the performance differences were not as clear; on some datasets the default sep size performed poorly while on other datasets it was competitive. However, on neural networks the SO method still tended to have the best performance.\footnote{An exception to the good performance of \texttt{Adam(SO)} was on the splice dataset, where on the second iteration the method converged numerically to a stationary point and halted. We found that this degenerate behaviour was not observed if we added regularization.}

\begin{figure}
\includegraphics[width=0.24\textwidth]{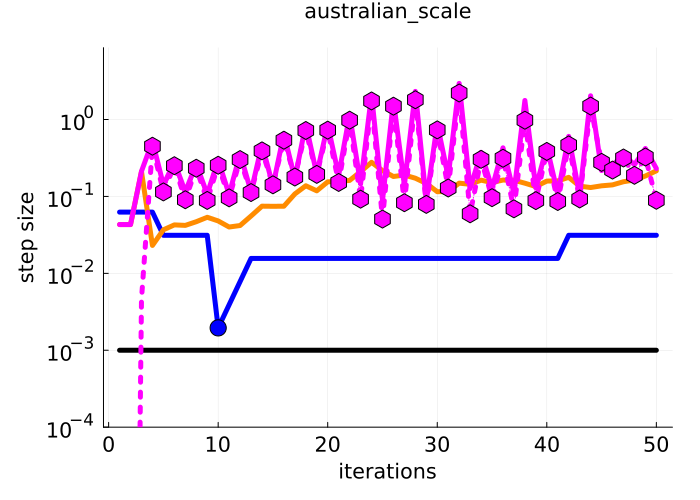}
\includegraphics[width=0.24\textwidth]{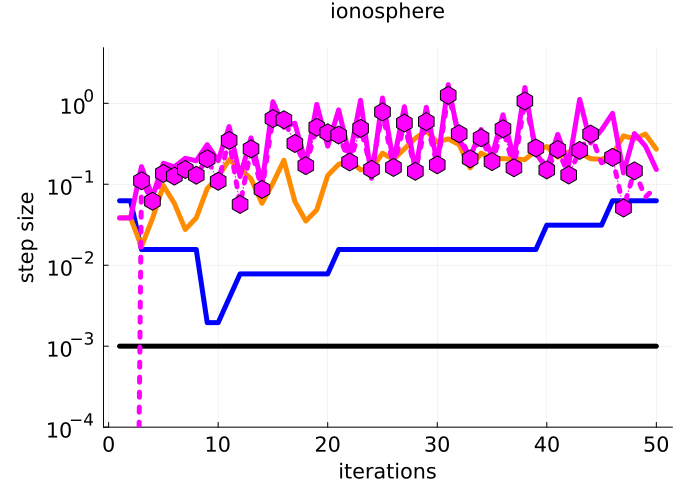}
\includegraphics[width=0.24\textwidth]{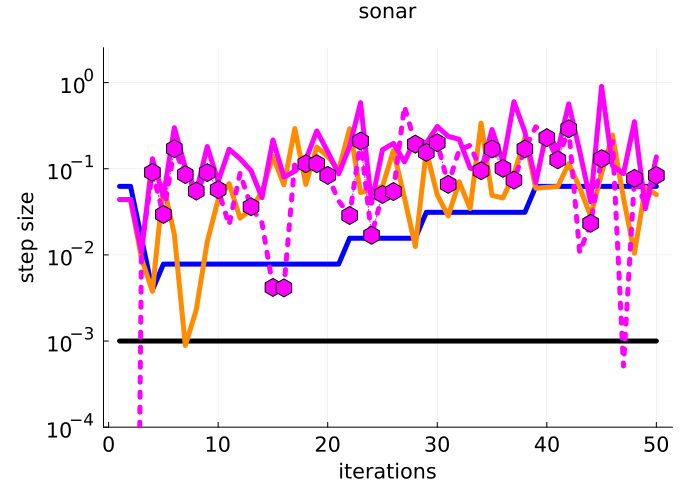}
\includegraphics[width=0.24\textwidth]{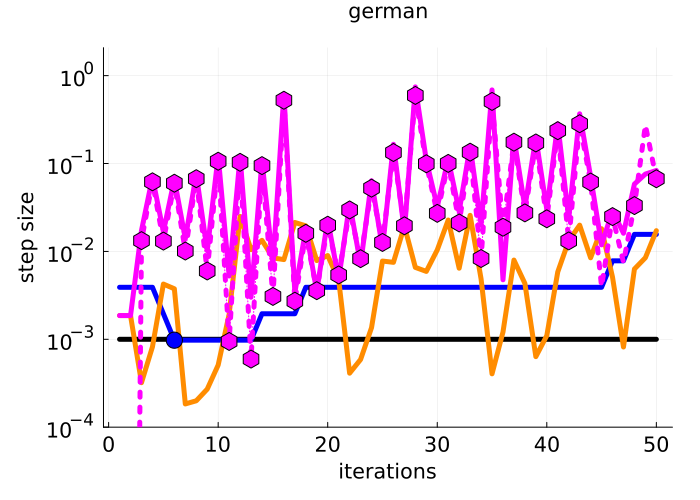}
\includegraphics[width=0.24\textwidth]{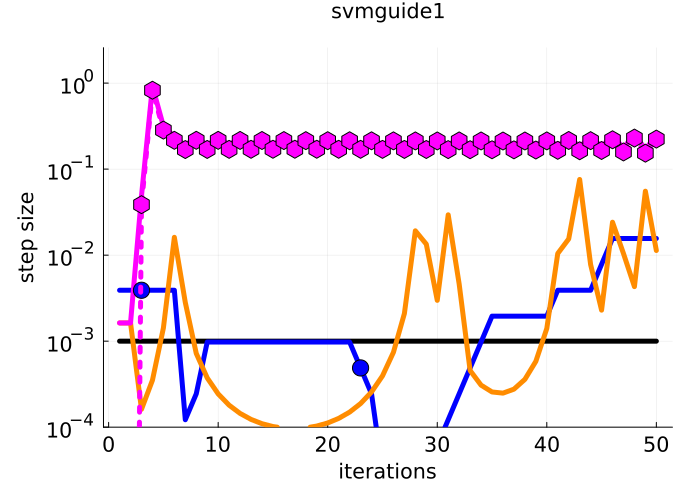}
\includegraphics[width=0.24\textwidth]{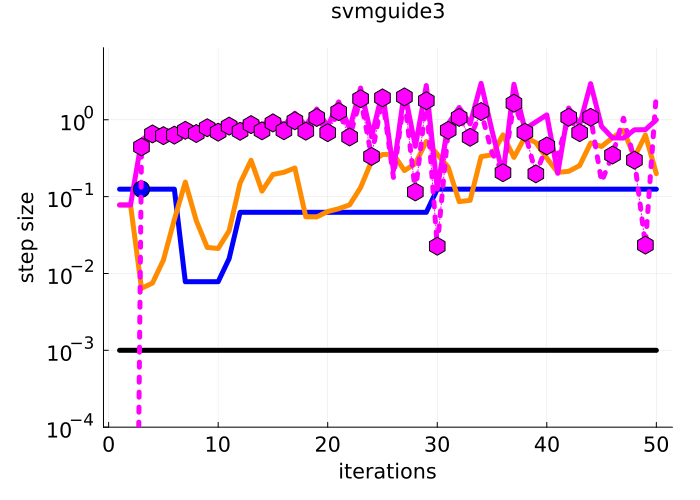}
\includegraphics[width=0.24\textwidth]{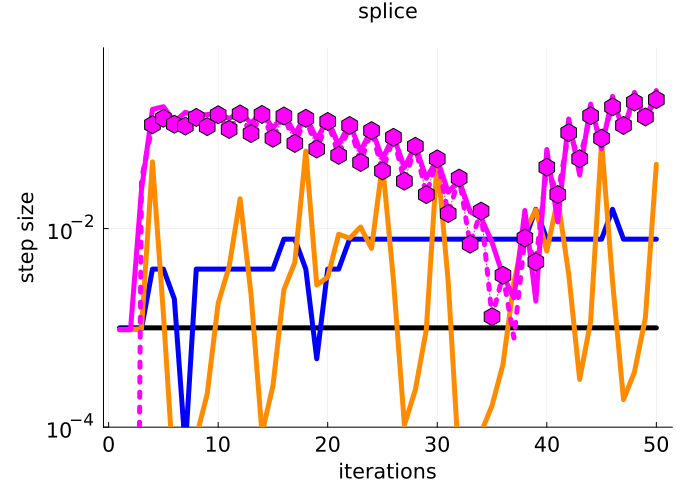}
\includegraphics[width=0.24\textwidth]{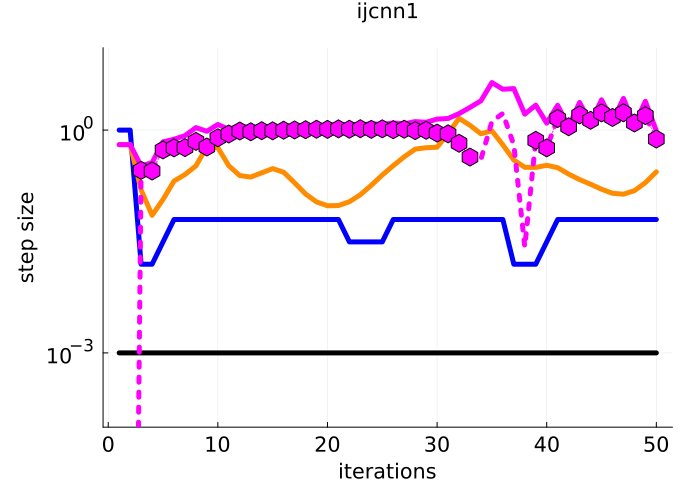}
\includegraphics[width=0.24\textwidth]{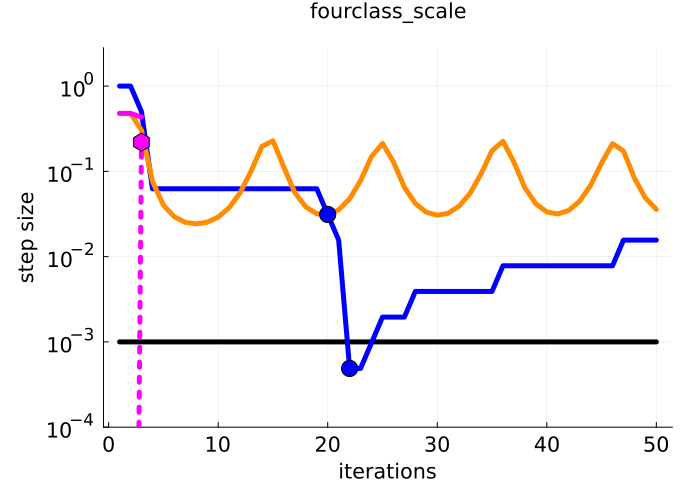}
\includegraphics[width=0.24\textwidth]{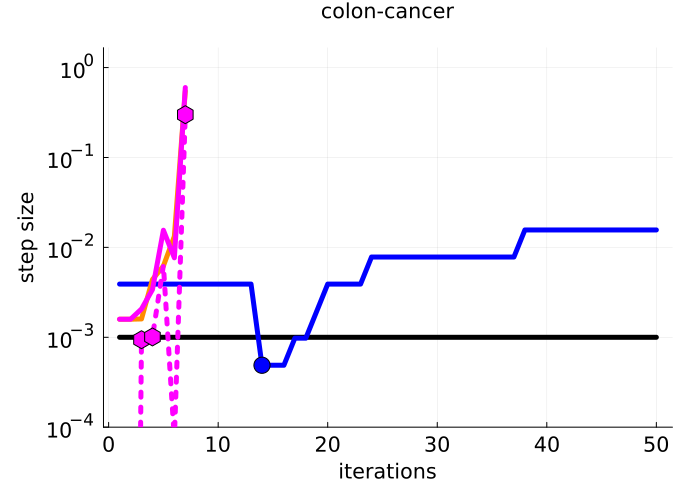}
\includegraphics[width=0.24\textwidth]{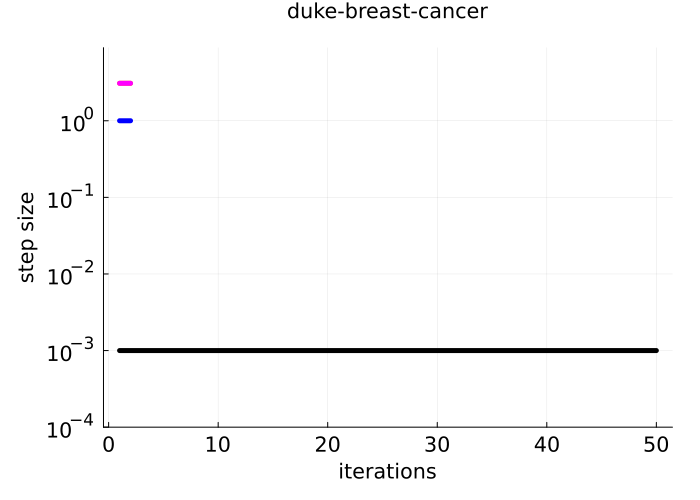}
\includegraphics[width=0.24\textwidth]{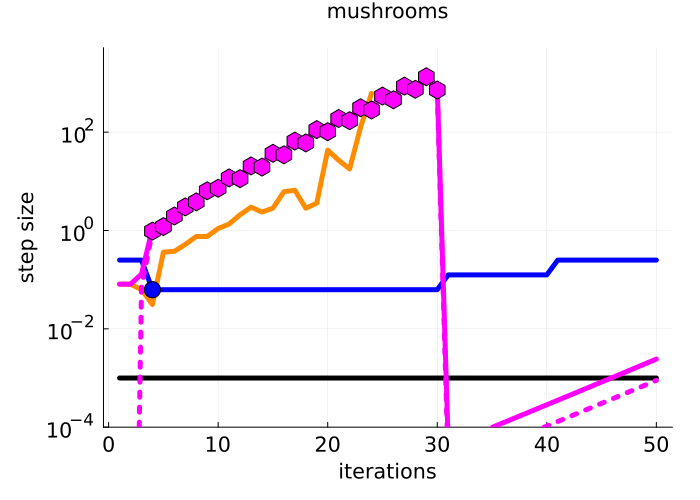}
\includegraphics[width=0.24\textwidth]{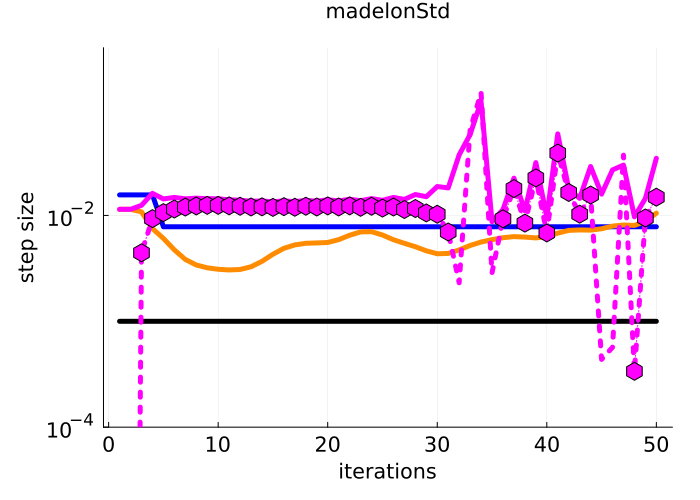}
\includegraphics[width=0.24\textwidth]{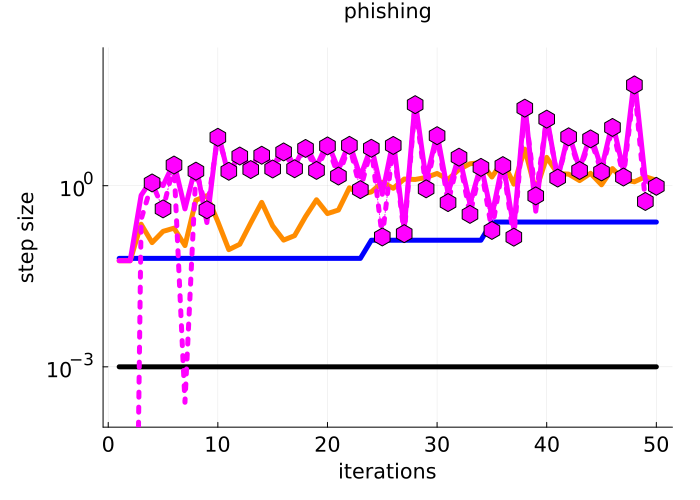}
\includegraphics[width=0.24\textwidth]{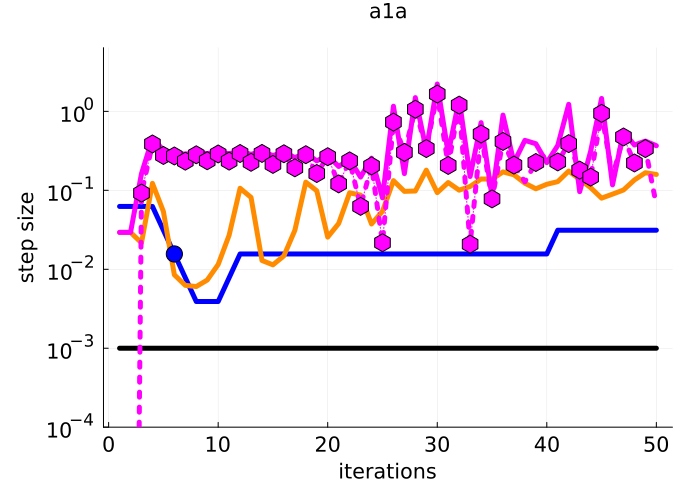}
\includegraphics[width=0.24\textwidth]{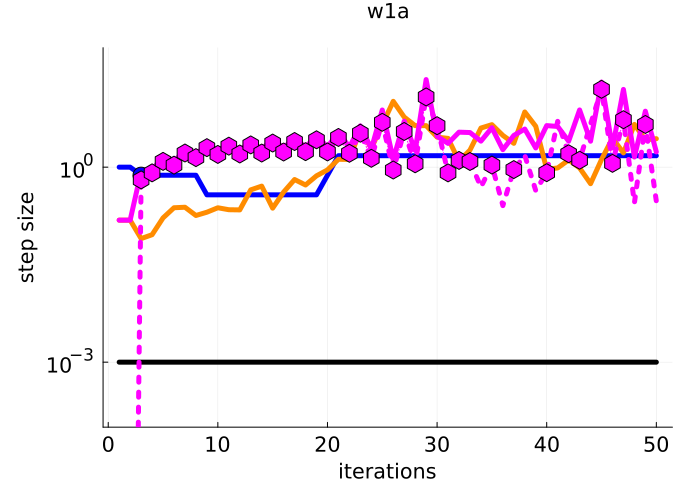}
\centering
\includegraphics[width=.67\textwidth]{AdamLegend.png}
\caption{Step sizes of different variations on Adam for fitting logistic regression models. We plot the absolute value, and include markers to indicate iterations where the step was negative. The dashed line is for the step size on $d_{k-1}$ in the SO method. We see that the default step size tends to be too small, that the line search tends to use larger constant values over several iterations, that the LO seems to show periodic behaviour with a period longer than 2, and that the SO method typically shows periodic behaviour with a period of 2 where positive and negative step sizes of similar magnitudes are used.}
\label{fig:AdamsteplogReg}
\end{figure}

\begin{figure}
\includegraphics[width=0.24\textwidth]{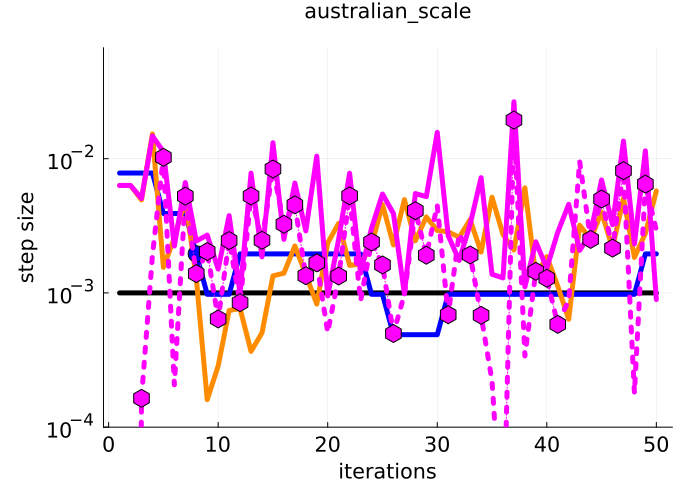}
\includegraphics[width=0.24\textwidth]{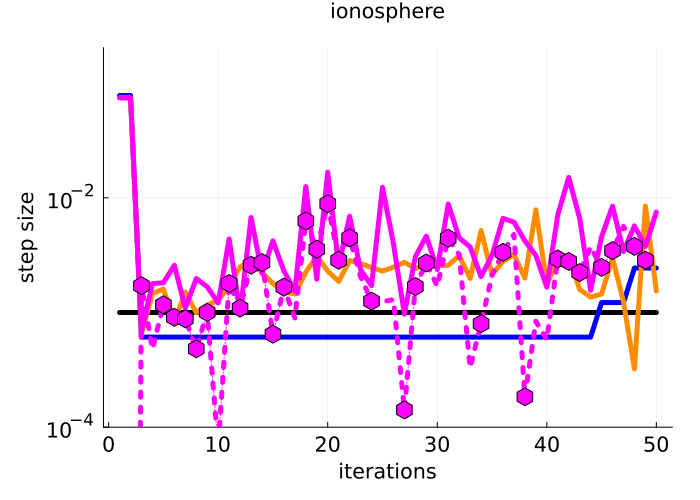}
\includegraphics[width=0.24\textwidth]{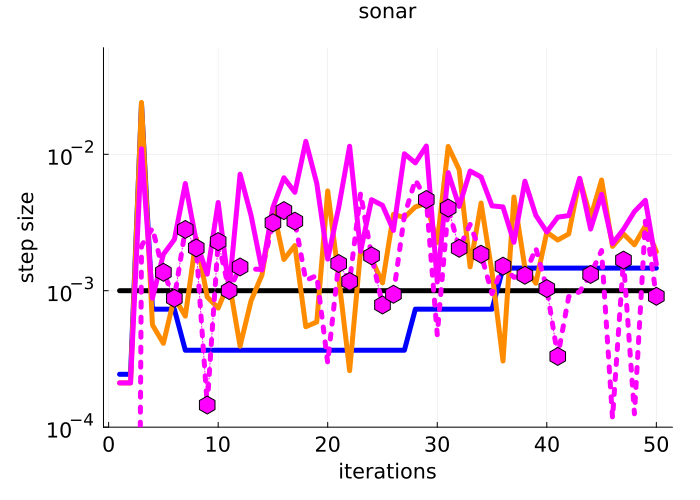}
\includegraphics[width=0.24\textwidth]{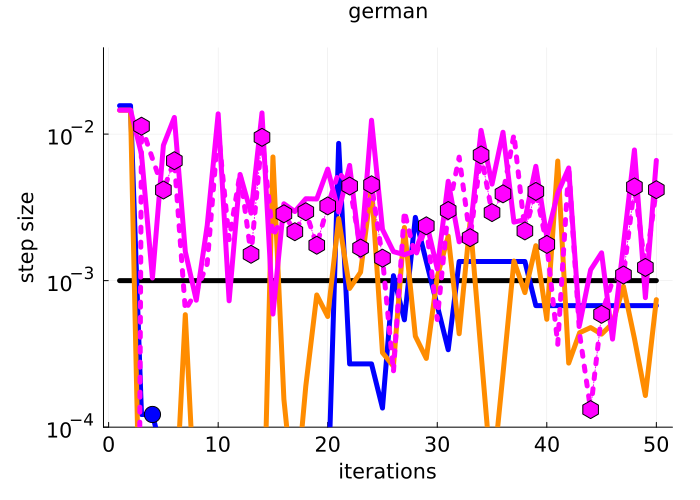}
\includegraphics[width=0.24\textwidth]{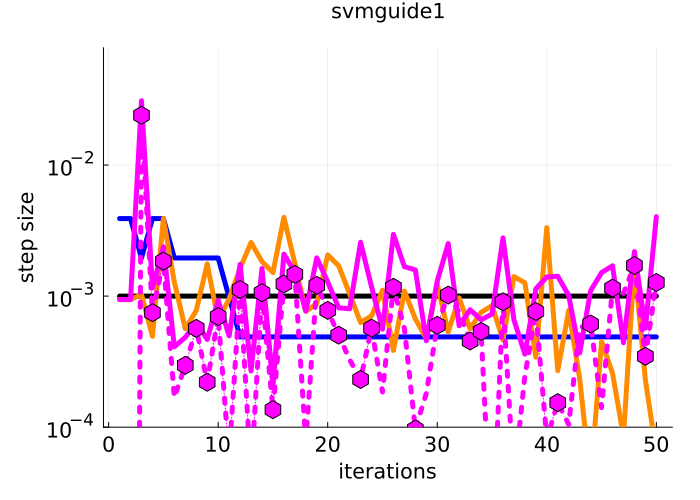}
\includegraphics[width=0.24\textwidth]{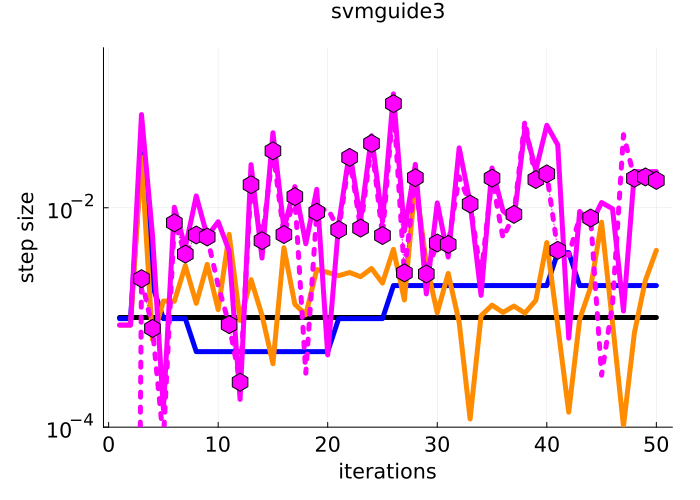}
\includegraphics[width=0.24\textwidth]{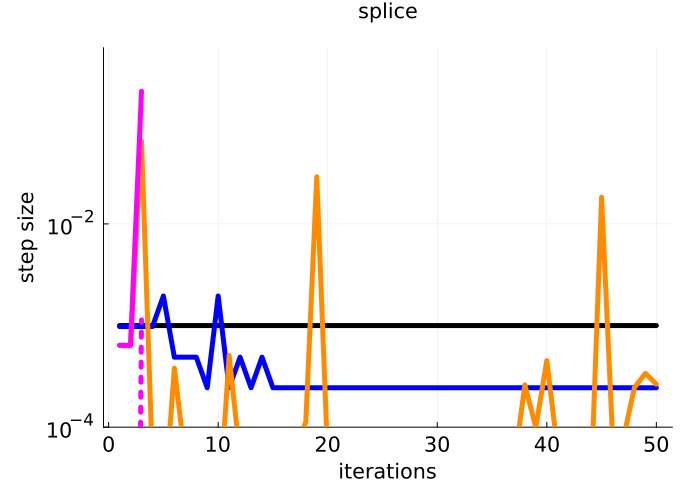}
\includegraphics[width=0.24\textwidth]{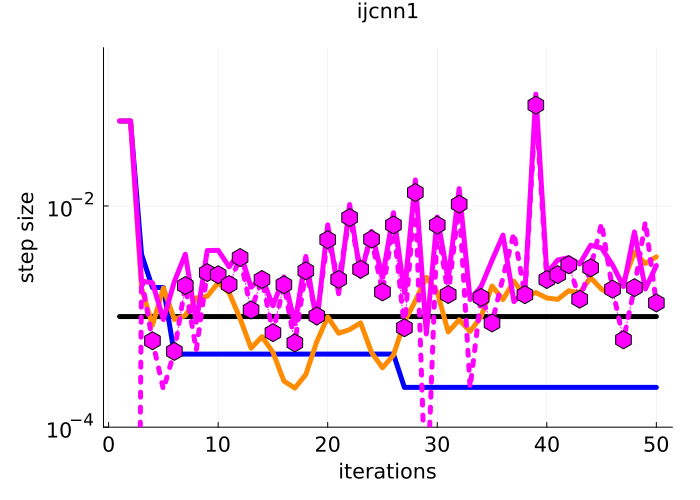}
\includegraphics[width=0.24\textwidth]{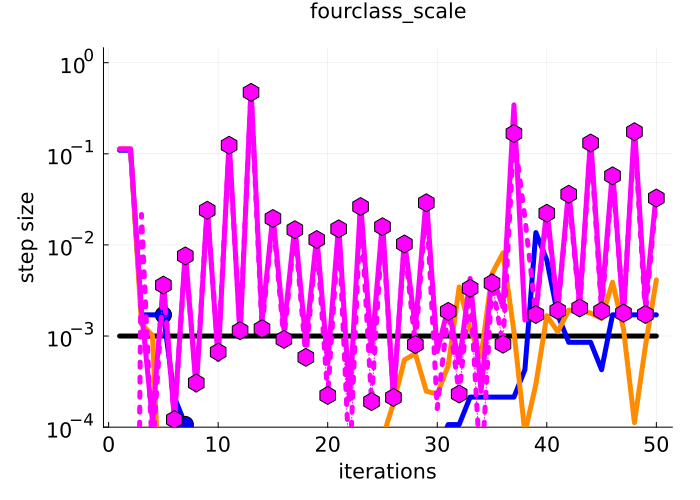}
\includegraphics[width=0.24\textwidth]{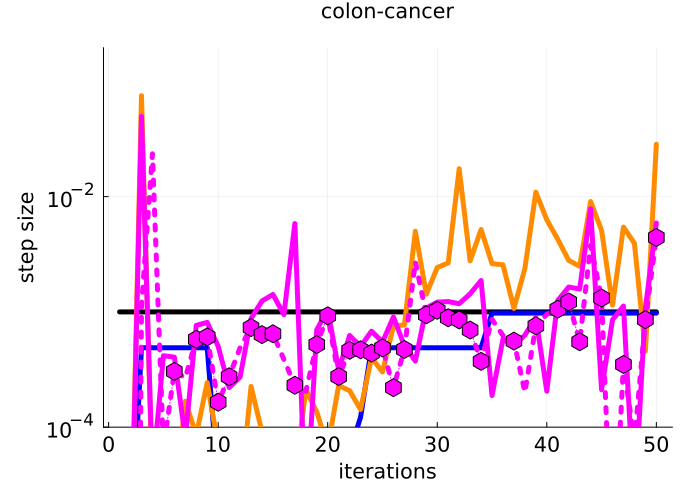}
\includegraphics[width=0.24\textwidth]{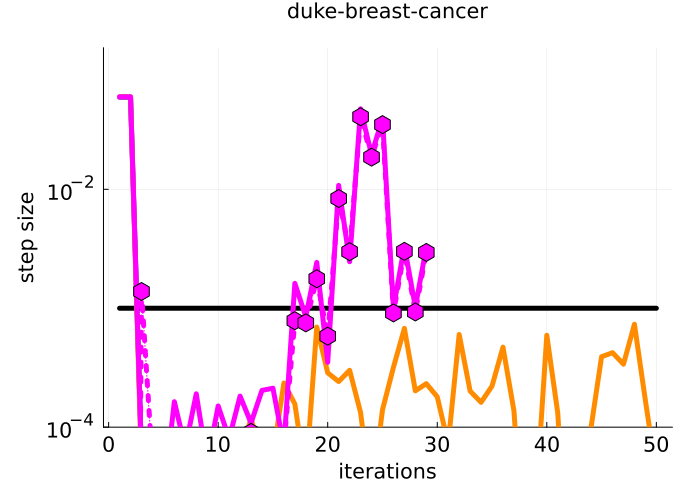}
\includegraphics[width=0.24\textwidth]{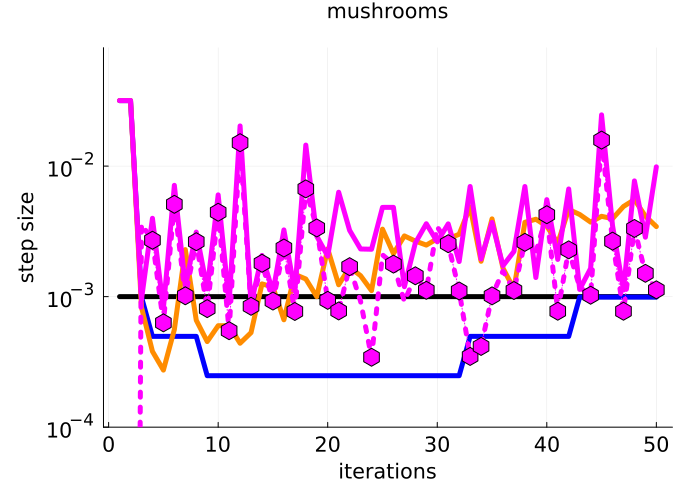}
\includegraphics[width=0.24\textwidth]{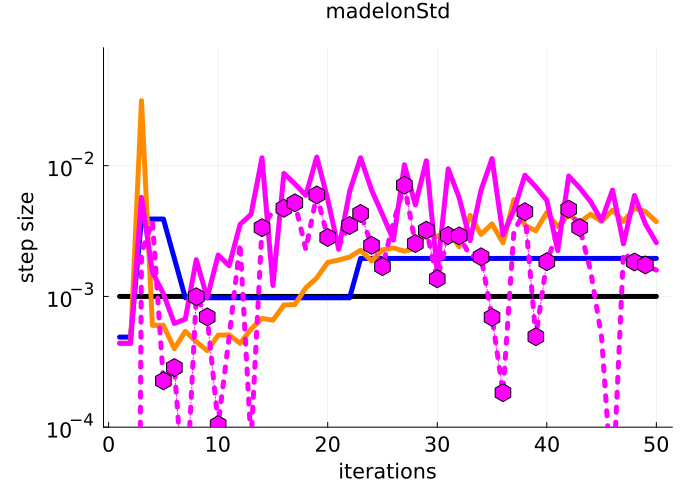}
\includegraphics[width=0.24\textwidth]{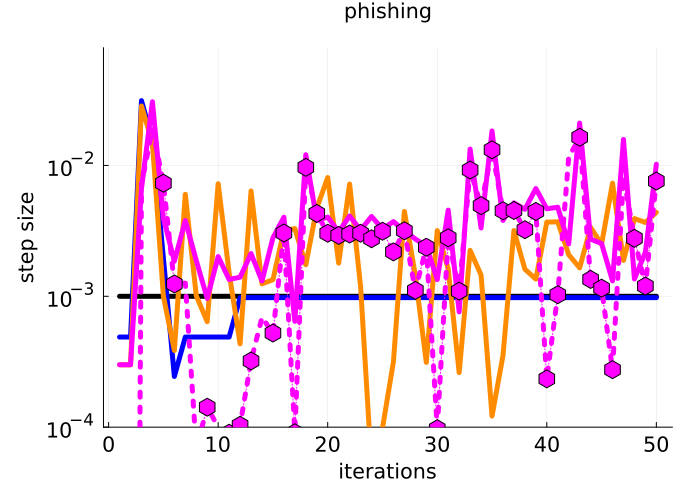}
\includegraphics[width=0.24\textwidth]{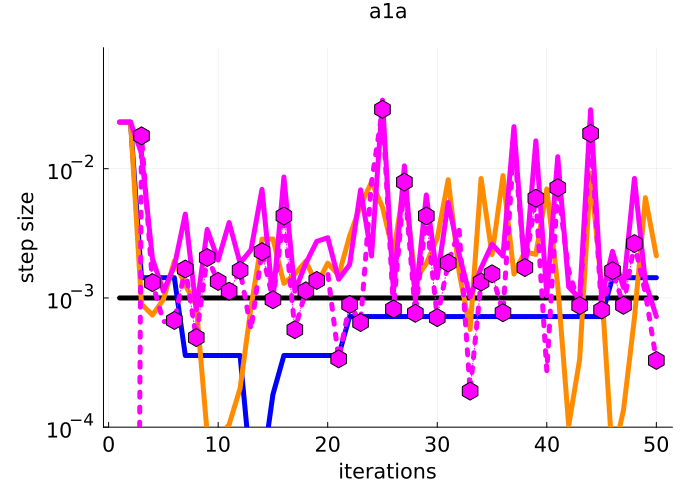}
\includegraphics[width=0.24\textwidth]{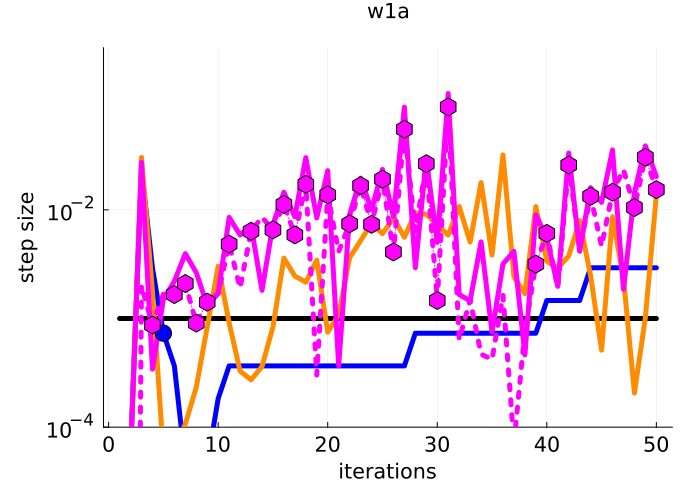}
\centering
\includegraphics[width=.67\textwidth]{AdamLegend.png}
\caption{Performance of different first-order methods for 2-layer networks. We plot the absolute value, and include markers to indicate iterations where the step was negative. The dashed line is for the step size on $d_{k-1}$ in the SO method. We see that the default step size tends to be in a similar range to the other methods, that the line search tends to use constant values over several iterations, the LO is consistently changing the step size, and that the SO method often shows periodic behaviour with a period of 2 where positive and negative step sizes of similar magnitudes are used.}
\label{fig:Adamstepnn100}
\end{figure}

The relative performance differences of the different Adam variants can partially be explained by looking at the step size plots Figures~\ref{fig:AdamsteplogReg} and~\ref{fig:Adamstepnn100}. For logistic regression we see that the default step is often too small while for neural networks the default step size appears more reasonable. We also see that the line search method often uses the same step size for many iterations (and occasionally used a negative step size), while the LO method was constantly changing the step size. Periodic behaviour is observed in the LO method, but the period appears larger than it did for gradient decent with LO. Finally, the SO method tended to use step sizes with similar magnitudes on $d_k$ and $d_{k-1}$, but on many iterations it used a positive step size on $d_k$ and a negative step size on $d_{k-1}$. Thus, the \texttt{Adam(SO)} method often seemed to approximate the difference between successive Adam directions in order to make more progress in decreasing the objective. 

\subsection{Gradient, Quasi-Newton, and Adam with[out] Subspace Optimization}

\begin{figure}
\includegraphics[width=0.24\textwidth]{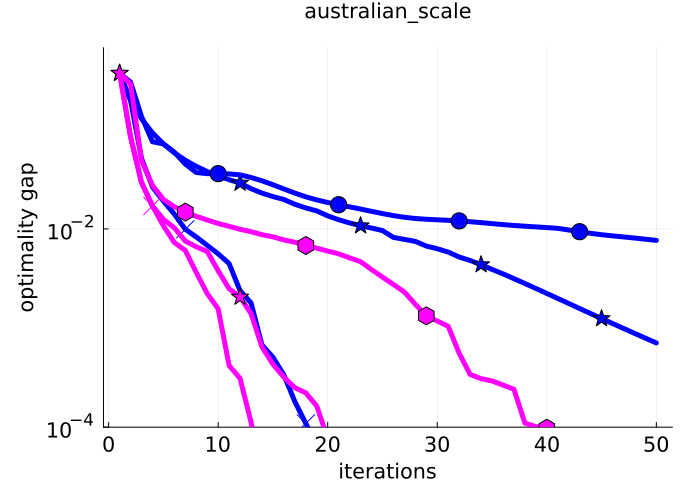}
\includegraphics[width=0.24\textwidth]{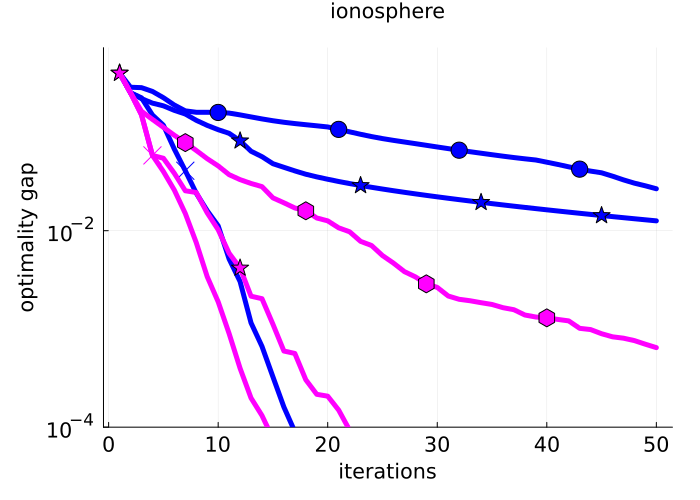}
\includegraphics[width=0.24\textwidth]{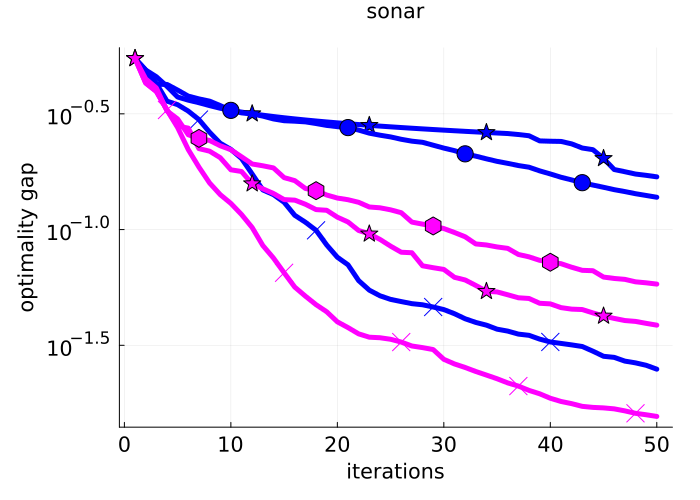}
\includegraphics[width=0.24\textwidth]{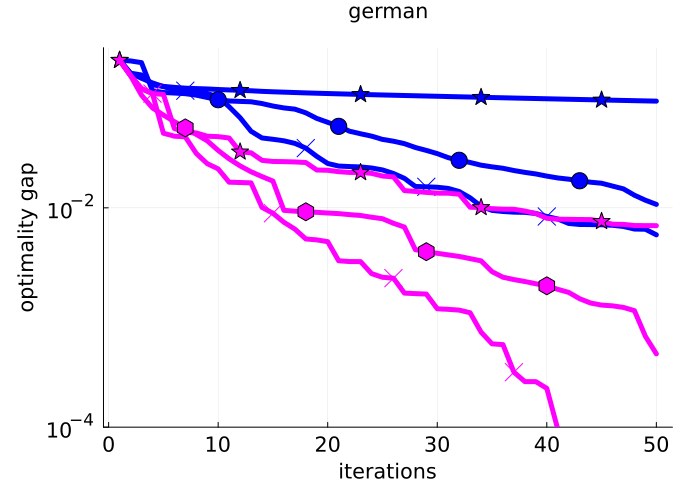}
\includegraphics[width=0.24\textwidth]{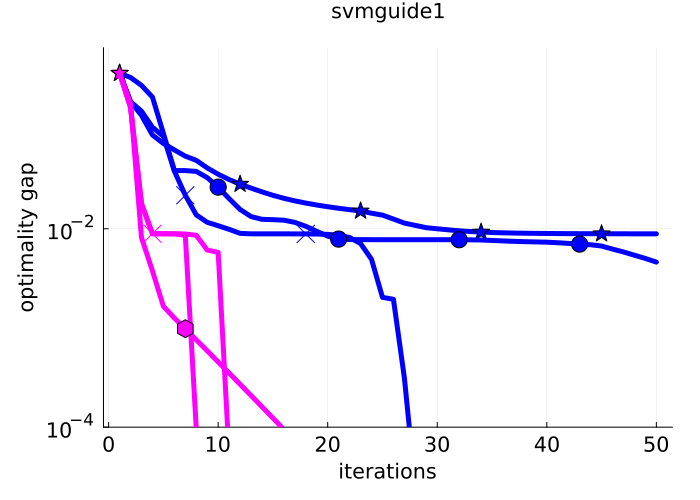}
\includegraphics[width=0.24\textwidth]{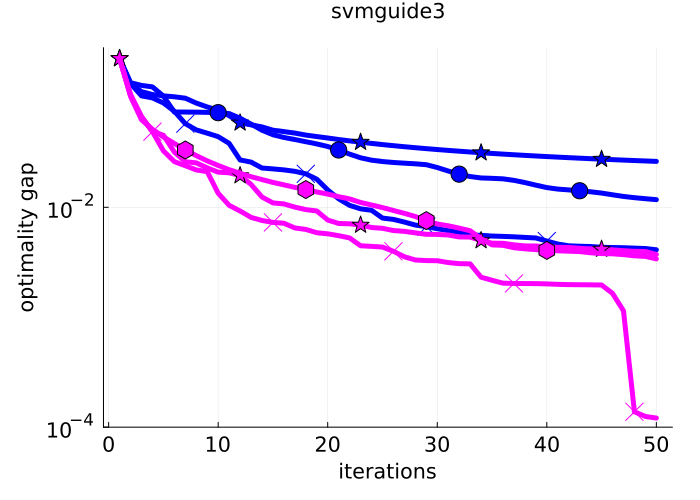}
\includegraphics[width=0.24\textwidth]{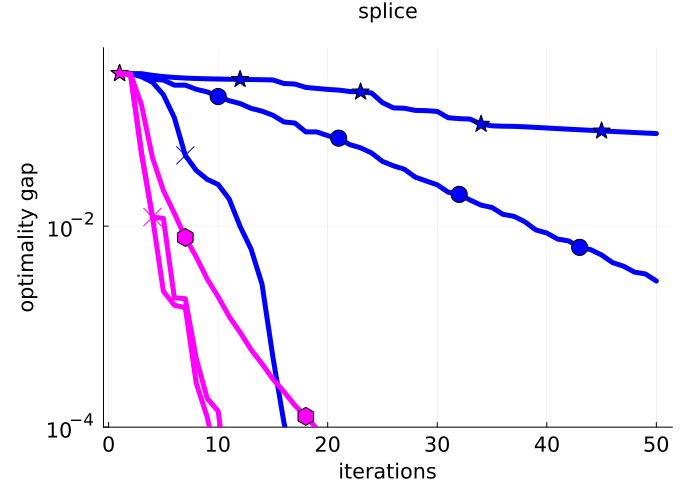}
\includegraphics[width=0.24\textwidth]{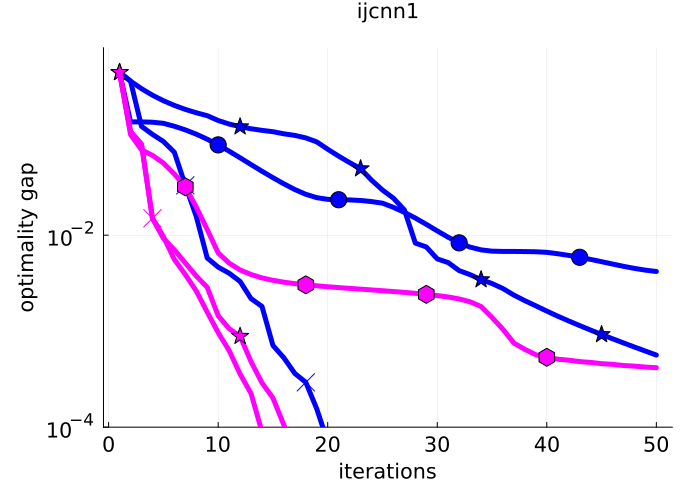}
\includegraphics[width=0.24\textwidth]{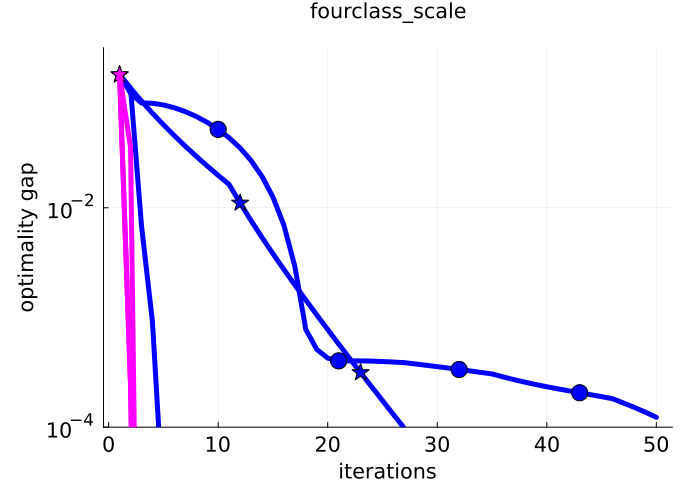}
\includegraphics[width=0.24\textwidth]{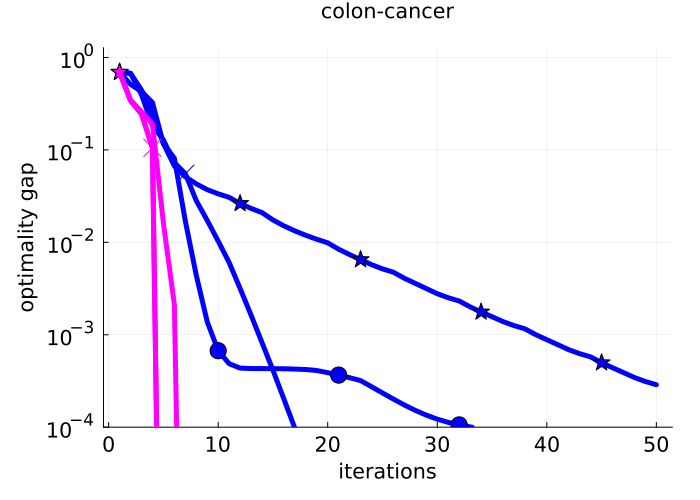}
\includegraphics[width=0.24\textwidth]{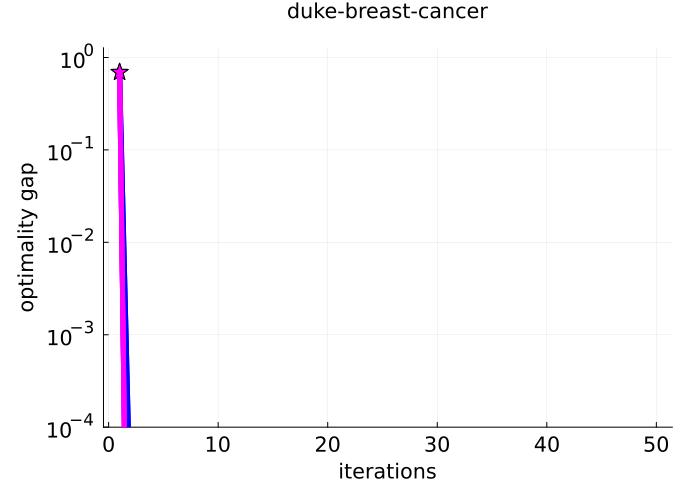}
\includegraphics[width=0.24\textwidth]{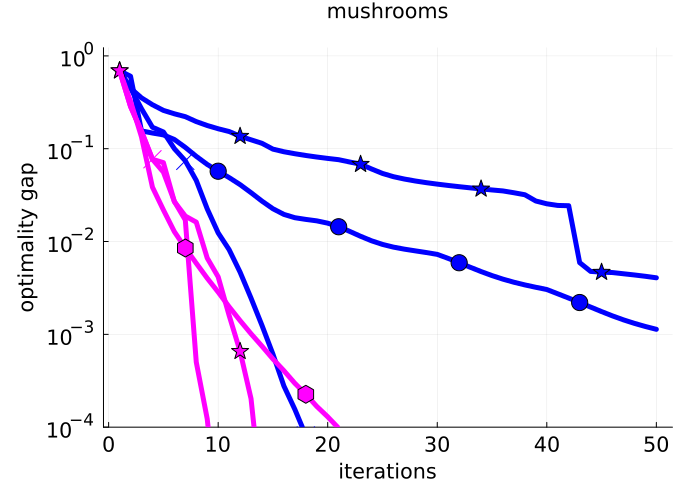}
\includegraphics[width=0.24\textwidth]{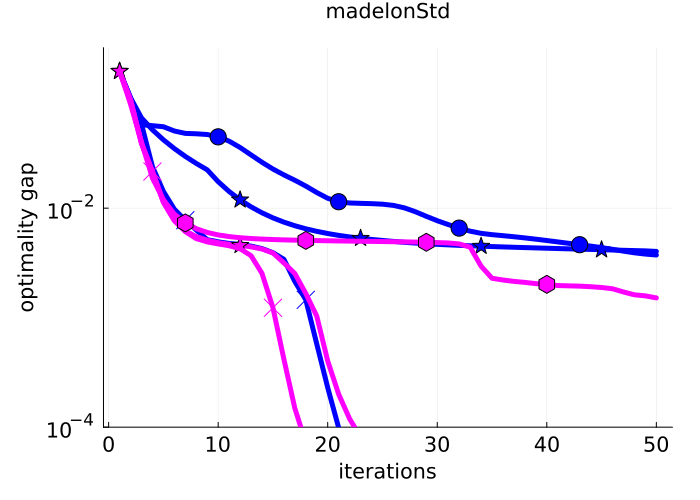}
\includegraphics[width=0.24\textwidth]{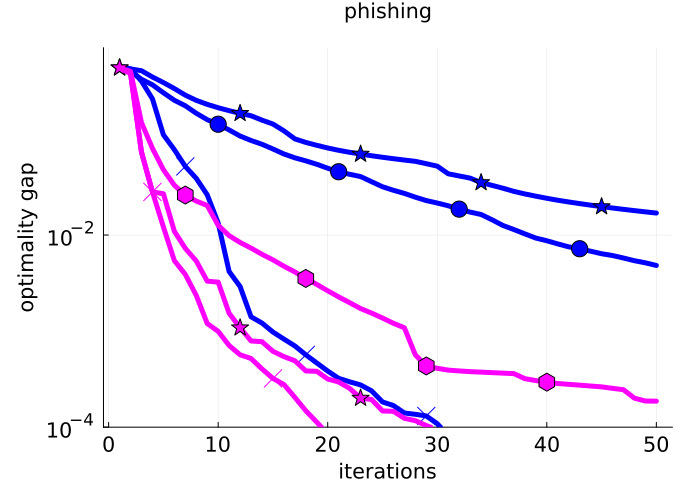}
\includegraphics[width=0.24\textwidth]{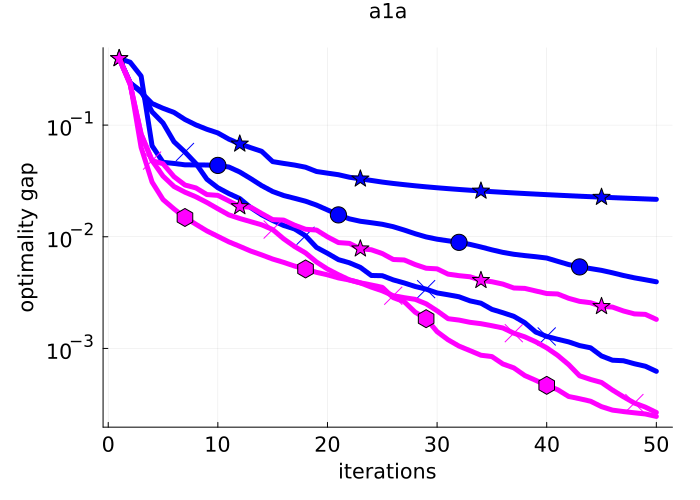}
\includegraphics[width=0.24\textwidth]{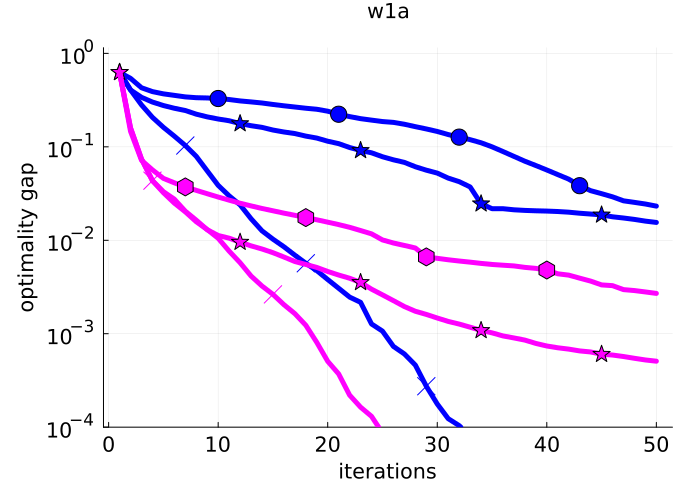}
\centering
\includegraphics[width=\textwidth]{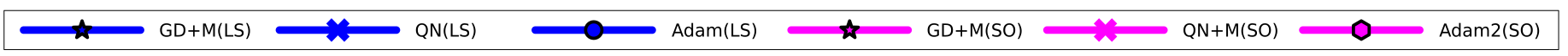}
\caption{Performance of different gradient-based, quasi-Newton, and Adam methods for fitting logistic regression models. The blue lines use a classic line search that does not exploit the problem structure while the purple lines exploit the problem structure to perform SO. We see that the SO methods tend to dominate, with only the \texttt{QN(LS)} method coming close to the performance of the SO methods. Among the SO methods, the \texttt{QN+M(SO)} method has among the best performance across datasets.}
\label{fig:bestlogReg}
\end{figure}

\begin{figure}
\includegraphics[width=0.24\textwidth]{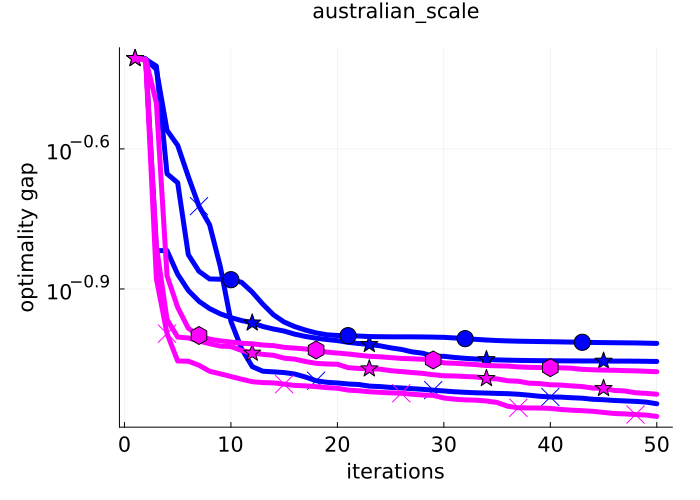}
\includegraphics[width=0.24\textwidth]{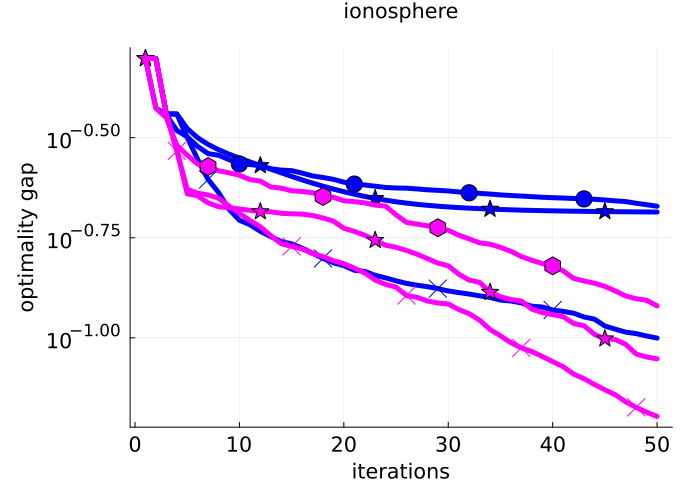}
\includegraphics[width=0.24\textwidth]{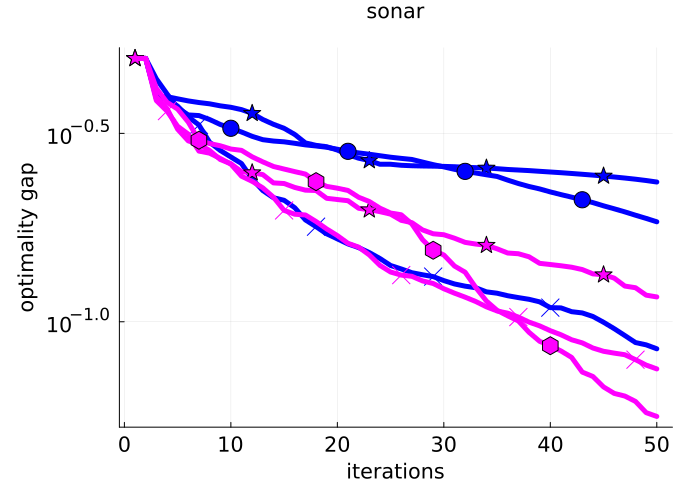}
\includegraphics[width=0.24\textwidth]{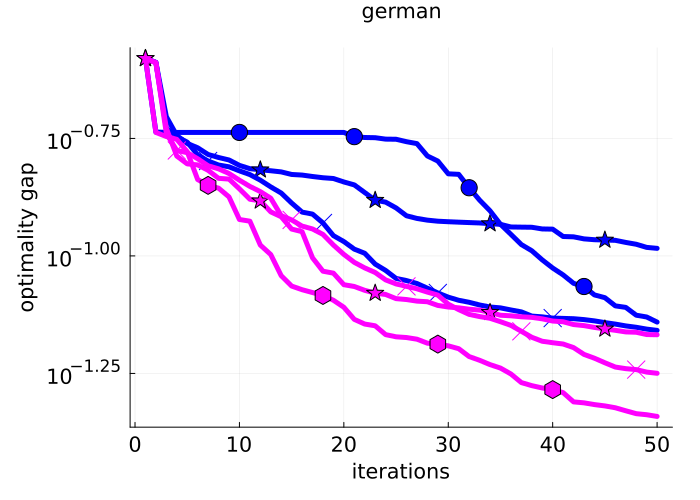}
\includegraphics[width=0.24\textwidth]{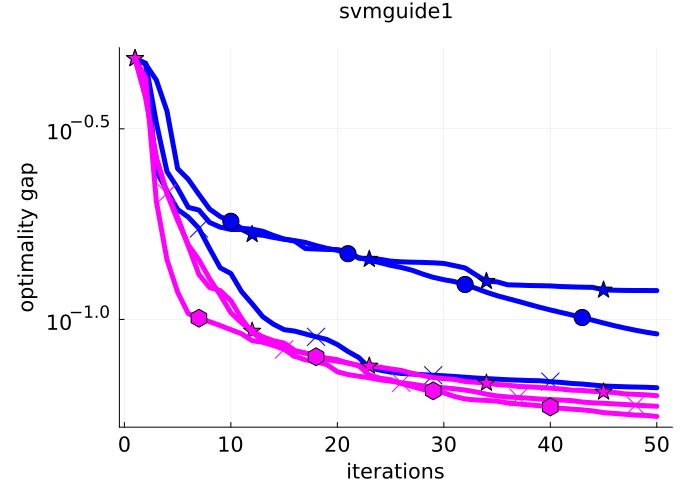}
\includegraphics[width=0.24\textwidth]{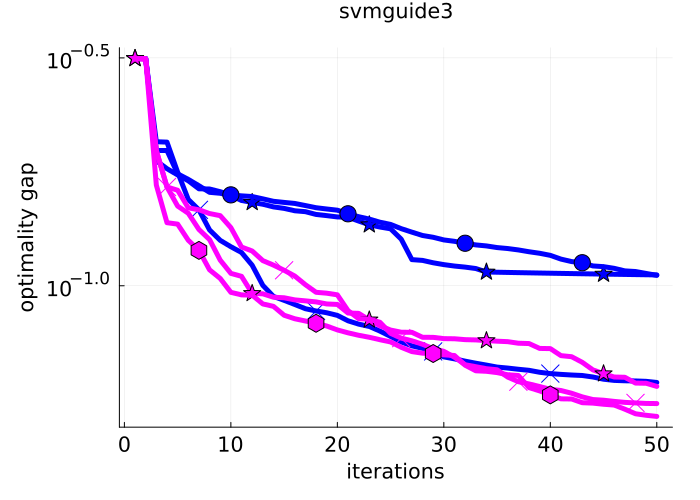}
\includegraphics[width=0.24\textwidth]{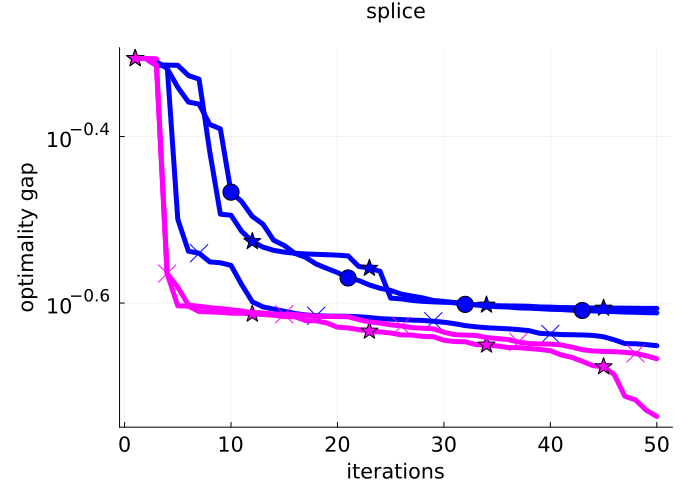}
\includegraphics[width=0.24\textwidth]{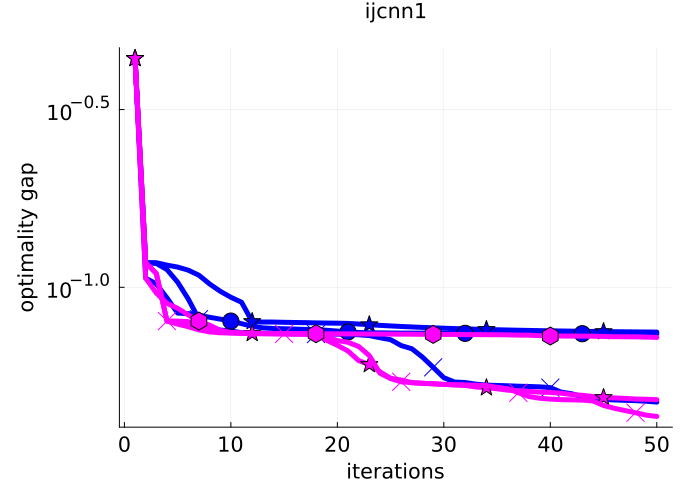}
\includegraphics[width=0.24\textwidth]{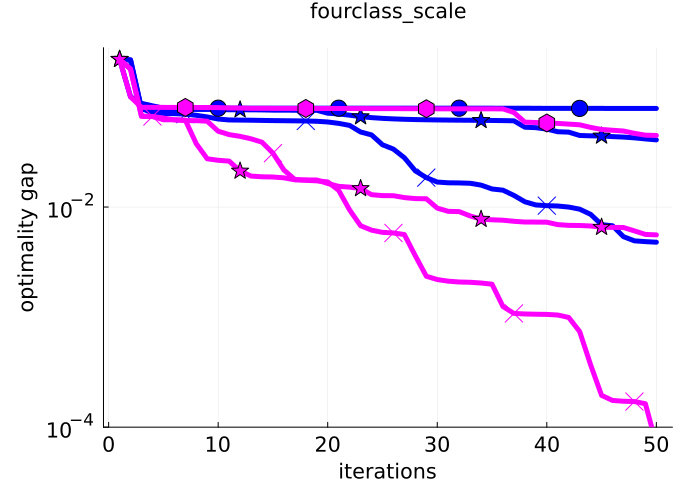}
\includegraphics[width=0.24\textwidth]{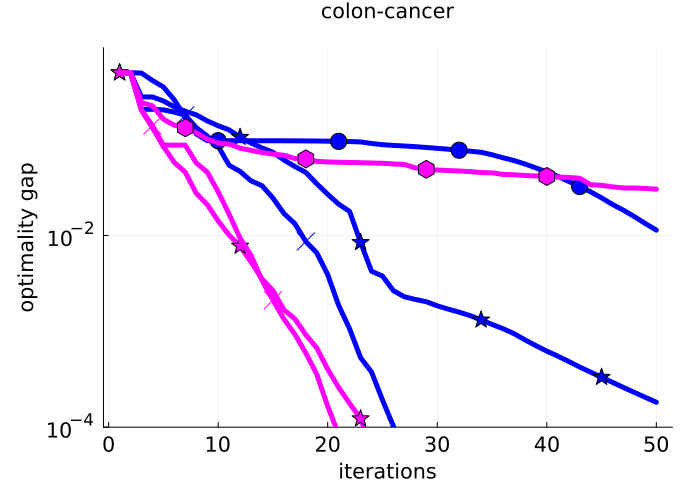}
\includegraphics[width=0.24\textwidth]{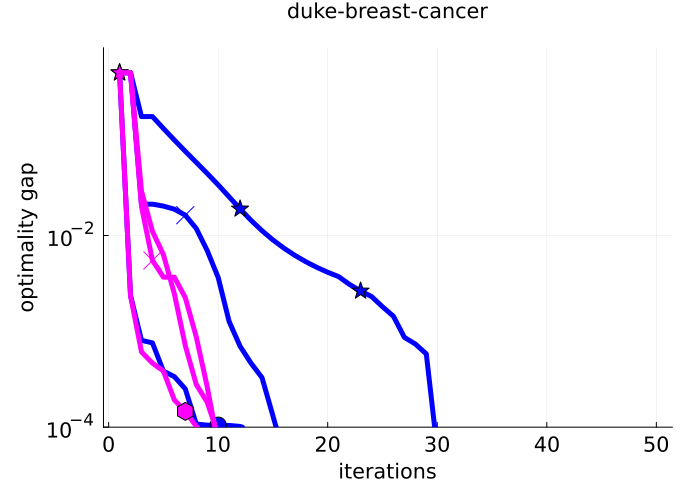}
\includegraphics[width=0.24\textwidth]{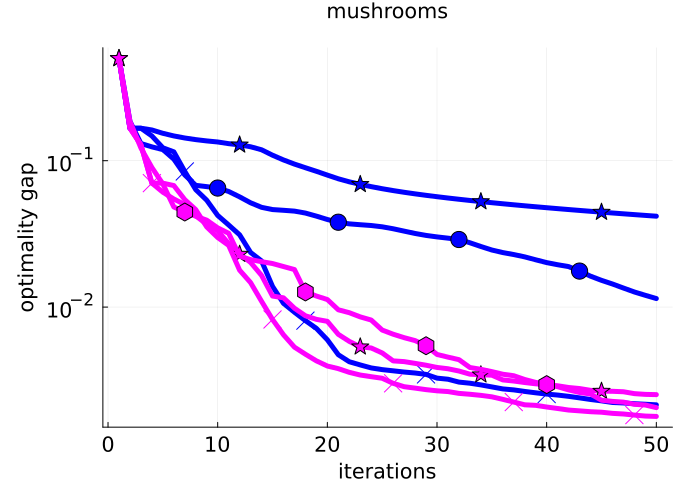}
\includegraphics[width=0.24\textwidth]{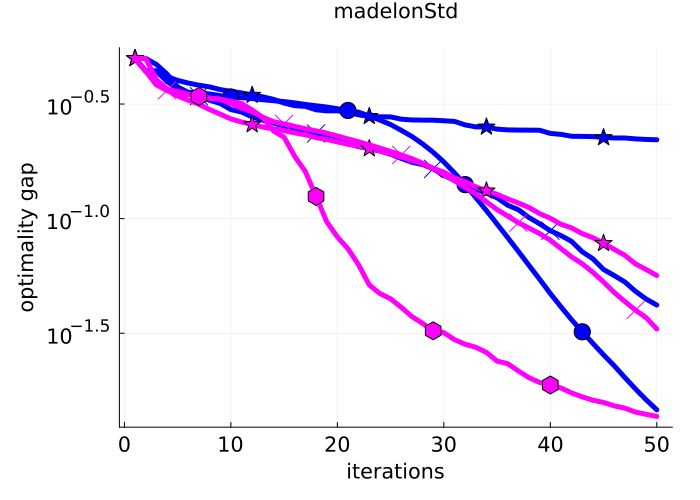}
\includegraphics[width=0.24\textwidth]{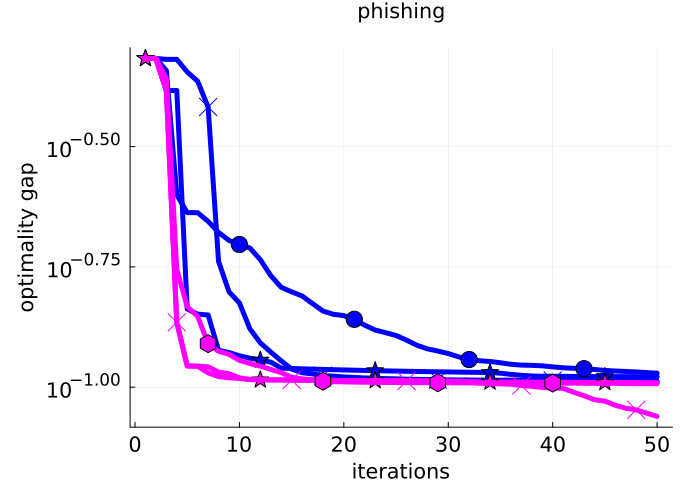}
\includegraphics[width=0.24\textwidth]{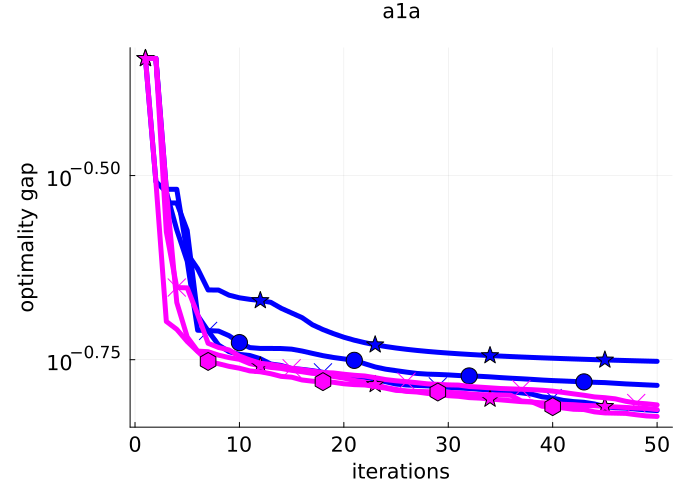}
\includegraphics[width=0.24\textwidth]{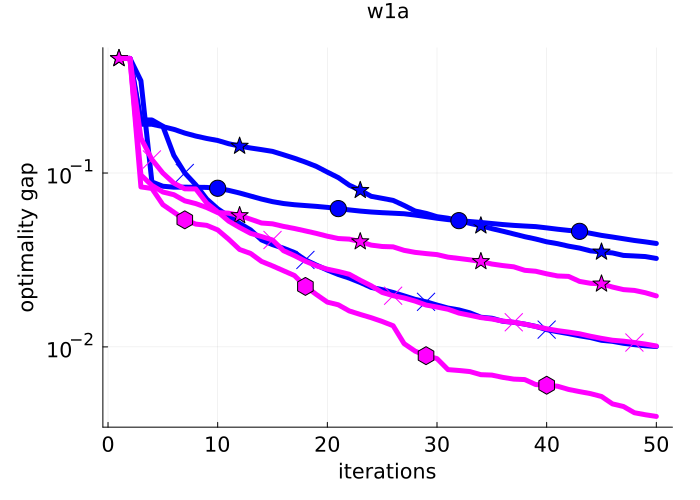}
\centering
\includegraphics[width=\textwidth]{BestLegend.png}
\caption{Performance of different gradient-based, quasi-Newton, and Adam methods for fitting 2-layer neural networks. The blue lines use a classic line search that does not exploit the problem structure while the purple lines exploit the problem structure to perform SO. We see that SO methods tend to perform better across most datasets. On most datasets the \texttt{QN+M(SO)} method has among the best performance, except on some datasets while the \texttt{Adam2(SO)} method performs better.}
\label{fig:bestnn100}
\end{figure}

Our experiments indicate that gradient descent with momentum benefit from LO and SO, that quasi-Newton methods benefit from LO while SO allows us to add an optimized momentum term, and that Adam often benefits from LO while using SO to combine multiple Adam directions yields further improvements. However, up to this point we have not compared the three styles of methods (gradient, quasi-Newton, and Adam) to each other. In Figures~\ref{fig:bestlogReg} and~\ref{fig:bestnn100} we compare the performance of what we view as the best-performing method of each of the 3 types if we exploit LO/SO and the best performing methods if we do not exploit the problem structure. 

These ``best methods'' plots indicate that the choice regarding \textbf{whether or not to use SO seems more important than choosing between gradient/quasi-Newton/Adam} methods. Among the methods not exploiting SO, only the \texttt{QN(LS)} method seems to consistently come close to the performance of the SO methods (this could be due to its natural step size of $\alpha_k=1$). On the logistic regression problems the \texttt{QN+M(SO)}  method seems to be the most effective, always performing at least as well or better than all other methods. On the neural network problems the \texttt{QN+M(SO)} was also consistently among the best methods, but on a small number of datasets the multi-direction \texttt{Adam2(SO)} method performed the best.

\section{Beyond Deterministic, SO-Friendly, and Differentiable Problems}
\label{sec:discuss}

In this work we show that SO can be efficiently applied to a restricted class of neural networks, and it seems to generically improve the performance of deterministic optimization algorithms. However, many deep learning architectures do not follow these structures and are trained with SGD variants. In this section
we discuss directions that future work could take to exploit SO within stochastic algorithms and to use SO to speed up training neural networks that are not SO-friendly.

\subsection{Using Subspace Optimization within SGD}

SGD methods use variants of the GD+M algorithm where $\nabla f(w_k)$ is computed based on a subset of the training examples, which can be advantageous if the number of training examples is large. If we use LO or SO to set the SGD step size(s), SGD does not converge in general. However, it has recently been shown that  some LS variants do converge for over-parameterized models and that this gives impressive performance in various settings~\citep{Vaswani2019,galli2023don}. We believe that SO could be used to further improve the performance of SGD for over-parameterized models. For under-parameterized models SO could be used with growing batch sizes~\citep{Friedlander2012,Bollapragada2018,De2016}, with alternating SGD steps~\citep{richardson2016seboost}, or 
to implement better models of the objective~\citep{Asi2019}. We could alternately simply use a very-large fixed batch size, such as the 3.2 million used by GPT-3~\citep{Brown2020}, since large batch sizes makes SGD behave like a deterministic algorithm up to a fixed accuracy.

\subsection{Using Subspace Optimization for Deep Learning}

Many deep learning models are not SO-friendly. Nevertheless, we may be able to use SO to implement methods that make more progress per iteration than SGD. For example, prox-linear methods partially linearize functions that are written as compositions (such as neural networks) and can guarantee more progress per iteration than gradient descent~\citep{Nesterov2007,Lewis2016,Duchi2018,Drusvyatskiy2019}. The iterations of the prox-linear method are LCPs, and thus this is a natural direction to explore to obtain better optimization algorithms for deep learning.

An alternative way to use SO for deep learning is with layer-wise training. It has been shown that layer-wise training allows building deep models that are competitive on ImageNet~\citep{belilovsky2019greedy}. Layer-wise training can be formulated as an LCP and may be a way to remove the frustrations associated with training deep networks (due to sensitivity to hyper-parameters). While the default layer-wise method would focus on optimizing a single layer, it may be possible to quickly optimize SO-friendly sub-networks of a deep network to yield a fast and robust training method.

\acks{We would like to thank Michael Zibulevsky for pointing us towards additional references on the topic.
Betty Shea is funded by an NSERC Canada Graduate Scholarship. The work was partially supported by the Canada CIFAR AI Chair Program and NSERC Discovery Grant RGPIN-2022-036669.}

\appendix

\section{Solving SO Problems}
\label{app:SOsolve}

\citet{Conn1994} discuss the details of a way to numerically solve the SO problem at every iteration. They suggest using a full-memory BFGS method, which finds a local optimum with a local superlinear convergence rate under suitable conditions. Using the full-memory variant is feasible since the SO problem is low-dimensional, and allows us to easily modify the BFGS matrix to prevent it becoming unrelated to the gradient direction (which can cause BFGS to not converge). \citet{Conn1994} suggest terminating the BFGS iterations when a lack of potential progress is detected. Despite solving the SO problem inexactly, they prove that the overall algorithm is convergent if the learning rate $\alpha$ is initialized with a line search and other step sizes are initialized to 0 (assuming that the BFGS iterations monotonically decrease $f$). 

In our experiments, we found that a simple approach to solving the SO problem led to good performance in practice. In particular, we solved the SO using the simple Barzilai-Borwein (BB)~\citep{Barzilai1988} approximate quasi-Newton method, augmented with a non-monotonic Armijo line-search~\citep{Grippo1986} as in~\citet{Raydan1997}. The BB method has a superlinear convergence rate for two-dimensional quadratic functions, so we expect it to work efficiently for low-dimensional problems. To initialize the method, we simply set all step sizes to zero. 

As opposed to quasi-Newton methods, SESOP uses a Newton-like method to numerically solve the SO~\citep{Narkiss2005}. But this approach requires extra implementation effort for new problems to compute the second derivatives on the subspace. Alternately, cutting plane methods generalize the bisection-style methods used in line-search codes, but this requires additional memory and these methods require careful implementation in cases where the SO is non-convex~\citep{Hinder2018}. We note that the best performance in terms of runtime might be obtained by increasing the accuracy that we solve the SO problem as the algorithm runs; when far from a solution it is unlikely that a high accuracy is needed but we can benefit from an accurate solution as we approach an optimum. However, it is not obvious how to optimally choose the accuracy at each step\footnote{One practical approach could be to vary the maximum number of iterations allowed for the subproblem solver depending on the stage of the main optimization problem. For example, if we are at the $k$th iteration of the optimization process, we can set the maximum number of iterations allowed in the inner loop \texttt{maxIterInnerLoop} to be some multiple of $k$.}.


\section{Matrix Factorization Problems}
\label{app:MF}

Matrix factorization (MF) problems can be written in the form $f(U,W) = g(UW^T)$ for a function $g$ and matrices $U \in \R^{n\times r}$ and $W \in \R^{d\times r}$ for some rank $r \leq \min\{n,d\}$. For this problem setting, SO becomes appealing if the matrix multiplication $UW^T$ is the dominant cost, which happens as the rank $r$ grows (in the typical case where evaluating $g$ is less expensive than the matrix multiplication).
A classic example is the principal component analysis (PCA) problem,  which can be written $f(U,W) = \half\norm{UW^T - X}_F^2$ for the data matrix $X$ where $\norm{\cdot}_F$ is the matrix Frobenius norm. For the PCA setting we have $g(M) = \half\norm{M - X}^2$, and thus evaluating $g$ becomes cheaper than computing the matrix multiplication $UW^T$ as the rank $r$ grows.

The MF gradient matrices have the form~$\nabla_U f(U,W) = \nabla g(UW^T)W$ and $\nabla_W f(U,W) = \nabla g(UW^T)^TU$. Thus, the GD+M update has the form
\begin{align*}
	U_{k+1} & = U_k - \alpha_k^1\nabla g(U_kW_k^T)W_k + \beta_k^1(U_k - U_{k-1}),\\    W_{k+1} & = W_k - \alpha_k^2\nabla g(U_kW_k^T)^TU_k + \beta_k^2(W_k - W_{k-1} ).
\end{align*}
We have written this update in terms of four step sizes: a learning rate $\alpha_k^1$ on $U$, a learning rate $\alpha_k^2$ on $W$, and separate a momentum rate $\beta_k^1$ and $\beta_k^2$ for each matrix. 
The bottleneck in this update is matrix multiplications costing $O(ndr)$.
With all rates fixed to constant values this update requires 3 matrix multiplications with this cost: $U_kW_k^T$ and the matrix $\nabla g(U_kW_k^T)$ multiplied by both $W_k$ and $U_k$. A common variation is to update only one of the two matrices on each iteration (``alternating minimization'')~\citep{bell2007scalable}, and in this case only 2 matrix multiplications are required.\footnote{If we only update one matrix on each iteration, it can make sense to replace $U_{k-1}$ and $W_{k-1}$ in the momentum term with the matrices from earlier iterations.}

There are several ways we could incorporate SO into a MF optimization method:
\begin{enumerate}[leftmargin=*]
	\item{\bf Alternating Minimization}: if we only update one matrix on each iteration, then the problem considered on each iteration is an LCP. Thus, we can use SO to set the learning rate and momentum rate with only 2 matrix multiplications.
	\item{\bf Simultaneous Gradient Descent}: if we do not use momentum terms, we can use SO to optimally set the two step sizes $\alpha_k^1$ and $\alpha_k^2$,
	\begin{align*}
	& \argmin_{\alpha^1,\alpha^2} f(U_k - \alpha^1\nabla_U f(U_k, W_k),W_k - \alpha^2\nabla_W f(U_k, W_k))\\
	\equiv & \argmin_{\alpha^1,\alpha^2} g((U_k - \alpha^1\nabla g(U_kW_k^T)W_k)(W_k - \alpha^2\nabla g(U_kW_k)^TU_k)^T)\\
	\equiv & \argmin_{\alpha^1,\alpha^2} g(U_kW_k^T - \alpha^1\underbrace{\nabla g(M_k)W_kW_k^T}_{D_k^1} - \alpha^2 \underbrace{U_kU_k^T\nabla g(M_k)}_{D_k^2} + \alpha^1\alpha^2\underbrace{\nabla g(M_k)W_kU_k^T\nabla g(M_k)}_{D_k^3})\\
	\equiv & \argmin_{\alpha^1, \alpha^2}g(M_k - \alpha^1D_k^1 - \alpha^2D_k^2 + \alpha^1\alpha^2D_k^3).
\end{align*}
	If we have stored $M_k$ from the previous iteration, this requires a total of 5 matrix multiplications (two to form $D_k^1$, two to form $D_k^2$, and one more to form $D_k^3$).
	This is 2 more multiplications than using fixed a step size, but we would expect that using two step sizes would lead to more progress per iteration. Further, since the asymptotic cost is increased by a factor of less than 2, using SO to set both step sizes on each iteration is faster than re-running with two different sets of fixed step sizes. Further, compared to methods that only update one matrix per iteration, updating both matrices with an optimal learning rate only requires one additional matrix multiplication for both matrices to be updated. Finally, we note that if only one learning rate was used ($\alpha^1=\alpha^2$), the LS would still require up to 5 matrix multiplications. Thus, when updating both matrices, SO is not ``free'' in the sense that it does require additional matrix multiplications. 
	\item{\bf Momentum on One Matrix}: If we use momentum on only one of the matrices, it requires a total of 7 matrix multiplications to set all three step sizes. If we choose the $U$ matrix, this leads to an update of the form
\begin{align*}
	U_{k+1} & = U_k - \alpha^1\nabla g(U_kW_k^T)W_k + \beta(U_k - U_{k-1}),\\    W_{k+1} & = W_k - \alpha^2\nabla g(U_kW_k^T)^TU_k).
\end{align*}
The function value for a given choice of $\{\alpha^1,\alpha^2,\beta\}$ is given by
\begin{align*}
	& f(U_k - \alpha^1\nabla_U f(U_k, W_k) + \beta(U_k - U_{k-1}),W_k - \alpha^2\nabla_W f(U_k, W_k))\\
	= & g(((1+\beta)U_k - \beta U_{k-1} - \alpha^1\nabla g(U_kW_k^T)W_k)(W_k - \alpha^2\nabla g(U_kW_k)^TU_k)^T)\\
	= & g((1+\beta)U_kW_k^T  - \alpha^1\nabla g(U_kW_k^T)W_kW_k^T - \alpha^2(1+\beta) U_kU_k^T\nabla g(U_kW_k^T) \\
	& + \alpha^1\alpha^2\nabla g(U_kW_k^T)W_kU_k^T\nabla g(U_kW_k^T)
	\red{- \beta U_{k-1}W_k^T + \alpha^2\beta U_{k-1}U_k^T\nabla g(U_kW_k^T))}).
\end{align*}
These 7 matrix multiplications are two more than the 5 we require if we do not include the momentum term (the extra matrix multiplications come from the terms highlighted in \red{red}). If we used momentum on $W$ instead of $U$, it would similarly require 7 matrix multiplications.
\item {\bf Momentum on Both Matrices - Exact SO}: the update if we use a momentum term on both matrices is given by
\begin{align*}
	U_{k+1} & = U_k - \alpha^1\nabla g(U_kW_k^T)W_k + \beta^1(U_k - U_{k-1}),\\    W_{k+1} & = W_k - \alpha^2\nabla g(U_kW_k^T)^TU_k + \beta^2(W_k - W_{k-1})).
\end{align*}
The function value for a given choice $\{\alpha^1,\alpha^2,\beta^1,\beta^2\}$ is given by
\begin{align*}
	& f(U_k - \alpha^1\nabla_U f(U_k, W_k) + \beta^1(U_k - U_{k-1}),W_k - \alpha^2\nabla_W f(U_k, W_k) + \beta^2(W_k - W_{k-1})\\
	= & g(((1+\beta^1)U_k - \beta^1 U_{k-1} - \alpha^1\nabla g(U_kW_k^T)W_k)((1+\beta^2)W_k - \beta^2W_{k-1} - \alpha^2\nabla g(U_kW_k)^TU_k)^T)\\
	= & g((1+\beta^1)(1+\beta^2)U_kW_k^T  - \alpha^1(1+\beta^2)\nabla g(U_kW_k^T)W_kW_k^T - \alpha^2(1+\beta^1) U_kU_k^T\nabla g(U_kW_k^T) \\
	& + \alpha^1\alpha^2\nabla g(U_kW_k^T)W_kU_k^T\nabla g(U_kW_k^T) + \beta^1\beta^2U_{k-1}W_{k-1}^T\\
	&\red{- \beta^1(1+\beta^2) U_{k-1}W_k^T - (1+\beta^1)\beta^2U_kW_{k-1}^T  + \alpha^2\beta^1U_{k-1}U_k\nabla g(U_kW_k^T) + \alpha_1\beta^2\nabla g(U_kW_k^T)W_kW_{k-1}^T}
\end{align*}
This requires 9 matrix multiplications, which is 4 more than the 5 we require if not include any momentum terms.
	\item {\bf Momentum on Both Matrices - Inexact SO}: nine matrix multiplications to perform SO compared to 3 matrix multiplications to use fixed learning and momentum rates may be a significant increase in the iteration cost. Rather than computing all 9 required matrix multiplications and precisely optimizing the step sizes, we could instead compute 2 matrix multiplications plus one matrix multiplication for each set of step sizes that we try. Specifically, as before we track $M_k = U_kW_k^T$ and use 2 matrix multiplications to compute $\nabla_U f(U_k,W_k) = \nabla g(M_k)W_k$ and $\nabla_Wf(U_k,W_k) = \nabla g(M_k)^TU_k$. We could then compute 
\begin{align*}
	& f(U_k - \alpha^1\nabla_U f(U_k, W_k) + \beta^1(U_k - U_{k-1}),W_k - \alpha^2\nabla_W f(U_k, W_k) + \beta^2(W_k - W_{k-1})\\
	= & g((U_k - \alpha^1\nabla_U f(U_k,W_k) + \beta^1(U_k - U_{k-1}))(W_k - \alpha^2\nabla_W f(U_k,W_k) + \beta^2(W_k - W_{k-1}))^T),
\end{align*}
which requires one matrix multiplication for each  prospective $\{\alpha^1,\alpha^2,\beta^1,\beta^2\}$ we test. Thus, this may be advantageous over the precise approach if we can find suitable step sizes using less than 7 guesses.
\end{enumerate}
Modern variations on MF problems often include regularizers on the matrices $U$ and $W$, or consider matrix completion settings where $g$ only depends on a subset of the entries in $UW^T$. For example, both of these variations appear in the  probabilistic matrix factorization framework that is widely used for recommender systems~\citep{Mnih2007}. Regularization does not complicate the use of SO in the typical case where (given $UW^T$) evaluating the regularizer is much less expensive than evaluating $UW^T$ such as penalizing $\norm{UW^T}_F^2$. Matrix completion similarly supports SO since the sparse dependency on $UW^T$ can be exploited in all the required matrix multiplications.

\section{Log-Determinant Problems}
\label{app:logdet}

We say that a problem is a log determinant (LD) problem if we have a symmetric matrix variable $V \in \R^{d \times d}$ that is constrained to be positive definite and the dominant cost in evaluating the objective is computing the logarithm of the determinant of $V$, $\log|V|$. A classic example is fitting the precision matrix of a multivariate Gaussian, where $f(V) = \text{Tr}(SV) - \log|V|$ for the empirical covariance matrix $S$. In this example the trace term in this objective is a linear composition but the log determinant term is not. Note that we also obtain trace and log-determinant terms in the more complex case of fitting a multivariate student $t$~\citep{Cornish1954,Lindsey2006}, and indeed log-determinants problems tend to arise when fitting distributions under a change of variables. In this section we discuss how SO is efficient for this type of problem under certain types of updates. 

A common approach to solving log-determinant problems is coordinate optimization and block coordinate descent~\citep{Banerjee2006,Friedman2008,Scheinberg2009,Hsieh2014}. On each iteration, these methods update either a single entry or a single column (and corresponding row) of the matrix. In this setting of updating a single column in each iteration, SO is efficient due to the multilinearity property of determinants: if all columns are fixed except one, then the determinant is linear with respect to that column. Thus, in block coordinate updates where all columns are fixed except one, the update acts as if the problem was an LCP.

However, rather than restricting to updating one element/column at a time, we can alternately consider SO along general rank-1 search directions. For example, if on iteration $k$ we consider the rank-1 direction $u_ku_k^T$, by applying the matrix determinant lemma we have 
\begin{align*}
	|V_{k+1}| & = |V_k + \alpha_k u_ku_k^T|\\
	& = (1 + \alpha_k u_k^TV_k^{-1}u_k)|V_k|\\
	& = (1 + \alpha_k u_k^T\tilde{u}_k)|V_k|,
\end{align*}
where we compute $\tilde{u}_k$ by solving $V_k\tilde{u}_k = u_k$.
By tracking the determinant $|V_k|$, we can thus evaluate $|V_{k+1}|$ for many potential values of $\alpha_k$ at the cost of solving a single linear system. We can additionally use the matrix inversion lemma to perform SO on two rank-1 directions,
\begin{align*}
	|V_{k+1}| & = |V_k + \alpha_k^1 u_ku_k^T + \alpha^2_k v_kv_k^T|\\
	& = (1 + \alpha^2_k v_k^T(V_k + \alpha^1_k u_ku_k^T)^{-1}v_k)
	|V_k + \alpha^1_k u_ku_k^T|\\
	& = (1 + \alpha^2_k v_k^T(V_k^{-1} - \frac{\alpha^1_k}{1 + \alpha^1_k u_k^TV_k^{-1}u_k}V_k^{-1}u_ku_k^TV_k^{-1})v_k)
	(1 + \alpha^1_k u_k^TV_k^{-1}u_k)|V_k|\\
	& = (1 + \alpha^2_k v_k^T\tilde{v}_k - \frac{\alpha^1_k\alpha^2_k}{1 + \alpha^1_k u_k^T\tilde{u}_k}v_k^T\tilde{u}_ku_k^T\tilde{v}_k)(1 + \alpha^1_k u_k^T\tilde{u}_k)|V_k|\\
	& = ((1 + \alpha^1_ku_k^T\tilde{u}_k)(1+\alpha_k^2v_k^T\tilde{v}_k) - \alpha_k^1\alpha_k^2v_k^T\tilde{u}_ku_k^T\tilde{v}_k)|V_k|,
\end{align*}
where $\tilde{v}_k$ solves $V_k\tilde{v}_k=v_k$. Thus, performing an SO to set the step sizes $\alpha^1$ and $\alpha^2$ for two rank-1 search directions can be done for the cost of solving two linear systems. Solving these linear systems has a similar or smaller cost than computing/updating a Cholesky factorization to compute the determinant and check positive-definiteness.

\pagebreak

\bibliography{SO-arXiv}

\end{document}